\documentclass[11pt]{article}
\usepackage[margin=0.75in]{geometry}
\usepackage[T1]{fontenc}
\usepackage{lmodern,microtype}
\usepackage{amsthm, amsfonts, amssymb, amsmath,enumitem, bbm, mathabx}
\usepackage{mathtools,natbib}
\mathtoolsset{showonlyrefs}
\usepackage{graphicx}
\usepackage{hyperref}
\hypersetup{
    colorlinks = true,
    citecolor  = blue,
    linkcolor  = red,
    urlcolor   = black
}
\usepackage{geometry, xcolor}
\usepackage{fancyhdr}
\usepackage{algorithm}
\usepackage{algorithmic}
\usepackage{booktabs} 
\usepackage{tabularx}

\usepackage{subcaption}

\usepackage{colortbl}
\newcommand{\greymidrule}{\arrayrulecolor{black!30}\midrule\arrayrulecolor{black}}

\makeatletter
\renewcommand\p@subfigure{Figure~\thefigure.}
\makeatother

\definecolor{maroon}{RGB}{128,0,0}

\newcommand{\R}{\mathbb{R}}      
\newcommand{\N}{\mathbb{N}}      
\newcommand{\IP}{\mathbb{P}} 
\newcommand{\Hcal}{\mathcal{H}}
\newcommand{\Ncal}{\mathcal{N}}
 
\newcommand{\Var}{\operatorname{Var}} 

\newcommand{\SC}{\texttt{SC}}
\newcommand{\CC}{\texttt{C}}
\newcommand{\QC}{\texttt{QC}}
\newcommand{\LC}{\texttt{LC}}
\newcommand{\St}{\textsc{St}}
\newcommand{\STC}{\texttt{StC}}
\newcommand{\LS}{\texttt{LS}}
\newcommand{\LSC}{\texttt{LSC}}
\newcommand{\DISS}{\texttt{DISS}}
\newcommand{\LG}{\texttt{LG}}

\newcommand{\RSI}{\texttt{RSI}}

\newcommand{\newname}{\texttt{CGF-like}}

\newcommand{\wh}{\widehat}
\newcommand{\wt}{\widetilde}

\newcommand{\dunif}{\mathrm{Unif}}

\DeclareMathOperator*{\argmin}{arg\,min}
\DeclareMathOperator*{\argmax}{arg\,max}

\def\IE{\mathbb{E}}
\def\R{\mathbb{R}}

\def\N{\mathbb{N}}
\def\bo{\mathbf{1}}

\newcommand{\Dcal}{\mathcal{D}}
\newcommand{\bv}{\mathbf{v}}

\newcommand{\Lcal}{\mathcal{L}}

\newcommand{\Mcal}{\mathcal{M}}
\newcommand{\DMcal}{\mathcal{DM}}

\newcommand{\Fcal}{\mathcal{F}}

\newcommand{\Scal}{\mathcal{S}}
\newcommand{\Vcal}{\mathcal{V}}
\newcommand{\Xcal}{\mathcal{X}}

\newcommand{\op}{\operatorname{op}}
\newcommand{\ind}[1]{\boldsymbol{1}\{#1\}}
\newcommand{\abs}[1]{\left\lvert #1 \right\rvert}

\newcommand{\Tr}{\mathrm{Tr}}

\newcommand{\given}{\hspace{1pt}\vert\hspace{1pt}}

\newcommand{\PL}{\texttt{P\L{}}}
\newcommand{\KLS}{\texttt{K\L{}$^\star$}}

\theoremstyle{plain}

\newtheorem{asmp}{Assumption}
\newtheorem{thm}{Theorem}[section]
\newtheorem{prop}{Proposition}
\newtheorem{lem}[thm]{Lemma}
\newtheorem{cor}[thm]{Corollary}

\newtheorem{defn}[thm]{Definition}

\usepackage[T1]{fontenc}
\usepackage{amsmath,amssymb}
\usepackage{xcolor}
\usepackage{tikz}
\usetikzlibrary{arrows.meta,positioning,calc, tikzmark, decorations.pathreplacing}

\definecolor{flice}{HTML}{DCEEFB}
\definecolor{flhole}{HTML}{1F3A5F}
\definecolor{flgoal}{HTML}{F2C14E}
\definecolor{flstart}{HTML}{B8E0D2}
\definecolor{flarrow}{HTML}{B22222}
\def\flH{H}\def\flG{G}\def\flS{S}
\newcommand{\flgrid}{%
  \foreach \row [count=\r from 0] in {{S,F,F,F},{F,H,F,H},{F,F,F,H},{H,F,F,G}}{
    \foreach \cell [count=\c from 0] in \row {
      \ifx\cell\flH\def\fc{flhole}\def\tc{white}\else
      \ifx\cell\flG\def\fc{flgoal}\def\tc{black}\else
      \ifx\cell\flS\def\fc{flstart}\def\tc{black}\else
      \def\fc{flice}\def\tc{black}\fi\fi\fi
      \fill[\fc] (\c,-\r) rectangle ++(1,-1);
      \node[text=\tc,font=\ttfamily\footnotesize] at (\c+0.84,-\r-0.82) {\cell};
    }
  }
  \draw[gray!70,thin] (0,0) grid[step=1] (4,-4);
  \draw[thick] (0,0) rectangle (4,-4);
}

\definecolor{geomfill}{HTML}{EAF2FB}
\definecolor{geomborder}{HTML}{356A9A}

\definecolor{boundfill}{HTML}{F2ECFA}
\definecolor{boundborder}{HTML}{7451A1}

\definecolor{driftfill}{HTML}{FFF2DF}
\definecolor{driftborder}{HTML}{B66A16}

\definecolor{regfill}{HTML}{E9F6F1}
\definecolor{regborder}{HTML}{2D7A67}

\tikzset{
  assumption/.style={
    rounded corners=2.2mm,
    minimum width=2.65cm,
    minimum height=1.05cm,
    align=center,
    inner sep=4.5pt,
    line width=0.9pt,
    font=\sffamily
  },
  geometry/.style={
    assumption,
    draw=geomborder,
    fill=geomfill
  },
  errorbound/.style={
    assumption,
    draw=boundborder,
    fill=boundfill
  },
  drift/.style={
    assumption,
    draw=driftborder,
    fill=driftfill
  },
  regularity/.style={
    assumption,
    draw=regborder,
    fill=regfill
  },
  implication/.style={
    -{Stealth[length=3.0mm,width=2.1mm]},
    line width=1.0pt,
    draw=black!72,
    shorten >=2pt,
    shorten <=2pt
  }
}

\title{A general recipe for data-adaptive anytime-valid envelopes for iterative algorithms}
\author{%
\begin{tabular}{cc}
Soham Bonnerjee & Aaditya Ramdas \\
\textit{University of Pennsylvania} & \textit{Stanford University} \\
\texttt{sohamb@wharton.upenn.edu} & \texttt{aramdas@stanford.edu}
\end{tabular}
}
\date{}

\begin{document}
\maketitle
\begin{abstract}
Most convergence guarantees for iterative stochastic algorithms are stated at a fixed iteration count and do not justify stopping at a data-dependent time. Recent anytime-valid envelopes do, but deterministic ones are typically too conservative for practical stopping, and existing data-adaptive ones are derived case-by-case, each concerned with a specific algorithm and a specific noise model and boundary construction. This paper provides a general recipe for sharp, data-adaptive anytime-valid envelopes for a large class of iterative algorithms. Our approach has two key steps: an optimization module derives a one-step recursion that performance measures of many such algorithms satisfy, reducing the error analysis to time-uniform concentration for predictably weighted sums of noise. Then, a statistical module ports the sub-$\psi$ condition to vector-valued processes, uniformly over predictable weights, and derives off-the-shelf boundaries via mixtures, stitching, and product betting martingales. Combining the two yields readymade envelopes for many iterative algoithms including but not limited to stochastic gradient descent (SGD), Oja's algorithm, temporal difference learning, Q-learning, and Bures-Wasserstein SGD, many with no prior data-adaptive anytime-valid guarantees. Specifically, for SGD, our framework recovers, often under weaker conditions, or significantly sharpens every existing envelope we are aware of. Our constructions come with explicit implementation details and are substantiated by numerical experiments.
\end{abstract}

\section{Introduction}
We consider the problem of deriving sharp, observable, anytime-valid envelopes to oracle performance measures associated with online stochastic optimization algorithms. Concretely, consider an algorithm $\{x_t, \xi_t\}_{t\ge1} \in \Hcal \otimes \Hcal$ characterized by the recursion $x_t = h_t(x_{t-1}, \xi_t)$ for some functions $\{h_t\}_{t\ge 1}$, where $\Hcal$ is some Hilbert space, and $\{\xi_t\}_{t=1}^\infty \sim P$ for some Borel measure $P$ on $\Hcal^{\infty}$. The associated performance measure $\{Z_t\}_{t\ge 1}$ usually concerns how close the algorithm chain $\{x_t\}_{t\ge 1}$ is to convergence, where the notion of convergence may differ based on applications. Common examples of $Z_t$, for algorithms such as stochastic gradient descent (SGD), include squared distance \(\|x_t-x^\star\|^2\) to an unknown optimum $x^\star$, or the objective suboptimality \(F(x_t)-F(x^\star)\), where $F$ is the objective function in question. Clearly, this measure is oracle in nature, which neither informs any stopping time-analysis, nor provides a valid confidence set around the optimum when the algorithm is stopped at an ad-hoc stopping time (such as, when $\|x_t - x_{t-1}\|$ falls below a certain threshold, or stopping based on asymptotic convergence guarantees at a pre-specified iteration). Both these issues can be addressed by constructing a non-asymptotic, anytime-valid upper bound to the measure $Z_t$. Specifically, for a prescribed \(\alpha\in(0,1)\) and a deterministic starting time \(t_0\geq1\), we seek an envelope \(\widetilde{\mathcal M}_t(\alpha)\) satisfying
\begin{align}
    \mathbb P\!\bigl(
Z_t\leq\widetilde{\mathcal M}_t(\alpha)
\text{ for all }t\geq t_0
\bigr)\geq1-\alpha.
\label{eq:anytime-valid}
\end{align} 
\begin{figure}[!tbp]
\centering
\begin{subfigure}[t]{0.43\textwidth}
\centering
\vspace{0pt}
\includegraphics[width=\linewidth]{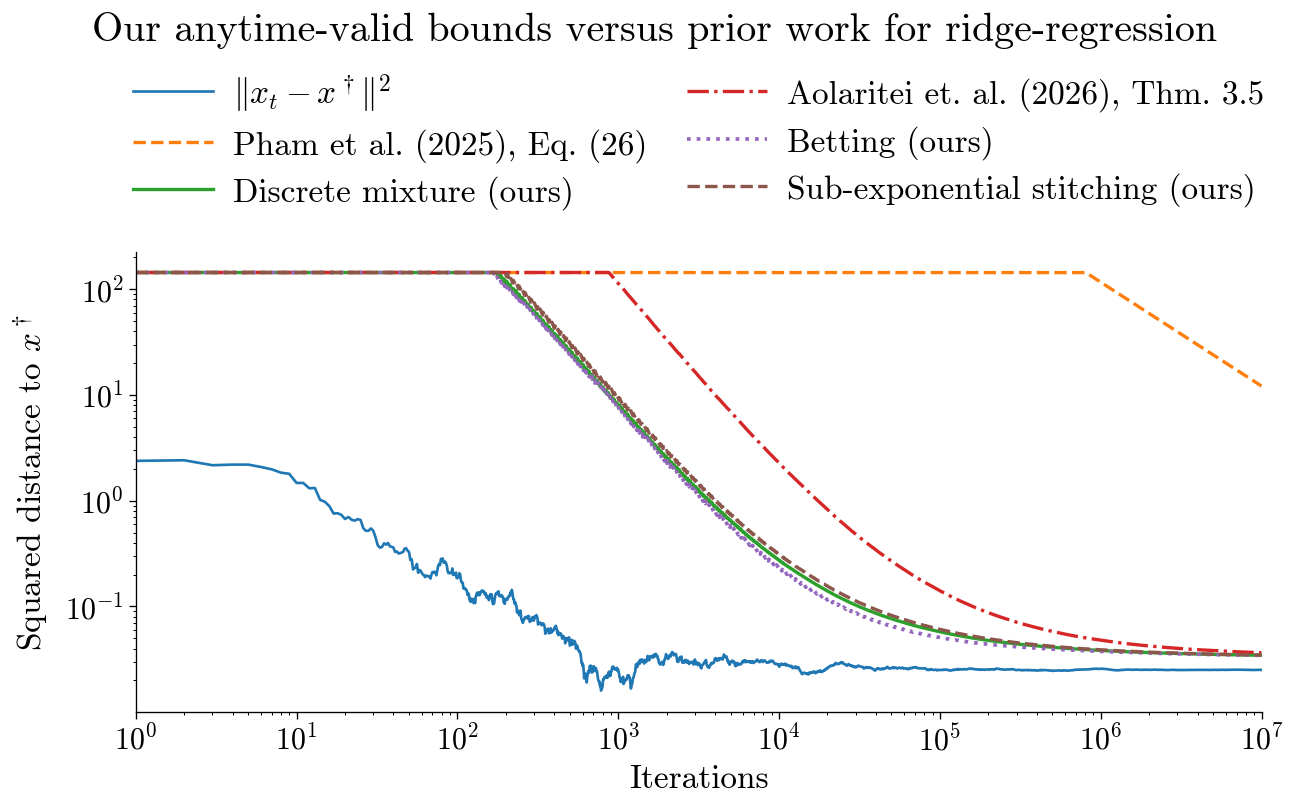}
\caption{}
\label{fig:ridge-prelim}
\end{subfigure}
\hfill
\begin{subfigure}[t]{0.54\textwidth}
\centering
\vspace{0pt}
\resizebox{\linewidth}{!}{%
\begin{tikzpicture}[x=0.75pt,y=0.75pt,yscale=-1,xscale=1]

\draw[fill={rgb,255:red,236;green,194;blue,194},fill opacity=1]
(33,39) -- (252.5,39) -- (252.5,137) -- (33,137) -- cycle;

\draw[fill={rgb,255:red,248;green,231;blue,28},fill opacity=1]
(360.5,51) -- (571.5,51) -- (571.5,117) -- (360.5,117) -- cycle;

\draw[fill={rgb,255:red,230;green,241;blue,218},fill opacity=1]
(360,198) -- (571.5,198) -- (571.5,282) -- (360,282) -- cycle;

\draw[fill={rgb,255:red,231;green,222;blue,208},fill opacity=1]
(690,57) -- (850.5,57) -- (850.5,101) -- (690,101) -- cycle;

\draw[dash pattern={on 4.5pt off 4.5pt}]
(23.5,28) -- (579.5,28) -- (579.5,151) -- (23.5,151) -- cycle;

\draw[dash pattern={on 4.5pt off 4.5pt}]
(682,107) -- (851.5,107) -- (851.5,50) -- (682,50) -- cycle;

\draw[fill={rgb,255:red,239;green,211;blue,245},fill opacity=1]
(34.5,274) -- (227.5,274) -- (227.5,222) -- (34.5,222) -- cycle;

\draw[dash pattern={on 4.5pt off 4.5pt}]
(27,216) -- (235.5,216) -- (235.5,281) -- (27,281) -- cycle;

\draw[fill={rgb,255:red,212;green,243;blue,236},fill opacity=1]
(360,359) -- (611.5,359) -- (611.5,408) -- (360,408) -- cycle;

\draw[-{Stealth[length=2.5mm,width=2mm]}]
(471,125) -- (471,190);

\draw[-{Stealth[length=2.5mm,width=2mm]},rounded corners=4pt]
(676.5,114) -- (576.5,196);

\draw[-{Stealth[length=2.5mm,width=2mm]}]
(131.5,143) -- (131.5,188);

\draw[-{Stealth[length=2.5mm,width=2mm]},rounded corners=5pt]
(240.5,287) --  (360.5,358);

\draw[-{Stealth[length=2.5mm,width=2mm]}]
(471.5,289) -- (471.5,350);

\draw (51,43) node[anchor=north west, inner sep=0.75pt, align=left]
{{\Large Optimization module}};

\draw (38,83) node[anchor=north west, inner sep=0.75pt, align=center]
{\hspace*{0.75cm}(eg. \SC/\CC/\PL/\LS)};

\draw (380,54) node[anchor=north west, inner sep=0.75pt, align=left]
{{\Large Noise assumptions}};

\draw (418,93) node[anchor=north west, inner sep=0.75pt, align=left]
{Definition~\ref{def:psi}};

\draw (399,204) node[anchor=north west, inner sep=0.75pt, align=center]
{Anytime-valid intervals\\ on $\displaystyle\sum_{s=1}^{t}\langle v_s,\xi_s\rangle$};

\draw (700,70) node[anchor=north west, inner sep=0.75pt, align=center]
{\footnotesize stitching/mixing/betting};

\draw (247,8) node[anchor=north west, inner sep=0.75pt, align=left]
{User assumptions};

\draw (725,27) node[anchor=north west, inner sep=0.75pt, align=left]
{User's choice};

\draw (41,226) node[anchor=north west, inner sep=0.75pt, align=center]
{{\large \hspace*{0.4cm} Performance measure}\\ $Z_t$};

\draw (90,191) node[anchor=north west, inner sep=0.75pt, align=left]
{User's goal};

\draw (363,365) node[anchor=north west, inner sep=0.75pt, align=center]
{Final anytime-valid intervals\\ on SGD iterates $x_t$ through recursion};

\end{tikzpicture}%
}
\caption{}
\label{fig:schematic}
\end{subfigure}

\caption{\textbf{(a)} Comparison of different anytime-valid upper bounds versus $t$ for a particular SGD run on a ridge-regression problem, where $x^\dagger$ is the true regression coefficient. The \cite{pham2025time}, Eqn.~(26) boundary, in \textcolor{orange}{orange}, is deterministic. The \cite{aolaritei2026beyond}, Thm. 3.5 boundary, in \textcolor{red}{red}, is recovered by our framework as the sub-Gamma stitching bound; see \eqref{eq:AJ}. The discrete mixture boundary, in \textcolor{green}{green} employs Corollary~\ref{corol:dm} with $f(\lambda)
=
{0.4\,\mathbf{1}_{0<\lambda\le 0.38}}
/{\lambda \log^{1.4}({0.38e}/{\lambda})}.$ The sub-exponential stitching in \textcolor{maroon}{maroon} employs Proposition~\ref{prop:stitch} and sharpens Theorem 3.5 of \cite{aolaritei2026beyond}; see \S\ref{se:sub-gamma-stitch}. The betting boundary in \textcolor{violet}{violet} employs Proposition~\ref{prop:betting-SGD}. Since we run projected SGD, every boundary is capped at the global bound implied by the projection set. The details can be found in \S\ref{se:simu-det-comparison}. \textbf{(b)} A schematic of our general approach; the user supplies structural assumptions on the algorithm and on the gradient noise together with a performance measure $Z_t$; the probabilistic module returns anytime-valid bounds on the weighted noise sum, which are then transferred through \eqref{eq:key-recursion} to yield the final envelope.}
\label{fig:placeholder}
\end{figure}
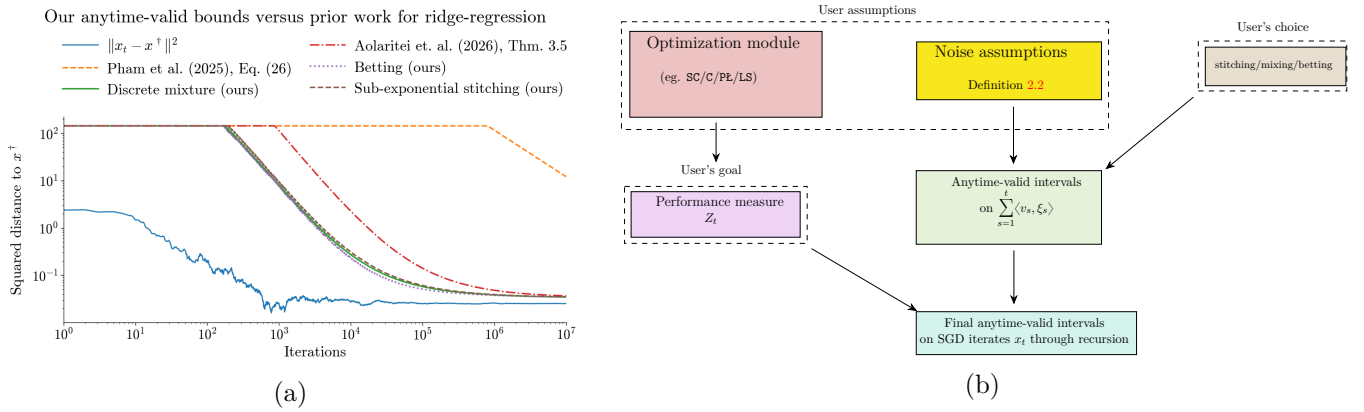
 The anytime-valid envelopes facilitate arbitrary and post-hoc data-dependent stopping times, allowing for potentially incorporating external domain-knowledge \citep{grunwald2024safe, ramdas2023game}. The existing literature on anytime-valid inference for optimization algorithms \citep{feng2023anytime, feng2025breaking, pham2025time, aolaritei2025stopping, aolaritei2026beyond} is nascent, and they have left two important questions open: 
 \begin{itemize}
 \item[(i)] Sharpness: \textit{how to ensure sharpness of the constructed envelopes?} 
 A substantial section of boundaries constructed thus far have placed particular emphasis on optimal boundaries matching the fixed iterate analysis in the $\Tilde{O}(\cdot)$ sense. These deterministic anytime-valid envelopes are tuned to worst-case control, and are usually too large to be useful in practice. Therefore fail to provide any reasonable data-dependent stopping times. Moreover, often such results require particular choices of deterministic step-sizes, such as $\eta_t\asymp 1/t$ in \citep{pham2025time, kar2026high} or $\eta_t \asymp 1/\log^2 t$ in \cite{feng2023anytime}, hindering generalization to the more widely used data-dependent/adaptive step-sizes. The works closest to ours, \citet{aolaritei2025stopping,aolaritei2026beyond}, partially address this issue by constructing data-adaptive time-uniform boundaries. Their boundaries, however, rely on sub-Gamma stitching, leaving open whether sharper boundaries can be obtained under the same assumptions through betting, much as \citet{waudby-jrssb} sharpened empirical-Bernstein confidence sequences by betting. We remark that the question of sharp envelopes, or early stopping in an online algorithm, is supremely important in myriad applications. In particular, for many machine-learning problems \citep{zoph2017neural, shao2024deepseekmath, shao2025deepseekmath} as well as other scientific applications \citep{pai2023, loeffler2024reinvent}, there exists a physical as well as computational cost of procuring new observations as well as gradient computation. In such situations, the role of a sharp envelope is paramount, as it provides an early stopping time while maintaining validity, potentially saving the user significant resources. 
     \item[(ii)] Generalizability: \textit{can we get a unifying framework that does not require us to procure boundaries on a case-by-case basis for different algorithms and starting assumptions?} 
   Another common aspect of the existing time-uniform analysis of stochastic optimization algorithms is, that they are largely case-by-case. Most of them concern stochastic gradient descent, and do not extend to other iterative algorithms such as TD-learning or Q-learning. Even within SGD, each analysis is tailored to a fixed combination of noise assumptions (typically sub-Gaussian), structural assumptions (convexity, or particular forms of non-convexity) and performance measures; see, e.g., \citet{pham2025time,feng2023anytime,feng2025breaking,li2026exact}. On a high level, and despite apparent similarity in their treatments, every new combination seems to call for a new analysis. 
 \end{itemize}
 We address both these questions by crystallizing the construction of the envelopes $\widetilde{\mathcal M}_t(\alpha)$ into two broad steps, motivated, respectively, by the key recursion \eqref{eq:key-recursion} satisfied by the performance measures, and by the burgeoning literature on anytime-valid testing and confidence sequences (starting from \citet{howard2018, howard2021time}; see \S\ref{se:lit-rev}). The former accommodates general, data-dependent step sizes and incorporates the noisy observations, such as the stochastic gradients in SGD; the latter, through the sub-$\psi$ analysis and variants of product betting martingales, handles a host of combinations of structural assumptions, noise assumptions and performance measures:
\begin{itemize}
    \item In the first step, through an \textit{optimization module}, we reduce the problem to constructing anytime-valid envelopes for arbitrary linear combinations of (dependent) random variables. Our approach is based on a variant of the Robbins--Siegmund \textit{almost-supermartingale} framework \citep{robbins-siegmund, lai1989, pham2025time, feng2025breaking}. In particular, for any iterative algorithm $(x_t, \xi_t)$ and the corresponding performance measure $Z_t$, we show that $Z_t$ satisfies a recursion
\begin{align}
Z_t \le a_t Z_{t-1} + b_t + \langle v_t, \xi_t \rangle,
\label{eq:key-recursion}
\end{align}
where $a_t$ and $v_t$ are \textit{predictable}, i.e., $\Fcal_{t-1}:=\sigma(\xi_s: s<t)$-measurable, and $b_t$ is observed or data-dependent. Unrolling the recursion \eqref{eq:key-recursion} reduces the error analysis to control of weighted noise $\sum_{s=1}^t\langle v_s, \xi_s \rangle$; see Lemma \ref{lem:observable-envelope}.
    \item In the second step, we build on the sub-$\psi$ framework of \cite{howard2018,howard2021time, whitehouse2026time} to implement different strategies of anytime-valid bound constructions, such as mixing, stitching, and betting \cite{waudby-jrssb, martinez2026Gaussianefficient}. Porting these constructions into our framework is not straightforward; in \eqref{eq:key-recursion}, the noise variables $\xi_t$ are not directly observable, which adds to the complications of the construction. Moreover, the weights $v_t$ are arbitrary predictable sequences, often depending on oracle quantities such as $x^\star$, and cannot be absorbed directly into the global sub-$\psi$ assumptions of \cite{howard2018,howard2021time, whitehouse2026time}. We circumvent these issues via adapting the slightly stronger sub-$\psi$ condition from (\cite{casgrain2024}) to vector-valued processes (see Definition~\ref{def:psi} herein), that ensures tractable tail behavior uniformly over predictable weights. This \textit{statistical module} allows us to construct envelopes, which we then port together with observable surrogates for the oracle quantities, built inductively from previous envelopes (Lemma~\ref{lem:observable-envelope}); for betting, we further exploit the boundedness of the observed gradients (Theorem~\ref{thm:obs-betting}).
    We defer the  details of boundary constructions to \S\ref{se:prob}.
\end{itemize}
Combining the two modules above furnishes a unified recipe of deriving anytime-valid boundaries for generic iterative algorithms. We document the two key benefits of our generalized perspective and  the resulting constructions.
\begin{itemize}
\item[(i)] Our modularized approach allows us to move away from a case-by-case analysis hitherto prevalent in the nascent literature of time-uniform boundary constructions for iterative algorithms; see the discussion in \S\ref{se:lit-rev}. In particular, decoupling the optimization module from the anytime-valid construction means that algorithm-specific tools are needed only to derive \eqref{eq:key-recursion}. This modularity lets us recover (often under weaker conditions) and sharpen the few existing time-uniform bounds in the SGD literature. It also yields entirely new bounds for iterative algorithms that have never been given an anytime-valid analysis. Table \ref{tab:literature-comparison} summarizes the existing results we recover and sharpen, along with the new results we establish for algorithms previously unexplored in the time-uniform setting.
    \item[(ii)] Our envelopes are not deterministic; they make use of the observed data-streams. Moreover, the use of exact product-betting martingales allows our envelopes to be much sharper than the existing stitching based envelopes.  Therefore, our envelopes provide the sharpest such constructions in this literature; see \ref{fig:ridge-prelim}.
\end{itemize}

\begin{table}[tbp]
\centering
\small
\setlength{\tabcolsep}{4pt}
\newcommand{\grpend}[1]{\tikzmark{#1}\\[-\normalbaselineskip]}

\begin{tikzpicture}[remember picture, overlay,
    grp/.style={decorate, decoration={brace, amplitude=4pt}, line width=0.6pt}]
  \draw[grp] ([xshift=-4pt, yshift=\the\ht\strutbox]pic cs:Ibot)
          -- ([xshift=-4pt, yshift=\the\ht\strutbox]pic cs:Itop)
          node[midway, left=5pt] {\textbf{I}};
  \draw[grp] ([xshift=-4pt, yshift=\the\ht\strutbox]pic cs:IIbot)
          -- ([xshift=-4pt, yshift=\the\ht\strutbox]pic cs:IItop)
          node[midway, left=5pt] {\textbf{II}};
  \draw[grp] ([xshift=-4pt, yshift=\the\ht\strutbox]pic cs:IIIbot)
          -- ([xshift=-4pt, yshift=\the\ht\strutbox]pic cs:IIItop)
          node[midway, left=5pt] {\textbf{III}};
\end{tikzpicture}%
\hspace*{3em}
\begin{tabularx}{\dimexpr\linewidth-3em\relax}{@{}X p{0.27\linewidth} *{3}{c}@{}}
\toprule
Existing results & Result in this paper & [A] & [B] & [C] \\
\midrule

\multicolumn{5}{@{}l}{\tikzmark{Itop}\textbf{Stochastic gradient descent}} \\
Strongly convex\newline
\citet[Thm.~2.3,~3.5]{aolaritei2026beyond}\newline \citet[Cor.~1]{pham2025time}
& \S\ref{se:rsi}; \S\ref{se:sub-gamma-stitch}; Prop.~\ref{prop:betting-SGD}
& $\checkmark$ & $\checkmark$ & --- \\
\midrule
\addlinespace[3pt]
Convex\newline
\citet[Cor.~2.2]{aolaritei2025stopping}
& Prop.~\ref{prop:convex-hybrid}; \S\ref{app:convex-stitching}
& $\checkmark$ & $\checkmark$ & --- \\
\midrule
\addlinespace[3pt]
Nonconvex\newline
\citet[Cor.~2.6]{aolaritei2025stopping}; \newline \citet[Cor.~3]{pham2025time}
& Prop.~\ref{prop:ls-hybrid}; \S\ref{se:Lipschitz-stitching}; \S\ref{se:kl-sgd}
& $\checkmark$ & $\checkmark$ & --- \\
\grpend{Ibot}
\midrule
\addlinespace[3pt]

\tikzmark{IItop}
Polyak--Ruppert averaging\newline
\citet[Thm.~4.4]{xie2024asymptotic}
& Thm.~\ref{thm:pr-coverage} and~\ref{thm:pr-rates}$^\dagger$
& --- & --- & $\checkmark$ \\
\midrule[\heavyrulewidth]
\multicolumn{5}{@{}l}{\textbf{Momentum-assisted SGD}} \\
Sub-Gaussian noise\newline  
\citet[Thm.~5--6]{feng2023anytime};
\newline
\citet[Thm.~1, Prop.~1]{feng2025breaking}
& Prop.~\ref{prop:mom-rec};
  Thm.~\ref{thm:mom-coverage},~\ref{thm:mom-rates}$^\dagger$
& --- & $\checkmark$ & $\checkmark$ \\
\midrule
\addlinespace[3pt]
Finite variance\newline
\citet[Thm.~3--4]{feng2023anytime}
& Prop.~\ref{prop:mom-quad}; Cor.~\ref{cor:mom-fv}$^\dagger$
& --- & $\checkmark$ & $\checkmark$ \\
\midrule
\addlinespace[3pt]
\textbf{Oja's algorithm}\newline
\citet[Prop.~4, Thm.~3]{pham2025time}
& Prop.~\ref{prop:oja-rec}; \S\ref{se:oja}
& --- & $\checkmark$ & $\checkmark$ \\
\grpend{IIbot}
\midrule[\heavyrulewidth]

\tikzmark{IIItop}\textbf{Temporal difference learning}
& Thm.~\ref{thm:td-last}
& --- & --- & $\checkmark$ \\
\midrule
\addlinespace[3pt]
\textbf{Q-learning}\newline
\citet[Algorithm 6]{evendar2006action}
& Prop.~\ref{prop:q-rec}; Thm.~\ref{thm:q}
& --- & --- & $\checkmark$ \\
\midrule
\addlinespace[3pt]
\textbf{Bures - Wasserstein SGD}
& Thm.~\ref{thm:bw}; Prop.~\ref{prop:bw-stat}
& --- & --- & $\checkmark$ \\
\grpend{IIIbot}
\bottomrule
\end{tabularx}
\caption[Comparison of our results with the relevant literature.]{%
  Comparison of our results with the relevant literature. For brevity, the table displays only results that are explicitly stated or discussed in this paper. To summarize, \textbf{I} compiles examples with existing anytime-valid, data-dependent boundaries. In contrast, existing time-uniform boundaries for the examples in group \textbf{II} have been deterministic, and in the case of PR-averaged SGD, only asymptotic. Finally, group \textbf{III} accumulates some cases where no time-uniform guarantees were previously available. Many further results, all of which are the first of their kind in the corresponding setting, follow directly by pairing the optimization modules of \S\ref{app:opt-sgd-boundaries} with Theorems~\ref{thm:obs-comp}--\ref{thm:obs-betting} and Lemma~\ref{lem:observable-envelope}. 
    \begin{minipage}[t]{\linewidth}
    \begin{enumerate}[leftmargin=2.5em, itemsep=0pt, topsep=2pt]
      \item[\textbf{[A]}] Recovers the cited guarantee on the stated comparison domain under the same or weaker conditions, or other assumptions on the noise distribution.
      \item[\textbf{[B]}] Improves the result by providing a sharper envelope with the same initialization and under the same assumptions.
      \item[\textbf{[C]}] Provides the \textit{first nonasymptotic, anytime-valid, and data-dependent} envelope in the corresponding literature or specific setting. Previous results in these settings either provided deterministic envelopes or did not exist.
    \end{enumerate}
  \end{minipage}
}
\vspace*{0.5cm}
\small $^\dagger$: The boundaries in Theorem~\ref{thm:pr-rates} also achieve the optimal rate in the fixed-$t$ sense.
\label{tab:literature-comparison}
\end{table}

We emphasize that on a structural level, we require no particular tail assumptions, such as sub-Gaussianity on the noise process driving the stochastic algorithms, nor any growth or smoothness conditions, such as convexity or its generalizations, on the optimization problem itself. Any number of these conditions can be applied on top of our constructions to derive the suitable recursion \eqref{eq:key-recursion}, and the resulting anytime-valid boundary. The co-ordination between these two steps, and the flexibility of this modularization, will be made clearer through the analyses in \S\ref{se: anytime-valid-sgd}, \S\ref{se:appl}, \S\ref{app:opt-sgd-boundaries} and \S\ref{app:appl}. We now situate our recipe within the broader literature on time-uniform inference and stochastic optimization.

\subsection{Prior Work} \label{se:lit-rev}
Iterative stochastic algorithms are one of the commonest forms of sequential experiments. Common examples include the ubiquitous stochastic gradient descent (SGD; \citet{robbins1951stochastic}) and its variants, including momentum-based and accelerated methods \citep{polyak1964some, nesterov1983method}, as well as linear stochastic approximation methods such as Temporal Difference learning \cite{sutton1988learning} and Q-learning \cite{watkins1992q}; online principal component analysis (PCA) using Oja's algorithm \cite{oja1982simplified}; stochastic mirror descent \cite{nemirovski1983problem, nemirovski2009robust}; and many others. Classical non-asymptotic analyses \citep{tong2004, moulines2011non, rakhlin, ghadimi2012optimal, shamir2013stochastic, harvey2019tight, liu2024revisiting, harvey2019simple} have mainly focused on deriving optimal high-probability or in-expectation rates of averaged or last-iterate performances of SGD and its variants in terms of Landau notations $O(\cdot)$, working mainly with the strongly convex optimization functions with various smoothness assumptions. Beyond strong convexity, complementary works such as \citet{ramdas13, schmidt2013fast, ghadimi2016accelerated, lei2017less, zaheer2018adaptive, bottou2018optimization, lan2020first, li2021page, kavis2022high, fatkhullin2025can} provide asymptotic convergence, and non-asymptotic performance control results for general convex and non-convex stochastic optimization with SGD and its variants under different regimes of schedules.

Despite sustained research on the convergence of SGD and stochastic approximation algorithms, anytime-valid envelopes for the relevant error metrics have only recently begun to attract attention. In particular, \cite{feng2023anytime} notes the limited development of such methods that are tailored for stochastic optimization, despite the extensive literature on anytime-valid testing, confidence sequences, and related inference \citep{darling1967confidence, lai1976confidence, ramdas2023game, ramdas2025hypothesis}. This being a work at the cross-roads of the anytime-valid inference and stochastic optimization, we take a quick scan of the anytime-valid inference literature, before shifting focus to their application in stochastic optimization, where we further position our contribution vis-à-vis related developments.

\paragraph{Anytime-valid inference.}
Time-uniform inference owes its germination to the sequential tests and the confidence sequences of \citet{wald1945sequential, darling1967confidence, lai1976confidence}. Moreover, nearly all modern analyses leverage Ville's inequality \citep{ville1939etude} which establishes a time-uniform control over non-negative supermartingales - a structure ubiquitous in iterative algorithms. This approach was popularized by \citet{howard2018, howard2021time} who discussed the \textit{stitching} method of joining linear boundaries into curved ones. They also employed the classical method of mixtures \citep{darling1968some, kaufmann2021mixture} to construct continuous and discrete \textit{mixture} boundaries. In particular, the stitching method was shown to achieve the optimal iterated logarithmic rates. This framework has since been extended to vector-valued processes \citep{whitehouse2026time}, as well as into  more application-specific refinements \citep{casgrain2024,chugg2023time,chugg-pac-bayes}. A closely related thread of work considers time-uniform empirical Bernstein confidence sequences \citep{howard2021time, waudby-jrssb}, an instance of which appears in the SGD envelopes of \citet{aolaritei2026beyond}. These results have recently been further adapted to myriad settings \citep{chugg2025closed, wang2026sharp, martinez2025sharp, martinez2026empirical}. In parallel, and rooted in game-theoretic probability \citep{shafer2021testing}, betting-based constructions build test supermartingales as the wealth of a gambler wagering against the null with predictable stakes. For bounded observations these yield confidence sequences that are often markedly tighter than their stitching or empirical-Bernstein counterparts \citep{waudby-jrssb, shekhar2023near, orabona2023tight}, with recent refinements in \citet{martinez2026Gaussianefficient, voracek2026star}. Both approaches fit within the broader theory of e-values and e-processes \citep{vovk2021evalues, grunwald2024safe}; see \citet{ramdas2023game, ramdas2025hypothesis} for comprehensive treatments. Our probabilistic module draws on both the sub-$\psi$ and betting constructions to instantiate mixture, stitching and betting boundaries in \S\ref{se:prob}.

\paragraph{Stochastic optimization and time-uniform boundaries.}
A small but growing body of work on nonasymptotic anytime-valid envelopes for stochastic optimization draws on these developments \citep{feng2023anytime,pham2025time,feng2025breaking,kar2026high,chen2026revisiting}. These works largely concern SGD under fixed noise and structural assumptions. The closest precedents to this paper, \citet{aolaritei2025stopping,aolaritei2026beyond}, obtain data-adaptive boundaries through sub-Gaussian and sub-Gamma stitching constructions. The key questions of \textit{generalizability} and \textit{sharpness} of the envelopes that they leave open, have already been discussed. Even on the asymptotic time-uniform front, \cite{xie2024asymptotic} employs time-uniform Gaussian approximation results in the spirit of \cite{waudby2023distribution,waudby2024time, waudby2025nonasymptotic}. In fact, the substantial similarity in the underlying approaches of these papers suggests the potential for a unifying framework that systematically organizes and improves them- a task we undertake in this paper. This is a natural point to offer a bird's-eye view of the contributions and organization of this paper.

\subsection{Organization of the paper} The general probabilistic toolbox, including the different approaches to constructions of boundaries, is discussed in \S\ref{se:prob}. Section~\ref{se: anytime-valid-sgd} then illustrates our approach in the canonical setting of SGD. Under conditions previously studied in the literature, such as convexity or Lipschitz gradients, we derive envelopes that are substantially sharper than existing ones. Finally, \S\ref{se:appl} moves beyond SGD to five further applications, providing the first instances of data-dependent anytime-valid boundaries in each setting. In particular, previous anytime-valid confidence sequences \citep{aolaritei2025stopping,aolaritei2026beyond}, derived for SGD, do not apply directly either, since these examples, such as TD or Q-learning algorithms, are not stochastic gradient methods. The ease and sharpness of these derivations stem from our modularized perspective, which lends a level of generalizability while enabling us to remove slack from hitherto overlooked aspects of any particular problem. 

\paragraph{Notations.}Here $\mathcal B(I)$ denotes the Borel sigma-field on $I$. For a measurable space $\mathcal X$, $\mathcal P(\mathcal X)$ denotes the set of probability measures on $\mathcal X$; when $\mathcal X$ is finite, it is identified with the probability simplex in $\R^{|\mathcal X|}$. Matrix-valued quantities are identified with vectors through $\mathrm{vec}(\cdot)$, so that $\langle\cdot,\cdot\rangle$ and $\|\cdot\|$ on matrices denote the Frobenius inner product and norm. A sequence of random objects $\{y_t\}$ is called \textit{predictable} for a filtration $\{\Fcal_t\}_{t\ge 1}$ if $y_t$ is $\Fcal_{t-1}$-measurable; it is called \textit{observable} if it is available to the iterative algorithm as it runs. We sometimes shorten stochastic processes $(M_t)_{t \geq 1}$ to just $M$.


\section{Probabilistic tools: sub-$\psi$ processes and boundaries} \label{se:prob}
In this section, we construct different anytime-valid boundaries for the generic performance measure $Z_t$ from \eqref{eq:key-recursion}. The key idea is to leverage \eqref{eq:key-recursion} to proceed inductively; a valid envelope for $Z_{t-1}$ is needed to procure one for $Z_t$. In particular, an envelope $R_{t-1}$ for $Z_{t-1}$ is fed forward to construct $M_t$ and $a_t$. The resulting construction of $R_t$, when unrolled,  illustrates why controlling $Y_t(\bv):=\sum_{s=1}^t \langle v_s, \xi_s \rangle, \ \bv=\{v_t\}_{t\ge1},$ uniformly over time, is sufficient for obtaining anytime-valid guarantees for $Z_t$.  
\begin{lem}[Observable recursive envelopes]\label{lem:observable-envelope}
Let $(Z_t)_{t\ge0}$ be a nonnegative adapted process, let $v_t$ be predictable, let $b_t$ and $\xi_t$ be observable and $\Fcal_{t}$-adapted respectively. Let $R_0$ be observable with $\IP(Z_0\le R_0)\ge1-\alpha_0$ for some $\alpha_0\in[0,1)$, and put $U_0=0$. For $t\ge1$, let $a_t>0$ be predictable and observable, possibly depending on $\{R_s\}_{s=0}^{t-1}$. Furthermore, let $U_t$ be observable, and define recursively
\[
A_0=1,\  A_t=\prod_{s=1}^ta_s,\  R_t:=A_t\Big(R_0+\sum_{s=1}^t\frac{b_s}{A_s}+U_t\Big).
\]
Suppose that, for every $t\ge1$, \eqref{eq:key-recursion} holds almost surely on $\{Z_{t-1}\le R_{t-1}\}$. Set $\widetilde{\bv}=(v_s/A_s)_{s\ge1}$, fix $\alpha\in(0,1)$, and let $\Mcal_\alpha$ satisfy $\IP\big(Y_t(\widetilde{\bv})\le\Mcal_\alpha(t;\widetilde{\bv})\text{ for all }t\ge1\big)\ge1-\alpha$. Suppose, for every $t\ge1$,
\begin{equation}
\bigcap_{s=0}^{t-1}\{Z_s\le R_s\}\cap\{Z_t>R_t\}\ \subseteq\ \big\{\Mcal_\alpha(t;\widetilde{\bv})\le U_t\big\}\ \text{almost surely}.
\label{eq:star}
\end{equation}
Then $(R_t)_{t\ge0}$ is observable, adapted, and satisfies $\IP\big(Z_t\le R_t\text{ for all }t\ge0\big)\ge1-\alpha-\alpha_0.$
\end{lem}
 The proof of Lemma~\ref{lem:observable-envelope} can be found in Appendix~\ref{app:se2proofs}. In particular, if $R_0\ge Z_0$ almost surely and \eqref{eq:key-recursion} holds almost surely for every $t$, the lemma applies with $\alpha_0=0$. Two features of Lemma~\ref{lem:observable-envelope} deserve emphasis. Crucially, we only require \eqref{eq:key-recursion} to hold on $\{Z_{t-1}\le R_{t-1}\}$, and $a_t$ may be built from past envelopes, eg. in Oja's algorithm \S\ref{se:oja} or SGD analysis under local self concordance property, in Appendix \S\ref{app:lsc-module}.

The rest of this section is devoted to time-uniform analysis of $Y_t$, which fulfills a key requirement of Lemma \ref{lem:observable-envelope}. A key bottleneck towards establishing anytime-valid boundaries for $Y_t$ is the presence of predictable but arbitrary $v_t$'s. The literature in this direction \citep{howard2018, howard2021time, whitehouse2026time} primarily considers families of exponential processes indexed by $\xi$ and assumes the existence of a function that plays the role of a cumulant-generating function. This hinders a direct adoption of such assumptions for our analysis, since the predictable sequence $\bv$ cannot be incorporated inside a global assumption. Therefore, in \S\ref{se:asmp}, we pursue a slightly stronger assumption which allows us to deliver valid uniform boundaries across a multitude of tail behaviors. 

\subsection{Assumptions on $\xi_t$}\label{se:asmp}
Our key condition is directly motivated by the sub-$\psi$ condition of \cite{casgrain2024} and \cite{whitehouse2026time}. Specifically, for some $\lambda_{\max}>0$, we call $\psi:[0, \lambda_{\max})\to \R_+$ a \newname\ function if $\psi$ is strictly convex, $\psi(0)= \psi'(0)=0$ and $\psi''(0)>0$. We collect a few examples of $\psi$ which are particularly useful in our applications.
\begin{align}
    \psi_N(\lambda)= \frac{\lambda^2}{2}, \ \psi_{E, c}(\lambda) = \frac{-\log (1-c\lambda) - c\lambda}{c^2}, \ \text{and } \psi_{\mathrm{G},c}(\lambda) = \frac{\lambda^2}{2(1-c\lambda)},
\end{align}
where $c\in \R$. For example, a vast majority of stochastic gradient descent analysis appears to assume the noise process to be sub-Gaussian \citep{feng2023anytime, madden2024, feng2025breaking, chen2025, aolaritei2025stopping, shen2026sgd, aolaritei2026beyond, agrawal2026concentration}, which corresponds to $\psi=\psi_N$. Note that the corresponding $\lambda_{\max}$ for $\psi_{E, c}$ and $\psi_{\mathrm{G}, c}$ are $1/(c \vee 0)$. In the following, we introduce a mildly strengthened notion of sub-$\psi$ processes.

\begin{defn}[Sub-$\psi$ process]\label{def:psi}
    Suppose $\psi:[0, \lambda_{\max})\to \R_+$ is a \newname\ function. Let $\{\xi_t\}_{t=0}^{\infty}$ and $\{\Delta V_t\}_{t=0}^{\infty}$ be $\R^d$ and $\R^{d \times d}$-valued random processes adapted to a filtration $(\Fcal_t)_{t\geq 0}$ such that $\Delta V_t \succeq 0.$ Then $(\xi_t, \Delta V_t)_{t=0}^{\infty}$ is called a \textit{sub-$\psi$ process} if for every $u \in S^{d-1}$ and every $\lambda \in [0, \lambda_{\max})$, it follows that
    \[\exp\bigg(\sum_{s=1}^t  \Big(\lambda \langle u, \xi_s \rangle - \psi(\lambda) \langle u, \Delta V_s u \rangle  \Big) \bigg)\]
    is a supermartingale. 
\end{defn}
Definition~\ref{def:psi} can be understood to be a mild strengthening of the sub-$\psi$ processes of \cite{whitehouse2026time}, similar to how \cite{casgrain2024} strengthened the sub-$\psi$ conditions of \cite{howard2018}. This strengthening is necessitated since we require anytime-valid envelopes not on $\sum_{s=1}^t \xi_s$, but on $\sum_{s=1}^t \langle v_s,\xi_s\rangle$ where $(v_s)$ are predictable.  In particular, we recover Definition~2.3 of \cite{whitehouse2026time} with $S_t = \sum_{s=1}^t \xi_s, \ V_t=\sum_{s=1}^t \Delta V_s$ by taking $v_s= u\in \mathbb{S}^{d-1}$. We note that \cite{whitehouse2026time}, Model~5.1 therein, also uses an assumption similar to Definition~\ref{def:psi} to analyze online linear regression. The key use of Definition \ref{def:psi} for our problem is captured in Lemma \ref{lem:suff}, presented in Appendix \ref{app:boundaries}, where we show that Definition \ref{def:psi} is equivalent to ensuring that for every $\lambda \in [0, \lambda_{\max})$ and for every random process $\mathbf{v}:=\{v_s\}_{s=1}^{\infty}$ adapted to $
(\Fcal_{s-1})_{s\ge 1}$ satisfying $\lambda\|v_s\|< \lambda_{\max}$ for all $s\ge 1$, 
\begin{align}
    M_t(\bv, \lambda):=\exp\bigg(\sum_{s=1}^t  \Big(\lambda \langle v_s, \xi_s \rangle - \Gamma_{\psi}(\Delta V_s, \lambda v_s)  \Big) \bigg) \label{eq:psi}
\end{align}
is itself a super-martingale, where 
$\Gamma_{\psi}: \R^{d\times d}\times \R^d\mapsto \R_{\ge0}$ is given by
\[ \Gamma_{\psi}(A, \theta)= \frac{\psi(\|\theta\|)}{\|\theta\|^2}\langle \theta, A\theta\rangle, \ \|\theta\|< \lambda_{\max}. \]
In Table~\ref{tab:sub-psi-examples}, we list a few practically relevant examples of sub-$\psi$ processes. In particular, our results do not require $\xi_t$ to have any finite moments, even though most of our applications involve martingale difference sequences $\{\xi_t\}$.
\begin{table}[t]
\centering
\small
\renewcommand{\arraystretch}{1.25}
\begin{tabular}{@{} l c l @{}}
\toprule
Conditions on $\xi_t$ & $\psi$ & $\Delta V_t$ \\
\midrule
$\begin{array}{@{}l@{}}
\|\xi_t\| \le c_t \\
\IE_{t-1}[\xi_t] = 0
\end{array}$
& $\psi_N$
& $c_t^2\, I_d$ \\ 
\hline
$\xi_t \overset{d}{=} -\xi_t \given \Fcal_{t-1}$
& $\psi_N$
& $\xi_t \xi_t^\top$ \\
\hline
$\begin{array}{@{}l@{}}
\IE_{t-1}[\|\xi_t\|^2] < \infty \\
\IE_{t-1}[\xi_t] = 0
\end{array}$
& $\psi_N$
& $\frac{1}{3}\,\xi_t\xi_t^\top + \frac{2}{3}\,\IE_{t-1}[\xi_t\xi_t^\top]$ \\
\hline
$\begin{array}{@{}l@{}}
\IE_{t-1}[\|\xi_t\|^3] < \infty \\
\IE_{t-1}[\xi_t] = 0
\end{array}$
& $\psi_{\mathrm{G},1/6}$
& $\xi_t\xi_t^\top + \IE_{t-1}[\,\|\xi_t\|\, \xi_t\xi_t^\top\,]$ \\
\hline
$\begin{array}{@{}l@{}}
\|\xi_t\| \le c \\
\IE_{t-1}[\xi_t] = 0
\end{array}$
& $\psi_{E,c}$
& $\xi_t\xi_t^\top$ \\
\hline
$\begin{array}{@{}l@{}}
\|\xi_t\| \le 1/2 \\
\IE_{t-1}[\xi_t] = 0
\end{array}$
& $\psi_{E,1}$
& $\begin{array}{@{}l@{}}
(\xi_t - \wh\mu_{t-1})(\xi_t - \wh\mu_{t-1})^\top, \\
\wh\mu_{t-1} = \frac{1}{t-1}\sum_{s=1}^{t-1} \xi_s
\end{array}$ \\
\hline
$\begin{array}{@{}l@{}}
\IE_{t-1}\big[\abs{\langle u, \xi_t \rangle}^k\big]
\le \frac{k!}{2}\, c^{k-2}\, \IE_{t-1}\big[\langle u, \xi_t \rangle^2\big] \\
 \  \text{for all } u \in \Scal^{d-1},\ k \ge 3 \\
\IE_{t-1}[\xi_t] = 0
\end{array}$
& $\psi_{\mathrm{G},c}$
& $\IE_{t-1}[\xi_t\xi_t^\top]$ \\
\bottomrule
\end{tabular}
\caption{Examples of pairs $(\psi, \Delta V_t)$ for which $(\xi_t, \Delta V_t)_{t \ge 0}$ is a sub-$\psi$ process (Definition~\ref{def:psi}). Here $c > 0$ is a constant, $c_t > 0$ is predictable, and $\wh\mu_0 := 0$ by convention.}
\label{tab:sub-psi-examples}
\end{table}
With Definition~\ref{def:psi} in place, we are ready to set up and state our uniform boundary results in the following section.

\subsection{Boundaries by mixing, stitching and betting} \label{subsec:induction}
Armed with the Definition \ref{def:psi}, in this section we derive uniformly valid and observable boundaries $(b_{\alpha, t})_{t\ge 1}$ for $Y_t(\bv)$, which one can then combine with Lemma \ref{lem:observable-envelope}. Our approach combines three related but separate procedures of constructing time-uniform envelopes:
\begin{itemize}
    \item \textit{Mixing} \citep{darling1968some, robbins1969probability, robbins1970boundary, lai1976boundary, kaufmann2021mixture, howard2021time, chugg2025closed, whitehouse2026time}. This proceeds via first mixing super-martingales via a mixture distribution over $\lambda$, before inverting it to obtain time-uniform boundaries. This is discussed in detail in \S\ref{app:mixing}.

    \item \textit{Stitching} \citep{howard2021time, chugg-pac-bayes, orabona2023tight, chugg2023time, whitehouse2026time}. Stitching resembles the chaining argument in empirical process theory. In particular, in the stitching construction, one places a geometric grid over $\lambda$, and constructs a linear boundary over each grid via Ville's inequality \cite{ville1939etude, howard2018} before combining them via union bound to provide a curved boundary. As a proponent of the stitching construction, \cite{howard2021time} noted the asymptotic optimality of stitching construction for sums of random variables, obtaining the $\log \log t$ rate instead of the $\log t$ rate that a mixing method often entails. \S\ref{app:stitching} contains a thorough discussion of the stitching construction.

    \item \textit{Betting} \citep{waudby-jrssb, shekhar2023near, voracek2026star, martinez2026Gaussianefficient}. If the random variables are additionally known to be bounded, then one can provide a sharper construction than both stitching and mixing. The key idea is the observation that Definition \ref{def:psi} entails that $\IE[\xi_t]=0$ if $\xi_t$ are bounded (see Lemma \ref{lem:subpsi-martingale-differences} in Appendix \S\ref{app:boundaries}). Therefore, if we bet against the mean not being zero, and end up with a lot of wealth, then we can safely conclude that the considered envelope is invalid. Following \cite{waudby-jrssb}, we formalize the above insight into valid envelopes via product-betting martingales and their variants. This is elaborated in \S\ref{app:betting}. 
\end{itemize}
The triplet of constructions sets us up for application onto \eqref{eq:key-recursion} to provide the final time-uniform boundaries for different iterative algorithms. However, there remain two key bottlenecks: (i) how to reach the recursion \eqref{eq:key-recursion} for any standard iterative algorithm, i.e. the description of the \textit{optimization module}; and (ii) it is not entirely clear if the boundaries are completely observable, since the variance process $\Delta V_s$ contains the oracle quantities $\xi_s$ or $\IE_{s-1} g_s$; moreover, often, the linear coefficients $v_s$ themselves may not be observable.  
Subsequently, we tackle the more straightforward second question. 

\subsubsection{Boundaries for $Y_t(\bv)$} \label{se:boundary-Yt}
This section collects observable, anytime-valid boundaries for $Y_t$ via different constructions. The construction of observable boundaries is motivated by the observation that in most applications, it turns out that one can replace the potentially unobservable $\|v_s\|$ by slightly larger yet observable quantities. This provides us with valid time-uniform envelopes, which we summarize in Theorems~\ref{thm:obs-comp} and \ref{thm:obs-betting}. 

In particular, we streamline exposition by providing observable boundaries to $Y_t(\bv)$ for each of the mixing, stitching or betting constructions. To describe the boundaries compactly, we get some mild notational bobbery out of the way here. For a \newname\ function $\psi$ with an associated $\lambda_{\max}$, denote
\begin{align}
    K_t(\lambda):=\sum_{s=1}^t\nu_s\psi(\lambda M_s), \ \lambda\in \Lambda_t(M):=\big[0,\lambda_{\max}\wedge\frac{\lambda_{\max}}{\max_{s\le t}M_s}\big), \nonumber
\end{align}
where $M=(M_s)_{s\ge 1}$. Note that Eqn \eqref{eq:psi} prescribes choices of $\lambda$ satisfying $\lambda \|v_s\| < \lambda_{\max}$. Often the vectors $\bv$ themselves may not be observable, but we may have access to observable upper bounds $(M_s)$, which motivates the definition of $\Lambda_t$ above. Furthermore, to emphasize the role of boundedness in betting constructions, we write $\Lambda_t(M)=\Lambda_t^G(M)$ when $\lambda_{\max}=1/G$, where $G$ is the deterministic bound on the random variables. The individual boundary constructions are discussed seriatim.
\begin{itemize}
    \item \textit{Mixing}: suppose $F$ denotes a probability distribution on $[0, \lambda_{\max})$.
    Define 
    \begin{align}
        \mathcal U^{\mathrm{mix}}_\alpha(t;M,\nu)&:=\sup\Big\{x\in\R:\int_{\Lambda_t(M)}e^{\lambda x-K_t(\lambda)}\,dF(\lambda)<\frac{1}{\alpha}\Big\}. \label{eq:boundary-mixing}
    \end{align}
    \item \textit{Discrete mixture and stitching}: fix deterministic support points $\lambda_k\in(0,\lambda_{\max}), \ k\ge 0,$
and deterministic weights $p_k>0$ satisfying $\sum_{k=0}^{\infty} p_k \le 1.$ Define
\begin{align}
\mathcal U^{\mathrm{dm}}_\alpha(t;M,\nu)&:=\sup\Big\{x\in\R:\sum_{k:\lambda_k\in\Lambda_t(M)}p_k e^{\lambda_kx-K_t(\lambda_k)}<\frac{1}{\alpha}\Big\},\\
\mathcal U^{\St}_\alpha(t;M,\nu)&:=\inf_{k:\lambda_k\in\Lambda_t(M)}\frac{\log(1/(\alpha p_k))+K_t(\lambda_k)}{\lambda_k}, \label{eq:boundary-dm-stitch}
\end{align}
with $\inf\emptyset:=+\infty$.
\item \textit{Betting}: define 
\begin{align}
    J_t(\lambda)
    :=\sum_{s=1}^t\psi_{E,1}(\lambda M_s\|g_s\|),
    \  \lambda\in\Lambda_t^G(M).\label{eq:jt}
\end{align}
Additionally, consider functions $H_t:[0,G^{-1})\times[0,\infty) \longrightarrow[0,\infty]$ such that the map  $z\mapsto H_t(\lambda,z)$ is non-decreasing, non-negative,
and lower semicontinuous, with $H_t(\lambda,0)=0$. Then, define 
\begin{align}
    b_t^H(y,z)
    &:=
    \sum_{k:\lambda_k\in\Lambda_t^G(M)}
    p_k\exp\bigl(
        \lambda_k y-J_t(\lambda_k)+H_t(\lambda_k,z)
    \bigr), \nonumber\\
    \mathcal U_\alpha^{\mathrm{bet}}(t;M,H,z)
    &:=
    \sup\Bigl\{
        y\in\R:b_t^H(y,z)<\frac1\alpha
    \Bigr\}. \label{eq:boundary-betting}
\end{align}
\end{itemize}
Subsequently, we present the triarchy of boundary constructions in Theorems \ref{thm:obs-comp} and \ref{thm:obs-betting}. Specifically, in Theorem \ref{thm:obs-comp}, we distinguish between mixing with continuous density, which sometimes have closed form solutions, and mixing with a discrete distribution, which can be implemented via numerical root-finding algorithms. The proof of Theorem \ref{thm:obs-comp} follows routine adaptation of Lemma 2 in \cite{howard2021time}; whenever $\|v_s\|<M_s$, the choice of $\lambda$ ensures that $\lambda \|v_s\| < \lambda_{\max}$, and Definition \ref{def:psi} and Eqn. \eqref{eq:psi} yields a supermartingale, to which Ville's inequality applies. Subsequently, via induction, we obtain tight observable envelopes for $Z_t$ in \eqref{eq:key-recursion}.
\begin{thm}[Mixing and stitching]\label{thm:obs-comp}
Let $\bv=(v_t)_{t\ge1}$ be predictable and fix $\alpha\in(0,1)$. Consider $M:=(M_t)_{t\ge1}$, where $M$ is nonnegative, observable and predictable, and let $\mathsf E_t(M):=\bigcap_{s=1}^t\{\|v_s\|\le M_s\}$ for $t\ge1$. Let $(\xi_t,\Delta V_t)_{t\ge0}$ be a sub-$\psi$ process such that $\Delta V_t\preceq\nu_tI$ almost surely for all $t\ge1$, where $\nu_t\ge0$ is observable. Recall $\mathcal U^{\diamond}$ for $\diamond\in\{\mathrm{mix},\mathrm{dm},\St\}$ from \eqref{eq:boundary-mixing}-\eqref{eq:boundary-dm-stitch}. Then, it holds that
\begin{align*}
    \IP\Big(\bigcup_{t\ge1}\big[\mathsf E_t(M)\cap\{Y_t(\bv)>\mathcal{U}^{\diamond}_\alpha(t;M,\nu)\}\big]\Big)\le\alpha.
\end{align*}
\end{thm}
Theorem \ref{thm:obs-comp} is proved in Appendix \S\ref{app:se2proofs}; to aid comprehension, the separate mixing and stitching based results are recorded in Theorem \ref{thm:prob} and Proposition \ref{prop:stitch}, which we then combine to deliver Theorem \ref{thm:obs-comp}. Note that, the set $\mathsf E_t(M)$ forces the boundary in Theorem \ref{thm:obs-comp} to work within the set $\{\|v_s\| \l  M_s\}$. Ideally, we would want $\mathsf{E}_t(M)$ to hold almost surely, so that $\|v_t\|\le M_t$ almost surely for all $t\ge1$, and then $\IP(Y_t(\bv)\le U_t \text{ for all } t\ge1)\ge1-\alpha$. In general, verifying this condition for a suitable choice of $M$ is non-trivial. In the examples of \S\ref{se: anytime-valid-sgd}, \S\ref{se:appl}, \S\ref{app:opt-sgd-boundaries} and \S\ref{app:appl}, we do so by constructing each $M_s$ recursively from the previous envelope $R_{s-1}$. 

Before proceeding, we record a useful corollary providing closed-form, easily computable versions of the Gaussian mixture and stitching bounds when $\psi=\psi_N$, and of the stitching bound when $\psi= \psi_{\mathrm G}$.
\begin{cor}\label{cor:closed-form}
Grant the assumptions in Theorem~\ref{thm:obs-comp}, and let $V_t:=\sum_{s=1}^t\nu_sM_s^2$. Then, under the choices of $\psi$ and tuning parameters in each row below, the corresponding boundary satisfies the stated bound.
\begin{center}
\upshape
\renewcommand{\arraystretch}{1.35}
\begin{tabularx}{\linewidth}{@{}c l l >{\raggedright\arraybackslash}X l@{}}
\toprule
 & Construction & $\psi$ & Tuning & Boundaries $\mathcal U_\alpha(t;M,\nu)\le$ \\
\midrule
(i) & Normal mixture, $\mathcal U^{\mathrm{mix}}_\alpha$
    & $\psi_N$
    & $F(\cdot;\rho)=N(0,\rho^{-1})$
    & $\Ncal_\alpha(V_t)$ \\
\midrule
(ii) & Sub-Gaussian stitching, $\mathcal U^{\St}_\alpha$
    & $\psi_N$
    & $p_k=6/[\pi^2(k+1)^2]$, $\lambda_k=e^{-k/2}$
    & $\min\{4\mathfrak S_\alpha(V_t),\,7\mathfrak S_\alpha(V_t/4)\}$ \\
\midrule
(iii) & Sub-gamma stitching, $\mathcal U^{\St}_\alpha$
    & $\psi_{E,G}$ or $\psi_{\mathrm G,G}$
    & $\{\lambda_k,p_k\}$ as in (ii); $\lambda_{\max}=1/G$
    & $\mathfrak G_\alpha(V_t,c_t)$ \\
\bottomrule
\end{tabularx}
\end{center}
Here $q_\alpha(x):=\max\{x,2\log(2/\alpha)+2\}$, $c_t:=G\max\{1,\max_{s\le t}M_s\}$, and
\begin{align*}
\Ncal_\alpha(V_t)&:=\sqrt{2(V_t+\rho)\log\Big(1+\frac1{2\alpha}\sqrt{\frac{V_t+\rho}{\rho}}\Big)},\\
\mathfrak S_\alpha(V_t)&:=\sqrt{q_\alpha(V_t)\bigl[\log(2/\alpha)+\log\log(e+q_\alpha(V_t))\bigr]}, \ \text{and}\\
\mathfrak G_\alpha(V_t,c_t)&:=4\mathfrak S_\alpha(V_t)+2\sqrt e\,c_t\Bigl[\log\frac{\pi^2}{6\alpha}+2\log\big(2+2\log_+(2c_t)\big)\Bigr].
\end{align*}
\end{cor}
In particular, the choice of $(\lambda_k, p_k)_{k\ge 1}$ in Corollary \ref{cor:closed-form}(ii) mimics the choices of \cite{aolaritei2025stopping} and \cite{aolaritei2026beyond} in their stitching constructions. We discuss them in more detail in \S\ref{se: anytime-valid-sgd} and Appendix \S\ref{app:opt-sgd-boundaries}. 

\paragraph{Betting constructions.}\label{par:betting} Moving beyond mixing and stitching, we proceed towards a betting-based boundary when the noise variables $\xi_t$ form a martingale difference sequence with $\|\xi_t\|\le c$. At the outset, given a sequence $(\wt{\lambda}_s)_{s\ge 1} \subset [0, \lambda_{\max})]$, one can follow \cite{waudby-jrssb} to procure a confidence set for zero via the product-betting martingale 
\begin{align}
    K_t(\bv)
    =\prod_{\substack{1\le s\le t\\ \wt{\lambda}_s \|v_s\| < 1/c}}
          (1+\wt{\lambda}_s\langle v_s,\xi_s\rangle),
    \ K_0(\bv)=1. \label{eq:naive-betting-martingale}
\end{align}
Roughly speaking, at round $s$ a gambler, betting against the (true) hypothesis that $\langle v_s,\xi_s\rangle$ has conditional mean zero, stakes $\wt{\lambda}_s$ fraction of her wealth. In contrast to the supermartingales from \eqref{eq:psi} underlying Theorem~\ref{thm:obs-comp}, one would expect a sharpening of the boundaries constructed with $K_t$. The supermartingale structure used previously can be understood to lead to a wastage of wealth that the exact martingale structure takes care of. However, unlike \cite{waudby-jrssb}, this construction does not directly yield confidence sets for the performance measures $Z_t$ we are concerned with. Recursion \eqref{eq:key-recursion} dictates one must find an upper bound to $Y_t(\bv)=\sum_{s=1}^t \langle v_s, \xi_s\rangle$. To that end, for any fixed $\lambda>0$, write
\begin{align}
    \log K_t(\bv) = \lambda Y_t(\bv) + \underbrace{\sum_{s=1}^t \big[ \log(1 + \wt{\lambda}_s \langle v_s, \xi_s \rangle) - \lambda \langle v_s, \xi_s\rangle\big]}_{:=A} \label{eq:log-bet}
\end{align}
In order to control $A$ in \eqref{eq:log-bet}, one might be tempted to choose $\wt{\lambda}_1=\wt{\lambda}_2=\ldots=\wt{\lambda}_t=\lambda$ to yield $\log K_t(\bv) = \lambda Y_t(\bv) - \sum_{s=1}^t \psi_{E,1}(-\lambda \langle v_s , \xi_s\rangle)$. However, this choice is sub-optimal in two different ways. Firstly, this choice forgoes the usual sharpness of betting-based confidence sets, which stems from a judicious, data-adaptive choice of the bets $\{\wt{\lambda}_s\}$. Moreover,  \eqref{eq:naive-betting-martingale} does not directly exploit the oft-available additional information that $g_t$ itself is bounded (i.e., in the case of bounded gradients for SGD). In particular, we would like to exploit the tighter bound $\|\xi_s\| \le c + \|\IE_{s-1} g_s\|$, which can be much smaller than $2c$ when $\|\IE_{s-1} g_s\|$ is small. For example:
\begin{itemize}
    \item from $\|\xi\| \le 2c$, we obtain $\wt{\lambda}_s \le 1/(2c \|v_s\|)$;
    \item using $\|g_s\| \le c$, we obtain a potentially much larger admissible bet $\wt{\lambda}_s \le 1/(c \|v_s\| + \langle v_s, \IE_{s-1}g_s \rangle)$.
\end{itemize}
Therefore, the gambler can afford to bet more when $\langle v_s, \IE_{s-1}g_s \rangle<0$, and must bet less when $\langle v_s, \IE_{s-1}g_s \rangle>0$. We formalize this by considering $\wt{\lambda}_s:={\lambda}/ ({1+\lambda\langle v_s, \IE_{s-1}g_s\rangle })$ for a fixed $\lambda>0$, so that \eqref{eq:naive-betting-martingale} yields
\begin{align*}
    K_t^1(\bv,\lambda)= \prod_{s=1}^t (1+\lambda_s\langle v_s,\xi_s\rangle) = \prod_{s=1}^t\frac{1+\lambda\langle v_s,g_s\rangle}{1+\lambda\langle v_s,\IE_{s-1}g_s\rangle},
\end{align*}
which is an \textit{exact} nonnegative martingale. Analogous to \eqref{eq:log-bet}, we can proceed as
 \begin{align}
     \log K_t^1(\bv, \lambda) = \lambda Y_t(\bv) - \sum_{s=1}^t \big(\psi_{E,1}(-\lambda \langle v_s, g_s\rangle ) - \psi_{E,1}(-\lambda \langle v_s, \IE_{s-1}g_s\rangle ) \big),  \nonumber
 \end{align}
 which leaves us to bound the two usually unobservable sums: $\sum_{s=1}^t \psi_{E,1}(-\lambda \langle v_s, g_s\rangle)$, and $\sum_{s=1}^t \psi_{E,1}(-\lambda \langle v_s, \IE_{s-1}g_s\rangle)$. Assuming that $\|v_s\|<M_s$ where $(M_s)_{s\ge 1}$ are observable, it follows 
 \[ \sum_{s=1}^t\psi_{E,1}(-\lambda \langle v_s, g_s\rangle ) \le \sum_{s=1}^t\psi_{E,1}(\lambda M_s\|g_s\|),  \]
 which motivates the definition of $J_t(\lambda)$ in \eqref{eq:jt}. The second sum is taken care of by a non-decreasing non-negative process $H_t$; see equation \eqref{eq:obs-bet-mean-correction}. Finally, the construction of $\lambda_s$ involves yet another fixed $\lambda$, and we are still required to guarantee optimality over our choice of this $\lambda$. Therefore, much like Stitching, we choose a grid $\{\lambda_k\}$ and probability mass $\{p_k\}$, and consider the martingale $\sum_k p_k  K_t^1(\bv,\lambda_k)$. This leads to Theorem \ref{thm:obs-betting}.
\begin{thm}[Betting]\label{thm:obs-betting}
Let $\bv$, $M$ and $\mathsf E_t(M)$ be as in Theorem~\ref{thm:obs-comp}. Additionally, suppose for some observable $g_s$ 
satisfying $\|g_s\|\le G$ for a constant $G>0$, one can write $\xi_s=g_s-\IE_{s-1}g_s$. Fix $\alpha\in(0,1)$ and consider $(\lambda_k, p_k)$ as in Theorem \ref{thm:obs-comp}. Recall $H_t$ from \eqref{eq:boundary-betting}, and assume that almost surely on $\mathsf E_t(M)$, as well as
simultaneously for $\lambda\in\Lambda_t^G(M)$ and
$0\le z\le Z_t$, it holds that 
\begin{equation}\label{eq:obs-bet-mean-correction}
    H_t(\lambda,z)
    \le
    \sum_{s=1}^t
    \psi_{E,1}\!\bigl(
        -\lambda\langle
            v_s,\IE_{s-1}g_s
        \rangle
    \bigr).
\end{equation}
Then, for $\mathcal{U}_{\alpha}^{\mathrm{bet}}$ from \eqref{eq:boundary-betting}, it follows that 
\[\IP\Big(\bigcup_{t\ge1}\Big[\mathsf E_t(M)\cap\bigcup_{0\le z\le Z_t}\big\{Y_t(\bv)>\mathcal U_\alpha^{\mathrm{bet}}(t;M,H,z)\big\}\Big]\Big) \le \alpha.\]
\end{thm}

Theorems~\ref{thm:obs-comp}-\ref{thm:obs-betting} are proved in Appendix~\ref{app:se2proofs}.  Our two-step approach can be summarized as follows: once the problem at hand has been reduced to the anvil of \eqref{eq:key-recursion}, Lemma~\ref{lem:observable-envelope} lets the hammers of Theorems \ref{thm:obs-comp}-\ref{thm:obs-betting} forge observable anytime-valid envelopes. In practice, for most applications it is straightforward to obtain observable $J_t$ and $H_t$; we will look at various examples of these computations for SGD in \S\ref{se: anytime-valid-sgd}, and for a few other interesting applications, in \S\ref{se:appl}. 
\allowdisplaybreaks
\section{Anytime-valid boundaries in stochastic gradient descent}\label{se: anytime-valid-sgd}
As a canonical example of an iterative algorithm, we consider the problem of stochastic gradient descent. Consider minimizing a differentiable function $F:\R^d\to \R$, $F\in \mathcal{C}^1$ over a set $\Xcal\subseteq \R^d$, given by: 
\[ x^{\star}\in \argmin_{x \in \Xcal} F(x) \subseteq \operatorname{int}(\Xcal),\]
and the corresponding SGD algorithm:
\begin{align}\label{eq: general SGD}
    x_t = \Pi_{\Xcal}(x_{t-1}-\eta_t g_{t}),  \  x_0 \in \R^d.
\end{align}
 For convenience, throughout this paper, we let $\Xcal=\R^d$. If $\Xcal$ is a closed convex set, almost the entirety of our analysis carries over via the contractive property of a projection. Typically, an analysis of \eqref{eq: general SGD} consists of some assumptions on the smoothness and growth of the objective function $F$, which then drives the performance metric to be considered for the error analysis of the recursion. A detailed discussion on a number of such assumptions and their relationship to the performance measures, is provided in Appendix~\ref{app:opt}. This analysis leads to a recursion of the form \eqref{eq:key-recursion}, on top of which we can employ the anytime-valid boundaries derived in the previous section via Lemma~\ref{lem:observable-envelope}. In the following, we describe this entire procedure through three separate analyses of SGD under three different sets of assumptions. A discussion on the derivation of \eqref{eq:key-recursion} under a diverse set of assumptions, along with implications between these assumptions themselves, is included in Appendix \S\ref{app:opt}. Additional derivations of recursion \eqref{eq:key-recursion} and corresponding anytime-valid envelopes under many different assumptions are discussed in \S\ref{app:opt-sgd-boundaries}. These examples also serve as practical illustrations of our general approach; we hope they will guide readers in carrying it over to the many other applications that space does not permit us to cover, where it should likewise yield diverse and new anytime-valid constructions.

\subsection{Case I: Restricted secant inequality and bounded stochastic gradient}\label{se:rsi}
Assume that $F$ satisfies the restricted secant inequality (\RSI; \cite{vaswani2019painless}): there exists $\mu>0$, such that 
 \[ \langle\nabla F(x) - \nabla F(x^\star), x-x^\star\rangle \geq \mu \|x-x^\star\|^2, \ x \in \Xcal. \]
 Note that \RSI\ is strictly weaker than strong convexity (\SC). In fact, any $\mu$-strongly convex function satisfies \RSI\ with the same parameter $\mu$, but the converse fails.\footnote{The function $f(x)=x^2/2+8\log(1+x^2)$ satisfies \RSI\ with respect to its unique minimizer $x^\star=0$, but is not even convex.} Throughout this section, we additionally assume that the stochastic gradients are bounded, $\|g_t\|\le G$ almost surely, and $\IE_{t-1}[\xi_t]=0$. While we work with a weaker condition in \RSI, this set-up resembles that of \cite{aolaritei2026beyond}, who derived time-uniform boundaries under strong convexity and bounded gradients, and we treat it as our baseline. Consider the performance metric $Z_{t,1} := \|x_t - x^\star\|^2$. 
 \subsubsection{Optimization module for \RSI}\label{se:opt-rsi}
 In this section, we briefly discuss the derivation of \eqref{eq:key-recursion}; this is also present in Lemma C.2 of \cite{aolaritei2026beyond}.
 Define ${A}_t=\prod_{\ell=1}^t(1-2\mu \eta_{\ell})$, $s<t$, and assume $2\mu\eta_t<1$. Since $\nabla F(x^\star)=0$ and $\xi_t=\nabla F(x_{t-1})-g_t$, expanding $\|x_{t-1}-\eta_tg_t-x^\star\|^2$ and applying \RSI\ gives
\begin{align}
    Z_{t,1}\le(1-2\mu\eta_t)Z_{t-1,1}+\eta_t^2\|g_t\|^2+\langle2\eta_t(x_{t-1}-x^\star),\xi_t\rangle, \label{eq:stronglyconvex-recursion}
\end{align}
which is \eqref{eq:key-recursion} with $a_t=1-2\mu\eta_t$, $b_t=\eta_t^2\|g_t\|^2$ and $v_t=2\eta_t(x_{t-1}-x^\star)$; its unrolled form is Lemma~C.2 of \cite{aolaritei2026beyond}. Moreover, \RSI, $\nabla F(x^\star)=0$ and Jensen's inequality give $$\mu\|x_t-x^\star\|^2\le\langle\nabla F(x_t),x_t-x^\star\rangle\le\|\IE_{t}g_{t+1}\|\,\|x_t-x^\star\|\le G\|x_t-x^\star\|,$$ whence $Z_{t,1}\le R_\star:=G^2/\mu^2$ almost surely. Throughout this section, we take $R_0:=R_\star$ and write $D_t:=R_0+\sum_{s=1}^t\eta_s^2\|g_s\|^2/ A_s$. On $\{Z_{t-1,1}\le R_{t-1}\}$, one can control $v_t$ as $\|v_t\| \le 2\eta_t\sqrt{R_{t-1}}$.
Thereafter, Theorems~\ref{thm:obs-comp} and \ref{thm:obs-betting} provide an observable anytime-valid envelope in conjunction with Lemma~\ref{lem:observable-envelope}.

\subsubsection{Betting construction under \RSI }\label{app:betting-sgd}
As a typical demonstration of our modularized approach of combining the optimization module and Theorems \ref{thm:obs-comp}-\ref{thm:obs-betting}, in this section we discuss a betting construction under \RSI\ and bounded gradients. We separately discuss a sub-gamma stitching boundary in \S\ref{se:sub-gamma-stitch}; while the sub-gamma stitching boundary has the benefit of a closed form expression, it uses the inequality $\psi_{E,G} \le \psi_{\mathrm{G}, G}$. Clearly, this can be understood to introduce some slackness in the bound, which can be uniformly improved in a straightforward manner. For example, at the cost of an analytic computation, one can directly perform a sub-exponential stitching boundary with the same choices of $p_{k,\ell}$ and $\lambda_{k, \ell}$ to obtain a uniformly sharper boundary. In our simulation studies~\ref{fig:ridge-prelim}, the exponential discrete mixture boundary (i.e. $\psi=\psi_{E,G}$ along with Corollary~\ref{corol:dm})  with the choices as in Figure 4 of \cite{howard2021time} outperforms the stitching bound; see the discussion immediately preceding Proposition 2 in \cite{howard2021time}. If we emphasize upon obtaining a closed form bound, then there are other alternatives as well. For example, in Appendix \S\ref{app:GE-mixture}, we show that the gamma-exponential mixture boundary can also be accompanied by a sharp closed form upper bound, facilitating computation. 

However, a common feature of the aforementioned boundaries is that they make use of the bounded gradients only through the use of $\psi_{E,G}$ function in the Definition \ref{def:psi} formulation. As discussed before Theorem \ref{thm:obs-betting}, and later in more detail in \S\ref{app:betting} (also see \cite{waudby-jrssb}), with the bounded gradients we can sharpen boundaries via the product-betting martingales compared to these supermartingale-based boundaries. In the following, we leverage Theorem~\ref{thm:obs-betting} and Lemma~\ref{lem:observable-envelope} to derive an observable betting boundary for stochastic gradient descent under \RSI\ and bounded gradient assumptions. First, some notational niceties are warranted. Recall $J_t$ from \eqref{eq:jt}, and define
\begin{align}
    H_t^{(w)}(\lambda,z)&:=\sum_{s=1\vee(t-w)}^t\psi_{E,1}\Big(\frac{2\lambda\mu\eta_s}{A_s}\big(\sqrt z-\|x_t-x_{s-1}\|\big)_+^2\Big) \label{eq:Ht-W},
\end{align}
which plays the role of $H_t$ from Theorem \ref{thm:obs-betting}.
Proposition \ref{prop:betting-SGD} mimics the construction of $\mathcal{U}_\alpha$ in \eqref{eq:boundary-betting}.
\begin{prop}\label{prop:betting-SGD}
Fix $\alpha\in(0,1)$, $w\in\N_0$, and deterministic $\lambda_k\in(0,1/G)$ and $p_k>0$, $k\ge0$, such that $\lambda_k\downarrow0$ and $\sum_{k\ge0}p_k=1$. With $R_0^w:=R_\star$, define recursively, for $t\ge1$, $M_t:=2\eta_t\sqrt{R_{t-1}^w}/{A}_t$. Moreover, for $\lambda\in\Lambda_t^G(M)$ and $z\ge0$, recall \eqref{eq:Ht-W} and define $W_t^{(w)}(z):= b_t^H({z}/{{A}_t}-D_t, z)$, where $b_t^H$ is as in \eqref{eq:boundary-betting} with $H_t^{(w)}$ playing the role of $H_t$ therein. Then,
\begin{equation}\label{eq:betting-observable-costs}
\begin{aligned}
    R_t^w&:=\sup\Big(\{0\}\cup\big\{z\in[0,R_\star]:W_t^{(w)}(z)\le1/\alpha\big\}\Big)
\end{aligned}
\end{equation}
satisfies $\IP(Z_{t,1}\le R_t^w\text{ for all }t\ge0)\ge1-\alpha$.
\end{prop}
Proposition \ref{prop:betting-SGD} is proved in Appendix \ref{app:se3.1proofs}. For $\lambda>0$, consider the following elementary inequality:
\begin{equation}\label{eq:betting-quadratic-relaxation}
    \psi_{E,1}(x)\le\frac{x^2}{b^2}\psi_{E,1}(b),
    \  |x|\le b<1,
\end{equation}
Note that, \eqref{eq:betting-quadratic-relaxation} along with $\psi_{E,c}\le \psi_{\mathrm{G},c}$ entails that
\begin{equation}
   J_t(\lambda) \le 
    \sum_{s=1}^t\psi_{E,G}(\lambda M_s)\|g_s\|^2 \le 
     \psi_{\mathrm{G}, \ G\Tilde{M}_t}(\lambda)\sum_{s=1}^t M_s^2 \|g_s\|^2, \ \Tilde{M}_t = \max_{1\le s\le t} M_s. \label{eq:ineq}
\end{equation} The inequalities \eqref{eq:ineq} enable a short comparison between the different envelopes. Throughout the comparison, we use the same mixtures $\{p_k\}$ and parameters $\{\lambda_k\}$. Fix $w>0$. 
\begin{itemize}
    \item The exponential discrete mixture boundary, denoted by $R_t^{\mathrm{dm}}$, is obtained by replacing $J_t$ and $H_t^{(w)}(\lambda_k, z)$ in $W_t^{(w)}$ by $\sum_{s=1}^t\psi_{E,G}(\lambda M_s)\|g_s\|^2$, and $0$, respectively. The resultant $W_t(z)$ satisfies $W_t(z) \le W_t^{(w)}(z)$, and therefore, $R_t^w \le  R_t^\mathrm{dm}$. 
    \item Furthermore, the exponential stitching boundary, denoted by $R_t^{\St}$ is derived by replacing the sum over $\{k\ge 0: \lambda_k \in \Lambda_t^G\}$ in $W_t(z)$ by taking a supremum over the same set. Let $\tilde{W}_t$ denote the resulting threshold, which satisfies $\tilde{W}_t(z) \le W_t(z)$. Consequently, $R_t^{\mathrm{dm}} \le R_t^\St$. 
            \item Finally, the sub-gamma stitching boundary, denoted by $\tilde{R}_t^{\St}$, can be conjured by replacing $\psi_{E,G}(\lambda_k M_s)\|g_s\|^2$ in $\Tilde{W}_t(z)$ by $\psi_{\mathrm{G}, \ G\Tilde{M}_t}(\lambda_k)\sum_{s=1}^t M_s^2 \|g_s\|^2$. In particular, the empirical-Bernstein boundary in \citet[Theorem~3.5]{aolaritei2026beyond} (henceforth denoted by $R_t^{\mathrm{ALBJ}}$) provides a closed-form upper bound to the sub-Gamma stitching boundary with the choice
            \begin{align}
                p_{k,\ell}=\frac{1}{(k+1)(k+2)(\ell+1)(\ell+2)}, \ \lambda_{k, \ell} = \frac{1}{2^{\ell} + \sqrt{2^{k-1}/ \log(1/(\alpha p_{k,\ell}))}}. \label{eq:AJ}
            \end{align} Consequently, under \eqref{eq:AJ}, the sub-gamma stitching boundary satisfies $R_t^\St \le \tilde R_t^{\St}\le R_t^{\mathrm{ALBJ}}$ for all $t$. This is further detailed in Appendix \S\ref{se:sub-gamma-stitch}.
\end{itemize}
In particular, with choices of $p_k$ and $\lambda_k$ in \eqref{eq:AJ}, it holds that for any $w>0$, $R_t^w \le R_t^{\mathrm{ALBJ}}$. To summarize, we provide a class of boundaries each uniformly sharper than \cite{aolaritei2026beyond}.

\subsection{Case II: Lipschitz gradient}
\label{se:Lipschitz-sgd} \label{se:LS}
\RSI\  in \S\ref{se:rsi} forces every stationary
point of $F$ to be the global minimizer, and thereby excludes a large
class of nonconvex objectives routinely encountered in modern machine
learning. In this section, we work with potentially non-convex $F$, and repeat the procedure of \S\ref{se:rsi} by only assuming that $F$ has an $L$-Lipschitz gradient (\LS)  i.e.,
\[
 \|\nabla F(x)-\nabla F(y)\|\le L\|x-y\|,\  x,y\in\mathbb R^d .
\]
Equivalently, we assume that $F$ satisfies the descent lemma, \cite[Proposition~A.24]{bertsekas1999nonlinear}:
\begin{align}\label{eq:descent}
    F(y) \le F(x) + \langle \nabla F(x) , y-x\rangle + \frac{L}{2}\|x-y\|^2.
\end{align}
In this case, there might not be any unique global minima; therefore instead of ${Z}_{t,1}$, an appropriate performance metric would approximate stationarity
\citep{ghadimi2016accelerated,bottou2018optimization}. Accordingly, the performance metric we consider is:
\[\tilde{Z}_{t,2}:= F(x_t) -  F^\star +  \sum_{s=1}^t \eta_s\| \nabla F(x_{s-1})\|^2, \]
where $F^\star=\min F(x)$. The reason for using the notation $\tilde{Z}$ will be apparent in the Appendix \S\ref{app:performance}; $\tilde{Z}_{t,2}$ turns out to be the sum of two measures often considered separately in the literature. In particular, \citet{aolaritei2025stopping} provided anytime-valid control of one-component of $\tilde{Z}_{t,2}$: the weighted
stationarity measure ${Z}_{t,5}:= S_t^{-1}\sum_{s=1}^t \eta_s\| \nabla F(x_{s-1})\|^2$, where $S_t=\sum_{s=1}^t\eta_s$. Subsequently, we derive the optimization module for $\tilde{Z}_{t,2}$, with which we subsequently discuss a self-normalized as well as a betting-based time-uniform envelope.
\subsubsection{Optimization module for \LS}\label{se:ls-opt}
Deriving a recursion of the type \eqref{eq:key-recursion} for $\tilde{Z}_{t,2}$ is trivial via \eqref{eq:descent}. Let $R_0\ge \tilde{Z}_{0,2}$ almost surely. 
Put $b_t:=L\eta_t^2\|g_t\|^2/2$ and $D_t:=R_0+\sum_{s=1}^t b_s$.
Applying \eqref{eq:descent} to the step $x_t=x_{t-1}-\eta_tg_t$, we obtain
\begin{equation*}
 \tilde{Z}_{t,2}\le \tilde{Z}_{t-1,2}+L \eta_t^2 \|g_t\|^2/2+\langle v_t,\xi_t\rangle,
\end{equation*}
where $v_t= \eta_t \IE_{t-1}g_t$. Thereafter,
Lemma~\ref{lem:observable-envelope} along with the boundaries of \S\ref{se:prob} provides a large class of anytime-valid boundaries for $\tilde{Z}_{t,2}$, and by extension, ${Z}_{t,5}$. In particular, in the Appendix \S\ref{se:Lipschitz-stitching}, we discuss how particular stitching constructions recover the envelope from Corollary 2.6 of \cite{aolaritei2025stopping}; therein, we also provide an alternative construction which can be sharper when the SGD chain is nearing the local optima. In the following, we explore some new constructions based on self-normalization (Table \ref{tab:sub-psi-examples}~row 3), and betting, and discuss the relationship between themselves, and with the stitching constructions in Appendix \S\ref{se:Lipschitz-stitching}. 
\subsubsection{Self-normalized and betting envelopes under bounded gradients}
\label{se:ls-betting}

In this section, we work with the additional assumption that $\|\nabla F(x)\|\le G$ for all $x\in\R^d$, where $G>0$ is known, similar to \citet[Corollary~2.6]{aolaritei2025stopping}. This assumption enables us to construct several mixing and stitching-based envelopes via Table \ref{tab:sub-psi-examples}, third row. Following \cite{whitehouse2026time}, we call these the \textit{self-normalized} envelopes. Subsequently, we further sharpen these envelopes via a quadratic product-betting supermartingale. 

In addition to $\|\nabla F(x)\|\le G$, assume that $\xi_t$ are conditionally sub-Gaussian with variance proxy $\sigma^2$; i.e., in Definition~\ref{def:psi}, $\psi=\psi_N$, $\lambda_{\max}=\infty$, and $\Delta V_s = \sigma^2 I$. Therefore, with $\xi_t=\nabla F(x_{t-1})-g_t$, and $\psi=\psi_N$, Lemma~\ref{lem:subpsi-martingale-differences} in the appendix entails
\begin{equation}\label{eq:ls-moment-assumption}
    \IE_{t-1}\xi_t=0,\  \IE_{t-1}[\xi_t\xi_t^\top]\preceq\sigma^2I,\  t\ge1.
\end{equation}
Consider Table~\ref{tab:sub-psi-examples}, row 3. Equation \eqref{eq:ls-moment-assumption} implies that $(\xi_t,\Delta V_t^{\mathrm{SN}})_{t\ge1}$ with $\Delta V_t^{\mathrm{SN}}:=\frac13\xi_t\xi_t^\top+\frac23\IE_{t-1}[\xi_t\xi_t^\top]$ is a sub-$\psi_N$ process. Since $\|\xi_t\|\le G+\|g_t\|$, \eqref{eq:ls-moment-assumption} gives $\Delta V_t^{\mathrm{SN}}\preceq\nu_tI$ with the observable $\nu_t:=\frac13(G+\|g_t\|)^2+\frac23\sigma^2$. Consequently, the choices in Theorem~\ref{thm:obs-comp} read, 
\begin{equation*}
    K_t(\lambda)=\lambda^2Q_t^{\mathrm{SN}}/2, \ \text{where } Q_t^{\mathrm{SN}}:=\sum_{s=1}^t\nu_sM_s^2=\frac{G^2}{3}\sum_{s=1}^t\eta_s^2\big[(G+\|g_s\|)^2+2\sigma^2\big],
\end{equation*}
where $M_s=\eta_s G$.
  Let the discrete-mixture and stitching boundaries of Theorem~\ref{thm:obs-comp} be denoted as $U_t^{\mathrm{SN,DM}}$ and $U_t^{\mathrm{SN,St}}$ respectively.  Denote by $R_t^{\mathrm{SN,DM}}$ and $R_t^{\mathrm{SN,St}}$ the corresponding envelopes $R_0+L\sum_{s=1}^t\eta_s^2\|g_s\|^2/2+U_t$. By Theorem~\ref{thm:obs-comp},
\begin{equation}
    \IP\big(\tilde{Z}_{t,2}\le R_t\text{ for all }t\ge0\big)\ge1-\alpha,\ R_t\in\{R_t^{\mathrm{SN,DM}},R_t^{\mathrm{SN,St}}\}.
    \label{eq:ls-sn-coverage}
\end{equation}

\paragraph{Betting envelope.} Similar to the proof of Lemma~3(f) of \citet{howard2018}, the self-normalized construction can be further sharpened by retaining the positive quadratic factor underlying the finite-variance inequality. In particular, we proceed along the lines of Corollary~\ref{corol:betting}, however, through a quadratic product-betting supermartingale in lieu of the linear product-betting martingale. Write
\begin{align}
     \chi(u):=u-\log(1+u+u^2/3),\ u\in\R,\ j(b):=\max\big\{\chi(b),\chi(-\min\{b,1\})\big\}\le\frac{b^2}{6},\ b\ge0. \label{eq:chi}
\end{align}
Analogous to \eqref{eq:jt}, define
\begin{align}
\begin{aligned}
    J_t^{\mathrm Q}(\lambda) &:=\sum_{s=1}^t\Big[j\big(\lambda\eta_sG(G+\|g_s\|)\big)+\log\Big(1+\frac{\lambda^2\sigma^2G^2\eta_s^2}{3}\Big)\Big],\\
    W_t^{\mathrm Q}(y)&:=\sum_{k\ge0}p_k\exp\bigl(\lambda_ky-J_t^{\mathrm Q}(\lambda_k)\bigr).
\end{aligned} \label{eq: jt-quad}
\end{align}
The following result establishes a betting-based anytime-valid envelope for $\tilde{Z}_{t,2}$ under the bounded-gradient assumption.
\begin{prop}[Observable quadratic betting envelope]
\label{prop:ls-quadratic-betting}
Grant \eqref{eq:ls-moment-assumption} and $\|\nabla F\|\le G$, and let $\{\lambda_k,p_k\}_{k\ge0}$ be as above. For $t\ge1$ and $\lambda>0$, define
\begin{equation}
\begin{aligned}
    U_t^{\mathrm Q}
    &:=\sup\{y\in\R:W_t^{\mathrm Q}(y)<1/\alpha\},\ 
    R_t^{\mathrm Q}:=R_0+L\sum_{s=1}^t\eta_s^2\|g_s\|^2/2+U_t^{\mathrm Q},\  R_0^{\mathrm Q}:=R_0,
\end{aligned}
\end{equation}
where we recall $W_t^{\mathrm Q}$ from \eqref{eq: jt-quad}. 
Then $(R_t^{\mathrm Q})_{t\ge0}$ is observable and adapted, and
\begin{equation*}
    \IP\big(\tilde{Z}_{t,2}\le R_t^{\mathrm Q}\text{ for all }t\ge0\big)\ge1-\alpha.
\end{equation*}
\end{prop}
Proposition~\ref{prop:ls-quadratic-betting} is proved in Appendix~\ref{app:se3proofs}. Its proof uses the nonnegative supermartingale
\begin{equation*}
    K_t^{\mathrm Q}(\bv,\lambda):=\prod_{s=1}^t\frac{1+\lambda\langle v_s,\xi_s\rangle+\lambda^2\langle v_s,\xi_s\rangle^2/3}{1+\lambda^2\sigma^2\|v_s\|^2/3},\  K_0^{\mathrm Q}(\bv,\lambda)=1.
\end{equation*}
We emphasize that unlike in \S\ref{app:betting}, the numerator in $K_t^{\mathrm Q}$ is positive for every $\langle v_s, \xi_s\rangle $ and every $\lambda>0$, so that no restriction on $\lambda$ or boundedness of $\|g_t\|$ is required. Moreover, for comparison, $j(x) \le x^2/6$ and $\log(1+x)\le x$, $x\ge0$, give $J_t^{\mathrm Q}(\lambda)\le\lambda^2Q_t^{\mathrm{SN}}/2$. Consequently, for the same $\{\lambda_k,p_k\}$, $\alpha$ and $R_0$, it holds that $R_t^{\mathrm Q}\le R_t^{\mathrm{SN,DM}}\le R_t^{\mathrm{SN,St}},$ for $t\ge 1$.

\subsubsection{Combining the self-normalized and the betting envelope}\label{se:betting-combn}
In this section, we provide an envelope that combines the benefits of both the self-normalized and betting constructions. Interestingly, this leads to a uniformly sharper boundary than the sub-Gaussian stitching boundary of \cite{aolaritei2025stopping}, which we also discuss in the Appendix \S\ref{se:Lipschitz-stitching}.

It is evident that the self-normalized envelope uses sub-Gaussian assumption only through Lemma~\ref{lem:subpsi-martingale-differences}, thereby foregoing its full efficacy; on the other hand, the betting envelope uses only the bounded gradient assumption. Instead of naively taking a maximum among the two envelopes to preserve validity, or using a union bound, we can combine both constructions into a single mixture before inversion. Define
\begin{align*}
    W_t^{\mathrm H}(y)
    &:=\frac12\sum_{k\ge0}p_k\Big[\exp\Big(\lambda_ky-\frac{\lambda_k^2}{2}Q_t^G\Big)+\exp\bigl(\lambda_ky-J_t^{\mathrm Q}(\lambda_k)\bigr)\Big],\\
    U_t^{\mathrm H}
    &:=\sup\{y\in\R:W_t^{\mathrm H}(y)<1/\alpha\},\ 
    R_t^{\mathrm H}:=R_0+L\sum_{s=1}^t\eta_s^2\|g_s\|^2/2+U_t^{\mathrm H},\  R_0^{\mathrm H}:=R_0,
\end{align*}
where $Q_t^G:= \sigma^2G^2\sum_{s=1}^t\eta_s^2$.
The following result derives a hybrid envelope that combines the benefits of both the sub-Gaussian discrete mixture, and the betting envelope.
\begin{prop}[Hybrid envelope]\label{prop:ls-hybrid}
Under the assumptions of \S\ref{se:Lipschitz-sgd} and $\|\nabla F(x)\|\le G$, it follows that
\begin{align*}
    \IP\big(\tilde{Z}_{t,2}\le R_t^{\mathrm H}\text{ for all }t\ge0\big)\ge1-\alpha.
\end{align*}
\end{prop}
For a comparison with existing literature, we let $\alpha\in(0,2/e)$, $\lambda_k=e^{-k/2}$, $p_k=6/[\pi^2(k+1)^2]$ and denote $q(x)=\max\{x,2\log(2/\alpha)+2\}$. These choices enable an application of Corollary \ref{cor:closed-form}(ii); the corresponding boundary is exactly the one in Corollary 2.6 of \cite{aolaritei2025stopping}. This boundary is henceforth denoted by $\bar{R}_t^{\mathrm{AJ}}$; for details, we refer the reader to \S\ref{se:Lipschitz-stitching}. As in \S\ref{app:betting}, one can show that if $\bar R_0^{\mathrm{AJ}}=R_0$, then, 
\begin{equation}
    R_t^{\mathrm H}< \bar R_t^\mathrm{AJ},\  t\ge1.
\label{eq:ls-hybrid-comparison}
\end{equation}
Proposition~\ref{prop:ls-hybrid}, along with \eqref{eq:ls-hybrid-comparison} is proved in Appendix~\ref{app:se3.2proofs}. To conclude, \S\ref{se:rsi} and \S\ref{se:Lipschitz-sgd} jointly provide a glimpse of the generality of our approach, as we are able to recover, improve and provide alternatives to existing results in the literature. In Appendix~\S\ref{app:opt-sgd-boundaries}, we derive many more anytime-valid boundaries under various combinations of assumptions and performance measures, most of which come with no prior precedents in the corresponding set-ups.

\subsection{Case III: Polyak--Ruppert averaging (strong convexity, Lipschitz gradient)}\label{se:pr}
A weak smoothness assumption such as \LS\ only allows us to control $\tilde{Z}_{t,2}$ or $Z_{t,5}$. In order to ascertain convergence, which should lead to decaying envelopes for either $Z_{t,1}$ or $Z_{t,2}$, one often needs to pair a growth assumption such as \SC\ or \RSI, and a smoothness assumption together. We defer a detailed discussion on different growth and smoothness assumptions to \S\ref{app:opt}; in this section, we consider \SC\ along with \LS. Moving beyond the final iterate analysis of \S\ref{se:rsi}, we establish anytime-valid boundaries for the Polyak-Ruppert averaged version. Our motivation for considering the PR-averaged iterates is two-fold. Firstly, averaging the iterates, $\bar x_t:=t^{-1}\sum_{s=1}^tx_s$, has been the classical remedy for the sensitivity of the last iterate to the step-size schedule \citep{ruppert1988efficient,polyak1992acceleration}; with $\eta_t\asymp t^{-\beta}$, $\beta\in(1/2,1)$, $\|\bar x_t-x^\star\|$ attains the parametric rate $t^{-1/2}$, whereas $\|x_t-x^\star\|$ is of order $t^{-\beta/2}$ \citep[Theorems~1 and~3]{moulines2011non}. Secondly, this simple setting allows us to investigate the optimality properties of the different envelopes hitherto proposed. In fact, the combination of \SC\ and \LS\ has been among the most common assumptions in the analysis of SGD, from the classical works of \citet{polyak1992acceleration, nemirovski2009robust, moulines2011non} to more recent analyses in convergence and inference \citep{gorbunov2020unified, garrigos2023handbook, liu2024revisiting, chen2020statistical, sheshukova2026Gaussian, wei2026general}. Existing time-uniform guarantees for $\bar x_t$ are, however, asymptotic \citep{xie2024asymptotic}. 

Subsequently, we follow the same script as above, of first obtaining an optimization module before calling on Theorems \ref{thm:obs-comp} or \ref{thm:obs-betting} to provide the anytime-valid boundaries. 

\subsubsection{Optimization module for PR-averaged SGD under \SC\ and \LS}\label{se:opt-pr}
This section is devoted to a derivation of \eqref{eq:key-recursion} for SGD under \SC\ and \LS. Concretely, consider $F\in\mathcal C^1(\R^d)$ satisfying \SC\ and \LS\ with known constants $0<\mu\le L<\infty$, i.e.,
\begin{equation}\label{eq:pr-sc-ls}
\langle\nabla F(x)-\nabla F(y),x-y\rangle\ge\mu\|x-y\|^2,\ \|\nabla F(x)-\nabla F(y)\|\le L\|x-y\|,\  x,y\in\R^d,
\end{equation}
and $x^\star$ denotes its unique minimizer. We also assume that the gradients $\|g_t\| \le G$ almost surely. A naive anytime-valid envelope for $\bar x_t$ already follows from \S\ref{se:rsi}: on the event of Proposition~\ref{prop:betting-SGD} (under its step-size condition $2\mu\eta_t<1$), $\|\bar x_t-x^\star\|\le t^{-1}\sum_{s\le t}\sqrt{R^w_s}$ for all $t\ge1$. This bound averages error bounds rather than errors, and thereby forfeits the cancellation of the noise across iterations that motivates PR-averaging in the first place. In our analysis, we retain this cancellation by isolating the cumulative noise, which satisfies \eqref{eq:key-recursion} with $a_t=1$, and by transferring its anytime-valid envelope to $\bar x_t$ through an observable linearization of the averaged gradient. The performance measure considered is $\bar Z_t=\|\bar x_t-x^\star\|^2$. Maintaining the notations as in \S\ref{se:rsi}, it follows $\xi_t=\nabla F(x_{t-1})-g_t$, so that $\IE_{t-1}\xi_t=0$. An anytime-valid envelope for $\bar Z_t$ proceeds through a guarantee for $\|\Xi_t\|^2$, where $\Xi_t:= \sum_{s=1}^t \xi_s$, $\Xi_0=0$. To that end, we follow the recipe of \S\ref{se:prob}. Since $\xi_t=\nabla F(x_{t-1})-g_t$, direct expansion gives
\begin{equation}\label{eq:pr-rec}
\|\Xi_t\|^2=\|\Xi_{t-1}\|^2+\|g_t\|^2-\|\nabla F(x_{t-1})\|^2+\langle v_t,\xi_t\rangle\le\|\Xi_{t-1}\|^2+\|g_t\|^2+\langle v_t,\xi_t\rangle,
\end{equation}
and $v_t:=2\bigl(\Xi_{t-1}+\nabla F(x_{t-1})\bigr)$. This recovers \eqref{eq:key-recursion} for $Z_t:=\|\Xi_t\|^2$. 

\subsubsection{Boundaries for PR-averaged iterates}\label{se:PR-boundaries}
Having discussed the optimization module for PR-averaged iterates $\bar x_t$, in this section we collect a catalog of different boundaries furnished by Theorems \ref{thm:obs-comp} and \ref{thm:obs-betting}. 

To that end, observe that $\|\nabla F(x_{t-1})\|=\|\IE_{t-1}g_t\|\le G$ by Jensen's inequality. Therefore, for a sequence $\{B_t\}$ such that $\|\Xi_s\| \le B_s$, it follows that $\|v_t\|\le M_t:=2(B_{t-1}+G)$ on $\{\|\Xi_{t-1}\|\le B_{t-1}\}$. To streamline notations, $U_t=U_t^b$ will refer to one of the following cases: $b=\mathrm{NM}$ for the Gaussian mixture boundary in Corollary \ref{cor:closed-form}(i); $b=\mathrm{GS}$ refers to the sub-Gaussian stitching boundary, i.e. \eqref{eq:boundary-dm-stitch} with $\psi=\psi_N$ (or its upper bound in Corollary \ref{cor:closed-form}(ii)); similarly, $b=\mathrm{\Gamma S}$ denotes the sub-Gamma stitching boundary, and $b=\mathrm{Bet}$ denotes the betting boundary \eqref{eq:boundary-betting} with $H_t\equiv0$. In particular, for $b\ne\mathrm{NM}$, we use the choices of $(\lambda_k, p_k)$ in Corollary \ref{cor:closed-form}. 

\begin{thm}[Anytime-valid envelopes for the averaged iterate]\label{thm:pr-coverage}
Grant \eqref{eq:pr-sc-ls}, and fix $\alpha\in(0,1)$ and $b\in\{\mathrm{NM},\mathrm{GS},\Gamma\mathrm S,\mathrm{Bet}\}$. Additionally, assume that \begin{equation}\label{eq:pr-hl}
\|\nabla^2F(x)-\nabla^2F(y)\|_{\op}\le L_H\|x-y\|,\ L_H>0, \ x,y\in\R^d.
\end{equation}
Put $B_0:=0$ and, for $t\ge1$, define recursively
\begin{equation}\label{eq:pr-B}
M_t:=2(B_{t-1}+G),\  B_t:=\Big(\sum_{s=1}^t\|g_s\|^2+U_t\Big)^{1/2},
\end{equation}
where $U_t=U^b_t$. Define $\Dcal_{t,j}=t^{-1}\sum_{s=1}^t\|x_{s-1}-\wt x_t\|^j$, $j\in\{1,2\}$, where $\wt x_t:=t^{-1}\sum_{s=1}^tx_{s-1}$. Finally, let  $R_t^{\mathrm{PR}}$ be given by
\begin{equation}\label{eq:pr-transfer}
R^{\mathrm{PR}}_t:=\Big[\min\Big\{\frac G\mu,\ \frac{\|\sum_{s=1}^tg_s\|+B_t}{\mu t}+\frac{\min\{L\Dcal_{t,1},L_H\Dcal_{t,2}/2 \}}{\mu}+\frac{\|x_t-x_0\|}{t}\Big\}\Big]^2,\  t\ge1.
\end{equation}
 Then  it follows that $\IP\Big(\bar Z_t\le R^{\mathrm{PR}}_t \text{ for all }t\ge1\Big)\ge1-\alpha.$
Moreover, along the same trajectory, $R^{\mathrm{PR},\mathrm{Bet}}_t\le R^{\mathrm{PR},\Gamma\mathrm S}_t$ for all $t\ge1$.
\end{thm}
Theorem~\ref{thm:pr-coverage} is proved in Appendix~\ref{app:pr-proofs}. The proof proceeds via transferring the problem of deriving a time-uniform boundary for $\bar Z_t$ to deriving a time-uniform boundary for $\|\Xi_t\|^2$. 

Beyond constructing new anytime-valid boundaries, and recovering and sharpening existing ones, a key question has so far remained open: \textit{are any of these envelopes optimal in an asymptotic sense?} While the plethora of assumptions and performance measures treated in this paper precludes a case-by-case optimality analysis within the confines of time and space, the ubiquitous setting of \SC\ and \LS\ offers a natural starting point. 

\subsubsection{Optimality properties of different boundaries}
In this section, we consider the power law step-size schedule $\eta_t \asymp t^{-\beta}, \ \beta \in (1/2, 1)$, and explore the asymptotic behavior of each of the mixing, stitching and betting constructions considered in \S\ref{se:PR-boundaries}.

In particular, for a fixed $t$, we are interested in the $O(\cdot)$ rate of $R_t^b$ for different choices of $b$. We note that the intuition that different boundaries may have different optimality properties is also discussed in \cite{howard2021time}; in particular, if the mixture distribution has a positive and continuous density around the origin (which includes the Gaussian mixture model), they show that the mixture boundary is supposed to behave logarithmically ($\sqrt{x\log x}$), instead of the sharper iterated logarithmic rate of $\sqrt{x\log \log x}$ (achieved by, eg. stitching). However, their results are derived for the sum of sub-$\psi$ random variables, which is not directly applicable in our somewhat different setting of PR-averaged SGD iterates. This begs the question of whether the Gaussian mixture density indeed has sub-optimal boundaries over the other options for this case. We conclusively answer this question in Theorem~\ref{thm:pr-rates}.
\begin{thm}[Rates of the averaged-iterate envelopes]\label{thm:pr-rates}
Put $h_{\mathrm{NM}}(t):=\log(e+t)$ and $h_b(t):=1+\log\log(e+t)$ for $b\in\{\mathrm{GS},\Gamma\mathrm S,\mathrm{Bet}\}$. Grant the conditions of Theorem~\ref{thm:pr-coverage} with $\eta_t=\eta_\circ(t+t_0)^{-\beta}$, where $\eta_\circ>0$, $t_0\ge0$ and $\beta\in(1/2,1)$. Then there is a constant $C$, depending only on $(\alpha,G,\mu,L, L_H, \eta_\circ,t_0,\beta)$, and additionally on $\rho$ if $b=\mathrm{NM}$ such that, for all $t\ge1$, $\IE R^{\mathrm{PR}}_t\le Ch_b(t)/t$.
\end{thm}
Theorem~\ref{thm:pr-rates} is proved in Appendix~\ref{app:pr-proofs}. It shows that the stitching and betting envelopes achieve iterated logarithmic rate, while the Gaussian-mixture envelope is sub-optimal, and of the order ${\log t/t}$. Intuitively, as the accumulated variance $v$ increases, the useful
values of $\lambda$ approach zero, but a fixed Gaussian mixing
distribution assigns exponentially decreasing mass to successive
geometric scales of $\lambda$. In fact, in light of \cite[Proposition~2]{howard2021time}, it can be shown that any mixing distribution $F$ which is continuous with positive density at and around the origin, will have the $\log t/t$ rate with power law step-sizes. In contrast, our stitching weights $p_k\asymp(k+1)^{-2}$ allocate the
error budget polynomially across these scales.

Moreover, Theorem~\ref{thm:pr-rates} also highlights that modulo the logarithmic penalties coming from ensuring validity over all $t\ge 1$, our envelopes achieve the known optimal rates in a fixed $t$-sense. While the result is derived under a specific set of assumptions for SGD, we expect such optimality to hold across the diverse settings and corresponding envelopes considered in this paper. To summarize, beyond simply furnishing time-uniform boundaries, our results are tight in an asymptotic sense.

Moving on, in \S\ref{se:appl} we move beyond SGD to analyze three separate iterative stochastic algorithms, and identify how the tools developed in \S\ref{se:prob} can be employed to provide anytime-valid boundaries. 

\section{Applications beyond SGD}\label{se:appl}
We now apply the recipe of~\ref{fig:schematic} to three iterative algorithms beyond SGD: Oja's algorithm (\S\ref{se:oja}), temporal difference (TD) learning (\S\ref{se:rl}), and Bures - Wasserstein SGD for Gaussian variational inference (\S\ref{se:bw}). In the following, we primarily focus on completing the \textit{optimization module}, deriving \eqref{eq:key-recursion}, so that Theorems~\ref{thm:obs-comp}-\ref{thm:obs-betting} can be employed to derive time-uniform boundaries. The resulting boundaries provide the \textit{first} instances of data-dependent, anytime-valid boundaries, for each of the three applications. 

Throughout, step sizes $\eta_t>0$ are predictable and observable. In addition, $U_t$  will be a generic placeholder for any one of the observable envelopes $\mathcal U^{\mathrm{mix}}_\alpha$, $\mathcal U^{\mathrm{dm}}_\alpha$, $\mathcal U^{\St}_\alpha$, or, $\mathcal U^{\mathrm{bet}}_\alpha$ from Theorems~\ref{thm:obs-comp} and \ref{thm:obs-betting}. The realized boundary for the considered performance measure, corresponding to $U_\alpha(t;M,\nu)$, will be denoted  $R_t$. 

\subsection{Oja's algorithm}\label{se:oja}

As a first illustration beyond SGD, we consider the online estimation of a leading principal component. Concretely, for a sequence of vectors $\{ X_n\}_{n \ge 1}\in\R^d$, let $u^\star$ be the leading unit eigen-vector of $\Sigma=\IE[X_tX_t^\top]$; when $d$ is large, storing the stream, or even maintaining the sample covariance matrix, is computationally prohibitive, and one instead seeks an online estimate of $u^\star$. Oja's algorithm \citep{oja1982simplified,oja1985stochastic, krasulina1969method} constitutes the canonical procedure for this task. For a fixed iteration $t$, the asymptotic behavior of Oja's algorithm is well-understood \citep{balsubramani2013fast,shamir2016convergence, jain2016streaming, lunde2021bootstrapping,kumar2025beyond}. Beyond convergence analyses, a key bottleneck to the development of anytime-valid boundaries for Oja's algorithm, has been that the random matrix product structure $\prod_{s\le t}(I+\eta_sX_sX_s^\top)u_0$; products, unlike sums, do not readily furnish an exponential supermartingale. To the best of our knowledge, the first time-uniform guarantee for Oja's algorithm appeared in \citet{pham2025time}, who bypass the product structure through a quantitative Robbins--Siegmund argument on the sine-squared error. While their envelope attains the optimal $O(\log\log t/t)$ width, it suffers from a few limitations from a practitioner's perspective. It (i) is deterministic, (ii) is tied to a particular deterministic schedule $\eta_t\asymp1/t$, and (iii) the analysis is confined to bounded observations. 

Our recipe overcomes each of these obstacles via a direct derivation of a sharp recursive inequality \eqref{eq:key-recursion}, and employing the results Theorem~\ref{thm:obs-comp}-\ref{thm:obs-betting}. In particular, assuming bounded observations $X_t$ yields normal-mixture, stitching and betting envelopes, whereas the self-normalized construction of Table~\ref{tab:sub-psi-examples} requires only a known bound on the conditional fourth moment, and thereby covers Gaussian and other unbounded data. Therefore, through our generalized approach we are able to significantly relax the assumptions of \cite{pham2025time}. Importantly, the resulting envelopes are data-dependent, valid for any predictable step sizes, require only $\IE_{t-1}[X_tX_t^\top]=\Sigma$ rather than independence. These developments represent, to the best of our knowledge, the \textit{first, data-dependent} anytime-valid guarantees for streaming PCA, while facilitating significant improvements and relaxations over the corresponding guarantees of \cite{pham2025time}.

\subsubsection{Set-up and optimization module for Oja's algorithm} \label{se:oja-opt}
Let $X_t\in\R^d$, $d\ge2$, be $\Fcal_t$-measurable with $\IE_{t-1}[X_tX_t^\top]=\Sigma$ for a fixed $\Sigma\succeq0$ with eigenvalues $\lambda_1(\Sigma)\ge\ldots\ge\lambda_d(\Sigma)$. Let $u^\star$ be a unit leading eigenvector of $\Sigma$, and suppose a known $\delta>0$ satisfies $\lambda_1(\Sigma)-\lambda_2(\Sigma)\ge\delta$. With $\widehat\Sigma_t:=X_tX_t^\top$ and $u_0\in\mathbb S^{d-1}$, Oja's algorithm reads
\[
u_t=\frac{(I+\eta_t\widehat\Sigma_t)u_{t-1}}{\|(I+\eta_t\widehat\Sigma_t)u_{t-1}\|},
\]
which is well defined since $\widehat\Sigma_t\succeq0$. Note that the noise variable can be recognized as $\xi_t:=\mathrm{vec}(\widehat\Sigma_t-\Sigma)$. We consider $Z_t:=1-\langle u_t,u^\star\rangle^2\in[0,1]$ as the performance measure.  For independent observations with $\|X_t\|\le B$, this is the measure analyzed by \citet{pham2025time}, whose Proposition~4 and Theorem~3 furnish a scalar recursion and a warm-started, deterministic time-uniform bound. In  the following, we derive a sharper recursion that produces \eqref{eq:key-recursion} for Oja's algorithm.

\begin{prop}\label{prop:oja-rec}
Grant the notations and conditions in~\ref{se:oja-opt}. For every $t\ge1$, it almost surely holds that
\begin{equation}\label{eq:oja-rec}
Z_t\le\big(1-2\delta\eta_t(1-Z_{t-1})\big)Z_{t-1}+3\eta_t^2\|\widehat\Sigma_tu_{t-1}\|^2+\langle\eta_t\,\mathrm{vec}\,\mathcal H(u_{t-1}),\xi_t\rangle,
\end{equation}
where $\mathcal H(u):=2\langle u,u^\star\rangle uu^\top-\langle u,u^\star\rangle(uu^{\star\top}+u^\star u^\top)$. Consequently, for any predictable $r_{t-1}\in[0,1]$ and any predictable $a_t\ge1-2\delta\eta_t(1-r_{t-1})$, \eqref{eq:key-recursion} holds almost surely on $\{Z_{t-1}\le r_{t-1}\}$ with
\[
b_t:=3\eta_t^2\|\widehat\Sigma_tu_{t-1}\|^2,\ v_t:=\eta_t\,\mathrm{vec}\,\mathcal H(u_{t-1}).
\]
\end{prop}
Proposition~\ref{prop:oja-rec} is proved in Appendix \S\ref{app:se4.1proofs}. Regarding the choice of $a_t$, one can simply choose $a_t=\exp\bigl(-2\delta\eta_t(1-r_{t-1})\bigr)$ or $a_t=1-2\delta\eta_t(1-r_{t-1})$, which is sharper whenever it is positive. The recursion in Proposition~\ref{prop:oja-rec} crucially differs from the corresponding recursion derived in Proposition 4 of \cite{pham2025time} in the coefficient of the second-order term $\eta_t^2$. While \cite{pham2025time} includes the global control $5B^4 + 2\eta_t B^6$ under the assumption $\|X_t\|\le B$, Proposition~\ref{prop:oja-rec} replaces this by the much sharper, data-dependent control $3\|\widehat\Sigma_tu_{t-1}\|^2= 3 \|X_t\|^2 \langle X_t , u_{t-1}\rangle^2$. In fact, under the bounded observations assumption, it trivially follows that $3\|\widehat\Sigma_tu_{t-1}\|^2 \le 3B^4 < 5B^4 + 2\eta_t B^6$. 

\subsubsection{Boundaries for Oja's algorithm} \label{se:oja-boundary}

The sharpened recursion \eqref{eq:oja-rec} directly leads to observable boundaries via Theorem~\ref{thm:obs-comp} upon a verification of the conditions therein. To that end, observe that $\|\mathcal H(u)\|^2=2\langle u,u^\star\rangle^2(1-\langle u,u^\star\rangle^2)\le2(1-\langle u,u^\star\rangle^2)$. Hence, on $\{Z_{t-1}\le R_{t-1}\}$,
\[
\frac{\|v_t\|}{A_t}\le M_t:=\frac{\eta_t\sqrt{2R_{t-1}}}{A_t}.
\]
Since \eqref{eq:key-recursion} admits predictable contraction coefficients, we feed the preceding envelope back into $a_t$ through Lemma~\ref{lem:observable-envelope}, so that the time-uniform envelope sharpens its own contraction as the run proceeds. For the noise, either a sub-$\psi$ proxy may be used, or, under bounded observations $\|X_t\|\le B$, the product-betting boundary: in that case $\mathrm{vec}\,\widehat\Sigma_t$ is observable, $\|\mathrm{vec}\,\widehat\Sigma_t\|=\|X_t\|^2\le B^2$ and $\xi_t=\mathrm{vec}\,\widehat\Sigma_t-\IE_{t-1}\mathrm{vec}\,\widehat\Sigma_t$, which is precisely the setting of Corollary~\ref{corol:betting} with $c=B^2$. A large class of anytime-valid envelopes follow suit.  For the reader's convenience, we list a few concrete envelopes in the Appendix \S\ref{app:oja}.

\subsection{Temporal difference learning}\label{se:rl}
Reinforcement learning furnishes two of the most widely deployed stochastic approximation schemes: temporal difference (TD) learning \citep{sutton1988learning} and Q-learning \cite{watkins1992q}. They look at two different aspects of the same problem: while TD learning evaluates a learned policy, Q-learning learns an optimal policy. Here, we discuss the derivation of time-uniform boundaries for TD learning; Appendix~\ref{app:q-learning} discusses the analogous derivation for Q-learning.

TD learning evaluates a fixed policy by solving a projected Bellman equation from sampled transitions. Asymptotic convergence of TD learning is classical, eg. the analysis of \citet{tsitsiklis1997analysis} for TD with linear approximation and the ODE method of \citet{borkar2000ode}. For TD learning with linear value approximation, finite-sample theory began with \citet{dalal2018finite, lakshminarayanan2018linear}, and was extended to  high-probability bounds \citep{mou2020linear,khamaru2021temporal,li2024highprob, durmus2021tight, durmus2025finite} as well as asymptotic inference\citep{samsonov2024Gaussian, wu2026statistical, wu2025uncertainty, zeng2026self}. Crucially, all of these guarantees are stated at a fixed, prespecified iteration; \citet{zeng2026self} themselves note that optional stopping and confidence sequences require a different inferential framework. The guarantees that do hold simultaneously over iterations are mainly of the type:  with some choice of step-sizes,  every iterate after a burn-in $N(\epsilon,\delta)$ lies in an $\epsilon$-ball around the TD fixed point with probability at least $1-\delta$ (\citet[Theorem~3.5]{dalal2018finite}). More broadly, anytime-valid inference in sequential decision making has been developed for off-policy evaluation in contextual bandits \citep{karampatziakis2021off,waudby2024anytime}, where importance-weighted estimators have an explicit martingale structure. However, the iterates of TD-learning obey a contraction-plus-noise recursion around an unknown fixed point, to which such constructions do not directly apply. To the best of our knowledge, no data-dependent anytime-valid envelopes are available for TD learning before this work.

We again play the familiar game of reducing this iterative algorithm to \eqref{eq:key-recursion}, and employing Theorems~\ref{thm:obs-comp}-\ref{thm:obs-betting} and Lemma \ref{lem:observable-envelope}. For TD learning, the positive-definiteness of the mean dynamics \citep[Lemma~3]{bhandari2018finite} yields the contraction directly (Proposition~\ref{prop:td-rec}). The resulting envelopes are data-dependent, valid for any predictable step sizes, and require no burn-in. Additionally, the envelopes transfer simultaneously to the error in value function $V$, separated from the bias or the approximation error of the linear architecture (Theorem~\ref{thm:td-last}). 

\subsubsection{Set-up and optimization module for TD learning}\label{se:td}
This section discusses the formal set-up of TD learning, and goes on to derive the key recursion \eqref{eq:key-recursion} for TD learning.
Consider a finite discounted Markov decision process (MDP) \citep{bertsekas2017,sutton2018} $\mathcal M=(\mathcal S,\mathcal A,P,\mathcal R,\gamma)$, where $\mathcal S$ and $\mathcal A$ denote the \textit{state space} of the environment and the \textit{action space} available to the agent, both assumed for simplicity to be finite, $P:\mathcal S\times\mathcal A\to\mathcal P(\mathcal S)$ is the transition kernel, $\mathcal R:\mathcal S\times\mathcal A\to\mathcal P([-r_{\max},r_{\max}])$ is the reward law with mean $\bar r(s,a)$, and $\gamma\in[0,1)$ is the discount factor. For a known, fixed policy $\pi:\mathcal S\to\mathcal P(\mathcal A)$, consider the Markov chain on $\mathcal S$ with transition kernel and mean reward
\[
P_\pi(s,\cdot):=\sum_{a\in\mathcal A}\pi(a\mid s)\,P(\cdot\mid s,a),\ r_\pi(s):=\sum_{a\in\mathcal A}\pi(a\mid s)\,\bar r(s,a),
\]
and we let $d_\pi$ denote a stationary distribution of $P_\pi$, i.e., $d_\pi P_\pi=d_\pi$. 

Following the set-up of \citet{dalal2018finite,wu2026statistical}, at each iteration, independently of $\Fcal_{t-1}$ and across iterations, one draws $S_t\sim d_\pi$, $A_t\sim\pi(\cdot\mid S_t)$, $r_t\sim\mathcal R(\cdot\mid S_t,A_t)$ and $S'_t\sim P(\cdot\mid S_t,A_t)$, and only the triple $(S_t,r_t,S'_t)$ is used by the algorithm. In particular, $S'_t\mid S_t\sim P_\pi(S_t,\cdot)$, $\IE[r_t\mid S_t]=r_\pi(S_t)$ and $|r_t|\le r_{\max}$. The value function of $\pi$, $\Vcal^\pi(s):=\IE[\sum_{k\ge0}\gamma^kr_k\mid S_0=s]$ with actions $A_k\sim\pi(\cdot\mid S_k)$, is the unique solution of the Bellman equation $\Vcal^\pi=r_\pi+\gamma P_\pi \Vcal^\pi$. The statistical goal is to compute $\mathcal{V}^\pi$.

In practice, it is computationally inconvenient to compute $\Vcal^\pi$ directly due to the usually prohibitively large size of the state space $\Scal$. To mitigate this, one often resorts to a feature map $\phi:\mathcal S\to\R^d$ satisfying
\[
\sup_s\|\phi(s)\|\le1,\ C:=\IE_{S\sim d_\pi}[\phi(S)\phi(S)^\top]\succeq c_0I
\]
for a known $c_0>0$. Denote $V_\theta(s):=\phi(s)^\top\theta$. The goal of the temporal difference learning is to obtain an accurate approximation of $\Vcal^\pi$ via learning $\theta^\star \in \argmin_{\theta} \| V_\theta - \Vcal^\pi\|_{d^\pi}$, where for a measure $\mu$, $\|f\|_{\mu}:= (\IE_{X \sim \mu}[f(X)^2])^{1/2}$ denotes the $\Lcal_2(\mu)$ norm. Since $V_\theta$ is a linear function of $\theta$, this is also known as the linear value approximation. 
 TD learning proceeds via the recursion $\theta_t=\theta_{t-1}-\eta_tg_t$, where,
\begin{gather*}
g_t:=\phi(S_t)\big[(\phi(S_t)-\gamma\phi(S'_t))^\top\theta_{t-1}-r_t\big].
\end{gather*}
To identify the noise sequence $\xi_t$ in the TD recursion, we introduce a few more notations. Define
\[A_\pi:=\IE[\phi(S)(\phi(S)-\gamma\phi(S'))^\top],\ \text{and }   b_\pi:=\IE[r\phi(S)]. \]
Let $\kappa>0$ be such that $e^\top A_\pi e\ge\kappa\|e\|^2$; for example, Lemma 3 in \cite{bhandari2018finite} yields the choice $\kappa=(1-\gamma)c_0$, although sharper choices exist. Subsequently, to simplify computation, we assume $\theta^\star = A_\pi^{-1} b_{\pi}$, so that $\Vcal_{\theta^\star}= \Vcal^\pi$.
  The noise variable can be recognized as
\begin{align}
    \xi_t:=\IE_{t-1}g_t-g_t=A_\pi(\theta_{t-1}-\theta^\star)-g_t, \label{eq:xi-td}
\end{align}
and $\|g_t\|\le (1+\gamma)\|\theta_{t-1}\|+r_{\max}$. We consider the squared parameter error $Z_t:=\|\theta_t-\theta^\star\|^2$ as the performance measure. In the following, we derive the corresponding version of \eqref{eq:key-recursion}.
\begin{prop}\label{prop:td-rec}
With the notation and conditions of \S\ref{se:td}, for every $t\ge1$, it almost surely holds that
\begin{equation}\label{eq:td-rec}
Z_t\le(1-2\kappa\eta_t)Z_{t-1}+\eta_t^2\|g_t\|^2+\langle2\eta_t(\theta_{t-1}-\theta^\star),\xi_t\rangle.
\end{equation}
\end{prop}
Proposition~\ref{prop:td-rec} is proved in Appendix~\ref{app:se4proofs}. Equation \eqref{eq:td-rec} can be considered a sharpened counterpart of the expected drift underlying the analysis of \citet{bhandari2018finite}. Indeed, the last term vanishes upon taking the conditional expectation $\IE_{t-1}$, whereas we retain it together with the realized, observable second-order term $\eta_t^2\|g_t\|^2$. 

\subsubsection{Boundaries for TD-learning}
Proposition \ref{prop:td-rec} drives the application of Lemma~\ref{lem:observable-envelope} to provide sharp, data-dependent envelopes. A general boundary is summarized in the following result. 
\begin{thm}\label{thm:td-last}
With $\xi_t$ as in \eqref{eq:xi-td}, let $(\xi_t,\Delta V_t)_{t\ge0}$ be a sub-$\psi$ process with $\Delta V_t\preceq\nu_tI$ as in Theorem~\ref{thm:obs-comp}. Let $a_t=1-2\kappa \eta_t$, $b_t=\eta_t^2 \|g_t\|^2$, and define recursively $A_t:=\prod_{s=1}^ta_s$ and
\[
R_t:=A_t\Big(R_0+\sum_{s=1}^t\frac{b_s}{A_s}+U_t\Big) \ \text{with } R_0 \ge Z_0 \ \text{and } M_t= 2\eta_t \sqrt{R_{t-1}}/A_t.
\]
Then, with probability at least $1-\alpha$, simultaneously for all $t\ge0$, $\|\Vcal_{\theta_t}-\Vcal_{\theta^\star}\|^2_{d_\pi}\le R_t$.
\end{thm}
Even though Theorem~\ref{thm:td-last} follows almost trivially from Proposition~\ref{prop:td-rec} and Lemma~\ref{lem:observable-envelope}, for the sake of completeness, we present a proof in Appendix \S\ref{app:se4.2proofs}. As for the noise assumptions, either a sub-$\psi$ proxy may be used, or, for projected TD learning, the product-betting boundary. A few concrete envelopes are presented in Appendix \S\ref{app:td}. 

\subsection{Bures - Wasserstein SGD for Gaussian variational inference}\label{se:bw}

Variational inference (VI), with Gaussian VI as its canonical instance, replaces sampling from a posterior by optimization over a tractable family of distributions \citep{wainwright2008graphical,blei2017variational}. A recent line of work views Gaussian VI through the lens of Wasserstein gradient flows \citep{jordan1998variational,otto2001geometry,ambrosio2008gradient,wibisono2018sampling}. Restricted to Gaussians, the Wasserstein distance induces the Bures-Wasserstein geometry \citep{takatsu2011wasserstein,bhatia2019bures}, which already underpins provably convergent algorithms for Bures-Wasserstein barycenters \citep{chewi2020gradient,altschuler2021averaging}. Building on this geometry, \citet{lambert2022variational} derived Gaussian VI as the Bures - Wasserstein gradient flow of the KL divergence and proposed a stochastic gradient scheme (BW-SGD) whose expected squared Wasserstein error contracts geometrically up to a step-size-dependent floor when $\pi$ is strongly log-concave. For a fixed iteration $t$, the behavior of these schemes is well-understood \citep{diao2023forward, domke2023provable, kim2023convergence, luu2025stochastic}; crucially, none of the guarantees are time-uniform.

In practice, default implementations of stochastic VI stop at a data-dependent time, i.e. based on the relative change of a noisy estimate of the evidence lower bound as in Algorithm 1 of \citep{kucukelbir2017automatic}. Similarly, \citet{welandawe2024framework} adapt the learning rate by detecting stationarity of the iterates, and terminate by weighing an estimated symmetrized KL divergence to the optimal approximation against the remaining computational cost. On the other hand, \citet{dhaka2020robust} showed that common stochastic optimizers can return inaccurate approximations already in moderate dimensions, and proposed to treat the iterates as a Markov chain, with convergence diagnostics and a stopping rule that are robust to noisy objective evaluations.  Despite their practical use, none ensures the validity of a data-dependent stopping time at some prescribed level. Our recipe addresses this gap via a direct derivation of \eqref{eq:key-recursion} for the Bures-Wasserstein SGD. The resulting envelopes are data-dependent and work for any predictable step sizes below the stability threshold of \citet{lambert2022variational}. To the best of our knowledge, these are the first time-uniform guarantees for stochastic variational inference.

\subsubsection{Set-up and optimization module for Bures - Wasserstein SGD}\label{se:bw-opt}

Stochastic variational inference considers a target density $\pi\propto e^{-F}$ on $\R^d$, and aims to find the Gaussian measure $q^\star=N(m^\star,\Sigma^\star)$ that minimizes $\mathrm{KL}(q\|\pi)$ over nondegenerate Gaussians $q$. The primary motivation for such approximations is to facilitate sampling. Concretely, suppose $F\in C^2$ and $\mu I\preceq\nabla^2F(x)\preceq I$ for all $x$, with a known $\mu\in(0,1]$. By \citet[\S3.3]{lambert2022variational}, $\IE_{q^\star}\nabla F=0$ and $(\Sigma^\star)^{-1}=\IE_{q^\star}\nabla^2F\succeq\mu I$, and consequently $\Sigma^\star\preceq\mu^{-1}I$. We consider the Bures-Wasserstein SGD (Algorithm~1 of \citet{lambert2022variational}).

Concretely, given $q_{t-1}=N(m_{t-1},\Sigma_{t-1})$, draw $X_t\mid\Fcal_{t-1}\sim q_{t-1}$, and set $g_t:=\nabla F(X_t)$, $J_t:=\nabla^2F(X_t)$,
\begin{gather*}
m_t=m_{t-1}-\eta_tg_t,\  q_t=N(m_t,\Sigma_t),\\
\text{and }  \Sigma_t=\mathrm{clip}_{1/\mu}\big((I-\eta_t(J_t-\Sigma_{t-1}^{-1}))\Sigma_{t-1}(I-\eta_t(J_t-\Sigma_{t-1}^{-1}))\big),
\end{gather*}
where for $\tau>0$ and a symmetric matrix $\Sigma$ with spectral decomposition $\Sigma=\sum_{i=1}^d\lambda_iu_iu_i^\top$, we write $\mathrm{clip}_\tau(\Sigma):=\sum_{i=1}^d\min\{\lambda_i,\tau\}\,u_iu_i^\top$. Consider $Z_t:=W_2^2(q_t,q^\star)$ as the performance measure. For Gaussian random vectors, \cite{dowson1982frechet} provides a particularly simple representation for $Z_t$:
\begin{equation}\label{eq:bw-w2}
Z_t = W_2^2\big(N(m_t,\Sigma_t),N(m^\star,\Sigma^\star)\big)=\|m_t-m^\star\|^2+\Tr\big(\Sigma_t+\Sigma^\star-2(\Sigma_t^{1/2}\Sigma^\star\Sigma_t^{1/2})^{1/2}\big).
\end{equation}
To prepare for the optimization-module recursion \eqref{eq:key-recursion}, we require some more notation. For $a \in \R^{d}, B\in \R^{d\times d}$, write $(a :B) \in \R^{d \times (d+1)}$ imbued with the inner product \[ \langle a_1: B_1, \ a_2 : B_2\rangle= a_1^\top a_2 + \Tr(B_1^\top B_2). \]
Then, let $G_t:=(g_t-\nabla F(m_{t-1}),J_t)$ and, denote the noise $\xi_t:=\IE_{t-1}G_t-G_t$. In the following, we derive \eqref{eq:key-recursion} for BW-SGD.
\begin{prop}\label{prop:bw-rec}
Consider the notation and conditions of \S\ref{se:bw-opt}. Let the step sizes $\eta_t$ be predictable with $0<\eta_t\le\mu^2/60$ almost surely. Then, for every $t\ge1$, it almost surely holds that
\begin{equation}\label{eq:bw-rec}
Z_t\le(1-2\mu\eta_t)Z_{t-1}+\eta_t^2\bigl[\|g_t\|^2+\|(J_t-\Sigma^{-1}_{t-1})\Sigma^{1/2}_{t-1}\|^2\bigr]+\langle v_t,\xi_t\rangle,
\end{equation}
where $v_t:=2\eta_t\big(m_{t-1}-m^\star,\mathrm{sym}\bigl[(I-T_{\Sigma_{t-1}})\Sigma_{t-1}\bigr]\big)$, with $$T_\Sigma:=\Sigma^{-1/2}(\Sigma^{1/2}\Sigma^\star\Sigma^{1/2})^{1/2}\Sigma^{-1/2},$$ so that $x\mapsto m^\star+T_\Sigma(x-m)$ is the optimal transport map from $N(m,\Sigma)$ to $q^\star$. Here, $\mathrm{sym}(D):=(D+D^\top)/2$.
\end{prop}
Proposition~\ref{prop:bw-rec} is proved in Appendix~\ref{app:bw-proof}, where we also recall some background on the Bures - Wasserstein geometry of Gaussian variational inference. Specifically, our approach proceeds by retaining the centered first-order term in the coupling argument of \citet{lambert2022variational}. Note that the second-order term of the recursion is observable; moreover, $v_t$ is controlled by the preceding envelope without knowledge of $m^\star$, $\Sigma^\star$ or the optimal transport map. 

\subsubsection{Boundaries for BW-SGD}
As before, recursion \eqref{eq:bw-rec} directly leads to observable boundaries via Theorem~\ref{thm:obs-comp}, upon a verification of the conditions therein. Firstly, note that by \eqref{eq:bw-w2}, $Z_{t-1}$ can be written as $Z_{t-1}=\|m_{t-1}-m^\star\|^2+\|C\Sigma_{t-1}^{1/2}\|^2$ with $C=I-T_{\Sigma_{t-1}}$. As $\|\mathrm{sym}(D)\|\le\|D\|$ and $\|C\Sigma_{t-1}\|\le\|C\Sigma_{t-1}^{1/2}\|\,\|\Sigma_{t-1}\|_{\mathrm{op}}^{1/2}$, hence
\[
\|v_t\|^2\le4\eta_t^2\bigl[\|m_{t-1}-m^\star\|^2+\|C\Sigma_{t-1}\|^2\bigr]\le4\eta_t^2\max\{1,\|\Sigma_{t-1}\|_{\mathrm{op}}\}Z_{t-1}.
\]
Consequently, on $\{Z_{t-1}\le R_{t-1}\}$, 
\begin{align}
    \|v_t\|/A_t\le M_t:=2\eta_tA_t^{-1}\sqrt{\max\{1,\|\Sigma_{t-1}\|_{\mathrm{op}}\}R_{t-1}}. \label{eq:bw-m-defn}
\end{align}
Thus, Lemma~\ref{lem:observable-envelope} leads to the following result.
\begin{thm}\label{thm:bw}
Grant the notation and conditions of Proposition \ref{prop:bw-rec}, and fix $\alpha\in(0,1)$. Recall $M_t$ from \eqref{eq:bw-m-defn}. At every iteration $t$, choose
\[
\nu_t\in\Big\{2\max\Big\{\|\Sigma_{t-1}\|_{\mathrm{op}},\frac{(1-\mu)^2d}4\Big\},\ \frac23\big(\|G_t\|^2+\Tr(\Sigma_{t-1})+d\big)\Big\}.
\]
 Finally, let $$R_0:=\mu^{-2}(\|\nabla F(m_0)\|+\sqrt{d/\mu})^2+\Tr(\Sigma_0)+d/\mu,$$ and define recursively, for $t\ge1$,
\[
\begin{gathered}
A_t:=A_{t-1}(1-2\mu\eta_t),\ A_0=1, \  b_t:=\eta_t^2\bigl[\|g_t\|^2+\|(J_t-\Sigma_{t-1}^{-1})\Sigma_{t-1}^{1/2}\|^2\bigr],\\
M_t:=\frac{2\eta_t}{A_t}\sqrt{\max\{1,\|\Sigma_{t-1}\|_{\mathrm{op}}\}R_{t-1}}, \ V_t:= \sum_{s=1}^t \nu_s M_s^2, \text{ and } R_t:=A_t\Big(R_0+\sum_{s=1}^t\frac{b_s}{A_s}+U_t\Big),
\end{gathered}
\]
where $U_0:=0$, and $U_t\in\big\{\Ncal_\alpha(V_t),\ 4\mathfrak S_\alpha(V_t)\big\}$ is either the Gaussian-mixture or the sub-Gaussian stitching boundary from Corollary \ref{cor:closed-form}(i)-(ii). 
Then, $\IP(W_2^2(q_t,q^\star)\le R_t\text{ for all }t\ge0)\ge1-\alpha$.
\end{thm}
Theorem \ref{thm:bw} records the first anytime-valid control for the convergence errors of Bures-Wasserstein SGD. Notably, in the spirit of the posterior error bounds of \citet{huggins2020validated}, the anytime-valid Wasserstein error bound also transfers to posterior summaries and decisions. 
\begin{prop}\label{prop:bw-stat}
On the event of Theorem~\ref{thm:bw}, simultaneously for all $t\ge0$, the following hold.
\begin{itemize}
    \item For every $L_f$-Lipschitz $f:\R^d\to\R$, $|\int f\,dq_t-\int f\,dq^\star|\le L_f\sqrt{R_t}$. Moreover, for every $w\in\R^d$ and $p\in(0,1)$,
\[ |Q_p^t(X) - {Q}_p^{\star}(X)| \le \|w\|\sqrt{(1+z_p^2)R_t},  \]
where $Q_p^t$ and ${Q}_p^{\star}$ denote the $p$-quantiles of $w^\top X$ under $X\sim q_t$ and $X\sim q^\star$ respectively, and $z_p$ the $N(0,1)$ $p$-quantile.
\item If a loss $\mathcal L(u,x)$ satisfies
\[ \sup_u|\mathcal L(u,x) - \mathcal L(u,x')| \le L_{\mathcal L}|x-x'|, \ \IE[|\Lcal(u,x)|]<\infty, \]
then every $u_t\in\arg\min_u\int\mathcal L(u,x)\,dq_t(x)$ satisfies $$\int\mathcal L(u_t,x)\,dq^\star(x)-\min_u\int\mathcal L(u,x)\,dq^\star(x)\le2L_{\mathcal L}\sqrt{R_t}.$$
\end{itemize}
\end{prop}
We end with a remark that these are computational guarantees concerning the projection $q^\star$ onto the space of Gaussian distributions. For the target $\pi$ itself, $|\int f\,dq_t-\int f\,d\pi|\le L_f\sqrt{R_t}+|\int f\,dq^\star-\int f\,d\pi|$, requires the second term be controlled by \citet[Theorem~2.1]{katsevich2023approximation} under some additional assumptions (or be estimated post hoc through the bounds of \citet{huggins2020validated}). In general, one may use Proposition \ref{prop:bw-stat} to stop BW-SGD as soon as $\|w\|\sqrt{(1+z_p^2)R_t}$ falls below a user-specified tolerance for the posterior summaries of interest. Thus, our results combine to give informed, decision-theoretic stopping rules for Bures-Wasserstein SGD, providing first-of-its-kind results in stochastic VI.
\section{Numerical Experiments}\label{se:simu}
In this section, we illustrate the envelopes of \S\ref{se: anytime-valid-sgd} and \S\ref{se:appl} through two numerical examples. In \S\ref{se:simu-det-comparison}, we begin with a warm-up example: projected SGD for ridge regression, which falls within the strongly convex setting of \S\ref{se:rsi}. This is one of the settings in which time-uniform guarantees are already available, both deterministic \citep{pham2025time} and data-dependent \citep{aolaritei2026beyond}, which allows a direct comparison of our envelopes with the existing literature. Moving on, \S\ref{se:simu-td} turns to reinforcement learning, where we evaluate a fixed policy on the FrozenLake benchmark via TD(0) learning (\S\ref{se:rl}). In contrast to the first example, no data-dependent anytime-valid envelopes were previously available for TD learning. Across both examples, all envelopes are computed at level $\alpha=0.05$ from the observed iterates along a single trajectory of $T=10^7$ iterations.
\subsection{Warm-up: a ridge regression example}\label{se:simu-det-comparison}
We begin with the strongly convex setting of \S\ref{se:rsi}. Consider independent pairs $(X_t,y_t)$ from the linear model
\begin{equation*}
    y_t=X_t^\top x^\dagger+\varepsilon_t,\ x^\dagger=(0.7,-1.1,0.4,0.9,-0.6)^\top\in\R^5,
\end{equation*}
where we write $X_t$ for the covariates to reserve $x_t$ for the SGD iterates. The covariates are isotropic\footnote{Specifically, $X_t=L_tU_t$, where $U_t\sim\dunif(\mathbb S^4)$ and $L_t=2.7+0.3V_t$ with $V_t\sim\mathrm{Beta}(2,2)$, independently.} with $2.7\le\|X_t\|\le3$, so that $\IE[X_tX_t^\top]=mI_5$ with $m\approx1.625$, and the errors are independent truncated Gaussians,
\begin{equation*}
    \varepsilon_t\sim N(0,\sigma^2)\,\big|\,\{|\varepsilon_t|<3\}, \ \Var(\varepsilon_t)=1.
\end{equation*}
We run projected SGD \eqref{eq: general SGD}, initialized at the origin, for $T=10^7$ iterations on the ridge objective
\begin{equation*}
    F(x)=\frac12\IE\big(y_t-X_t^\top x\big)^2+\frac{\lambda}{2}\|x\|^2,\ \|x\|\le10,
\end{equation*}
with $\lambda=0.1m$ and $\eta_t=2/[m(t+32)]$. Here $F$ is strongly convex, and its minimizer is the ridge solution $x^\star=x^\dagger/1.1$ rather than the regression coefficient $x^\dagger$. Note that \cite{pham2025time} proposed deterministic envelopes for $\|x_t-x^\dagger\|^2$. Therefore, to aid comparison, the envelopes $R_t$ obtained via \S\ref{se: anytime-valid-sgd} for $\|x_t-x^\star\|^2$ are transferred to the error $\|x_t-x^\dagger\|^2$ as $R_t^\circ=(\sqrt{R_t}+\delta_\lambda)^2$, where $\delta_\lambda=\lambda S/(m+\lambda)\ge\|x^\star-x^\dagger\|$ with $S=2$. Moreover, the stochastic gradients are bounded by $G\approx118.6$, and since $\|x_t\|\le10$ and $\|x^\dagger\|\le S$, we cap the envelopes $R_t^\circ$ at the global bound $(10+S)^2=144$ for display.

We compare five envelopes, each at level $\alpha=0.05$. The following re-uses the notation of \S\ref{app:betting-sgd}.
\begin{itemize}
    \item[(i)] The deterministic envelope of \citet[Eq.~(26)]{pham2025time}, whose assumptions hold here with $B=3$, $D=20$ and $B_1=B^2(D+1)+\lambda(D+S)\approx192.6$;
    \item[(ii)] the empirical-Bernstein envelope of \citet[Theorem~3.5]{aolaritei2026beyond} $R_t^{\mathrm{ALBJ}}$, which gives a closed-form upper bound to the sub-gamma stitching envelope with the grid \eqref{eq:AJ};
    \item[(iii)] (ours) the exponential discrete mixture envelope $R_t^{\mathrm{dm}}$ of \S\ref{app:betting-sgd}, i.e., Corollary~\ref{corol:dm} with $\psi=\psi_{E,G}$ and $\Delta V_t=g_tg_t^\top$ as in \S\ref{se:sub-gamma-stitch}, using $\bar\lambda=0.38/G$, $r=1.1$ and the density $f$ of \ref{fig:ridge-prelim}; this choice follows \citet[Figure~4]{howard2021time}, see Appendix A.6 therein;
    \item[(iv)] (ours) the sub-exponential stitching envelope $R_t^\St$;
    and,
    \item[(v)] (ours) the betting envelope $R_t^w$ of Proposition~\ref{prop:betting-SGD} with $w=32$ and the same $(\lambda_k,p_k)$ as in (iii).
\end{itemize}
\ref{fig:ridge-prelim} displays the practical gap between deterministic and data-dependent envelopes. The realized error $\|x_t-x^\dagger\|^2$ settles around the squared ridge bias $\|x^\star-x^\dagger\|^2\approx0.025$ within $10^3$ iterations, whereas the envelope of \citet{pham2025time} remains at the global bound until $t=807{,}518$, and still equals $12.1$ at $t=10^7$. Moreover, our discrete-mixture, sub-exponential stitching, and betting based envelopes are sharper over $R_t^{\mathrm{ALBJ}}$ by orders of magnitude. For example, the envelope $R_t^{\mathrm{ALBJ}}$ falls below the global bound at $t=876$, while our envelopes do so by $t=193$, and remain below $R_t^{\mathrm{ALBJ}}$ throughout. In particular, the betting envelope is about $20$ times smaller than $R_t^{\mathrm{ALBJ}}$ at $t=10^4$ and $10^5$, and still about $6$ times smaller at $t=10^7$. Moreover, \ref{fig:ridge-prelim} also demonstrates the theoretically derived ordering $R_t^w\le R_t^{\mathrm{dm}}\le R_t^{\St}$ of \S\ref{app:betting-sgd} at every iteration. Since the common bias correction increasingly dominates all data-dependent envelopes in \ref{fig:ridge-prelim} and masks their differences, Table~\ref{tab:ridge} reports the data-dependent envelopes $R_t$ for $\|x_t-x^\star\|^2$ before the correction. We emphasize that the betting-based envelopes require the additional restriction of bounded gradients, which is satisfied in this example. 
\begin{table}[!htbp]
\centering
\caption{Envelopes $R_t$ for $\|x_t-x^\star\|^2$, before the bias correction, along the run in \ref{fig:ridge-prelim}.}
\label{tab:ridge}
\begin{tabular}{@{}rccccc@{}}
\toprule
$t$ & $\|x_t-x^\star\|^2$ & $R_t^{\mathrm{ALBJ}}$ & $R_t^{\St}$ & $R_t^{\mathrm{dm}}$ & $R_t^w$ (betting) \\
\midrule
$10^3$ & $1.91\times10^{-3}$ & $109.8$ & $8.23$ & $6.93$ & $6.79$ \\
$10^4$ & $7.15\times10^{-4}$ & $1.765$ & $0.141$ & $0.116$ & $0.0869$ \\
$10^5$ & $4.72\times10^{-5}$ & $3.75\times10^{-2}$ & $4.19\times10^{-3}$ & $3.40\times10^{-3}$ & $1.89\times10^{-3}$ \\
$10^6$ & $7.29\times10^{-6}$ & $1.38\times10^{-3}$ & $2.39\times10^{-4}$ & $1.97\times10^{-4}$ & $1.44\times10^{-4}$ \\
$10^7$ & $4.68\times10^{-7}$ & $8.56\times10^{-5}$ & $1.97\times10^{-5}$ & $1.65\times10^{-5}$ & $1.49\times10^{-5}$ \\
\bottomrule
\end{tabular}
\end{table}

\subsection{Temporal difference learning on FrozenLake}\label{se:simu-td}
We next illustrate the envelopes of \S\ref{se:rl} on FrozenLake\footnote{\href{https://gymnasium.farama.org/environments/toy_text/frozen_lake}{https://gymnasium.farama.org/environments/toy\_text/frozen\_lake}}, a standard tabular benchmark of the OpenAI Gym and Gymnasium suites \citep{brockman2016openai,towers2024gymnasium}, which has also been used to study TD-type policy evaluation \citep{wang2021online}.
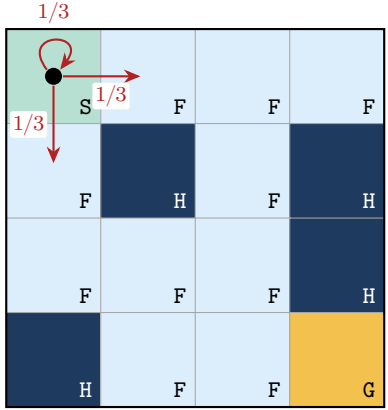
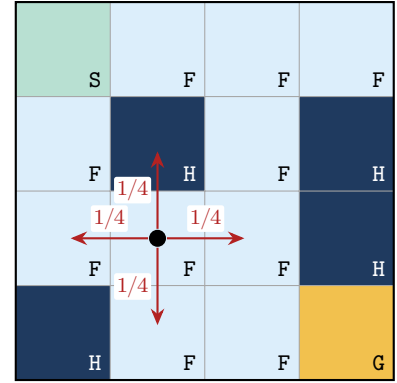
\begin{figure}[!htbp]
\centering
\begin{subfigure}[t]{0.42\textwidth}
\centering
\begin{tikzpicture}[scale=1.25,
    mv/.style={-{Stealth[length=2.2mm,width=1.8mm]},thick,flarrow},
    pr/.style={font=\scriptsize,fill=white,inner sep=1pt,rounded corners=1pt}]
  \flgrid
  \fill[black] (0.5,-0.5) circle (0.09);
  \draw[mv] (0.6,-0.5) -- (1.42,-0.5) node[pr,pos=0.62,below=2pt] {$1/3$};
  \draw[mv] (0.5,-0.6) -- (0.5,-1.42) node[pr,midway,left=2pt] {$1/3$};
  \draw[mv] (0.43,-0.43) .. controls (0.12,-0.02) and (0.88,-0.02) .. (0.57,-0.43);
  \node[pr,text=flarrow] at (0.5,0.17) {$1/3$};
\end{tikzpicture}
\caption{}
\label{fig:fl-slip}
\end{subfigure}
\hfill
\begin{subfigure}[t]{0.42\textwidth}
\centering
\begin{tikzpicture}[scale=1.25,
    mv/.style={-{Stealth[length=2.2mm,width=1.8mm]},thick,flarrow},
    pr/.style={font=\scriptsize,fill=white,inner sep=1pt,rounded corners=1pt}]
  \flgrid
  \fill[black] (1.5,-2.5) circle (0.09);
  \draw[mv] (1.6,-2.5) -- (2.42,-2.5) node[pr,midway,above=2pt] {$1/4$};
  \draw[mv] (1.4,-2.5) -- (0.58,-2.5) node[pr,midway,above=2pt] {$1/4$};
  \draw[mv] (1.5,-2.6) -- (1.5,-3.42) node[pr,midway,left=2pt] {$1/4$};
  \draw[mv] (1.5,-2.4) -- (1.5,-1.58) node[pr,midway,left=2pt] {$1/4$};
\end{tikzpicture}
\caption{}
\label{fig:fl-uniform}
\end{subfigure}
\caption{The FrozenLake map, with start $\mathtt S$, frozen tiles $\mathtt F$, holes $\mathtt H$ and goal $\mathtt G$. \textbf{(a)} Slippery dynamics for the action \emph{right} at $\mathtt S$: the agent moves right, or in either perpendicular direction, each with probability $1/3$; a move off the grid (here, up) leaves it in place. \textbf{(b)} Under the uniform policy with slippery dynamics, the agent moves in each of the four directions with probability $1/4$, staying in place at the boundary; entering $\mathtt H$ or $\mathtt G$ ends the episode.}
\label{fig:frozenlake}
\end{figure}
The agent moves on the $4\times4$ grid of Figure~\ref{fig:frozenlake} from the start $\mathtt S$ towards the goal $\mathtt G$, and an episode terminates upon entering a hole $\mathtt H$ or the goal. At each step, the policy chooses an intended action $a\in\{\mathtt{left},\mathtt{down},\mathtt{right},\mathtt{up}\}$, which the slippery ice then perturbs: the direction $D$ in which the agent actually moves is either $a$ itself or one of the two directions perpendicular to $a$, each with probability $1/3$. Formally, the realized direction and the reward satisfy
\begin{equation*}
    \IP(D=d\mid a)=
    \begin{cases}
        1/3, & d=a \text{ or } d\perp a,\\
        0, & d=\neg a,
    \end{cases}
    \
    r(s,a,s')=
    \begin{cases}
        1, & s'=\mathtt G,\\
        0, & \text{otherwise},
    \end{cases}
\end{equation*}
where $d\perp a$ denotes either direction perpendicular to $a$, and $\neg a$ the direction opposite to it. The agent stays in place if the move would leave the grid.

As a canonical example, we evaluate the uniform policy $\pi(a\mid s)=1/4$ with $\gamma=0.9$, using tabular features on the $d=11$ nonterminal states, and $\phi(s)=0$ for terminal $s$. Under $\pi$, the slippery dynamics reduce to a symmetric random walk: the agent moves in each of the four directions with probability $1/4$, and stays in place at the boundary (\ref{fig:fl-uniform}). The transitions $(S_t,r_t,S'_t)$ are sampled independently as in \S\ref{se:td}. As the stationary distribution of the episodic chain $P_\pi$ is concentrated on the absorbing terminal states, we draw $S_t$ from the stationary distribution of the chain that restarts each episode at $\mathtt S$; transitions from terminal states leave $\theta_t$ unchanged, since $\phi$ vanishes there.

We run TD(0) for $T=10^7$ iterations from $\theta_0=0$, with $\eta_t=2/[\kappa(t+1024)]$ and $\kappa=0.003$. Starting from $R_0=d$, we compare three envelopes, each at level $\alpha=0.05$, all derived from \eqref{eq:td-rec} via Lemma~\ref{lem:observable-envelope}:
\begin{itemize}
    \item[(i)] the sub-gamma stitching envelope $\tilde R_t^{\St}$ with the grid of Corollary~\ref{cor:closed-form}(iii);
    \item[(ii)] the closed-form upper bound $\Mcal_\alpha^{GE,2}$, with $c=(1+\gamma)d + 1$ and $\rho=1$, to the gamma-exponential mixture envelope $R_t^{\mathrm{GE}}$; see \S\ref{app:GE-mixture}; and,
    \item[(iii)] the betting envelope $R_t^w$ of Proposition~\ref{prop:betting-SGD}, with $\mu$ replaced by $\kappa$ in \eqref{eq:Ht-W}, $w=32$, and the same $(\lambda_k,p_k)$ as in (i).
\end{itemize}

\begin{figure}[!htbp]
\centering
\includegraphics[width=0.8\textwidth]{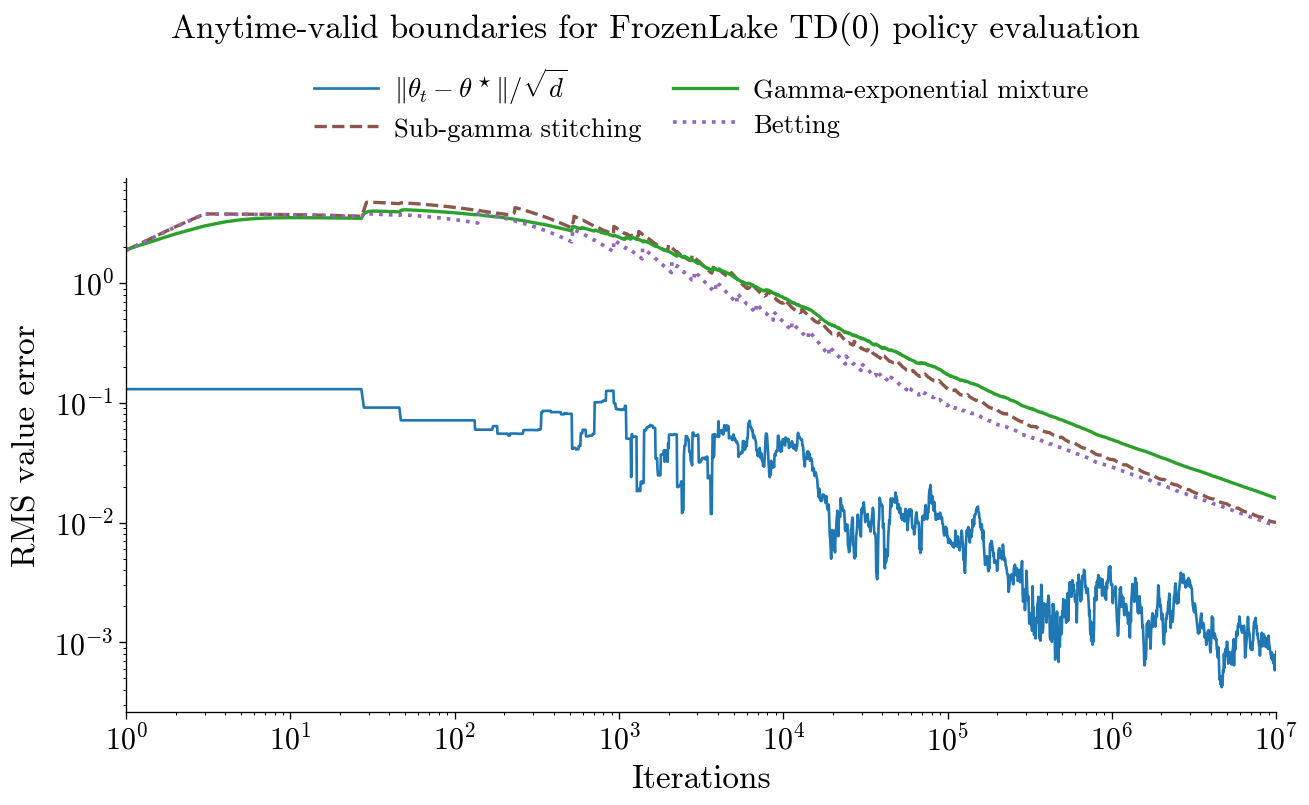}
\caption{Error $\|\theta_t-\theta^\star\|/\sqrt{d}$ of TD(0) on FrozenLake, together with the sub-gamma stitching, gamma-exponential mixture and betting envelopes, each at level $\alpha=0.05$.}
\label{fig:td-frozenlake}
\end{figure}

Figure \ref{fig:td-frozenlake} displays the realized error together with the three envelopes.  The betting envelope is the smallest throughout, in agreement with the ordering $R_t^w\le\tilde R_t^{\St}$ of \S\ref{app:betting-sgd} for a common grid: it is $52\%$ smaller than the stitching envelope at $t=10^4$, and still $12\%$ smaller at $t=10^7$. The gamma-exponential mixture envelope is competitive early on, but is overtaken by stitching after a few thousand iterations, and is $2.6$ times larger at $t=10^7$. This is the familiar trade-off between mixing with a fixed prior and stitching: the former grows like $\sqrt{V_t\log V_t}$, whereas the latter attains the iterated-logarithm rate $\sqrt{V_t\log\log V_t}$ \citep{howard2021time}. 

To the best of our knowledge, these are the first data-dependent anytime-valid envelopes for TD learning, and can be used along with, or as, data-dependent stopping rules. In practice, TD learning is typically stopped either based on thresholds derived from sample-complexity bounds \citep{bhandari2018finite,dalal2018finite}, or heuristically, once successive iterates or the empirical TD errors stabilize \citep{puterman1994markov}. Neither approach yields level-$\alpha$ validity, while also often being deterministic. In contrast, our envelopes directly yield the stopping time $\tau_\varepsilon:=\inf\{t\ge0:R_t\le\varepsilon\}$, at which $\|\theta_{\tau_\varepsilon}-\theta^\star\|^2\le\varepsilon$ with probability at least $1-\alpha$. 

\section{Conclusion}
This paper presented a general recipe for constructing nonasymptotic, anytime-valid envelopes for the performance of iterative stochastic algorithms, computed from the observed trajectory alone. Since they are valid at arbitrary stopping times, the envelopes allow practitioners to continuously monitor an algorithm and employ a data-dependent stopping time. The key idea is simple: once the performance measure of an algorithm satisfies the recursion \eqref{eq:key-recursion}, controlling it uniformly over time reduces to bounding a weighted sum of martingale differences, $\sum_{s\le t}\langle v_s,\xi_s\rangle$. This is a task for which nonnegative supermartingales, built via mixing, stitching or product-betting martingales, are well suited, and Lemma~\ref{lem:observable-envelope} transfers the resulting boundaries back to the iterates. The envelopes do not bank on any particular choices of parameters of the corresponding iterative algorithms. To help a practitioner, we provide extensive demonstrations of applying our constructions to explicit algorithms such as SGD or TD learning. The resulting boundaries recover, and often sharpen existing time-uniform guarantees in some of the settings, while yielding the first data-dependent guarantees in several others. Specifically, we show, both theoretically, and empirically, that the betting-based envelopes often outperform the stitching and mixing-based envelopes. In summary, the practitioner need only supply structural assumptions on the algorithm, which usually determines the performance measure, and the noise; our framework then yields a large class of observable, anytime-valid boundaries depending on the preferred choice of construction.

Beyond the settings of this paper, several interesting questions remain open.  For example, for SGD, while we allow step-sizes $\eta_t$ to be predictable, for many adaptive methods such as AdaGrad \citep{duchi2011adaptive} or Adam \citep{kingma2015adam}, step sizes are $\mathcal F_t$-adapted, i.e. they depend on the current stochastic gradient. Consequently, the weighted noise sum is no longer a martingale. Additionally, extending the recipe to cover Markovian noise constitutes another natural direction for future work.
\section*{AI usage statement\footnote{\href{https://leidendeclaration.ai/}{https://leidendeclaration.ai/}}}
The authors acknowledge the use of ChatGPT-Pro in improving exposition, and simplifying some proofs, such as the tedious calculations in \S\ref{app:lemma-proofs} that were verified independently. The numerical experiments were implemented by Codex. The authors take responsibility for the veracity of all claims and technical details in the paper.

\renewcommand{\bibfont}{\small}
\bibliographystyle{plainnat}
\bibliography{references}

\newpage
\begin{center}
    \textbf{\Large Appendix}
\end{center}
This appendix is devoted to extensive discussions on the constructions of the different boundaries, and their application to various optimization modules under diverse settings. In particular, \S\ref{app:boundaries} expands upon \S\ref{se:prob} to include a detailed case-by-case treatment of the different boundaries. Moving on, \S\ref{app:opt} analyzes the optimization module, and \S\ref{app:opt-sgd-boundaries} contains the derivation of the key recursion for stochastic gradient descent under different assumptions. Subsequently, \S\ref{app:appl} contains two more applications in addition to the examples in \S\ref{se:appl}; as in that section, the examples presented in \S\ref{app:appl} also provide the first anytime-valid data-dependent envelopes in the corresponding cases. Finally, in \S\ref{app:se2proofs}-\S\ref{app:lemma-proofs}, we present the proofs of our results, modularized section-wise.  

\section{Discussion on anytime-valid boundaries}\label{app:boundaries}
In this section, we include a detailed discussion of the \textit{statistical module}, which concerns all our subsequent boundary constructions. Firstly, in \S\ref{app:connotations}, we provide an equivalent
characterization of the sub-$\psi$ condition that is immensely useful for our cause. Then, we leverage this to introduce the boundary constructions via mixing \S\ref{app:mixing}, stitching \S\ref{app:stitching} and betting \S\ref{app:betting}, all the while discussing the merits and pitfalls of each construction and how they are inter-related.
\subsection{Connotations of Definition \ref{def:psi}}\label{app:connotations}
The close similarity between Definition \ref{def:psi} and the other definition of sub-$\psi$ processes merits a further discussion of their correspondence. We start off by providing a correspondence between Definition \ref{def:psi} and \eqref{eq:psi}.
\begin{lem}\label{lem:suff}
  Consider $\Fcal_t$-adapted random processes $\{\xi_t\}_{t\ge 0}$ and $\{\Delta V_t\}_{t\ge 0}$. Then, \eqref{eq:psi} holds iff Definition~\ref{def:psi} holds. 
\end{lem}
Equation \ref{eq:psi} embodies the predictable-weight formulation of Definition \ref{def:psi}, and facilitates immediate application to the boundary constructions we consider here. Another consequence of Definition \ref{def:psi} is that it holds under a mild condition on the variance process $\Delta V_t$ that $\xi_t$ itself is a martingale-difference sequence. These results firmly place Definition \ref{def:psi} in the broader context of other similar sub-$\psi$ assumptions in the literature.
\begin{lem}
\label{lem:subpsi-martingale-differences}
Suppose $(\xi_t,\Delta V_t)_{t\ge0}$ satisfies
Definition~\ref{def:psi}, and
\[
\mathbb E\operatorname{tr}(\Delta V_t)<\infty
\ \text{for every }t\ge1.
\]
Then $(\xi_t)_{t\ge1}$ is a square-integrable martingale difference
sequence with respect to $(\mathcal F_t)_{t\ge0}$. More precisely,
almost surely, for every $t\ge1$,
\[
\mathbb E_{t-1}[\xi_t]=0,
\ 
\mathbb E_{t-1}[\xi_t\xi_t^\top]
\preceq
\psi''(0)\,\mathbb E_{t-1}[\Delta V_t].
\]
\end{lem}
Lemmas~\ref{lem:suff} and \ref{lem:subpsi-martingale-differences} are proved in the Appendix~\ref{app:boundaryproofs}. 
\subsection{Boundary by mixing method}\label{app:mixing}
In this section, we discuss a class of uniform boundaries using the method of mixtures \citep{ville1939etude, wald1945sequential, darling1968some, robbins1969probability, robbins1970boundary, lai1976boundary, kaufmann2021mixture}. This approach builds on the conjugate mixture boundaries of \cite{howard2021time} by incorporating the predictable process $\bv$ into its boundaries. Mixture-based boundaries have been commonplace in iterative learning algorithms \cite{whitehouse2023sublinear, krishnamurthy2018semiparametric, whitehouse2026time}. In the context of Definition~\ref{def:psi} and \eqref{eq:psi}, the key idea underlying this approach is the observation that $\int M_t(\bv, \lambda) \ \text{d}F(\lambda)$ is an $\Fcal_t$-supermartingale for any mixture distribution $F$, which then can be inverted to either give a closed form or analytically computable anytime-valid envelope. Crucially, the domain of integration must adapt to the sequence $\bv$, which governs the tail behavior of $Y_t(\bv)$. This precludes a ready application of the boundaries derived in \cite{howard2021time}; in Theorem~\ref{thm:prob}, we provide a new result tuned to $Y_t(\bv)$. Throughout, $(\xi_t, \Delta V_t)_{t=0}^{\infty}$ denotes a sub-$\psi$ process and $\mathbf{v}=(v_s)_{s=1}^{\infty}\in \R^d$ is predictable. Consider a fixed $\alpha\in(0,1)$.
\begin{thm}\label{thm:prob}
   Grant Definition \ref{def:psi}, and suppose $F$ denotes a probability distribution on $[0, \lambda_{\max})$ and $\alpha\in (0,1)$.   Then, 
    \[ \IP\big(\text{for all $t\geq 1$:} \sum_{s=1}^t \langle \xi_s, v_s\rangle \leq \Mcal_{\alpha}(t;\mathbf{v})\big) \geq 1-\alpha,  \]
    where,
    \[ \Mcal_{\alpha}(t;\mathbf{v}, F) := \sup\Big\{x \in \R : m_{\alpha}(x, t; \bv, F) < \frac{1}{\alpha} \Big\}, \]
    with 
    \begin{align}
       m_{\alpha}(x, t; \bv, F)&:= \int_{\Lambda_t(\mathbf{v})} \exp\big(\lambda x - \sum_{s=1}^t \Gamma_{\psi}(\Delta V_s, \lambda v_s)  \big)  \ \textnormal{d}F(\lambda) \ \text{and}, \nonumber\\
        \ \Lambda_t(\mathbf{v})&:= \Big[0, \lambda_{\max} \wedge  \frac{\lambda_{\max}}{\sup_{1\leq s \leq t} \|v_s\|}\Big) \nonumber.
    \end{align}
\end{thm}
Theorem~\ref{thm:prob} is proved in Appendix~\ref{app:boundaryproofs}. It can be considered a direct analogue of Lemma 2 of \cite{howard2021time}. Corollary \ref{cor:closed-form}(i) provides a sharp, closed form upper bound for $\Mcal_{\alpha}(t,\bv)$, when $\psi=\psi_N$ and $F(\cdot; \rho)$ denotes the $N(0, \rho^{-1})$ density. In the following, we also discuss the gamma-exponential mixture boundary as another example and derive a sharp, easily computable upper bound using results from \cite{pinelis2005}. Following that, we discuss the discrete mixture boundary which facilitates numerical root-finding algorithms when closed form expressions or upper-bounds are not available.

\subsubsection{A new closed-form Gamma-Exponential mixture boundary}\label{app:GE-mixture}
In this case, we assume that $(\xi_t,\Delta V_t)_{t\ge0}$ is a sub-$\psi_{E, c}$ process and $\lambda_{\max}=1/c$, and let $\bv$ be as in Theorem~\ref{thm:prob}. Let 
\begin{align}
m_{\rho,c}(x;v,b)
:={}&
\frac{
\Big(\frac{\rho}{c^2}\Big)^{\rho/c^2}
\exp\big(
\frac{x+c}{cb}
+\frac{v+(\rho-c^2)_+}{c^2b^2}
-\frac{\rho}{c^2}\big)
\Gamma\Big(
1+\frac{v+(\rho-c^2)_+}{c^2b^2}
\Big)}{
b\,
\Gamma\Big(\frac{\rho}{c^2}\Big)
\gamma\Big(\frac{\rho}{c^2},\frac{\rho}{c^2}\Big)
\Big(
\frac{x+c}{cb}
+\frac{v+(\rho-c^2)_+}{c^2b^2}
\Big)^{
1+\frac{v+(\rho-c^2)_+}{c^2b^2}
}
}
\nonumber\\
&\ \cdot
\gamma\Big(
1+\frac{v+(\rho-c^2)_+}{c^2b^2},
\frac{x+c}{cb}
+\frac{v+(\rho-c^2)_+}{c^2b^2}
\Big) \label{eq:GE-root}
\end{align}
be defined for $v, b> 0$, where $\gamma(a,x)
:=
\int_0^x u^{a-1}e^{-u}\,du/ \Gamma(a)$ is the regularized lower incomplete gamma function. The expression for $m_{\rho, c}$ is from Proposition 9 of ~\cite{howard2021time}.

Let $\overline{V}_0:=0, \overline{B}_0:=1$. Define
\begin{align}
    \overline{V}_t
&=
\overline{V}_{t-1}
+
\langle v_t,\Delta V_t v_t\rangle, \nonumber\\
\overline{B}_t
&=
\overline{B}_{t-1}
\vee
\|v_t\|, \ \text{and}, \nonumber\\
\Mcal_{\alpha}^{GE,1}(t;\bv,\rho)
&=
\sup \big\{x: m_{\rho, c}(x; \overline{V}_t, \overline{B}_t)< 1/\alpha\big\}.
\end{align}
Then, from Theorem \ref{thm:prob}, we have $\IP\big(\text{for all $t\geq 1$:}\ \sum_{s=1}^t\langle\xi_s,v_s\rangle \leq \Mcal_{\alpha}^{GE,1}(t;\bv,\rho)\big) \geq 1-\alpha$. 

Computing $\Mcal_{\alpha}^{GE,1}$ requires numerical root-finding at every iteration $t$, which poses a significant computational barrier to its practical implementation. The main difficulty in inverting $m_{\rho,c}$ arises from the presence of the $\Gamma$ and $\gamma$ functions. Leveraging some tight upper and lower bounds for the incomplete gamma functions \citep{pinelis2005}, we circumvent this issue by deriving sharp closed-form upper bounds for $\Mcal_{\alpha}^{GE,1}$. 

In particular, write 
\begin{align}
 \mathfrak{a}(v,b)&:=1+[v+(\rho-c^2)_+]/(c^2b^2),    \nonumber \\
 q(v,b)&:=\mathfrak{a}(v,b)^{\mathfrak{a}(v,b)}e^{-\mathfrak{a}(v,b)}/\Gamma(\mathfrak{a}(v,b)), \ \text{and,} \nonumber\\
 \kappa&:=
\frac{(\rho/c^2)^{\rho/c^2}e^{-\rho/c^2}}
{\Gamma(\rho/c^2)\gamma(\rho/c^2,\rho/c^2)}, \label{eq:GE-defn}
\end{align}
 where $z_+:=\max\{z,0\}$. Let $\mathfrak{L_0}(v,b):=[\log(bq(v,b)/(\alpha\kappa))]_+$ and
\begin{equation}\label{eq:GE-level}
\mathfrak{L}(v,b):=
\Biggl[
\log\biggl(
q(v,b)\Bigl[
\frac{b}{\alpha\kappa}
+
\frac{1}{1+\sqrt{2\mathfrak{a}(v,b)\mathfrak{L_0}(v,b)}+\frac23\mathfrak{L_0}(v,b)}
\Bigr]
\biggr)
\Biggr]_+.
\end{equation}
For $a \ge 0, L\ge 0$, further define
\begin{align}
    D_a(L):= &\min\bigg\{\sqrt{2aL}+L, \sqrt{2aL}+\frac{2L}{3} \nonumber\\
    &+ \big(1 + \frac{a}{ \sqrt{2aL}+{2L}/{3}}\big)\bigg(\frac{L}{3} - \sqrt{2aL} + a \log\big(1 + \sqrt{\frac{2L}{a}} + \frac{2L}{3a}\big) \bigg)   \bigg\}, \label{eq:GE-D-defn}
\end{align}  
and $u_{\alpha, \rho, c}(v,b):=
c(b-1)+cbD_{\mathfrak{a}(v,b)}(\mathfrak{L}(v,b))$. Then Lemma~\ref{lem:GE-explicit} instructs that 
\[
\Mcal_{\alpha}^{GE,2}(t;\bv,\rho)
:=
u_{\alpha, \rho, c}(\overline{V}_t,\overline{B}_t)\ge\Mcal_{\alpha}^{GE,1}(t;\bv,\rho)
\]
satisfies $\IP\big(\text{for all $t\geq 1$:}\ \sum_{s=1}^t\langle\xi_s,v_s\rangle \leq \Mcal_{\alpha}^{GE,2}(t;\bv,\rho)\big) \geq 1-\alpha$.
\subsubsection{Discrete mixture boundary}

Closed-form expressions are not always available, in which case the corresponding boundary must be obtained numerically via root-finding. To further facilitate numerical implementation, we also introduce discrete-mixture boundaries, which empirically retain much of the sharpness of their continuous-mixture counterparts.
\begin{cor}[Discrete mixture boundary] \label{corol:dm}
 Fix deterministic support points $\lambda_k\in(0,\lambda_{\max}), \ k\ge 0,$
and deterministic weights $p_k>0$ satisfying $\sum_{k=0}^{\infty} p_k \le 1.$ Then, 
\[ \IP\big(\text{for all $t\geq 1$:} \sum_{s=1}^t \langle \xi_s, v_s\rangle \leq \DMcal_{\alpha}(t;\mathbf{v})\big) \geq 1-\alpha,  \]
    where,
    \[ \DMcal_{\alpha}(t;\mathbf{v}) := \sup\Big\{x \in \R : \sum_{\substack{k\ge 0\\ \lambda_k \in \Lambda_t(\mathbf{v})}} p_k \exp\big(\lambda_k x - \sum_{s=1}^t \Gamma_{\psi}(\Delta V_s, \lambda_k v_s)  \big) \  < \frac{1}{\alpha} \Big\}.  \]
\end{cor}
\begin{proof}
    Follows directly from Theorem~\ref{thm:prob}. 
\end{proof}
In particular, \cite{howard2021time} suggests the choice
    \[ \lambda_k= \frac{\bar{\lambda}}{r^{k+1/2}}, \ p_k = \frac{\bar{\lambda}(r-1)}{r^{k+1}} f(\frac{\bar{\lambda}}{r_k}), \]
    for some $\bar{\lambda}\in (0, \lambda_{\max})$, $r>1$, and non-increasing density function $f:(0, \bar{\lambda}) \to \R_+$. 

In the absence of any sharp upper bounds, both Theorem~\ref{thm:prob} and Corollary~\ref{corol:dm} require
the boundary to be obtained by inverting
$m_\alpha(x,t;\bv,F) = 1/\alpha$ in~$x$, a one-dimensional
root-finding problem that must be solved at every monitoring time.

\subsection{Boundary by stitching method}\label{app:stitching}
The mixture boundaries of the preceding section require either
closed-form inversion, as in the normal-mixture case, or numerical
root-finding at every iteration. The stitching method, introduced in
\cite{howard2021time} for the sub-Gaussian and sub-gamma settings,
avoids both by replacing the continuous mixture with a union bound
over geometrically spaced epochs, each governed by a single linear
boundary from Ville's inequality. The resulting envelope admits a
closed-form expression whose evaluation requires only the accumulated
compensator $\sum_{s=1}^t \Gamma_\psi(\Delta V_s, \lambda_k v_s)$
at each support point, with no root-finding step.  The resulting stitching boundary recovers, and in some cases sharpens, the
empirical-Bernstein envelopes derived in
\cite{aolaritei2026beyond}; see Appendix~\ref{app:opt-sgd-boundaries} for the
detailed comparisons.  The price of this analytical simplicity is a
loss of tightness: the stitching boundary replaces the weighted sum
in the discrete-mixture construction of
Corollary~\ref{corol:dm} by its largest single-account term,
discarding the mutual cancellation among accounts that the mixture
retains.
\begin{prop} \label{prop:stitch}
    Grant the conditions of Corollary~\ref{corol:dm}. Then, 
    \[ \IP\big(\text{for all $t\geq 1$:} \sum_{s=1}^t \langle \xi_s, v_s\rangle \leq \St_{\alpha}(t;\mathbf{v})\big) \geq 1-\alpha,  \]
    where,
    \[ \St_{\alpha}(t;\mathbf{v}) := \inf_{\substack{k\ge 0\\ \lambda_k \in \Lambda_t(\mathbf{v})}} \frac{\log \frac{1}{\alpha p_k} + \sum_{s=1}^t \Gamma_{\psi}(\Delta V_s, \lambda_k v_s) }{\lambda_k},\ \Lambda_t(\mathbf{v}):= \Big[0, \lambda_{\max} \wedge  \frac{\lambda_{\max}}{\sup_{1\leq s \leq t} \|v_s\|}\Big).  \]
\end{prop}
For explicit choices of $\lambda_k$ and $p_k$, $\St_{\alpha}$ admits a closed-form expression or bound, eg. Corollary \ref{cor:closed-form}(ii) and (iii), thereby avoiding the numerical procedures typically required to compute mixture boundaries. In particular, in \S\ref{se:Lipschitz-sgd} and \S\ref{se:Lipschitz-stitching}, we discuss recovering the stitching constructions corresponding to \cite{aolaritei2025stopping} and \cite{aolaritei2026beyond}, which showcase how our general approach can lead to their boundaries as special cases. However, this ease of computation comes at a cost; 
due to the presence of the infimum in $\St_\alpha$, it follows  $\DMcal_\alpha(t;\bv) \le \St_\alpha(t;\bv)$ for the same support
points $\{\lambda_l\}$, weights $\{p_k\}$, and same \newname\  function $\psi$. On the other hand, both the mixture and stitching constructions above apply under the full
generality of Definition~\ref{def:psi}, accommodating sub-Gaussian,
sub-gamma, sub-Poisson, and sub-exponential tails alike.  When the
noise is additionally known to be bounded, however, a sharper
construction is available, as we describe next.
\subsection{Boundary by betting method}\label{app:betting}
Throughout this section, we assume bounded noise, specifically $|\xi_t|\le c$. Under this assumption, one may either take $\psi=\psi_N$ with $\lambda_{\max}=\infty$ and employ the normal-mixture or stitching boundary, or take $\psi=\psi_{E,c}$ with $\lambda_{\max}=1/c$ and use the different mixture or stitching boundaries. However, it turns out that a sub-exponential mixture boundary itself can be improved via the product-betting martingales of \citet[Proposition~2]{waudby-jrssb}; see also \citet[Theorem~1]{orabona2023tight} for related concentration bounds derived from portfolio wealth. The following theorem describes a general anytime-valid envelope constructed via betting strategies.
\begin{thm}[Betting boundary for bounded noise]\label{thm:betting-bounded-noise}
Let $(\xi_t)_{t\ge1}$ satisfy
$\IE_{t-1}\xi_t=0$ and $\|\xi_t\|\le c$.
Fix $(p_k)$ and $(\lambda_k)$ from Corollary \ref{corol:dm}.
Recall $\Lambda_t(\bv)$ from Theorem~\ref{thm:prob} with $\lambda_{\max}=1/c$.
Then
\begin{equation*}
\begin{split}
    &\mathcal B^0_\alpha(t;\bv):=\sup\bigg\{y\in\R:
   b_{\alpha}^0(y; t, \bv)
       <\frac1\alpha\bigg\}, \\
    &\text{where, } b_{\alpha}^0(y; t, \bv):=\sum_{\substack{k\ge0\\\lambda_k\in\Lambda_t(\bv)}}p_k
    \exp\bigg(\lambda_k y-
       \sum_{s=1}^t\psi_{E,1}(-\lambda_k\langle v_s,\xi_s\rangle)\bigg) 
\end{split}
\end{equation*}
satisfies
\[
    \IP\big(Y_t(\bv)\le\mathcal B^0_\alpha(t;\bv)
       \text{ for all }t\ge1\big)\ge1-\alpha.
\]
\end{thm}
We provide a succinct proof of this theorem in Appendix~\ref{app:se2proofs}. We remark that, despite its resemblance to the discrete mixture boundary in Corollary~\ref{corol:dm}, Theorem~\ref{thm:betting-bounded-noise} produces a sharper boundary. This improvement comes from the use of exact martingales via the product betting construction $K_t$ in \eqref{eq:naive-betting-martingale},
    while the exponentiated supermartingale used in mixing and stitching method may be thought of as introducing some slackness over an exact martingale property. The sharpness of $\mathcal B^0_\alpha(t;\bv)$
    can also be intuited from \eqref{eq:betting-quadratic-relaxation}, which yields $\psi_{E,1}(-\lambda\langle v_s,\xi_s\rangle)
    \le\Gamma_{\psi_{E,c}}(\xi_s\xi_s^\top,\lambda v_s).$ 
 Informally, $\mathcal{B}^0_\alpha$ employs $\psi_{E,1}(-\lambda\langle v_s,\xi_s\rangle)$ directly based on the realized noise projections, whereas the mixture and stitching constructions replace each such term by a uniform upper bound which depends on
$\xi_s$ solely through the variance proxy $\Delta V_s$. When the noise
is light-tailed relative to its range - that is, when $|\langle
v_s,\xi_s\rangle|$ is typically much smaller than its worst-case bound
$c\|v_s\|$ - this sharper control can yield a substantially
tighter envelope.  

\subsubsection{A sharper betting envelope}
In practice, often, the information provided is more than $\|\xi_t\|\le c$; instead, one often has $\xi_t = g_t - \IE_{t-1}g_t$, where $g_t$ are observables, and it may be known that $\|g_t\|\le c$. In this case, we can adapt Theorem~\ref{thm:betting-bounded-noise} to provide a sharper anytime-valid control than naively applying Theorem~\ref{thm:betting-bounded-noise} with $c$ replaced by $2c$. The following corollary can be established by considering the product betting martingale 
\begin{align}
    K_t^1(\bv, \lambda) = \prod_{\substack{1\le s\le t\\s<\tau(\lambda)}}
          \frac{1+\lambda\langle v_s,g_s\rangle}{1+\lambda\langle v_s,\IE_{s-1} g_s\rangle},
    \ K_0^1(\bv,\lambda)=1 \label{eq:new-betting-martingale}.
\end{align}
We explain the intuition behind this choice in detail while discussing the betting constructions in \S\ref{par:betting}.
\begin{cor} \label{corol:betting}
    Let $(g_t)_{t\ge1}$ be $\Fcal_t$-adapted such that $\|g_t\|\le c$ almost surely, where $c>0$ is
constant. Let $\xi_t:= g_t- \IE_{t-1}g_t$, and consider a predictable sequence $\bv=(v_s)_{s\ge1}$ . Fix $\alpha\in(0,1)$, $\{\lambda_k\}_{k\ge 0}\in(0,1/c)$ such that $\lambda_k \downarrow 0$, and deterministic weights $p_k>0$ with $\sum_{k\ge0}p_k=1$.
Recall $\Lambda_t(\bv)$ from Theorem~\ref{thm:prob} with $\lambda_{\max}=1/c$.
Then
\begin{equation*}
\begin{split}
    \mathcal B^1_\alpha(t;\bv)&:=\sup\bigg\{y\in\R:
   b_{\alpha}^1(y; t, \bv)
       <\frac1\alpha\bigg\}, \text{where, } \nonumber\\  b_{\alpha}^1(y; t, \bv)&:=\sum_{\substack{k\ge0\\ \lambda_k\in\Lambda_t(\bv)}}
p_k
\exp\Biggl[
    \lambda_k y
    -\sum_{s=1}^t
       \psi_{E,1}\!\bigl(-\lambda_k\langle v_s, g_s\rangle\bigr)
    +\sum_{s=1}^t
       \psi_{E,1}\!\bigl(-\lambda_k\langle v_s, \IE_{s-1}g_s\rangle\bigr)
\Biggr] 
\end{split}
\end{equation*}
satisfies
\[
    \IP\big(Y_t(\bv)\le\mathcal B^1_\alpha(t;\bv)
       \text{ for all }t\ge1\big)\ge1-\alpha.
\]
\end{cor}
Corollary \ref{corol:betting} can be made into an observable envelope, as discussed in Theorem \ref{thm:obs-betting}.

\subsection{Proofs of results in \S\ref{app:boundaries}}\label{app:boundaryproofs}
\begin{proof}[Proof of Lemma~\ref{lem:suff}]
For notational convenience, let \[\kappa_t (u, \lambda):= \exp(\lambda \langle u, \xi_t\rangle - \psi(\lambda) u^\top \Delta V_t u).\]
  If \eqref{eq:psi} holds, then, for a fixed $t$, and $u \in S^{d-1}$, choose $v_s = u \bo_{s=t}$. Clearly, $M_{t-1}(\bv, \lambda)=1$ implies that $$\IE_{t-1}[\kappa_t(u, \lambda)]= \IE_{t-1}[M_t(\bv, \lambda)]\le 1,$$
  where in the final inequality we use the supermartingale structure of $M_t$. This shows that $\prod_{s=1}^t \kappa_s(u, \lambda), \ K_0(u, \lambda)=1$ is a supermartingale, establishing Definition \ref{def:psi}.
  
  For the converse, let
 $$K_t(\theta)=\prod_{s=1}^t \exp\bigg(\sum_{s=1}^t  \Big( \langle \theta, \xi_s \rangle - \Gamma_{\psi}(\Delta V_s, \theta)  \Big) \bigg) , \ K_0(\theta)=1,$$ for each $\theta$ satisfying $\|\theta\|<\lambda_{\max}$. Proposition 7.20 of \cite{ramdas2025hypothesis} provides that $K_t(\theta)$ is a non-negative supermartingale. Moreover, $(K_t(\theta))_{\theta: \|\theta\|< \lambda_{\max}}$ is jointly measurable, and continuous in $\theta$; in particular, continuity at $\theta=0$ follows from 
  \[
0\le\Gamma_\psi(\Delta V_t,\theta)
\le\|\Delta V_t\|_{\text{op}}\psi(\|\theta\|)
\longrightarrow0
\ \text{as }\theta\to0.
\]
Consequently, an application of Lemma 2.4 of \cite{casgrain2024}, along with the observation that 
\[ \prod_{s=1}^t \int_{\theta: \|\theta\|<\lambda_{\max}}\frac{K_s(\theta)}{K_{s-1}(\theta)} \ \delta_{\lambda v_s}(\text{d}\theta) = \prod_{s=1}^t \kappa_s(\lambda v_s)= M_t(v, \lambda) \]
is a non-negative supermartingale, which completes the proof.
\end{proof}
\begin{proof}[Proof of Lemma~\ref{lem:subpsi-martingale-differences}]
We first prove that $\IE[|\xi_t|]<\infty$. Fix $t\ge1$ and $u\in\mathbb S^{d-1}$. 
Put $X(u)=\langle u,\xi_t\rangle$ and
$Y(u)=u^\top\Delta V_tu$. 
Choose a deterministic sequence
$\lambda_n\in(0,\lambda_{\max})$ with $\lambda_n\downarrow0$.
Definition \ref{def:psi}
gives, almost surely, for every $n\ge1$,
\begin{equation}
 \mathbb E_{t-1}
\Bigl[e^{\pm \lambda_nX(u)-\psi(\lambda_n)Y(u)}\Bigr]\le1, \label{eq:one-sided}
\end{equation}
which directly yields
\begin{align}
&\mathbb E_{t-1}\!\Biggl[
e^{-\psi(\lambda_n)Y(u)}\bigl(\frac{e^{\lambda_n X(u)}+ e^{-\lambda_n X(u)}}{2}-1\bigr)
\Biggr]\le
\mathbb E_{t-1}\!\Bigl[1-e^{-\psi(\lambda_n)Y(u)}\Bigr]
\le
\psi(\lambda_n)\mathbb E_{t-1}[Y(u)] \nonumber\\
\implies &  \mathbb E_{t-1}\!\Biggl[ \liminf_{n\to \infty}
e^{-\psi(\lambda_n)Y(u)} 2 \lambda_n^{-2}\bigl(\frac{e^{\lambda_n X(u)}+ e^{-\lambda_n X(u)}}{2}-1\bigr)
\Biggr] \overset{(a)}{\le} \liminf_{n\to \infty} 2 \lambda_n^{-2} \psi(\lambda_n)\mathbb E_{t-1}[Y(u)]\nonumber\\
\overset{(b)}{\implies} & \IE_{t-1}[X(u)^2] \le \psi''(0) \IE_{t-1}[Y(u)], \label{eq:subpsi-directional-second-moment}
\end{align}
where $(a)$ follows from Fatou's Lemma and $(b)$ employs $\psi(\lambda_n)\to 0$ and $2\psi(\lambda_n)/\lambda_n^2 \to \psi''(0)$.
Letting $u$ vary over the Euclidean basis vectors $(e_j)_{j=1}^d$, \eqref{eq:subpsi-directional-second-moment} leads to 
\[ \IE_{t-1}[\|\xi_t\|^2]\le \psi''(0) \IE_{t-1}[\Tr(\Delta V_t) ].\]
Therefore, since $\xi_t$ is square integrable, we can expand \eqref{eq:one-sided} to write
\begin{equation}
1 \pm \lambda_n \IE_{t-1} X(u) - \psi(\lambda_n) \IE_{t-1}Y(u) \le  \mathbb E_{t-1}
\Bigl[e^{\pm \lambda_nX(u)-\psi(\lambda_n)Y(u)}\Bigr]\le1, \nonumber
\end{equation}
from which it is immediate that
\[ \big | \IE_{t-1} X(u) \big|\le \frac{\psi(\lambda_n)}{\lambda_n} \IE_{t-1} Y(u) \to 0, \]
where the limiting assertion follows by the definition of \newname\ function. In view of $\xi_t$ being $\Fcal_{t}$-adapted, this establishes that $(\xi_t)$ constitutes a martingale difference sequence.
Finally,  $\mathbb E_{t-1}[\xi_t\xi_t^\top]
\preceq
\psi''(0)\,\mathbb E_{t-1}[\Delta V_t]$ follows trivially by noting that \eqref{eq:subpsi-directional-second-moment} holds for arbitrary $u\in \mathbb S^{d-1}$. This completes the proof.
\end{proof}

\begin{proof}[Proof of Theorem~\ref{thm:prob}]
    Fix $\lambda \in [0, \lambda_{\max})$, and let \[ \tau(\lambda) := \inf\{t\geq 1: \lambda \|v_t\| \geq  \lambda_{\max}\} ,\]
    with $\inf \emptyset:=\infty$ by convention. Denote $\Tilde{v}_s(\lambda):= v_s I\{s < \tau(\lambda)\}.$
    Then, by definition of stopping time $\tau(\lambda)$, $\Tilde{v}_s$ is predictable, and $\lambda \| \Tilde{v}_s(\lambda)\| < \lambda_{\max}$. Therefore, Definition~\ref{def:psi}, via \eqref{eq:psi} and Lemma \ref{lem:suff} entails that 
\[ M_n(\lambda):= \exp\bigg(\sum_{s=1}^n \big(\lambda \langle \Tilde{v}_s(\lambda), \xi_s \rangle - \Gamma_{\psi}( \Delta V_s,  \lambda \Tilde{v}_s(\lambda) )\big) \bigg), \]
is a supermartingale. Similar to Lemma 2 of \cite{howard2021time}, it follows that 
\[ M_n' := \int_0^{\lambda_{\max}} M_n(\lambda) \ \textnormal{d}F(\lambda) \]
is an $\Fcal_{n}$-adapted super-martingale\footnote{Here, we mention a measure-theoretic titbit- nominally, we require that $(M_t)$ is product-measurable when the underlying probability space is augmented with $F$.}. Therefore, Ville's inequality (Lemma 1, \cite{howard2018}) implies that
\begin{align}
    \IP(\mathcal{E}_{\alpha}) \leq \alpha, \ \mathcal{E}_{\alpha}:=\Big\{M_n' \geq 1/\alpha \ \text{for some $n\in \N$} \Big\}.  \label{eq:appl-of-Ville}
\end{align}
 Now, fix $t \geq 1$. Consequently, for $\lambda \in \big[0, \lambda_{\max} / \sup_{1\le s \le t} \|v_s\| \big),$ it follows $\tau(\lambda) > t$. Therefore, for all $s \leq t$, it holds that $v_s = \Tilde{v}_s(\lambda),$ which, in conjunction with \eqref{eq:appl-of-Ville} yields that
 \begin{align*}
     \IP\Bigg( \int_{\Lambda_t(\mathbf{v})} \exp\bigg( \sum_{s=1}^t\Big(\lambda \langle v_s, \xi_s \rangle - \Gamma_{\psi}(\Delta V_s, \lambda v_s)\Big)\bigg)  \ \textnormal{d}F(\lambda) \ge \frac{1}{\alpha} \Bigg) \le {\alpha}.
 \end{align*}
 The conclusion follows by definition of $\Mcal_{\alpha}(t; \mathbf{v}).$
\end{proof}
\begin{proof}[Proof of Theorem~\ref{thm:betting-bounded-noise}]
For $\lambda>0$, recall $\tau(\lambda)$ from the proof of Theorem~\ref{thm:prob}, and set
\begin{equation*}
    K_t^0(\bv,\lambda)
    =\prod_{\substack{1\le s\le t\\s<\tau(\lambda)}}
          (1+\lambda\langle v_s,\xi_s\rangle),
    \  K_0^0(\bv,\lambda)=1.
\end{equation*}
Note that $K_t^0(\bv,\lambda)$ is a nonnegative martingale. Since 
$|\lambda\langle v_s,\xi_s\rangle\ind{s<\tau(\lambda)}|\le 1$, therefore
conditional monotone convergence shows that
$K_t^0(\bv)=\sum_{k\ge0}p_kK_t^0(\bv,\lambda_k)$ is also a nonnegative
martingale, with initial value at most one. Finally, if
$\lambda_k\in\Lambda_t(\bv)$, then $\tau(\lambda_k)>t$, and
\[
    \log K_t^0(\bv,\lambda_k)
    =\lambda_kY_t(\bv)
      -\sum_{s=1}^t\psi_{E,1}(-\lambda_k\langle v_s,\xi_s\rangle).
\]
Thus, by definition, $b_{\alpha}^0(Y_t(\bv); t, \bv)\le K_t^0(\bv)$. Ville's inequality and monotonicity
in $y$ prove the assertion.
\end{proof}

\section{Optimization module for Stochastic gradient descent} \label{app:opt}
Having discussed the different constructions of the observable boundaries for $Y_t(\bv)$ in \S\ref{se:boundary-Yt} and \S\ref{app:boundaries}, this section sheds light onto the other component of a time-uniform envelope for an iterative algorithm-  namely, the \textit{optimization module}. Considering stochastic gradient descent as a canonical example of iterative algorithms, this section concerns itself with the derivation of the recursion \eqref{eq:key-recursion} under different assumptions, followed by their final conversion into anytime-valid boundaries through Theorems \ref{thm:obs-comp}-\ref{thm:obs-betting} and Lemma \ref{lem:observable-envelope}. 

To that end, recall the SGD recursion given by \eqref{eq: general SGD}. When discussing different assumptions, it is instructive to note that the assumptions often imposed on an SGD analysis can be broadly classified into two categories.
\begin{itemize}
    \item \textit{Growth} assumptions, which can be understood to impose a minimum curvature of the function $F$ around the local or global optima; and
    \item \textit{smoothness} assumptions, which ensures $F$ does not behave too erratically so that a reasonable initialization can be found.
\end{itemize}
We discuss many different assumptions belonging to each category in \S\ref{app:growth} and \S\ref{app:smoothness} respectively. Moreover, in Figure \ref{fig:objective-assumption-implications} we include a chart of the implications between the different assumptions, with detailed proofs deferred to \S\ref{app:implication-lemma-proofs}. In \S\ref{app:performance}, we provide the different performance measures, which we connect to the assumptions in \S\ref{app:opt-sgd-boundaries} to derive the recursions and finally, the corresponding time-uniform boundaries.
\subsection{Growth assumptions} \label{app:growth}
Let $A= \nabla^2 F(x^\star)$ denote the Hessian of the optimization problem \eqref{eq: general SGD}. A growth assumption typically posits that $F$ exhibits some notion of convexity around the local or global minima, so that once sufficiently close to the point, the SGD chain can ensure convergence. Formally, this translates to some lower bound on the growth of $\nabla F$. A standard example would be convexity. 
\begin{asmp}[\CC: Convexity]\label{ass:c}
    $F$ is convex in $\Xcal$:
    \[ F(x) - F(y) \leq \langle \nabla F(x), x-y \rangle, \ x,y\in \Xcal.  \]
\end{asmp}
Of course, much weaker conditions need to be imposed to ensure SGD analyses for non-convex objective functions, which have substantial applications in modern machine learning. In Table \ref{tab:growth}, we present a few more growth assumptions on the objective function commonly used in the analysis of SGD. Different analyses typically rely on different subsets of these assumptions.
\newcommand{\asmprow}[1]{\refstepcounter{asmp}\label{#1}\theasmp}
\begin{table}[ht]
\centering
\small
\renewcommand{\arraystretch}{1.35}
\begin{tabularx}{\linewidth}{@{}l l l >{\raggedright\arraybackslash}X@{}}
\toprule
Asm. & Name & Condition & Remarks \\
\midrule
\asmprow{ass:qc} & \QC$_\gamma$: Quasar-convexity
  & $F(x)-F^\star\le\gamma^{-1}\langle\nabla F(x),x-x^\star\rangle$
  & $\gamma\in(0,1]$, $x^\star$ unique; $\gamma=1$: star-convexity (\STC) \citep{hinder2020near} \\
   \midrule
\asmprow{ass:sc} & \SC: Strong convexity
  & $\langle\nabla F(x)-\nabla F(x'),x-x'\rangle\ge\mu\|x-x'\|^2$ & $\mu>0$\\
   \midrule
\asmprow{ass:rsi} & \RSI: Restricted secant
  & $\langle\nabla F(x), x-x^\star\rangle\ge\mu\|x-x^\star\|^2$
  & $\mu>0$; \citep{vaswani2019painless} \\
   \midrule
\asmprow{ass:lsc} & \LSC: Local self-concordance
  & $|h_{x,x'}'''(u)|\le C\|x-x'\|\,h_{x,x'}''(u)$
  & $F\in\mathcal C^3$, $h_{x,x'}(u):=F(x+u(x'-x))$, $u\in\R$ \citep{bach2010self} \\
   \midrule
\asmprow{ass:kl} & \KLS: Kurdyka--\L{}ojasiewicz
  & $\|\nabla F(x)\|\ge\varphi(F(x)-F^\star)$
  & $\varphi$ nondecreasing, $\varphi(0)=0<\varphi(s)$ for $s>0$; Example $\varphi(s)=\sqrt{\mu s}$: \PL\ \citep{scaman2022convergence,polyak1963gradient} \\
   \midrule
\asmprow{ass:diss} & \DISS: Dissipativity
  & $\langle x,\nabla F(x)\rangle\ge\mu\|x\|^2-b$ & $\mu, b>0$ \\
\bottomrule
\end{tabularx}
\caption{Growth assumptions; $F^\star=\inf_x F(x)$ when no unique global minimum exists.}
\label{tab:growth}
\end{table}
\subsection{Smoothness assumptions}\label{app:smoothness}
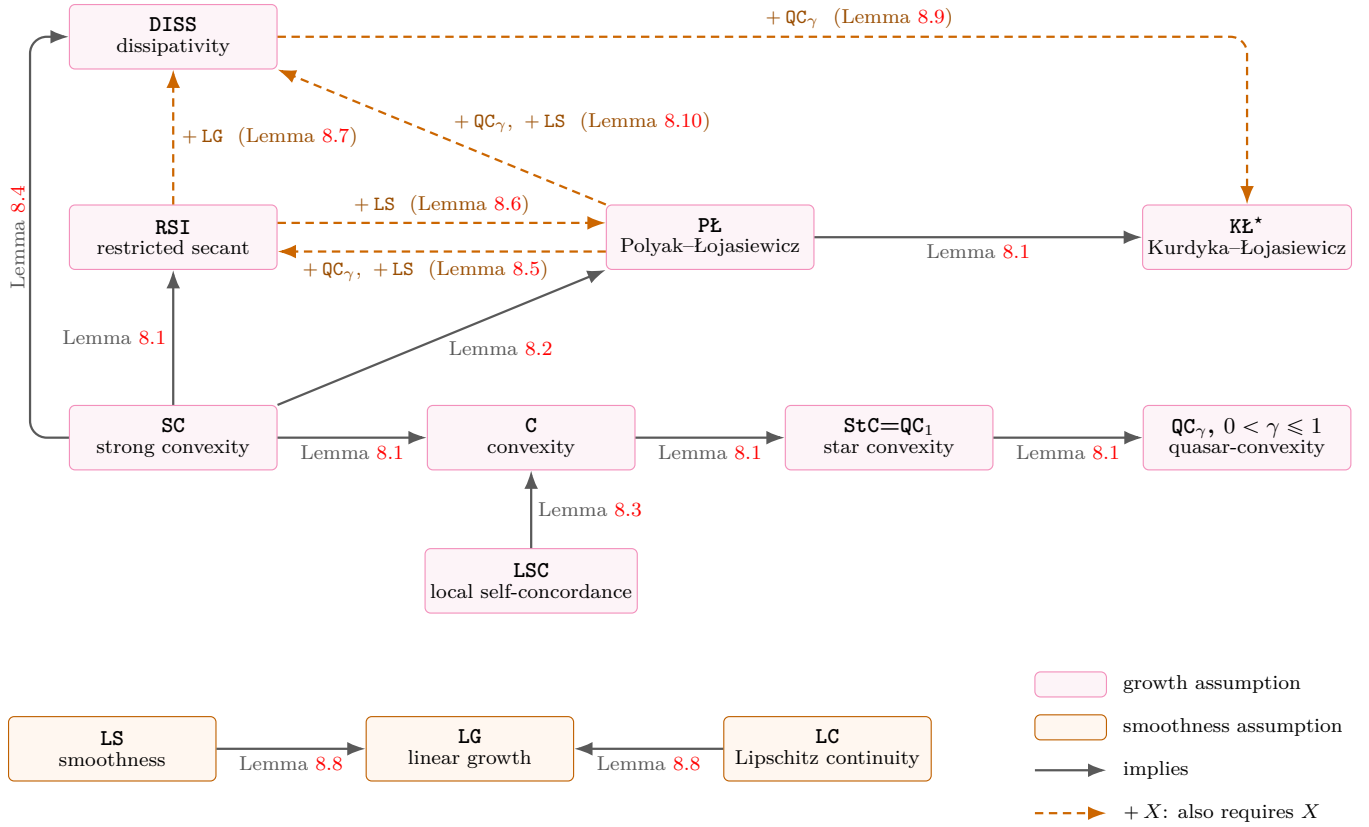
\begin{figure}[t]
\centering
\tikzset{
  box/.style        ={draw, rounded corners=2pt, align=center, inner sep=2pt,
                      minimum width=29mm, minimum height=9mm, font=\footnotesize},
  geometry/.style   ={box, draw=violet!55,      fill=violet!5},
  errorbound/.style ={box, draw=teal!70!black,  fill=teal!6},
  drift/.style      ={box, draw=magenta!55,     fill=magenta!5},
  regularity/.style ={box, draw=black!50,       fill=black!4},
  noise/.style      ={box, draw=orange!75!black,fill=orange!6},
  imp/.style        ={->, thick, black!65},
  cond/.style       ={->, thick, densely dashed, orange!80!black},
  lab/.style        ={font=\scriptsize, orange!60!black, fill=white,
                      inner sep=1.5pt, rounded corners=1pt},
  lem/.style        ={font=\scriptsize, black!65, fill=white,
                      inner sep=1pt, rounded corners=1pt},
  swatch/.style     ={minimum width=10mm, minimum height=3.5mm, inner sep=0pt},
}

\resizebox{\linewidth}{!}{%
\begin{tikzpicture}[>=Latex,scale=0.9,every node/.append style={transform shape}]
  \node[drift] (DISS) at (0,5.6)    {\bfseries \DISS\\[-1mm]\scriptsize dissipativity};
  \node[drift] (RSI)  at (0,2.8)    {\bfseries \RSI\\[-1mm]\scriptsize restricted secant};
  \node[drift] (PL)   at (7.5,2.8)  {\bfseries \PL\\[-1mm]\scriptsize Polyak--{\L}ojasiewicz};
  \node[drift] (KL)   at (15.0,2.8) {\bfseries \KLS\\[-1mm]\scriptsize Kurdyka--{\L}ojasiewicz};
  \node[drift] (SC)   at (0,0)      {\bfseries \SC\\[-1mm]\scriptsize strong convexity};
  \node[drift] (C)    at (5.0,0)    {\bfseries \CC\\[-1mm]\scriptsize convexity};
  \node[drift] (StC)  at (10.0,0)   {\bfseries \STC=\QC$_1$\\[-1mm]\scriptsize star convexity};
  \node[drift] (QC)   at (15.0,0)   {\bfseries \QC$_\gamma$, $0<\gamma\le1$\\[-1mm]\scriptsize quasar-convexity};
  \node[drift] (LSC)  at (5.0,-2.0) {\bfseries \LSC\\[-1mm]\scriptsize local self-concordance};

  \draw[imp] (SC)  -- node[lem,below=0.5mm]{Lemma~\ref{lem:imp-elementary}} (C);
  \draw[imp] (C)   -- node[lem,below=0.5mm]{Lemma~\ref{lem:imp-elementary}} (StC);
  \draw[imp] (StC) -- node[lem,below=0.5mm]{Lemma~\ref{lem:imp-elementary}} (QC);
  \draw[imp] (LSC) -- node[lem,right=0.5mm]{Lemma~\ref{lem:imp-lsc}} (C);
  \draw[imp] (SC)  -- node[lem,left=0.5mm]{Lemma~\ref{lem:imp-elementary}} (RSI);
  \draw[imp] (SC.north east) -- node[lem,below right=0mm and 0.5mm]{Lemma~\ref{lem:imp-sc-pl}} (PL.south west);
  \draw[imp] (PL)  -- node[lem,below=0.5mm]{Lemma~\ref{lem:imp-elementary}} (KL);
  \draw[imp, rounded corners=6pt] (SC.west) -- (-2.0,0) -- node[lem,sloped,above=0.3mm]{Lemma~\ref{lem:imp-sc-diss}} (-2.0,5.6) -- (DISS.west);
  \draw[cond] (RSI) -- node[lab,right=0.5mm]{$+\,$\LG\ \ (Lemma~\ref{lem:imp-rsi-diss})} (DISS);
  \draw[cond] ([yshift=2mm]RSI.east) -- node[lab,above=0.5mm]{$+\,$\LS\ \ (Lemma~\ref{lem:imp-rsi-pl})} ([yshift=2mm]PL.west);
  \draw[cond] ([yshift=-2mm]PL.west) -- node[lab,below=0.5mm,pos=0.55]{$+\,$\QC$_\gamma$,\, $+\,$\LS\ \ (Lemma~\ref{lem:imp-pl-rsi})} ([yshift=-2mm]RSI.east);
  \draw[cond] (PL.north west) -- node[lab,above right=0mm and 1mm]{$+\,$\QC$_\gamma$,\, $+\,$\LS\ \ (Lemma~\ref{lem:imp-pl-diss})} (DISS.south east);
  \draw[cond, rounded corners=6pt] (DISS.east) -- node[lab,above=0.5mm,pos=0.6]{$+\,$\QC$_\gamma$\ \ (Lemma~\ref{lem:imp-qc-diss-kl})} (15.0,5.6) -- (KL.north);
\end{tikzpicture}%
}

\vspace{4ex}

\resizebox{\linewidth}{!}{%
\begin{tikzpicture}[>=Latex,scale=0.9,every node/.append style={transform shape}]
  \node[noise] (LS)  at (0,-4.2)    {\bfseries \LS\\[-1mm]\scriptsize smoothness};
  \node[noise] (LC)  at (10.0,-4.2) {\bfseries \LC\\[-1mm]\scriptsize Lipschitz continuity};
  \node[noise] (LG)  at (5.0,-4.2)  {\bfseries \LG\\[-1mm]\scriptsize linear growth};

  \draw[imp] (LS) -- node[lem,below=0.5mm]{Lemma~\ref{lem:imp-lg}} (LG);
  \draw[imp] (LC) -- node[lem,below=0.5mm]{Lemma~\ref{lem:imp-lg}} (LG);

  \node[drift, swatch] at (13.4,-3.3) {};
  \node[font=\scriptsize, anchor=west] at (14.0,-3.3) {growth assumption};
  \node[noise, swatch] at (13.4,-3.9) {};
  \node[font=\scriptsize, anchor=west] at (14.0,-3.9) {smoothness assumption};
  \draw[imp]  (12.9,-4.5) -- (13.9,-4.5);
  \node[font=\scriptsize, anchor=west] at (14.0,-4.5) {implies};
  \draw[cond] (12.9,-5.1) -- (13.9,-5.1);
  \node[font=\scriptsize, anchor=west] at (14.0,-5.1) {$+\,X$: also requires $X$};
\end{tikzpicture}%
}

\caption{Universal implications among Assumptions~\ref{ass:c}--\ref{ass:LG} when $\mathcal X=\mathbb R^d$, $F\in\mathcal C^1$, and $x^\star$ is a minimizer of $F$ (unique wherever \QC$_\gamma$ is involved). A dashed arrow labelled $+X$ means the source assumption \emph{together with} $X$ implies the target. Each arrow is labelled with the lemma in Appendix~\S\ref{app:implication-lemma-proofs} that proves it.}
\label{fig:objective-assumption-implications}
\end{figure}

In contrast to the growth assumptions in \S\ref{app:growth}, smoothness assumptions ensure that $F$ possesses some sort of smoothness, so that a reasonable warm start can be guaranteed. This condition materializes through some upper-bound on the structure of $\nabla F(x)$ (vis-à-vis the lower bounds in the growth assumptions), which rules out pathological examples. Similar to Table \ref{tab:growth}, Table~\ref{tab:smooth} collects the regularity conditions on $F$.
\begin{table}[ht]
\centering
\small
\renewcommand{\arraystretch}{1.35}
\begin{tabularx}{\linewidth}{@{}l l l >{\raggedright\arraybackslash}X@{}}
\toprule
Asm. & Name & Condition & Remarks \\
\midrule
\multicolumn{4}{@{}l}{\emph{Regularity of $F$}} \\
\asmprow{ass:lsmooth} & \LC: Lipschitz continuity
  & $|F(x)-F(x')|\le L\|x-x'\|$ & $L>0$ \\
\asmprow{ass:lsmooth-grad} & \LS: $L$-smoothness
  & $\|\nabla F(x)-\nabla F(x')\|\le L\|x-x'\|$ &$L>0$ \\
\asmprow{ass:LG} & \LG: Linear growth
  & $\|\nabla F(x)\|\le L(1+\|x\|)$ &$L>0$ \\
\bottomrule
\end{tabularx}
\caption{Smoothness assumptions}
\label{tab:smooth}
\end{table}
We collect the various implications between Assumptions~\ref{ass:c}-\ref{ass:LG} in Figure~\ref{fig:objective-assumption-implications}. The corresponding results and proofs are present in \S\ref{app:implication-lemma-proofs}.

\subsection{Performance measures}\label{app:performance}
In the following, we also introduce the performance for which we seek to establish time-uniform control. We are primarily interested in the last iterate $x_t$; a separate analysis with Polyak-Ruppert averaged iterates is provided in \S\ref{se:pr}. Table \ref{tab:metrics} delineates the different metrics that we will henceforth consider. Note that the measures $Z_{t,4}$ and $Z_{t,5}$ were also analyzed by \cite{aolaritei2026beyond}.
\newcommand{\metricrow}[1]{\refstepcounter{metric}\label{#1}\themetric}
\begin{table}[ht]
\centering
\small
\renewcommand{\arraystretch}{1.4}
\begin{tabular}{@{}l l l l@{}}
\toprule
Metric & Measures & Last iterate $x_t$ \\
\midrule
\metricrow{metric-1} & Squared distance$^\dagger$
  & $Z_{t,1}:=\|x_t-x^\star\|^2$  \\
\metricrow{metric-2} & Suboptimality
  & $Z_{t,2}:=F(x_t)-F^\star$ \\
\metricrow{metric-3} & Gradient norm
  & $Z_{t,3}:=\|\nabla F(x_t)\|$  \\
\metricrow{metric-4} & Average suboptimality
  & \multicolumn{2}{l}{$Z_{t,4}:=S_t^{-1}\sum_{s=1}^t\eta_s\,\bigl(F(x_{s-1})-F^\star\bigr)$} \\
\metricrow{metric-5} & Average squared gradient norm
  & \multicolumn{2}{l}{$Z_{t,5}:=S_t^{-1}\sum_{s=1}^t\eta_s\,\|\nabla F(x_{s-1})\|^2$} \\
\bottomrule
\end{tabular}
\caption{Performance measures for SGD. Here $F^\star:=F(x^\star)$ and $S_t:=\sum_{s=1}^t\eta_s$. $^\dagger$Requires $x^\star$ to be unique.}
\label{tab:metrics}
\end{table}
Naturally, not every combination of assumptions and performance measure yields a recursion of the form \eqref{eq:key-recursion}. Each performance measure must instead be paired with structural assumptions that are compatible with it. In \S\ref{app:opt-sgd-boundaries}, we discuss many such different combinations of assumptions and performance measures, and elaborate on how to leverage Theorems \ref{thm:obs-comp}-\ref{thm:obs-betting} along with Lemma \ref{lem:observable-envelope} to provide different anytime-valid boundaries. 
\subsection{Proofs for implications between different growth and smoothness assumptions}\label{app:implication-lemma-proofs}
Throughout, $F\in\mathcal C^1(\R^d)$ has a minimizer $x^\star$, and we write $F^\star=F(x^\star)$, $h(x):=F(x)-F^\star$ and $r(x):=\|x-x^\star\|$; in particular, $\nabla F(x^\star)=0$. As in Table~\ref{tab:growth}, \QC$_\gamma$, and hence \STC$=$\QC$_1$, includes uniqueness of $x^\star$; we retain this convention wherever either condition is used. In particular, $\CC\Rightarrow\STC$ requires it, since convexity alone yields the star-convex inequality, but not uniqueness. We write $\SC_m$ for Assumption~\ref{ass:sc} with constant $m>0$, $\RSI_\mu$, $\LS_L$, $\LG_L$, $\LC_{L_0}$ and $\DISS_{a,b}$ for the corresponding rows of Tables~\ref{tab:growth} and \ref{tab:smooth}, and $\PL_\mu$ for $\|\nabla F(x)\|^2\ge\mu h(x)$, i.e., \KLS\ with $\varphi(s)=\sqrt{\mu s}$; thus the \PL\ parameter used here is twice that of \cite{karimi2016linear}. Finally, \KLS\ means precisely Assumption~\ref{ass:kl}; no integrability requirement on $1/\varphi$ is imposed.

\begin{lem}[Elementary implications]\label{lem:imp-elementary}
$\SC\Rightarrow\CC$, $\CC\Rightarrow\STC\Rightarrow\QC_\gamma$ ($0<\gamma\le1$), $\SC_m\Rightarrow\RSI_m$ and $\PL\Rightarrow\KLS$. If $\PL_\mu$ holds, one may take $\varphi(s)=\sqrt{\mu s}$.
\end{lem}
\begin{proof}
Integration along $x+u(y-x)$, $u\in[0,1]$, gives
\[
F(y)-F(x)-\langle\nabla F(x),y-x\rangle=\int_0^1\langle\nabla F(x+u(y-x))-\nabla F(x),y-x\rangle\,du\ge\frac m2\|y-x\|^2,
\]
and dropping the last term yields convexity. Setting $y=x^\star$ in the first-order inequality gives $h(x)\le\langle\nabla F(x),x-x^\star\rangle$, and $\gamma h(x)\le h(x)$ gives $\QC_\gamma$. Setting $x'=x^\star$ in Assumption~\ref{ass:sc} and using $\nabla F(x^\star)=0$ gives $\RSI_m$. The \PL\ implication follows by taking square roots.
\end{proof}

\begin{lem}[Appendix~B of \cite{karimi2016linear}]\label{lem:imp-sc-pl}
If $\SC_m$ holds, then $\|\nabla F(x)\|^2\ge2m\,h(x)$, $x\in\R^d$; that is, $\SC_m\Rightarrow\PL_{2m}$.
\end{lem}

\begin{lem}[$\LSC\Rightarrow\CC$]\label{lem:imp-lsc}
Assumption~\ref{ass:lsc} implies convexity.
\end{lem}
\begin{proof}
For $x\ne x'$, the inequality $|h_{x,x'}'''(u)|\le C\|x-x'\|h_{x,x'}''(u)$ with $C>0$ forces $h_{x,x'}''(u)\ge0$. Hence every restriction of $F$ to a line is convex.
\end{proof}

\begin{lem}[$\SC\Rightarrow\DISS$]\label{lem:imp-sc-diss}
If $\SC_m$ holds, then $\DISS_{a,b}$ holds with $a=m/2$ and any $b>0$ satisfying $b\ge\|\nabla F(0)\|^2/(2m)$.
\end{lem}
\begin{proof}
Strong monotonicity and Young's inequality give
$\langle x,\nabla F(x)\rangle\ge m\|x\|^2+\langle x,\nabla F(0)\rangle\ge\frac m2\|x\|^2-\frac{\|\nabla F(0)\|^2}{2m}$.
\end{proof}

\begin{lem}[$\PL+\QC_\gamma+\LS\Rightarrow\RSI$]\label{lem:imp-pl-rsi}
If $\PL_\mu$, $\QC_\gamma$ and \LS\ hold, then $\RSI_q$ holds with $q=\gamma\mu/4$.
\end{lem}
\begin{proof}
By $\nabla F(x^\star)=0$, quasar-convexity and Theorem~2 of \cite{karimi2016linear},
$\langle\nabla F(x)-\nabla F(x^\star),x-x^\star\rangle\ge\gamma h(x)\ge q\,r(x)^2$.
\end{proof}

\begin{lem}[$\RSI+\LS\Rightarrow\PL$]\label{lem:imp-rsi-pl}
If $\RSI_\mu$ and $\LS_L$ hold, then $\PL_{2\mu^2/L}$ holds.
\end{lem}
\begin{proof}
By $\nabla F(x^\star)=0$ and the Cauchy--Schwarz inequality, $\RSI_\mu$ gives $\mu r(x)^2\le\allowbreak\|\nabla F(x)\|r(x)$, i.e., $\|\nabla F(x)\|\ge\allowbreak\mu r(x)$. Moreover, \eqref{eq:descent} at $x^\star$ gives $h(x)\le\allowbreak\frac L2r(x)^2$. Therefore, $\|\nabla F(x)\|^2\ge\allowbreak\mu^2r(x)^2\ge\allowbreak\frac{2\mu^2}{L}h(x)$.
\end{proof}

\begin{lem}[$\RSI+\LG\Rightarrow\DISS$]\label{lem:imp-rsi-diss}
Suppose $\RSI_\mu$ and $\LG_L$ hold, and put $c=\|x^\star\|$. Then $\DISS_{a,b}$ holds with $a=\mu/4$ and any $b>0$ satisfying $b\ge\mu c^2+cL+c^2L^2/\mu$.
\end{lem}
\begin{proof}
By $\nabla F(x^\star)=0$, $\RSI_\mu$ and $\LG_L$,
\begin{align*}
\langle x,\nabla F(x)\rangle
&=\langle x-x^\star,\nabla F(x)\rangle+\langle x^\star,\nabla F(x)\rangle
\ge\mu r(x)^2-cL(1+\|x\|)\\
&\overset{(a)}{\ge}\frac\mu2\|x\|^2-\mu c^2-cL-cL\|x\|
\overset{(b)}{\ge}\frac\mu4\|x\|^2-\Big(\mu c^2+cL+\frac{c^2L^2}{\mu}\Big),
\end{align*}
where $(a)$ uses $\|x\|^2\le2r(x)^2+2c^2$, and $(b)$ follows from Young's inequality $cL\|x\|\le\frac\mu4\|x\|^2+\frac{c^2L^2}{\mu}$.
\end{proof}

\begin{lem}[$\LS\Rightarrow\LG$ and $\LC\Rightarrow\LG$]\label{lem:imp-lg}
Both \LS\ and \LC\ imply \LG.
\end{lem}
\begin{proof}
Under $\LS_L$, $\|\nabla F(x)\|\le L\|x\|+\|\nabla F(0)\|\le\max\{L,\|\nabla F(0)\|\}(1+\|x\|)$. Under $\LC_{L_0}$,  $|F(x+tu)-F(x)|\le L_0|t|$ immediately leads to $|\langle\nabla F(x),u\rangle|\le L_0$ for every unit vector $u$, which completes the proof.
\end{proof}
\begin{lem}[$\QC_\gamma+\DISS\Rightarrow\KLS$]\label{lem:imp-qc-diss-kl}
Suppose $\QC_\gamma$ and $\DISS_{a,b}$ hold. Choose $R>\allowbreak\max\{1,\sqrt{2b/a}\}$ and put $D=\allowbreak R+\allowbreak\|x^\star\|$. Then \KLS\ holds with
\[
\varphi(s)=\min\Big\{\frac{\gamma s}{D},\frac{aR}{2}\Big\}.
\]
\end{lem}
\begin{proof}
If $\|x\|\le R$, quasar-convexity gives $\gamma h(x)\le\|\nabla F(x)\|\|x-x^\star\|\le D\|\nabla F(x)\|$. If $\|x\|\ge R$, dissipativity gives
$\|\nabla F(x)\|\ge a\|x\|-\frac{b}{\|x\|}\ge\frac{aR}{2}$.
\end{proof}
\begin{lem}[$\PL+\QC_\gamma+\LS\Rightarrow\DISS$]\label{lem:imp-pl-diss}
Suppose $\PL_\mu$, $\QC_\gamma$ and $\LS_L$ hold. Set $q=\gamma\mu/4$ and $c=\|x^\star\|$. Then $\DISS_{a,b}$ holds with
\[
a=\frac q4, \  b=1+\Big(\frac q2+\frac{L^2}{2q}\Big)c^2.
\]
\end{lem}
\begin{proof}
Lemma~\ref{lem:imp-pl-rsi} gives $\langle\nabla F(x),x-x^\star\rangle\ge q\,r(x)^2$, while smoothness gives $\|\nabla F(x)\|\le L\,r(x)$. Thus
\[
\langle x,\nabla F(x)\rangle\ge q\,r(x)^2-Lc\,r(x)\overset{(a)}{\ge}\frac q2r(x)^2-\frac{L^2c^2}{2q}\overset{(b)}{\ge}\frac q4\|x\|^2-\Big(\frac q2+\frac{L^2}{2q}\Big)c^2,
\]
where $(a)$ follows from Young's inequality, and $(b)$ uses $\|x\|^2\le2r(x)^2+2c^2$.
\end{proof}

\section{From assumptions and performance measures to  envelopes}\label{app:opt-sgd-boundaries}
In this section, we analyze some of the different combinations of assumptions and performance measures that lead to the key recursion \eqref{eq:key-recursion} for some corresponding choices of $(a_t)_{t\ge 1}, (b_t)_{t\ge 1}$ and $(v_t)_{t\ge 1}$. To start off, we discuss the conditions of \S\ref{se:rsi} and \S\ref{se:Lipschitz-sgd}, for both of which we have discussed the derivation of \eqref{eq:key-recursion} with the corresponding measures $Z_{t,1}$, $Z_{t,2}$ and $Z_{t,5}$, and the subsequent betting constructions.  Here, we briefly discuss some stitching constructions through which we also recover the envelopes discussed in \cite{aolaritei2025stopping} and \cite{aolaritei2026beyond}.
\subsection{sub-Gamma stitching boundary for \RSI\ with $Z_{t,1}$}\label{se:sub-gamma-stitch}
Recall that the optimization module for \RSI\ is present in \S\ref{se:opt-rsi}, where we additionally assumed that $\|g_t\|\le G.$ Subsequently, we also discussed a betting-based boundary $R_t^{\mathrm{bet}}$, along with comparisons to exponential discrete mixture boundary $R_t^{\mathrm{dm}}$, exponential stitching boundary $R_t^{\St}$, and sub-gamma stitching boundary $\tilde R_t^{\St}$ in \S\ref{app:betting-sgd}; in particular, $R_t^{\mathrm{bet}}$ is strictly sharper than the stitching boundary $R_t^{\St}$ we are going to discuss in the following. Nonetheless, for the sake of completeness, we include a discussion here to highlight how our general theorems recover existing results simply through a direct choice of the parameters.

Observe that $(\xi_t,g_tg_t^\top)_{t\ge1}$ is a sub-$\psi_{E,G}$ process and $\lambda_{\max}=1/G$, and $\Delta V_t:=g_tg_t^\top\preceq\nu_tI$ with $\nu_t:=\|g_t\|^2$. In the notation of Theorem \ref{thm:obs-comp}, $K_t(\lambda):=\sum_{s=1}^t \|g_s\|^2 \psi_{\mathrm{G}, G}(\lambda M_s).$ Following Theorem \ref{thm:obs-comp}, define
\[
    U_t^\St:=\inf_{\substack{k\ge0\\ \lambda_k\in\Lambda_t(M)}}\frac{\log\frac{1}{\alpha p_k}+\sum_{s=1}^t\|g_s\|^2\psi_{\mathrm G,G}(\lambda_kM_s)}{\lambda_k},
\]
with $\Lambda_t(M)$ as in Theorem~\ref{thm:obs-comp}. Define $\tilde{R}_t^\St$ as in Lemma \ref{lem:observable-envelope}, and $M_t:={2\eta_t\sqrt{\Tilde R_{t-1}^{\St}}}/{A_t}$. Therefore, with $\wt\bv=(v_s/A_s)_{s\ge1}$, Theorem~\ref{thm:obs-comp} gives
\[
\IP\Big(\bigcup_{t\ge1}\big[\mathsf E_t(M)\cap\{Y_t(\wt\bv)>U_t^{\St}\}\big]\Big)\le\alpha,\ \mathsf E_t(M)=\bigcap_{s=1}^t\Big\{\Big\|\frac{v_s}{A_s}\Big\|\le M_s\Big\}.
\]
It remains to verify \eqref{eq:star}. On $\bigcap_{s=0}^{t-1}\{Z_{s,1}\le\Tilde R_s^{\St}\}$, for every $s\le t$,
\[
\Big\|\frac{v_s}{A_s}\Big\|=\frac{2\eta_s\sqrt{Z_{s-1,1}}}{A_s}\le\frac{2\eta_s\sqrt{\Tilde R_{s-1}^{\St}}}{A_s}=M_s,
\]
that is, $\mathsf E_t(M)$ holds, and there $\Mcal_\alpha(t;\wt\bv)=U_t^{\St}$. This is \eqref{eq:star}, and Lemma~\ref{lem:observable-envelope} yields 
$$\IP\big(Z_{t,1}\le\Tilde R_t^{\St}\text{ for all }t\ge0\big)\ge1-\alpha.$$
In particular, letting 
 \[ p_{k,\ell}=\frac{1}{(k+1)(k+2)(\ell+1)(\ell+2)}, \ \lambda_{k, \ell} = \frac{1}{2^{\ell} + \sqrt{2^{k-1}/ \log(1/(\alpha p_{k,\ell}))}}, \]
 we recover the empirical Bernstein boundary of Theorem 3.5 of \cite{aolaritei2026beyond}. Note that we do not compare with Theorem 2.3 therein, as their Theorem 3.5 already provides a sharpened boundary; however, even their Theorem 2.3 can be recovered via a sub-Gaussian stitching with the choices $p_k = (k+1)^{-1}(k+2)^{-1}$, and $\lambda_k=\sqrt{\log(1/\alpha p_k)/ 2^{k+1}}$. We emphasize that other choices of $\lambda_k$ and $p_k$ produces other anytime-valid boundaries. 
 \subsection{sub-Gaussian stitching boundary for \LS\ and $\tilde{Z}_{t,2}$} \label{se:Lipschitz-stitching}
 This section complements the self-normalized and betting constructions in \S\ref{se:ls-betting} and \S\ref{se:betting-combn} by providing a sub-Gaussian stitching construction. Along the way, much akin to \S\ref{se:sub-gamma-stitch}, we recover results from \cite{aolaritei2025stopping} while also providing a class of alternative results which can be sharper than \cite{aolaritei2025stopping} even under weaker conditions.
 
Recall \eqref{eq:ls-moment-assumption}. Moreover, under bounded gradients, $v_t=\eta_t\nabla F(x_{t-1})$ satisfies $\|v_t\|\le M_t:=\eta_tG$. Clearly, with this choice of $M$, $\mathsf E_t(M)$ of Theorem~\ref{thm:obs-comp} holds surely for every $t$.
 In particular, Corollary \ref{cor:closed-form}(ii) along with Lemma \ref{lem:observable-envelope} recovers the observable boundary in Corollary 2.6 of \cite{aolaritei2025stopping}:
\begin{align}
\bar{R}_t^{\mathrm{AJ}} &:= \bar{R}_{0}^{\mathrm{AJ}} + L \sum_{s=1}^t \eta_s^2\|g_s\|^2/2 + \min\Big\{ 4\mathfrak S_\alpha\Big(\sigma^2G^2\sum_{s=1}^t\eta_s^2\Big), 7 \mathfrak S_\alpha\Big(4^{-1}\sigma^2G^2\sum_{s=1}^t\eta_s^2\Big)\Big\}. \nonumber
\end{align} 
In the following, we recover an alternate bound without the deterministic control on the gradient. In fact, we observe that from \eqref{eq:descent} it follows directly that
\begin{align*}
    \|v_t\|^2 = \eta_t^2 \| \nabla F(x_{t-1})\|^2 \le 2 \eta_t^2 L(F(x_{t-1}) - F^\star)\le 2\eta_t^2 L Z_{t-1,2},
\end{align*}
and consequently, one can recursively define
\begin{align*}
     R_t^\dagger &:= R_{0}^\dagger + L \sum_{s=1}^t \eta_s^2\|g_s\|^2/2 + \min \Big\{4\mathfrak S_\alpha\Big(2L\sigma^2  \sum_{s=1}^t\eta_s^2R_{s-1}^\dagger\Big), 7\mathfrak S_\alpha\Big(2^{-1}L\sigma^2  \sum_{s=1}^t\eta_s^2R_{s-1}^\dagger\Big) \Big\}. \nonumber
\end{align*}
Clearly, $R_t^\dagger$ is also anytime-valid via Corollary \ref{cor:closed-form} and Lemma \ref{lem:observable-envelope}. In particular, this does not require gradients to be bounded. In general, $R_t^\dagger$ and  $\bar{R}_t^{\mathrm{AJ}}$ are not directly comparable; however, $R_t^\dagger$ can be sharper than the boundary $\bar{R}_t^{\mathrm{AJ}}$, which does not use the information of the chain's position in controlling $\|v_t\|^2$. On the other hand, as previously discussed in \S\ref{app:betting-sgd}, a provably sharper boundary can be obtained by maintaining the choices of $p_k$ and $\lambda_k$, and invoking the discrete mixture boundary, Corollary~\ref{corol:dm}.

\subsection{Boundaries under \LS\ and \KLS}
\label{se:kl-sgd}
Much like the combination of growth and smoothness assumptions (\SC\ and \LS\ respectively), in this section we pair \LS\ with the Kurdyka–Łojasiewicz or \KLS\ condition. Concretely, suppose that $F$ attains $F^\star=F(x^\star)$ and satisfies the gap inequality of Assumption~\ref{ass:kl}, written as
\begin{equation}
\label{eq:kl-gap-condition}
    \|\nabla F(x)\|\ge\varphi(F(x)-F^\star),\  x\in\mathbb R^d,
\end{equation}
where the known $\varphi:[0,\infty)\to[0,\infty)$ is nondecreasing with $\varphi(0)=0$ and $\varphi(u)>0$ for $u>0$. This includes the celebrated Polyak--\L{}ojasiewicz condition $\varphi(u)=\sqrt{\mu u}$ \citep{polyak1963gradient} and the power-type conditions of \citet{scaman2022convergence}. Interestingly, this modest growth assumption enables us to control the performance measure $Z_{t,2}$ directly.

\subsubsection{Optimization module}
As is the norm throughout the examples in this paper, we start off by deriving a nonlinear recursion. Applying \eqref{eq:descent} with $x=x_{t-1}$ and $y=x_t=x_{t-1}-\eta_tg_t$, and substituting $g_t=\nabla F(x_{t-1})-\xi_t$,
\[
    Z_{t,2}\le Z_{t-1,2}-\eta_t\|\nabla F(x_{t-1})\|^2+b_t+\langle v_t,\xi_t\rangle,\  b_t=\frac{L\eta_t^2\|g_t\|^2}{2}, \  v_t=\eta_t\nabla F(x_{t-1}),
\]
which, in conjunction with \eqref{eq:kl-gap-condition}, provides
\begin{equation}
\label{eq:kl-nonlinear-recursion}
    Z_{t,2}\le Z_{t-1,2}-\eta_t\varphi(Z_{t-1,2})^2+b_t+\langle v_t,\xi_t\rangle.
\end{equation}
Let $R_{\max}\in(0,\infty]$ be a known constant with $Z_{t,2}\le R_{\max}$ almost surely, and put $\bar R_{t-1}:=R_{t-1}\wedge R_{\max}$. Suppose $a_t\in(0,1]$ is predictable and observable, and define
\begin{equation}
\label{eq:kl-affine-majorant}
    \beta_t:=\sup_{0\le r\le\bar R_{t-1}}\big\{(1-a_t)r-\eta_t\varphi(r)^2\big\}\in\big[0,(1-a_t)\bar R_{t-1}\big].
\end{equation}
On $\{Z_{t-1,2}\le R_{t-1}\}$, we have $Z_{t-1, 2}\le\bar R_{t-1}$, whence $Z_{t-1,2}-\eta_t\varphi(Z_{t-1,2})^2\le a_tZ_{t-1}+\beta_t$, and \eqref{eq:kl-nonlinear-recursion} yields \eqref{eq:key-recursion} on $\{Z_{t-1,2 }\le R_{t-1}\}$.

\subsubsection{General boundary construction}
 Applying \eqref{eq:descent} with $y=x-\nabla F(x)/L$ gives $\|\nabla F(x)\|^2\le2L(F(x)-F^\star)$. Hence, on $\{Z_{t-1,2}\le R_{t-1}\}$,
\[
    \frac{\|v_t\|}{A_t}\le M_t:=\frac{\eta_t\sqrt{2L\bar R_{t-1}}}{A_t};
\]
if, in addition, $\|\nabla F\|\le G$, one may take $M_t:=\eta_t\min\{G,\sqrt{2L\bar R_{t-1}}\}/A_t$. With these choices, Lemma \ref{lem:observable-envelope} provides a general anytime-valid envelope
\[
    R_t:=A_t\Big(R_0+\sum_{s=1}^t\frac{b_s+\beta_s}{A_s}+U_t\Big),
\]
where $U_t$ can be replaced by the corresponding mixing, stitching or betting construction. For the conditionally sub-Gaussian noise of \S\ref{se:Lipschitz-stitching}, $\nu_t=\sigma^2$, and Corollary~\ref{cor:closed-form} gives the closed forms $U_t=\Ncal_\alpha(\sigma^2\sum_{s=1}^tM_s^2)$ and $U_t=4\mathfrak S_\alpha(\sigma^2\sum_{s=1}^tM_s^2)$; under \eqref{eq:ls-moment-assumption} and $\|\nabla F\|\le G$, the self-normalized proxy $\nu_t=\frac13(G+\|g_t\|)^2+\frac23\sigma^2$ of \S\ref{se:ls-betting} applies verbatim.
\subsubsection{Computation of $\beta_t$}
For any predictable grid $0=r_0<\ldots<r_m=\bar R_{t-1}<\infty$, $1-a_t\ge0$ and monotonicity of $\varphi$ give $(1-a_t)r-\eta_t\varphi(r)^2\le(1-a_t)r_{j+1}-\eta_t\varphi(r_j)^2$ for $r\in[r_j,r_{j+1}]$. Therefore, we can conclude
\[
    \beta_t\le\max\Big\{0,\max_{0\le j<m}\big[(1-a_t)r_{j+1}-\eta_t\varphi(r_j)^2\big]\Big\};
\]
Additionally, for two canonical choices of $\varphi$, $\beta_t$ is available in closed form.
\begin{itemize}
\item \emph{Polyak--\L{}ojasiewicz.} For $\varphi(u)=\sqrt{\mu u}$ and $\mu\eta_t<1$, the choice $a_t=1-\mu\eta_t$ gives $\beta_t=0$, recovering the linear contraction of the optimality gap. If, in addition, $\|\nabla F\|\le G$, then \eqref{eq:kl-gap-condition} gives $Z_t\le G^2/\mu$, so that one may take $R_{\max}=G^2/\mu$. Consequently, Corollary \ref{cor:closed-form} can be applied to provide, for example, the normal mixture boundary:
\[
    R_t=A_t\Big(R_0+\frac L2\sum_{s=1}^t\frac{\eta_s^2\|g_s\|^2}{A_s}+\Ncal_\alpha\Big(2L\sigma^2\sum_{s=1}^t\frac{\eta_s^2\bar R_{s-1}}{A_s^2}\Big)\Big),\  A_t=\prod_{s=1}^t(1-\mu\eta_s).
\]
\item \emph{Power-type K\L{}.} For $\varphi(u)=cu^\theta$, where $c>0$ and $1/2<\theta<1$, set $q=2\theta\in(1,2)$ and choose a predictable $r_t\in[0,\bar R_{t-1}]$ with $q\eta_tc^2r_t^{q-1}<1.$ Then
\begin{equation}
    a_t=1-q\eta_tc^2r_t^{q-1},\  \beta_t=(q-1)\eta_tc^2r_t^q \nonumber
\end{equation}
also provide a valid choice satisfying \eqref{eq:kl-affine-majorant}.
\end{itemize}

Moving on, we discuss the SGD analysis under \CC, which is the setting of \cite{aolaritei2025stopping} Corollary 2.1. As with the other examples, in this case as well we are able to recover and sharpen the existing results in the literature.
\subsection{Boundaries under \CC}\label{app:convex-sgd}
In this section, we grant only Assumption~\ref{ass:c} on $F$; the corresponding performance measure turns out to be the weighted suboptimality $Z_{t,4}$. We also work with projected SGD over a closed convex set $\Xcal$. In the sequel, in \S\ref{app:convex-opt}, we reduce the problem to \eqref{eq:key-recursion} with $a_t=1$, and feed the envelope back into Theorems \ref{thm:obs-comp}-\ref{thm:obs-betting}, and Lemma~\ref{lem:observable-envelope}. Subsequently, we discuss the stitching constructions in \S\ref{app:convex-stitching}, along with a hybrid construction in \S\ref{app:convex-hybrid}.

\subsubsection{Optimization module under \CC }\label{app:convex-opt}
 
Let $\Xcal\subseteq\R^d$ be nonempty, closed and convex, let $F$ be differentiable and satisfy Assumption~\ref{ass:c}, and fix $x^\star\in\argmin_{x\in\Xcal}F(x)$, $F^\star:=F(x^\star)$. Consider \eqref{eq: general SGD} with initialization $x_0\in\Xcal$, and $\eta_t>0$ is predictable and observable. As in \S\ref{se: anytime-valid-sgd}, $\xi_t=\nabla F(x_{t-1})-g_t$. We assume $\IE_{t-1}\xi_t=0$. Consider 
\[
Z_t^{\CC}:=Z_{t,1}+2S_tZ_{t,4},\ t\ge0,
\]
so that $Z_0^{\CC}=Z_{0,1}$. Lemma 4.1 of \cite{aolaritei2025stopping} yields
\begin{equation}\label{eq:convex-rec}
Z_t^{\CC}\le Z_{t-1}^{\CC}+\eta_t^2\|g_t\|^2+\langle2\eta_t(x_{t-1}-x^\star),\xi_t\rangle,
\end{equation}
which is \eqref{eq:key-recursion} with $a_t=1$, $b_t=\eta_t^2\|g_t\|^2$ and $v_t=2\eta_t(x_{t-1}-x^\star)$. Compared with \eqref{eq:stronglyconvex-recursion}, the contraction $1-2\mu\eta_t$ available under \RSI\ is replaced by the accumulated suboptimality inside $Z_t^{\CC}$. Consequently, $A_t=1$ and $\wt\bv=\bv$ in Lemma~\ref{lem:observable-envelope}.

\subsubsection{Stitching boundaries under \CC} \label{app:convex-stitching}
 Assume that $\|x-x^\star\|\le R_x<\infty$ for all $x\in\Xcal$. Assume that in Definition \ref{def:psi}, $\psi=\psi_N$ and $\Delta V_s = \sigma^2I$. Subsequently, Corollary \ref{cor:closed-form} asserts $\IP\big(Z_{t}^{\CC}\le R_t^{\mathrm{AJ}} \text{ for all }t\ge1\big)\ge1-\alpha$, where
\begin{equation}
R_t^{\mathrm{AJ}}:=R_x^2+\sum_{s=1}^t\eta_s^2\|g_s\|^2+\min \Big\{4 \mathfrak S_\alpha\Big(4\sigma^2R_x^2\sum_{s=1}^t\eta_s^2\Big), 7 \mathfrak S_\alpha\Big(\sigma^2R_x^2\sum_{s=1}^t\eta_s^2\Big) \Big\}.\label{eq:stitching-AJ}
\end{equation}
In particular, \eqref{eq:stitching-AJ} recovers the boundary in Corollary~2.2 of \citet{aolaritei2025stopping} (which, at the cost of being pedantic, uses only the term $7 \mathfrak S_\alpha(\sigma^2R_x^2\sum_{s=1}^t\eta_s^2)$). 
Via the discrete-mixture with the same choices of $p_k$ and, $\lambda_k$ as in Corollary \ref{cor:closed-form}(ii), one can provide an even sharper envelope compared to $\tilde{R}_t$.

Now suppose we assume additionally that $\|g_t\|\le G$ almost surely for a known $G>0$, as in \S\ref{se:rsi}. Then, as observed in \S\ref{se:sub-gamma-stitch}, $(\xi_t,g_tg_t^\top)_{t\ge0}$ is a sub-$\psi_{E,G}$ process and $\lambda_{\max}=1/G$. Hence, Corollary~\ref{cor:closed-form}(iii) gives the closed-form envelope
\[
R_t^{\Gamma\mathrm S}:=R_x^2+\sum_{s=1}^t\eta_s^2\|g_s\|^2 + \mathfrak G_\alpha\Big(4\sum_{s=1}^t\|g_s\|^2\eta_s^2R_x^2\Big).
\]
 We emphasize that other choices of $\lambda_k$ and $p_k$, e.g., the double-indexed grid of \S\ref{se:sub-gamma-stitch}, produce other anytime-valid boundaries.

\subsubsection{Combining the sub-Gaussian and the betting envelope}\label{app:convex-hybrid}
In this section, much like \S\ref{se:betting-combn}, we provide an envelope that combines the benefits of the sub-Gaussian stitching of \S\ref{app:convex-stitching} and the betting construction of \S\ref{app:betting}. Let $(p_k, \lambda_k)_{k \ge 1}$ be as in Corollary \ref{cor:closed-form}, and recall $\Lambda_t^G(M)$ from Theorem~\ref{thm:obs-betting} and $J_t$ from \eqref{eq:jt} for $M_s=2\eta_s R_x$. For $\lambda_k\in(0,1)$, define
\begin{equation}\label{eq:convex-hybrid-W}
W_t^{\mathrm H}(y):=\frac34\sum_{k\ge0}p_k\exp\Big(\lambda_ky-\frac{\lambda_k^2\sigma^2}{2}\sum_{s=1}^tM_s^2\Big)+\frac14\sum_{k:\,\lambda_k/G\in\Lambda_t^G(M)}p_k\exp\Big(\frac{\lambda_ky}{G}-J_t(\lambda_k/G)\Big),
\end{equation}
$U_t^{\mathrm H}:=\sup\{y\in\R:W_t^{\mathrm H}(y)<1/\alpha\}$ and $R_t^{\mathrm H}:=R_x^2+\sum_{s=1}^t\eta_s^2\|g_s\|^2+U_t^{\mathrm H}$. The following result proves the validity of $R_t^{\mathrm H}$ as a time-uniform envelope, and ascertains the improvement over $\tilde R_t$.
\begin{prop}[Hybrid envelope]\label{prop:convex-hybrid}
Fix $\alpha\in (0,1)$. Grant the notation and conditions of \S\ref{app:convex-stitching}, and let $\|g_t\|\le G$ almost surely. Then $R_t^{\mathrm{H}}$  satisfies $\IP(Z_t^{\CC}\le R_t^{\mathrm{H}} \text{ for all $t\ge 1$})\ge 1-\alpha$. Moreover, $R_t^{\mathrm H}\le \tilde{R}_t,\ t\ge1.$
\end{prop}

The preceding examples share a common pattern: once a growth assumption (sometimes along with a smoothness assumption) is matched with the performance measure it naturally controls, the derivation of an anytime-valid envelope proceeds through two broad steps: the reduction to \eqref{eq:key-recursion}, and obtaining an observable control through Theorems \ref{thm:obs-comp} or \ref{thm:obs-betting}. In the sequel, we carry out this reduction for the remaining assumptions of Tables~\ref{tab:growth} and \ref{tab:smooth}, which also illustrates how the assumptions relate to the performance measures.

\subsection{Optimization modules for other assumptions}\label{app:opt-modules}
 In this section, we focus on the assumptions not discussed until now. In several cases, we will recall \SC\ or \LS\ along with some other assumptions; the additional structure often permits a sharper recursion over \S\ref{se:rsi} or \S\ref{se:Lipschitz-sgd}, and we point out the source of each improvement along the way. In general, let $U_t$ denote a generic observable boundary from Theorems~\ref{thm:obs-comp}--\ref{thm:obs-betting}; the final observable boundaries $R_t$ will always be defined as $R_t=A_t(R_0+\sum_{s=1}^tb_s/A_s+U_t)$, with $A_t=\prod_{s=1}^ta_s$ following from Lemma \ref{lem:observable-envelope}. With this standardization, we go over several different assumptions in the following sections, where we first provide the optimization module, followed by the choices of $M$ which enables the applications of Theorem \ref{thm:obs-comp} or \ref{thm:obs-betting}.

\subsubsection{Analysis under \QC}\label{app:qc-module}
Convexity is often too stringent for the objectives of modern machine learning, whereas \RSI\ still insists on a quadratic growth of $F$ around $x^\star$. Quasar-convexity or \QC\ (Assumption~\ref{ass:qc}) sits between the two: it only asks the negative gradient to point towards $x^\star$:
\[F(x)-F^\star\le\gamma^{-1}\langle\nabla F(x),x-x^\star\rangle.\]
This assumption accommodates nonconvex objectives arising, e.g., in the identification of linear dynamical systems; see \citet{hinder2020near} and the references therein. Similar to \S\ref{app:convex-opt}, the natural performance measure is $Z_{t,4}$, which we track through the potential
\[
    Z_t^{\QC}:=Z_{t,1}+2\gamma S_tZ_{t,4},\ Z_0^{\QC}=Z_{0,1}, \ \text{where } S_t=\sum_{s=1}^t \eta_s.
\]
Note that $Z_t^{\QC}=Z_t^{\CC}$ if $\gamma=1$.
By definition of \QC\, it follows that
\begin{equation}\label{eq:qc-rec}
    Z_t^{\QC}\le Z_{t-1}^{\QC}+\eta_t^2\|g_t\|^2+\langle2\eta_t(x_{t-1}- x^\star),\xi_t\rangle,
\end{equation}
which is \eqref{eq:key-recursion} with $a_t=1$, $b_t=\eta_t^2\|g_t\|^2$ and $v_t=2\eta_te_{t-1}$. Moreover, since $Z_{t-1,1}\le Z_{t-1}^{\QC}$, on $\{Z_{t-1}^{\QC}\le R_{t-1}\}$,
\[
    \|v_t\|\le M_t:=2\eta_t\sqrt{R_{t-1}}.
\]
Therefore, pairing assumptions on $\psi$ along with choices on $\Delta V_t,$ Lemma~\ref{lem:observable-envelope}  yields that $Z_{t,4}\le R_t/(2\gamma S_t)$ holds simultaneously over $t\ge 1$ with probability $\ge 1-\alpha$.   Finally, under \SC, the inequality $\langle\nabla F(x_{t-1}),e_{t-1}\rangle\ge F(x_{t-1})-F^\star+\mu Z_{t-1,1}/2$ yields \eqref{eq:qc-rec} for the larger potential $Z_{t,1}+2S_tZ_{t,4}+\mu\sum_{s=1}^t\eta_sZ_{s-1,1}$, with the same $(b_t,v_t,M_t)$.

\subsubsection{Convexity and Lipschitz gradient}\label{app:cls-module}
Similar to the pairing of \SC\ and \LS\ in \S\ref{se:pr}, in this section we pair \CC\ with \LS, which is another common setting for smooth convex stochastic optimization \citep{moulines2011non,bottou2018optimization,lan2020first}. \CC\ and \LS\ together yield the interpolation inequality \citep[Theorem~4]{taylor2017smooth}
\begin{align}
    \langle\nabla F(x),x-x^\star\rangle\ge F(x)-F^\star+\frac1{2L}\|\nabla F(x)\|^2, \  x\in\R^d, \label{eq:interpolation}
\end{align}
As before, we first derive the optimization  module; then, under bounded stochastic gradients, we discuss a betting construction.

\paragraph{Optimization module.} Assume \CC, \LS\ and $0\le\eta_t\le1/L$. Recall $Z_t^{\CC}$ from \S\ref{app:convex-opt}, and let $e_t=x_t-x^\star$. Consider the representation
\begin{align}
    Z_{t,1}&=Z_{t-1,1}-2\eta_t\langle\nabla F(x_{t-1}),e_{t-1}\rangle+\eta_t^2\|g_t\|^2+\langle2\eta_te_{t-1},\xi_t\rangle\nonumber\\
    \overset{(a)}{\implies}Z_t^{\CC}&\le Z_{t-1}^{\CC}+\eta_t^2\|g_t\|^2-\frac{\eta_t}{L}\|\nabla F(x_{t-1})\|^2+\langle2\eta_te_{t-1},\xi_t\rangle,\label{eq:cls-rec}
\end{align}
where $(a)$ employs \eqref{eq:interpolation}. Finally, consider 
\[
    Z_t^{\CC,\LS}:=Z_t^{\CC}+\frac{S_t}LZ_{t,5}=Z_{t,1}+\sum_{s=1}^t\eta_s\Big(2\big(F(x_{s-1})-F^\star\big)+\frac1L\|\nabla F(x_{s-1})\|^2\Big),\ Z_0^{\CC,\LS}=Z_{0,1}.
\]
Then \eqref{eq:cls-rec} yields \eqref{eq:key-recursion} for $Z_t^{\CC,\LS}$ with $a_t=1$, $b_t=\eta_t^2\|g_t\|^2$ and $v_t=2\eta_te_{t-1}$. Note that, in contrast to the recursion in \S\ref{se:ls-opt}, adding \LS\ to \CC\ controls the $Z_{t,5}$ at no additional cost.
\paragraph{Bounded gradients and betting constructions.} Suppose now that $\|g_t\|\le G$ almost surely for a known $G>0$, as in \S\ref{se:rsi}.   Moreover, since $Z_{t-1,1}\le Z_{t-1}^{\CC,\LS}$, on $\{Z_{t-1}^{\CC,\LS}\le R_{t-1}\}$,
\[
    \|v_t\|\le M_t:=2\eta_t\sqrt{R_{t-1}},\ \text{and, } \|e_t\|\le\|e_{t-1}\|+\eta_t\|g_t\|\le \sqrt{R_{t-1}}+\eta_t\|g_t\|.
\]
With this choice, from Theorem \ref{thm:obs-betting} and Lemma \ref{lem:observable-envelope}, it is straightforward to derive the following betting construction.
\begin{prop}[Betting envelope under \CC\ and \LS]\label{prop:cls-betting}
Assume \CC, \LS\ and $\|g_t\|\le G$ almost surely. Fix $\alpha\in(0,1)$, an observable $R_0\ge Z_{0,1}$, and deterministic $\lambda_k\in(0,1/G)$ and $p_k>0$, $k\ge0$, such that $\lambda_k\downarrow0$ and $\sum_{k\ge0}p_k=1$. With $M_t$ as above, define $\Lambda_t(M)$ as in Theorem \ref{thm:obs-betting}. Define recursively, for $\lambda\in \Lambda_t(M)$ and $z\ge 0$,
\begin{equation}\label{eq:cls-betting}
\begin{aligned}
    J_t(\lambda)&:=\sum_{s=1}^t\psi_{E,1}\big(\lambda M_s\|g_s\|\big), \  H_t(\lambda,z):=t\,\psi_{E,1}\Big(\frac{\lambda}{t}\big(z-(\sqrt{R_{t-1}}+\eta_t\|g_t\|)^2\big)_+\Big),\\
    W_t(z)&:=\sum_{\substack{k\ge0\\\lambda_k\in\Lambda_t}}p_k\exp\bigl[\lambda_k(z-D_t)-J_t(\lambda_k)+H_t(\lambda_k,z)\bigr], \  R_t:=\sup\big\{z\ge0:W_t(z)\le1/\alpha\big\},
\end{aligned}
\end{equation}
where $D_t=R_0+\sum_{s=1}^t\eta_s^2|g_s|^2$.
Then $\IP\big(Z_t^{\CC,\LS}\le R_t\text{ for all }t\ge0\big)\ge1-\alpha$.
\end{prop}
The proof of Proposition~\ref{prop:cls-betting} is deferred to \S\ref{se:proof-opt-sgd}.

\subsubsection{Analysis under \LSC}\label{app:lsc-module}
Global strong convexity fails for many statistical objectives whose curvature degenerates away from the optimum, the canonical example being logistic regression. Nonetheless, many such examples remain convex; local self-concordance is one property that exploits the finer structures of such objectives.

For such objectives, \citet[Proposition~1]{bach2010self} showed that
\begin{align}
    e^{-C\|x-z\|}\nabla^2F(z)\preceq\nabla^2F(x)\preceq e^{C\|x-z\|}\nabla^2F(z),\ x,z\in\R^d. \label{eq:bach}
   \end{align} 
In the following, we derive the optimization module for \LSC.
   \paragraph{Optimization module for \LSC.} The key idea is to leverage \eqref{eq:bach} to show \SC\ and \LS\ hold on any known ball, with the radius determined by the Hessian $\nabla^2 F(x^\star)$. Concretely, 
let
$\Dcal:=\{x:\|x-x_0\|\le r_{\Dcal}\}$ with known radius
$r_{\Dcal}>0$ and $x^\star\in\Dcal$. Suppose that 
\[
m_0I_d\preceq\nabla^2F(x_0)\preceq M_0I_d,
\ 0<m_0\le M_0<\infty.
\]
Therefore, \eqref{eq:bach} implies that
\[
m_\star I_d\preceq\nabla^2F(x^\star)\preceq L_\star I_d,
\
m_\star:=e^{-Cr_{\Dcal}}m_0,
\
L_\star:=e^{Cr_{\Dcal}}M_0.
\]
Define
\[
\kappa_C(r):=\int_0^1e^{-Cur}\,du
=
\begin{cases}
(1-e^{-Cr})/(Cr), & r>0,\\
1, & r=0.
\end{cases}
\]
Clearly, $\kappa_C$ is positive,
continuous and nonincreasing on $[0,\infty)$.
Since $\nabla F(x^\star)=0$,
\eqref{eq:bach} gives for $x \in \R^d$
\begin{align}
\langle\nabla F(x),x-x^\star\rangle
&=\int_0^1
  (x-x^\star)^\top\nabla^2F\big(x^\star+u(x-x^\star)\big)\ (x-x^\star)\,du \nonumber\\
&\ge
  \kappa_C(\|x-x^\star\|)(x-x^\star)^\top\nabla^2F(x^\star)(x-x^\star)
\ge m_\star\|x-x^\star\|^2 \kappa_C\big(\|x-x^\star\|\big).
\label{eq:lsc-radial-drift}
\end{align}
Consequently, similar to \S\ref{se:opt-rsi}, it follows
\begin{equation}\label{eq:lsc-nonlinear-recursion}
Z_{t,1}
\le
\bigl[1-2m_\star\eta_t\kappa_C(\sqrt{Z_{t-1,1}})\bigr]
Z_{t-1,1}
+\eta_t^2\|g_t\|^2
+\langle2\eta_t(x_{t-1}-x^\star),\xi_t\rangle.
\end{equation}
Observe that \eqref{eq:lsc-nonlinear-recursion} is yet another instance of the coefficient $a_t$ depending on past observations, much like Oja's algorithm. Invoking the treatment in \S\ref{se:oja-boundary}, anytime-valid boundaries can be constructed for $Z_{t,1}$ under \LSC.

\subsubsection{Transfers between performance measures}\label{app:transfer-module}
Each analysis above targets the performance measure that its growth assumption controls naturally. However, often such analyses may carry over to other performance measures of interest. Rather than deriving a new recursion, one can transfer an envelope deterministically through the assumptions of Table~\ref{tab:smooth} while incurring no slackness (as one would sometimes, eg. see the discussion at the beginning of \S\ref{se:opt-pr}). Formally, let $Z_{t,i}\le R_t$, $i\in\{1,2,3, 4,5\}$, hold for all $t$ with probability at least $1-\alpha$. Then the bounds in Table~\ref{tab:transfers} hold for all $t$ with the same probability.

\begin{table}[t]
\centering
\small
\renewcommand{\arraystretch}{1.35}
\begin{tabularx}{\linewidth}{@{}l l >{\raggedright\arraybackslash\leavevmode}X@{}}
\toprule
Assumption & Transferred bound & Proof  \\
\greymidrule
\multicolumn{3}{@{}l}{\textbf{Base envelope:} $Z_{t,1}\le R_t$} 
\\
\bottomrule
\LC$_{L_0}$ & $Z_{t,2}\le L_0\sqrt{R_t}$ & \LC \\
\LS$_L$ & $Z_{t,2}\le LR_t/2$ & Equation \eqref{eq:descent} with $x=x^\star$ \\
 & $Z_{t,3}\le L\sqrt{R_t}$ & \LS \\
\LG$_K$ & $Z_{t,2}\le K\sqrt{R_t}\,(1+\|x_t\|+\sqrt{R_t})$ & \LG\\
 & $Z_{t,3}\le K(1+\|x_t\|)$ & Assumption~\ref{ass:LG} \\
$\|\nabla F\|^2\le K_\bullet(F-F^\star)$ & $Z_{t,2}\le K_\bullet R_t/4$, $Z_{t,3}\le K_\bullet\sqrt{R_t}/2$ & Lemma \ref{lem:self-bounding} \\
\bottomrule
\multicolumn{3}{@{}l}{\textbf{Base envelope:} $Z_{t,2}\le R_t$} \\
\greymidrule
\LS$_L$ & $Z_{t,3}\le\sqrt{2LR_t}$ & \eqref{eq:descent} with $y=x_t-\nabla F(x_t)/L$ \\
\PL$_\mu$ & $Z_{t,1}\le4R_t/\mu$ & \citet[Theorem~2]{karimi2016linear}; \citet[Corollary~12]{garrigos2023square} \\
\KLS, $\int_0^1dv/\varphi(v)<\infty$ & $Z_{t,1}\le\big(\int_0^{R_t}\varphi^{-1}(v) \ \text{d}v\big)^2$ & \citet[Theorem~2]{bolte2010characterizations} \\
& $Z_{t,4}\le S_t^{-1}\sum_{s=1}^t\eta_sR_{s-1}$ & Table~\ref{tab:metrics} \\
\bottomrule
\multicolumn{3}{@{}l}{\textbf{Base envelope:} $Z_{t,3}\le R_t$} \\
\greymidrule
\KLS & $Z_{t,2}\le\sup\{u\ge0:\varphi(u)\le R_t\}$ & \eqref{eq:kl-gap-condition} \\
\bottomrule
\multicolumn{3}{@{}l}{\textbf{Base envelope:} $Z_{t,4}\le R_t$} \\
\LS$_L$ & $Z_{t,5}\le2LR_t$ & \eqref{eq:descent} with $y=x_{s-1}-\nabla F(x_{s-1})/L$ \\
\bottomrule
\multicolumn{3}{@{}l}{\textbf{Base envelope:} $Z_{t,5}\le R_t$} \\
\PL$_\mu$ & $Z_{t,4}\le R_t/\mu$ & \eqref{eq:kl-gap-condition} with $\varphi(u)=\sqrt{\mu u}$ \\
\bottomrule
\end{tabularx}
\caption{Same-event transfers between performance measures. Bounds on or from $Z_{t,1}$ require $x^\star$ to be unique.}
\label{tab:transfers}
\end{table}

\subsection{Proofs for the results in \S\ref{app:opt-sgd-boundaries}}\label{se:proof-opt-sgd}
\begin{proof}[Proof of Proposition~\ref{prop:convex-hybrid}]
  Recall $v_s=2\eta_s(x_{s-1}-x^\star)$, and write $Y_t:=Y_t(\bv)$. As in the proof of Proposition~\ref{prop:ls-hybrid}, $E_t(\lambda):=\exp\bigl[\lambda Y_t-\lambda^2\sigma^2\sum_{s=1}^t\|v_s\|^2/2\bigr]$ is a nonnegative supermartingale with $E_0(\lambda)=1$, $\lambda\ge0$. Let $\tau(\lambda):=\inf\{t\ge1:\lambda\|v_t\|\ge1/G\}$, and consider \eqref{eq:new-betting-martingale} with $\bv$ replaced by $-\bv$,
\[
K_t^1(-\bv,\lambda)=\prod_{\substack{1\le s\le t\\ s<\tau(\lambda)}}\frac{1-\lambda\langle v_s,g_s\rangle}{1-\lambda\langle v_s,\nabla F(x_{s-1})\rangle},\ \lambda\in[0,1/G).
\]
Clearly, $K_t^1(-\bv,\lambda)$ is a nonnegative martingale with $K_0^1(-\bv,\lambda)=1$. By conditional monotone convergence, $K_t^{\mathrm H}:=\frac34\sum_{k\ge0}p_kE_t(\lambda_k)+\frac14\sum_{k\ge0}p_kK_t^1(-\bv,\lambda_k/G)$ is a nonnegative supermartingale with $K_0^{\mathrm H}=1$, and Ville's inequality (Lemma~1 of \citealp{howard2018}) gives $\IP(\Omega_\alpha)\ge1-\alpha$, where $\Omega_\alpha:=\{K_t^{\mathrm H}<1/\alpha\text{ for all }t\ge1\}$.

We next compare \(W_t^{\mathrm H}(Y_t)\) with \(K_t^{\mathrm H}\). Since $\log(1-u)=-u-\psi_{E,1}(u)$ for $u<1$ and $\xi_s=\nabla F(x_{s-1})-g_s$, on $\{t<\tau(\lambda)\}$,
\begin{align}
\log K_t^1(-\bv,\lambda)&=\lambda Y_t-\sum_{s=1}^t\psi_{E,1}(\lambda\langle v_s,g_s\rangle)+\sum_{s=1}^t\psi_{E,1}(\lambda\langle v_s,\nabla F(x_{s-1})\rangle)\overset{(a)}{\ge}\lambda Y_t-\sum_{s=1}^t\psi_{E,1}(\lambda\|v_s\|\|g_s\|), \label{eq:lambda_ineq}
\end{align}
where $(a)$ follows from $\psi_{E,1}(u)\le\psi_{E,1}(|u|)$ for $|u|<1$, and the monotonicity of $\psi_{E,1}$ on $[0,1)$. Since the projected iterates belong to $\Xcal$,
$\|v_s\|\le M_s=2\eta_sR_x$ almost surely for every $s\ge1$.
Hence $\lambda_k/G\in\Lambda_t^G(M)$ implies
$t<\tau(\lambda_k/G)$, and \eqref{eq:lambda_ineq} gives
\[
\exp\bigl(\lambda_kY_t/G-J_t(\lambda_k/G)\bigr)
\le K_t^1(-\bv,\lambda_k/G).
\]
Similarly, $\sum_{s\le t}\|v_s\|^2\le\sum_{s\le t}M_s^2$
bounds each Gaussian summand of $W_t^{\mathrm H}(Y_t)$
by its counterpart in $K_t^{\mathrm H}$.
Summing
yields $W_t^{\mathrm H}(Y_t)\le K_t^{\mathrm H}$.
Thus, on $\Omega_\alpha$, $Y_t\le U_t^{\mathrm H}$ for all $t\ge1$.
Unrolling \eqref{eq:convex-rec} now gives
\[
Z_t^{\CC}
\le R_x^2+\sum_{s=1}^t\eta_s^2\|g_s\|^2+Y_t
\le R_t^{\mathrm H},\qquad t\ge1,
\]
which establish anytime-validity.

Finally, we establish $R_t^{\mathrm H}\le\tilde R_t$ along the same
realized SGD trajectory. Observe that for every $k\ge0$ and
$y\in\R$,
\[
W_t^{\mathrm H}(y)
\ge \frac{3p_k}{4}
    \exp\Big(\lambda_ky-\frac{\lambda_k^2V_t}{2}\Big).
\]
Therefore, $W_t^{\mathrm H}(y)<1/\alpha$ implies
$y<\lambda_k^{-1}\log(4/(3\alpha p_k))+\lambda_kV_t/2$
for every $k\ge0$, from which we procure
\[
U_t^{\mathrm H}
\le \inf_{k\ge0}
\Big\{\frac{\log(4/(3\alpha p_k))}{\lambda_k}
      +\frac{\lambda_kV_t}{2}\Big\},
\]
This, in view of Corollary \ref{cor:closed-form}(ii), completes the proof. 
\end{proof}
\begin{proof}[Proof of Proposition~\ref{prop:cls-betting}]
Note that $Z_0^{\CC,\LS}\le R_0$ and \eqref{eq:key-recursion} holds with $a_t=1$, $b_t=\eta_t^2\|g_t\|^2$ and $v_t=2\eta_te_{t-1}$, while $\xi_s=-(g_s-\IE_{s-1}g_s)$. Therefore, assuming that $H_t$ satisfies \eqref{eq:obs-bet-mean-correction} on $\bigcap_{s<t}\{Z_s^{\CC,\LS}\le R_s\}$, we can proceed as in the proof of Proposition~\ref{prop:betting-SGD}, Theorem~\ref{thm:obs-betting} and Lemma~\ref{lem:observable-envelope} to yield the claim.  Subsequently, we verify \eqref{eq:obs-bet-mean-correction}. 

Clearly, $z\mapsto H_t(\lambda,z)$ is nondecreasing, nonnegative and lower semicontinuous, with $H_t(\lambda,0)=0$. On this event, $Z_{t,1}\le(\sqrt{R_{t-1}}+\eta_t\|g_t\|)^2$, and, with $m_s:=\langle v_s,\nabla F(x_{s-1})\rangle$, convexity and $\|\nabla F(x_{s-1})\|=\|\IE_{s-1}g_s\|\le G$ give $0\le\lambda m_s\le\lambda M_sG<1$ for $\lambda\in\Lambda_t$ and $s\le t$. Moreover, from \eqref{eq:interpolation} it follows that
\begin{align}
    m_s&=2\eta_s\langle\nabla F(x_{s-1}),e_{s-1}\rangle\ge2\eta_s\big(F(x_{s-1})-F^\star\big)+\frac{\eta_s}L\|\nabla F(x_{s-1})\|^2,\nonumber\\
    &\ \text{ and hence, }\ \sum_{s=1}^tm_s\ge Z_t^{\CC,\LS}-Z_{t,1}. \label{eq:m-s-ineq}
\end{align}
Hence, for $0\le z\le Z_t^{\CC,\LS}$,
\[
\begin{aligned}
    H_t(\lambda,z)\overset{(a)}{\le}t\,\psi_{E,1}\Big(\frac{\lambda}{t}\big(Z_t^{\CC,\LS}-Z_{t,1}\big)\Big)
&\overset{(b)}{\le}t\,\psi_{E,1}\Big(\frac\lambda t\sum_{s=1}^tm_s\Big)\\
&{\le}\sum_{s=1}^t\psi_{E,1}(\lambda m_s)=\sum_{s=1}^t\psi_{E,1}\big(\lambda\langle v_s,\IE_{s-1}g_s\rangle\big),
\end{aligned}
\]
where $(a)$ follows from the monotonicity of $\psi_{E,1}$ on $[0,1)$, and $(b)$ employs \eqref{eq:m-s-ineq}. This completes the proof.
\end{proof}

\section{Extended applications}\label{app:appl}
This section tabulates explicit boundary constructions corresponding to the examples in \S\ref{se:appl}. Furthermore, we supplement \S\ref{se:appl} by providing two more instances of our results yielding first-of-its-kind data-dependent time-uniform boundaries in settings with wide application.
\subsection{Explicit boundaries for Oja's algorithm}\label{app:oja}
Recall the Oja's algorithm from \S\ref{se:oja}.
For a candidate envelope $\{R_t\}$ satisfying $R_t \le 1$, consider $a_t:= 1-2\delta\eta_t(1-R_{t-1})$. Let $A_t:=\prod_{s=1}^ta_s$. Since $\|\widehat\Sigma_t\|=\|X_t\|^2$ and $M_s^2=2\eta_s^2R_{s-1}/A_s^2$, each noise assumption yields a concrete $U_t$.
\begin{itemize}
\item \emph{Bounded observations with mixing and stitching envelope.} If $\|X_t\|\le B$, Table~\ref{tab:sub-psi-examples} gives $\nu_t=B^4$, and Corollary~\ref{cor:closed-form}(i-(ii)) along with Table~\ref{tab:sub-psi-examples}, first row gives the normal-mixture and stitching envelopes
\[
\begin{aligned}
R_t&=A_t\Big(R_0+3\sum_{s=1}^t\frac{\eta_s^2\|\widehat\Sigma_su_{s-1}\|^2}{A_s}+U_t\Big),\\
\ U_t&\in\Big\{\Ncal_\alpha\Big(2B^4\sum_{s=1}^t\frac{\eta_s^2R_{s-1}}{A_s^2}\Big),\ 4\mathfrak S_\alpha\Big(2B^4\sum_{s=1}^t\frac{\eta_s^2R_{s-1}}{A_s^2}\Big)\Big\}.
\end{aligned}
\]
To replace the worst-case $B^4$ by the realized $\|X_s\|^4$, Table~\ref{tab:sub-psi-examples}, fifth row gives $\psi=\psi_{E,B^2}$ and $\nu_t=\|X_t\|^4$, and Corollary~\ref{cor:closed-form}(iii) gives the sub-gamma stitching envelope with
\[
U_t=\mathfrak G_\alpha\Big(2\sum_{s=1}^t\frac{\|X_s\|^4\eta_s^2R_{s-1}}{A_s^2},\ B^2\max\Big\{1,\max_{s\le t}\frac{\eta_s\sqrt{2R_{s-1}}}{A_s}\Big\}\Big).
\]

\item \emph{Bounded observations with betting envelope.} If $\|X_t\|\le B$, Theorem~\ref{thm:obs-betting} gives
\[
U_t=\sup\Big\{y:\sum_{k:\lambda_k\in\Lambda^{B^2}_t(M)}p_k\exp\Big(\lambda_ky-\sum_{s=1}^t\psi_{E,1}\big(\lambda_kM_s\|X_s\|^2\big)\Big)<\frac1\alpha\Big\},
\]
which is no larger than the sub-exponential discrete-mixture envelope from Theorem~\ref{thm:obs-comp} with $\nu_t=\|X_t\|^4$. The corresponding uniform boundary $R_t$ is constructed as:
\[
R_t:=A_t\Big(R_0+\sum_{s=1}^t\frac{b_s}{A_s}+\mathcal U_\alpha(t;M,\nu)\Big).
\]
\item \emph{Finite fourth moment with mixing and stitching.} If $\IE_{t-1}\|X_t\|^4\le K_4$ for a known constant $K_4>0$, Table~\ref{tab:sub-psi-examples}, third row with $Y_t=\mathrm{vec}\,\widehat\Sigma_t$ gives $\nu_t=\frac23(\|X_t\|^4+K_4)$; the same closed forms for mixing and stitching apply with $\psi=\psi_N$ and $V_t=\frac43\sum_{s=1}^t(\|X_s\|^4+K_4)\eta_s^2R_{s-1}/A_s^2$.
\end{itemize}

\subsection{Explicit boundaries for Temporal Difference learning}\label{app:td}
We follow \S\ref{app:oja} to provide explicit expressions for the boundaries constructed for TD-learning via Proposition \ref{prop:td-rec} and Theorem \ref{thm:td-last}.
\begin{itemize}
\item \emph{Sub-Gaussian with mixing and stitching envelope.} Recall that $M_s^2=4\eta_s^2R_{s-1}/A_s^2$, and let $G(\theta):=(1+\gamma)\|\theta\|+r_{\max}$, so that $\|g_t\|\le G(\theta_{t-1})$ and hence $\|\xi_t\|\le\|g_t\|+\|\IE_{t-1}g_t\|\le2G(\theta_{t-1})$. Table~\ref{tab:sub-psi-examples}, first row, with $c_t=2G(\theta_{t-1})$ gives $\psi=\psi_N$ and $\nu_t=4G(\theta_{t-1})^2$, and Corollary~\ref{cor:closed-form}(i)--(ii) gives the normal-mixture and stitching envelopes
\[
\begin{aligned}
R_t&=A_t\Big(R_0+\sum_{s=1}^t\frac{\eta_s^2\|g_s\|^2}{A_s}+U_t\Big),\\
\ U_t&\in\Big\{\Ncal_\alpha\Big(16\sum_{s=1}^t\frac{G(\theta_{s-1})^2\eta_s^2R_{s-1}}{A_s^2}\Big),\ 7\mathfrak S_\alpha\Big(4\sum_{s=1}^t\frac{G(\theta_{s-1})^2\eta_s^2R_{s-1}}{A_s^2}\Big)\Big\}.
\end{aligned}
\]
\item \emph{Self-normalized envelope.} Table~\ref{tab:sub-psi-examples}, third row, gives $\psi=\psi_N$ and $\Delta V_t=\frac13\xi_t\xi_t^\top+\frac23\IE_{t-1}[\xi_t\xi_t^\top]$. Since $\|\xi_t\|\le\|g_t\|+G(\theta_{t-1})$ and $\IE_{t-1}\|\xi_t\|^2\le\IE_{t-1}\|g_t\|^2\le G(\theta_{t-1})^2$, we have $\Delta V_t\preceq\nu_tI$ with the observable
\[
\nu_t:=\tfrac13\big(\|g_t\|+G(\theta_{t-1})\big)^2+\tfrac23G(\theta_{t-1})^2,
\]
and the same closed forms apply with \[
V_t=\frac43\sum_{s=1}^t\bigl[(\|g_s\|+G(\theta_{s-1}))^2+2G(\theta_{s-1})^2\bigr]\eta_s^2R_{s-1}/A_s^2.
\] Since $\nu_t\le2G(\theta_{t-1})^2$, this envelope is sharper than the sub-Gaussian envelope, and it adapts to small realized TD updates.
\item \emph{Betting envelope for projected TD learning.} Consider projected TD learning \citep{bhandari2018finite}: $\theta_t=\Pi_\Theta(\theta_{t-1}-\eta_tg_t)$, with $\Pi_\Theta$ being the projection onto $\{\|\theta\|\le\Theta\}$ and $\Theta\ge r_{\max}/\kappa$. Since $\|\theta^\star\|\le r_{\max}/\kappa\le\Theta$, hence the projection operator $\Pi_\Theta$ is Lipschitz with Lipschitz coefficient $1$. Therefore, $\|g_t\|\le (1+\gamma)\Theta+r_{\max}$. Theorem~\ref{thm:obs-betting} and $G=(1+\gamma)\Theta+r_{\max}$ then gives
\[
U_t=\sup\Big\{y:\sum_{k:\lambda_k\in\Lambda^{G}_t(M)}p_k\exp\Big(\lambda_ky-\sum_{s=1}^t\psi_{E,1}\big(\lambda_kM_s\|g_s\|\big)\Big)<\frac1\alpha\Big\},
\]
and Theorem~\ref{thm:td-last} continues to hold by Theorem~\ref{thm:obs-betting} and Lemma~\ref{lem:observable-envelope}.
\end{itemize}
\subsection{Momentum-assisted SGD}\label{se:mom}
As one of the most popular modifications of SGD, the stochastic heavy-ball method of \citet{polyak1964some}, also known as the momentum-assisted SGD or m-SGD, and the accelerated scheme of \citet{nesterov1983method} have become a mainstay of deep learning \citep{sutskever2013importance} owing to their increased efficiency. Accelerated stochastic approximation has been shown to attain the optimal rate for smooth convex stochastic programs \citep{lan2012optimal}, and such results have been extended to nonconvex ones \citep{ghadimi2016accelerated} as well. While the asymptotic behavior, almost sure convergence and high probability guarantees for mSGD has been well known \citep{gadat2018stochastic, sebbouh2021almost, liu2023high,liu2024revisiting}, none of these results facilitate a data-dependent stopping time.

The anytime question for momentum methods was first taken up by \citet{feng2023anytime}. Translating a Lyapunov function of the second-order ODE $\ddot X(t)+2t^{-1}\dot X(t)+2t^{-3/2}\nabla F(X(t))=0$ into discrete time, they constructed an exponential supermartingale to bound $F(x_t)-F^\star$ by a multiple of $(1+\log(1/\alpha))\log t/\sqrt t$, simultaneously over all $t$. Subsequently, \citet{feng2025breaking} distilled the argument into a large-deviation inequality for almost supermartingales in the sense of \citet{robbins-siegmund}; both works also refine the logarithmic factor to $\log^{(1+\varepsilon)/2}t$ through a slight modification of the step sizes \citep[Theorem~6]{feng2023anytime}, \citep[Proposition~1]{feng2025breaking}. To the best of our knowledge, these are the only anytime-valid guarantees available for a momentum method. However, from the standpoint of a practitioner, they share the drawbacks of deterministic envelopes discussed earlier. The envelopes do not respond to the realized trajectory, and the step sizes are tied to prescribed schedules $\eta_t\propto\log^{-2}(t+2)$ or $\eta_t\propto\log^{-(1+\varepsilon)}(t+2)$. Moreover, the noise is required to satisfy a norm-sub-Gaussian condition, and under a mere variance bound only $\Lcal_2$ and almost sure statements are available \citep[Theorems~3--4]{feng2023anytime}. 

Nonetheless, the structure uncovered by \citet{feng2023anytime} exactly provides the optimization module our recipe requires.  In \S\ref{se:mom-opt}, we revisit and establish this recursion for arbitrary predictable step sizes (Proposition~\ref{prop:mom-rec}). In \S\ref{se:mom-boundary}, we convert it into observable envelopes, both under the noise assumption of \citet{feng2023anytime,feng2025breaking} and under a finite conditional variance alone, where the self-normalized and quadratic betting constructions of \S\ref{se:ls-betting} apply (Theorem~\ref{thm:mom-coverage} and Proposition~\ref{prop:mom-quad}). Finally, \S\ref{se:mom-rates} shows that the realized envelope, and not merely the latent energy, attains the logarithmic rates of \citet{feng2023anytime,feng2025breaking}, with a constant that grows only polynomially in $1/\varepsilon$ (Theorem~\ref{thm:mom-rates}). To the best of our knowledge, the resulting envelopes are the first data-dependent, anytime-valid guarantees for a momentum-based stochastic optimizer.

\subsubsection{Set-up and optimization module for momentum-assisted SGD}\label{se:mom-opt}
Throughout this section, $F\in\mathcal C^1(\R^d)$ satisfies \CC\ (Assumption~\ref{ass:c}) and \LS\ with a known constant $L>0$, and attains its minimum $F^\star:=F(x^\star)$ at some $x^\star\in\R^d$. Starting from a deterministic $x_1=x_0\in\R^d$, the momentum scheme of \citet[eq.~(4)]{feng2025breaking}, equivalently \citet[eq.~(28)]{feng2023anytime}, iterates
\begin{equation}\label{eq:mom-update}
x_{t+1}=x_t+\frac{t}{t+2}\,(x_t-x_{t-1})-\frac{2\sqrt{\eta_t}}{(t+2)\sqrt t}\,g_t, \  t\ge1,
\end{equation}
where, for every $t\ge1$, the stochastic gradients $g_t$ are unbiased. Accordingly, the noise is $\xi_t:=\nabla F(x_t)-g_t$, so that $\IE_{t-1}\xi_t=0$. We remark that while \citet{feng2023anytime,feng2025breaking} analyze the deterministic schedule $\eta_t=\bigl[16L^2\log^2(t+2)\bigr]^{-1}$ and its variants, we allow arbitrary positive predictable step sizes, and impose a schedule only when comparing rates in \S\ref{se:mom-rates}. We also fix an observable $\eta_0>0$, which only enters the initial energy and the coefficient $a_1$ in \eqref{eq:key-recursion}.

The scheme \eqref{eq:mom-update} admits a transparent reformulation. Put $z_t:=(t+1)x_t-tx_{t-1}$ for $t\ge1$. Multiplying \eqref{eq:mom-update} by $t+2$, and unrolling the recursion through the definition of $z_t$, yields
\begin{equation}\label{eq:mom-averaging}
z_{t+1}=z_t-2\sqrt{\frac{\eta_t}{t}}\;g_t, \  x_t=\frac1{t+1}\Big(x_0+\sum_{s=1}^tz_s\Big), \  t\ge1.
\end{equation}
In other words, \eqref{eq:mom-update} runs a stochastic gradient recursion with step size $2\sqrt{\eta_t/t}$ on the auxiliary sequence $(z_t)$. The displacement of $z_t$ from the optimum,
\[
\varphi_t:=z_t-x^\star=(x_t-x^\star)+t(x_t-x_{t-1}), \  t\ge1,
\]
is the momentum-corrected displacement around which the analysis of \citet{feng2023anytime,feng2025breaking} revolves. Instead of directly working with $\|\varphi_t\|^2,$ we consider the performance measure
\begin{equation}\label{eq:mom-lyap}
Z_t:=\|\varphi_{t+1}\|^2+4\sqrt{(t+1)\eta_t}\,\bigl(F(x_t)-F^\star\bigr), \  t\ge0,
\end{equation}
which is the discrete Lyapunov function of \citet[eq.~(29)]{feng2023anytime}; see also \citet[eq.~(5)]{feng2025breaking}. Assume that $\|x_0-x^\star\|\le D$. Since $x_1=x_0$ gives $\varphi_1=x_0-x^\star$, and \eqref{eq:descent} with $\nabla F(x^\star)=0$ gives $F(x_0)-F^\star\le LD^2/2$, it follows that
\[
Z_0\le R_0:=(1+2L\sqrt{\eta_0})D^2.
\]
The following result provides the optimization module for $Z_t$.
\begin{prop}\label{prop:mom-rec}
Grant the notation and conditions of \S\ref{se:mom-opt}, and put
\begin{equation}\label{eq:mom-abv}
a_t:=\max\Big\{1,\sqrt{\eta_t/\eta_{t-1}}\Big\},\ b_t:=\frac{4\eta_t}{t}\,\|g_t\|^2,\ v_t:=4\sqrt{\frac{\eta_t}{t}}\;\varphi_t, \  t\ge1.
\end{equation}
Then, almost surely, for every $t\ge1$,
\begin{equation}\label{eq:mom-rec}
Z_t\le a_tZ_{t-1}+b_t-2\sqrt{\frac{\eta_t}{t}}\,\biggl[F(x_t)-F^\star+\frac1L\|\nabla F(x_t)\|^2\biggr]+\langle v_t,\xi_t\rangle.
\end{equation}
\end{prop}
Proposition~\ref{prop:mom-rec} is proved in Appendix~\S\ref{app:appl-proofs}. For nonincreasing step sizes, Proposition \ref{prop:mom-rec} coincides with \citet[Lemma~2]{feng2025breaking}, equivalently \citet[Lemma~3]{feng2023anytime}. In contrast, our result accommodates any predictable, possibly non-monotone and data-driven, step sizes.

\subsubsection{Boundaries for momentum-assisted SGD}\label{se:mom-boundary}
Proposition \ref{prop:mom-rec} can now be fed into Lemma~\ref{lem:observable-envelope}, upon verifying the conditions of Theorem~\ref{thm:obs-comp}. Since $\|\varphi_t\|^2\le Z_{t-1}$, on $\{Z_{t-1}\le R_{t-1}\}$ the normalized weights satisfy
\begin{equation}\label{eq:mom-M}
\frac{\|v_t\|}{A_t}\le M_t:=\frac4{A_t}\sqrt{\frac{\eta_tR_{t-1}}{t}}, \  t\ge1,
\end{equation}
where $M_t$ is predictable and observable. The following result provide general boundaries for m-SGD.

\begin{thm}\label{thm:mom-coverage}
Grant the notation and conditions of Proposition~\ref{prop:mom-rec}, and fix $\alpha\in(0,1)$. Let $\xi_t=\nabla F(x_t) - g_t$ satisfy the following: $(\xi_t,\Delta V_t)_{t\ge0}$ is a sub-$\psi$ process with $\Delta V_t\preceq\nu_tI$ for an observable, $\Fcal_t$-adapted $\nu_t\ge0$, and let $U_t$ be one of the boundaries $\mathcal U^{\mathrm{mix}}_\alpha(t;M,\nu)$, $\mathcal U^{\mathrm{dm}}_\alpha(t;M,\nu)$ and $\mathcal U^{\St}_\alpha(t;M,\nu)$ of Theorem~\ref{thm:obs-comp}, or an observable upper bound, such as those in Corollary~\ref{cor:closed-form}. 

Put $A_0:=1$ and $U_0:=0$, recall $R_0$ from \S\ref{se:mom-opt}, and define recursively, for $t\ge1$, $A_t:=A_{t-1}a_t$, $M_t$ as in \eqref{eq:mom-M}, and
\begin{equation}\label{eq:mom-envelope}
R_t:=A_t\Big(R_0+\sum_{s=1}^t\frac{4\eta_s\|g_s\|^2}{sA_s}+U_t\Big).
\end{equation}
Then 
\[
\begin{aligned}
&\IP\big(Z_t\le R_t\ \text{for all }t\ge0,\ \text{and}\ F(x_t)-F^\star\le\wh\Delta_t\ \text{for all }t\ge1\big)\ge1-\alpha,\\
&\ \wh\Delta_t:=\frac{R_t}{4\sqrt{(t+1)\eta_t}}, \  t\ge1.
\end{aligned}
\]
\end{thm}
Theorem~\ref{thm:mom-coverage} is proved in Appendix~\S\ref{app:appl-proofs}. As with the examples in \S\ref{se:appl}, we provide a few concrete boundary constructions.

All the constructions below take $\psi=\psi_N$.
\begin{itemize}
\item \emph{Conditionally sub-Gaussian noise.} Suppose that $\IE_{t-1}\exp\langle\theta,\xi_t\rangle\le\exp(\sigma_N^2\|\theta\|^2/2)$ for all $\theta\in\R^d$, with a known $\sigma_N>0$, i.e. $(\xi_t,\sigma_N^2I_d)_{t\ge0}$ is a sub-$\psi_N$ process. This enables us to take $\nu_t=\sigma_N^2$ and
\[
V^{N}_t:=\sigma_N^2\sum_{s=1}^tM_s^2=16\sigma_N^2\sum_{s=1}^t\frac{\eta_sR_{s-1}}{sA_s^2}.
\]
We denote by $U^{N,\mathrm{DM}}_t$ and $U^{N,\mathrm{St}}_t$ the boundaries analogous to self-normalized boundaries, with $Q^{\mathrm{SN}}_t$ replaced by $V^N_t$; let $R^{N,\mathrm{DM}}_t$ and $R^{N,\mathrm{St}}_t$ be the corresponding envelopes \eqref{eq:mom-envelope}. The normal-mixture bound $\Ncal_\alpha(V^N_t)$ of Corollary~\ref{cor:closed-form}(i) and the closed-form stitching bound of Corollary~\ref{cor:closed-form}(ii) are also valid choices of $U_t$. Note that this condition is weaker than the norm sub-Gaussian condition assumed by \citet{feng2023anytime, feng2025breaking}:
\begin{equation}
\IE_{t-1}\exp\big(\|\xi_t\|^2/\sigma^2\big)\le e, \ \IE_{t-1}\xi_t=0, \nonumber
\end{equation}

\item \emph{Finite variance and self-normalization.} Suppose only that $\IE_{t-1}\xi_t=0$ and $\IE_{t-1}\|\xi_t\|^2\le\sigma^2$ for a known $\sigma\ge0$, which, under unbiasedness, is the bounded-variance condition of \citet[Assumption~2]{feng2025breaking}. By \LS, $\|\nabla F(x_t)\|\le L\|x_t-x^\star\|\le G_t:=L(\|x_t-x_0\|+D)$, where $G_t$ is predictable and observable, and thus $\|\xi_t\|\le G_t+\|g_t\|$. Therefore, exactly as in \S\ref{se:ls-betting}, Table~\ref{tab:sub-psi-examples}, third row, gives $\Delta V^{\mathrm{SN}}_t\preceq\nu^{\mathrm{SN}}_tI$ with
\begin{equation}\label{eq:mom-sn}
\nu^{\mathrm{SN}}_t:=\frac{(G_t+\|g_t\|)^2+2\sigma^2}3, \  V^{\mathrm{SN}}_t:=\sum_{s=1}^t\nu^{\mathrm{SN}}_sM_s^2.
\end{equation}
Denoting by $R^{\mathrm{SN,DM}}_t$ and $R^{\mathrm{SN,St}}_t$ the envelopes \eqref{eq:mom-envelope} built analogously to self-normalized bounaries in \S\ref{se:ls-betting} with $Q^{\mathrm{SN}}_t$ replaced by $V^{\mathrm{SN}}_t$, Theorem~\ref{thm:mom-coverage} provides validity to both boundaries. To the best of our knowledge, this furnishes the first anytime-valid guarantee for \eqref{eq:mom-update} in the finite-variance regime of \citet[Theorems~3--4]{feng2023anytime}, where only in-expectation and almost sure convergence had been established previously.
\end{itemize}
Moreover, as in \S\ref{se:ls-betting} and \S\ref{se:betting-combn}, the self-normalized envelopes can be sharpened via the quadratic product-betting supermartingale. To that end, recall $j$ from \eqref{eq:chi}, fix deterministic $\lambda_k\downarrow0$ with $0<\lambda_0<\infty$ and $p_k>0$ with $\sum_{k\ge0}p_k\le1$, and define, for $\lambda>0$ and $y\in\R$,
\begin{equation}\label{eq:mom-quad}
\begin{aligned}
J^{\mathrm Q}_t(\lambda)&:=\sum_{s=1}^t\Big[j\big(\lambda M_s(G_s+\|g_s\|)\big)+\log\Big(1+\frac{\lambda^2\sigma^2M_s^2}3\Big)\Big],\\
\  W^{\mathrm Q}_t(y)&:=\sum_{k\ge0}p_k\exp\bigl[\lambda_ky-J^{\mathrm Q}_t(\lambda_k)\bigr],
\end{aligned}
\end{equation}
together with $U^Q_t:=\sup\{y\in\R:W^Q_t(y)<1/\alpha\}$. We denote by $R^{\mathrm Q}_t$ the recursion \eqref{eq:mom-envelope} with $U_t=U^{\mathrm Q}_t$.
\begin{prop}\label{prop:mom-quad}
Grant the notation and conditions of Proposition~\ref{prop:mom-rec}, fix $\alpha\in(0,1)$, and suppose that $\IE_{t-1}\xi_t=0$ and $\IE_{t-1}\|\xi_t\|^2\le\sigma^2$. Then the conclusion of Theorem~\ref{thm:mom-coverage} holds with $R_t=R^{\mathrm Q}_t$. Moreover, along a common trajectory, and with common $\{\lambda_k,p_k\}$, it holds for all $t\ge1$ that
\begin{equation}\label{eq:mom-ordering}
R^{\mathrm Q}_t\le R^{\mathrm{SN,DM}}_t\le R^{\mathrm{SN,St}}_t.
\end{equation}
\end{prop}
Proposition~\ref{prop:mom-quad} is proved in Appendix~\S\ref{app:appl-proofs}. In the following, we also discuss the asymptotic rates of the constructed envelopes.

\subsubsection{Rates of the observable envelopes}\label{se:mom-rates}
In the spirit of Theorem~\ref{thm:pr-rates}, in this section we quantify the asymptotic rates of the envelopes constructed under the step-size schedules of \citet{feng2023anytime,feng2025breaking}. For $\eta_\circ>0$ and $p>1$, put
\begin{equation}\label{eq:mom-constants}
H_p:=\eta_\circ\biggl[\frac1{(\log3)^p}+\frac{3(\log3)^{1-p}}{p-1}\biggr], \  C_\circ:=L^2\eta_\circ+\frac{L\sqrt{\eta_\circ}}2\,\zeta(3/2).
\end{equation}
The following theorem establishes the optimality of the envelopes constructed above in the sense of \cite{feng2023anytime, feng2025breaking}.
\begin{thm}\label{thm:mom-rates}
Grant the notation and conditions of Proposition~\ref{prop:mom-rec}, and suppose that $\IE_{t-1}\exp\langle\theta,\xi_t\rangle\le\exp(\sigma_N^2\|\theta\|^2/2)$ for all $\theta\in\R^d$, with a known $\sigma_N>0$. Choose the step sizes $\eta_t=\eta_\circ\log^{-p}(t+2)$ for $t\ge0$, where $\eta_\circ>0$ and $p>1$. Fix $\alpha\in(0,1)$, and let $R_t\in\{R^{N,\mathrm{DM}}_t,R^{N,\mathrm{St}}_t\}$ be level-$\alpha$ time-uniform envelopes with the grid
\[
\lambda_k=\frac{e^{-k/2}}{8\sigma_N^2H_p}, \  p_k=\frac{6}{\pi^2(k+1)^2}, \  k\ge0.
\]
With $H_p$ and $C_\circ$ as in \eqref{eq:mom-constants}, define, for $\delta\in(0,1)$,
\[
K_{\alpha,\delta}:=e^{8C_\circ+1}\biggl[R_0+16\sigma_N^2H_p\Big(d\log2+2\log\frac1\delta\Big)+8\sigma_N^2H_p\log\frac1{\alpha p_0}\biggr].
\]
Then, for every $\delta\in(0,1)$,
\[
\IP\Big(Z_t\le R_t\le K_{\alpha,\delta}\ \text{and}\ F(x_t)-F^\star\le\wh\Delta_t\le\frac{K_{\alpha,\delta}\log^{p/2}(t+2)}{4\sqrt{\eta_\circ(t+1)}}\ \text{ for all }t\ge1\Big)\ge1-\alpha-\delta.
\]
\end{thm}
Theorem~\ref{thm:mom-rates} is proved in Appendix~\S\ref{app:appl-proofs}. In particular, $\delta=\alpha$ yields the probability $1-2\alpha$. Additionally, the norm sub-Gaussian condition of \citet{feng2023anytime,feng2025breaking} implies the conditional sub-Gaussian condition with $\sigma_N^2=3\sigma^2/2$ by \citet[Lemma~4]{feng2025breaking}, and therefore, the same conclusion holds with the leading constant $$K_{a,\delta}':=e^{8C_\circ+1}\Bigl[R_0+8\sigma^2H_p(1+\log(1/\delta))+12\sigma^2H_p\log(1/(\alpha p_0))\Bigr].$$ 

For a fixed $d$, we now compare Theorem~\ref{thm:mom-rates} with the deterministic guarantees of \citet{feng2023anytime,feng2025breaking}. With $p=2$ and $\eta_\circ=1/(16L^2)$, Theorem~\ref{thm:mom-rates} recovers their rate $\log(t+2)/\sqrt{t+1}$. Likewise, with $p=1+\varepsilon$ with $\varepsilon\in(0,1]$ and $\eta_\circ=1/(16L^2)$, Theorem \ref{thm:mom-rates} recovers the rate of \citet[Theorem~6]{feng2023anytime}, while $\varepsilon\in(0,1/2)$ with $\eta_\circ=1/(16L^2C')$, $C'\ge1$, recovers the rate $\log^{(1+\varepsilon)/2}(t+2)/\sqrt{t+1}$ of \citet[Proposition~1]{feng2025breaking}. Therefore, our data-dependent, observable envelopes are also asymptotically optimal in the sense of \cite{feng2023anytime, feng2025breaking}.

We also derive the asymptotic properties of the envelope $R_t^{\mathrm{Q}}$ under the finite-variance condition.
\begin{cor}\label{cor:mom-fv}
Suppose that $\IE_{t-1}\xi_t=0$ and $\IE_{t-1}\|\xi_t\|^2\le\sigma^2$, and let $\eta_t=\eta_\circ\log^{-p}(t+2)$, $t\ge0$, with $\eta_\circ>0$ and $p>1$. Fix $\alpha\in(0,1)$. Then each of $R^{\mathrm{SN,DM}}_t$, $R^{\mathrm{SN,St}}_t$ and $R^{\mathrm Q}_t$ satisfies $\wh\Delta_t=O\big(\log^{p/2}(t+2)/\sqrt{t+1}\big)$ almost surely.
\end{cor}
Corollary~\ref{cor:mom-fv} is proved in Appendix~\S\ref{app:appl-proofs}. With $p=2$, which is the step-size choice in Theorem 3 of \cite{feng2023anytime}, Corollary \ref{cor:mom-fv} instructs that our boundaries match the asymptotic fixed-iterate rate of the actual performance measure $\IE[F(x_t) - F^\star]$- establishing the sharpness of our construction. As an aside, we emphasize that both the derivation of time-uniform envelope as well as its asymptotic rates are also new results, since only convergence was established previously.

We now turn to Q-learning.

\subsection{Q-learning}\label{app:q-learning}
Another application of reinforcement learning is the Q-learning algorithm \citep{watkins1992q}, which learns the optimal action-value function as the fixed point of the Bellman optimality operator. Q-learning has a long history of both asymptotic analyses based on stochastic-approximation arguments \citep{tsitsiklis1994asynchronous,jaakkola1994convergence} as well as non-asymptotic rates \citep{even2003learning,beck2012error,wainwright2019stochastic,chen2020finite,chen2021lyapunov,li2024qlearning}. Concentration bounds that hold for all iterations after a burn-in, or uniformly in time, are also available for contractive stochastic approximation and Q-learning \citep{borkar2021concentration,chandak2022concentration,chen2025}. Finally, asymptotic time-uniform inference has been furnished by functional central limit theorem and its extensions \citep{li2023statistical,bonnerjee2026sharp}. Closest in spirit to the present goal of time-uniform boundaries is the work of \citet{evendar2006action}, who construct confidence intervals around the Q-function to eliminate suboptimal actions and derive stopping conditions under which the learned policy is approximately optimal with high probability; their construction maintains separate upper and lower Q-tables rather than directly bounding the error of an ordinary Q-learning iterate.

In this section, we derive the optimization module for Q-learning, before deploying Theorems~\ref{thm:obs-comp}--\ref{thm:obs-betting} to provide anytime-valid envelopes for the Q-iterates or quality functions. For Q-learning, the $\ell_\infty$-contraction of the Bellman operator yields coupled coordinatewise quadratic recursions (Proposition~\ref{prop:q-rec}). Applying the probabilistic tools to these recursions gives observable bounds for the entire Q-table and the value of its greedy policy.

\subsubsection{Set-up and optimization module for Q-learning}\label{se:q}
In this section, we briefly describe the Q-learning algorithm, and derive the associated recursion \eqref{eq:key-recursion} for appropriate performance measures. It might be useful to recall that the $Q$ in Q-learning refers to the quality function $Q:\mathcal S\times\mathcal A\to\R$; to an interested reader, we refer \cite{sutton2018} for a detailed treatment. For the analysis, we work with a finite discounted MDP as in \S\ref{se:td}, with $\gamma\in[0,1)$, integrable reward laws, and $|\bar r(s,a)|\le r_{\max}$ for a known finite $r_{\max}\ge0$. In particular, denote $N:=|\mathcal S||\mathcal A|$, and index state-action pairs by $i\in[N]$. Therefore, we can also rewrite the Q-function as a vector $Q\in\R^N$. Let
\[
(\mathcal TQ)(s,a):=\bar r(s,a)
+\gamma\mathbb E_{S'\sim P(\cdot\mid s,a)}
\!\Bigl[\max_{a'}Q(S',a')\Bigr].
\]
Then $\mathcal T$ is a $\gamma$-contraction in $\|\cdot\|_\infty$ with unique fixed point $Q^\star$, and $\|Q^\star\|_\infty\le B^\star:=r_{\max}/(1-\gamma)$.

We consider both synchronous Q-learning \citep{li2023statistical,bonnerjee2026sharp}, where every coordinate is updated, and asynchronous Q-learning \citep{qu2020finite,li2020sample,chen2021lyapunov}, where only the current state-action coordinate is updated. Let $d_t(i)\in\{0,1\}$ indicate whether coordinate $i$ is activated at iteration $t$. Thus, $d_t(i)=1$ for every $i$ in the synchronous case, whereas $d_t(i)=\bo_{\{i=I_t\}}$ for the current state-action pair $I_t$ in the asynchronous case.  In particular, $d_t$, $Q_{t-1}$ and the finite step size $\eta_t>0$ are observable and $\Fcal_{t-1}$-measurable. On $\{d_t(i)=1\}$, the conditional marginal laws of $r_t(i)$ and $S'_t(i)$ given $\Fcal_{t-1}$ are $\mathcal R(\cdot\mid i)$ and $P(\cdot\mid i)$, respectively. 
Let $Y_t=(Y_t(i))_{i\in[N]}$ be the integrable observed increment, where
\[
Y_t(i):=
\begin{cases}
r_t(i)+\gamma\max_{a'}Q_{t-1}(S'_t(i),a')-Q_{t-1}(i),
&d_t(i)=1,\\
0,&d_t(i)=0.
\end{cases}
\]
With this notation, the Q-learning algorithm proceeds as follows:
\[
Q_t=Q_{t-1}+\eta_tY_t.
\]
Noting that
$\IE_{t-1}Y_t=\operatorname{diag}(d_t)(\mathcal TQ_{t-1}-Q_{t-1})$,
the corresponding noise term can be identified as
$\xi_t:=Y_t-\IE_{t-1}Y_t$. In particular,
$\IE_{t-1}\xi_t=0$ and $\xi_t(i)=0$ whenever $d_t(i)=0$.

Interestingly, since $\mathcal T$ contracts in $\|\cdot\|_\infty$ rather than in $\|\cdot\|$, $\|Q_t-Q^\star\|^2$ need not satisfy a contractive recursion; moreover, $\|Q_t-Q^\star\|_\infty^2$ is not smooth. We instead retain the coordinatewise quadratic functions $\Phi_i(z):=z_i^2/2$, and consider
\[
e_t:=Q_t-Q^\star,\ 
Z_t(i):=\Phi_i(e_t),\ 
\mathbf{Z}_t:=\max_iZ_t(i)=\frac12\|Q_t-Q^\star\|_\infty^2.
\]
With the notational niceties out of our way, we derive the optimization module.
\begin{prop}\label{prop:q-rec}
Grant the notation and conditions of \S\ref{se:q}. For every $t\ge1$ and $i\in[N]$, it almost surely holds that
\begin{equation}\label{eq:q-coordinate-rec}
\begin{split}
Z_t(i)\le{}&
\bigl[1-(2-\gamma)\eta_td_t(i)\bigr]Z_{t-1}(i)
+\gamma\eta_td_t(i)\mathbf{Z}_{t-1}
+\frac{\eta_t^2}{2}Y_t(i)^2
+v_t(i)\xi_t(i),\\
v_t(i):={}&\eta_td_t(i)e_{t-1}(i).
\end{split}
\end{equation}
\end{prop}

\subsubsection{Boundaries for Q-learning}
Now, similar to Theorem~\ref{thm:td-last}, we move on to constructing observable, anytime-valid boundaries. We maintain a bound $R_t(i)$ for each $Z_t(i)$, and put $\mathbf{R}_t:=\max_iR_t(i)$. Choose observable, predictable $a_t(i)\in (0, \infty)$ satisfying
\[
a_t(i)\ge1-(2-\gamma)\eta_td_t(i),\
A_0(i):=1,\  A_t(i):=\prod_{s=1}^ta_s(i).
\]
 On $\bigcap_j\{Z_{t-1}(j)\le R_{t-1}(j)\}$, Proposition~\ref{prop:q-rec} gives \eqref{eq:key-recursion} for coordinate $i$ with
\begin{equation}\label{eq:q-observable-coefficients}
b_t(i):=\gamma\eta_td_t(i)\mathbf{R}_{t-1}
+\frac{\eta_t^2}{2}Y_t(i)^2,
\ 
\frac{|v_t(i)|}{A_t(i)}
\le M_t(i):=
\frac{\eta_td_t(i)\sqrt{2R_{t-1}(i)}}{A_t(i)}.
\end{equation}
The coupling through $\mathbf{R}_{t-1}$ is handled by Lemma~\ref{lem:observable-envelope}.

\begin{thm}\label{thm:q}
Grant the notation and conditions above. Fix $\alpha\in(0,1)$ and deterministic $\alpha_i>0$ with $\sum_i\alpha_i\le\alpha$. Suppose that $(\xi_t(i),\Delta V_t(i))_{t\ge1}$ is a sub-$\psi$ process with $0\le\Delta V_t(i)\le\nu_t(i)$ for a finite observable adapted $\nu_t(i)$, and choose $U_t(i)$ from Theorem~\ref{thm:obs-comp} or Theorem~\ref{thm:obs-betting} at level $\alpha_i$ with $M(i):=(M_s(i))_{s\ge1}$, $g_t=Y_t(i)$, $H_t\equiv0$ and $\nu(i):=(\nu_s(i))_{s\ge1}$. 

Let $R_0(i):=\frac12(|Q_0(i)|+B^\star)^2$ and $\mathbf{R}_0:=\max_iR_0(i)$. With $a_t(i)$, $A_t(i)$, $b_t(i)$ and $M_t(i)$ as in \eqref{eq:q-observable-coefficients}, define recursively
\begin{equation}\label{eq:q-coordinate-envelope}
R_t(i):=
A_t(i)\Big(R_0(i)+\sum_{s=1}^t\frac{b_s(i)}{A_s(i)}
+U_t(i)\Big),
\ 
\mathbf{R}_t:=\max_iR_t(i).
\end{equation}
Here $U_t(i)=0$ until coordinate $i$ is first activated. Then, with probability at least $1-\alpha$, simultaneously for every $t\ge0$ and $i\in[N]$,
\[
Z_t(i)\le R_t(i),\ 
\|Q_t-Q^\star\|_\infty\le\sqrt{2\mathbf{R}_t}.
\]
On the same event, every greedy policy $\pi_t$ with respect to $Q_t$ satisfies
$\|\Vcal^\star-\Vcal^{\pi_t}\|_\infty\le2\sqrt{2\mathbf{R}_t}/(1-\gamma)$.
\end{thm}
Crucially, even beyond the usual application of $\mathbf{R}_t$ to procure a data-dependent stopping time via some threshold, there exists a more direct construction of a stopping time. Let $\hat a_t(s)\in\argmax_aQ_t(s,a)$ and $\varepsilon_t:=\sqrt{2\mathbf{R}_t}$. On the event of Theorem~\ref{thm:q}, if $Q_t(s,\hat a_t(s))-Q_t(s,a)>2\varepsilon_t$ for every state $s$ and every $a\ne\hat a_t(s)$, then
\[
Q^\star(s,\hat a_t(s))
\ge Q_t(s,\hat a_t(s))-\varepsilon_t
>Q_t(s,a)+\varepsilon_t
\ge Q^\star(s,a).
\]
Consequently, every policy selecting $\hat a_t(s)$ is optimal, including when the criterion is first met at a data-dependent time.

For the noise, either a sub-$\psi$ proxy may be used, or, under bounded rewards, the product-betting boundary. For the reader's convenience, we present a few concrete envelopes in the following. Recall $M_s(i)^2=2\eta_s^2d_s(i)R_{s-1}(i)/A_s(i)^2$.
\begin{itemize}
\item \emph{Finite second moment.}
If $\int r^2\,\mathcal R(dr\mid i)\le K_i$ for known finite $K_i$, $i\in[N]$, then
\[
\IE_{t-1}[Y_t(i)^2]\le K_t^{\mathrm{or}}(i)
:=2d_t(i)\bigl[K_i+(1+\gamma)^2\|Q_{t-1}\|_\infty^2\bigr].
\]
Table~\ref{tab:sub-psi-examples}, third row, gives the observable choice
$\nu_t(i)=\frac23\bigl[Y_t(i)^2+K_t^{\mathrm{or}}(i)\bigr]$ with $\psi=\psi_N$, and one may take $r_{\max}=\max_i\sqrt{K_i}$. The normal-mixture choice in \eqref{eq:q-coordinate-envelope} is
\[
U_t(i)=\Ncal_{\alpha_i}\!\Biggl(
\frac43\sum_{s=1}^t
\frac{\eta_s^2d_s(i)R_{s-1}(i)}{A_s(i)^2}
\bigl[Y_s(i)^2+K_s^{\mathrm{or}}(i)\bigr]\Biggr).
\]
Similarly, stitching envelopes may be given.

\item \emph{Bounded rewards, mixing/stitching.}
If $|r_t(i)|\le r_{\max}$ on activated coordinates,
$\|Q_0\|_\infty\le B:=r_{\max}/(1-\gamma)$ and $\eta_t\le1$, then $\|Q_t\|_\infty\le B$ pathwise and $Z_t(i)\le2B^2$. Thus, each $R_t(i)$ may be capped at $2B^2$, using the capped values subsequently in \eqref{eq:q-observable-coefficients}. If $B=0$, then $Q_t=Q^\star=0$ and one may take $\mathbf{R}_t=0$; below assume $B>0$. Moreover, $|Y_t(i)|\le2Bd_t(i)$ and $|\xi_t(i)|\le2Bd_t(i)$. Hence one may use $\psi=\psi_N$ with $\nu_t(i)=4B^2d_t(i)$. Alternatively, the bounded-observation construction underlying Corollary~\ref{corol:betting} gives $\psi=\psi_{E,2B}$ with $\nu_t(i)=Y_t(i)^2$. Theorem~\ref{thm:obs-comp} and Corollary~\ref{cor:closed-form} then give mixing, sub-exponential stitching, or sub-gamma stitching envelopes.

\item \emph{Bounded rewards, betting.}
Let $G_Q:=2B$. Under the same conditions,
Theorem~\ref{thm:obs-betting} with $g_t=Y_t(i)$ and $H_t\equiv0$ gives
\[
U_t(i)=
\sup\Big\{y\in\R:
\sum_{k:\lambda_k\in\Lambda_t^{G_Q}(M(i))}
p_k\exp\Big(
\lambda_ky-\sum_{s=1}^t
\psi_{E,1}\big(\lambda_kM_s(i)|Y_s(i)|\big)
\Big)<\frac1{\alpha_i}
\Big\},
\]
where $\lambda_k\in(0,G_Q^{-1})$, $\lambda_k\downarrow0$,
$p_k>0$ and $\sum_kp_k=1$ are deterministic. With the same initialization, coefficients $a_t(i)$, levels $\alpha_i$, and choices of $(\lambda_k,p_k)$, these betting envelopes are no larger than the corresponding exponential or sub-gamma stitching envelopes with $\nu_t(i)=Y_t(i)^2$.
\end{itemize}
\subsection{Proofs of results in \S\ref{app:appl}}\label{app:appl-proofs}
This section first collects the proofs of the results in \S\ref{se:mom}, and then those for \S\ref{app:q-learning}. We begin with an auxiliary lemma that records two consequences of \eqref{eq:mom-averaging}.
\begin{lem}\label{lem:mom-aux}
Grant the notation of \S\ref{se:mom-opt}. Then, for every $t\ge1$, the following hold.
\begin{enumerate}
\item[(i)] $\|x_t-x^\star\|\le\max\{\|x_0-x^\star\|,\max_{1\le s\le t}\|\varphi_s\|\}\le\max\{D,\max_{0\le s<t}Z_s^{1/2}\}$.
\item[(ii)] $F(x_t)-F^\star\le\bigl[t(F(x_{t-1})-F^\star)+L\|\varphi_t\|^2/2\bigr]/(t+1)$, and consequently
\[
\|\nabla F(x_t)\|^2\le c_tZ_{t-1},\ c_t:=\frac{L^2}{t+1}+\frac{L\sqrt t}{2(t+1)\sqrt{\eta_{t-1}}}.
\]
\end{enumerate}
\end{lem}
\begin{proof}[Proof of Lemma~\ref{lem:mom-aux}]
By \eqref{eq:mom-averaging} and $\varphi_s=z_s-x^\star$, it follows $x_t-x^\star=(t+1)^{-1}\bigl[(x_0-x^\star)+\sum_{s=1}^t\varphi_s\bigr]$; therefore, (i) follows from the triangle inequality, $\|x_0-x^\star\|\le D$ and $\|\varphi_s\|^2\le Z_{s-1}$. For (ii), the identity $(t+1)x_t=tx_{t-1}+(x^\star+\varphi_t)$ exhibits $x_t$ as a convex combination of $x_{t-1}$ and $x^\star+\varphi_t$. Convexity of $F$, followed by \eqref{eq:descent} with $x=x^\star$, $y=x^\star+\varphi_t$ and $\nabla F(x^\star)=0$, gives the first inequality. Next, \eqref{eq:descent} with $y=x-\nabla F(x)/L$ yields $\|\nabla F(x)\|^2\le2L(F(x)-F^\star)$ for every $x\in\R^d$. Therefore,
\[
\|\nabla F(x_t)\|^2\le\frac{L^2}{t+1}\|\varphi_t\|^2+\frac{L\sqrt t}{2(t+1)\sqrt{\eta_{t-1}}}\cdot4\sqrt{t\eta_{t-1}}\,\bigl(F(x_{t-1})-F^\star\bigr)\le c_tZ_{t-1},
\]
since both components of $Z_{t-1}$ in \eqref{eq:mom-lyap} are nonnegative.
\end{proof}
\begin{proof}[Proof of Proposition~\ref{prop:mom-rec}]
Fix $t\ge1$. By \eqref{eq:mom-averaging}, $\varphi_{t+1}=\varphi_t-2\sqrt{ \eta_t/t}\,g_t$. Therefore, substituting $g_t=\nabla F(x_t)-\xi_t$ one obtains
\begin{equation}\label{eq:mom-pf-expand}
\|\varphi_{t+1}\|^2=\|\varphi_t\|^2-4\sqrt{ \eta_t/t}\,\langle\nabla F(x_t),\varphi_t\rangle+b_t+\langle v_t,\xi_t\rangle.
\end{equation}
Since $\varphi_t=(x_t-x^\star)+t(x_t-x_{t-1})$, the interpolation inequality \eqref{eq:interpolation}, together with convexity in the form $\langle\nabla F(x_t),x_t-x_{t-1}\rangle\ge F(x_t)-F(x_{t-1})$, gives
\[
\langle\nabla F(x_t),\varphi_t\rangle\ge F(x_t)-F^\star+\frac1{2L}\|\nabla F(x_t)\|^2+t\big(F(x_t)-F(x_{t-1})\big).
\]
Plugging this into \eqref{eq:mom-pf-expand}, by definition of $Z_t$ we can write
\begin{align*}
Z_t-Z_{t-1}\le{}&b_t+\langle v_t,\xi_t\rangle-\frac2L\sqrt{ \eta_t/t}\,\|\nabla F(x_t)\|^2+\underbrace{4\sqrt{\eta_t}\Big(\sqrt{t+1}-\sqrt t-\frac1{\sqrt t}\Big)\big(F(x_t)-F^\star\big)}_{:=A}\\
&+\underbrace{4\sqrt t\big(\sqrt{\eta_t}-\sqrt{\eta_{t-1}}\big)\big(F(x_{t-1})-F^\star\big)}_{:=B}.
\end{align*}
Now we deliver the coup de grâce for this proof. Firstly, $\sqrt{t+1}-\sqrt t\le1/(2\sqrt t)$ controls $A$. Secondly, by definition of $a_t$, 
\[ B \le  4(a_t-1)\sqrt{t\eta_{t-1}}\big(F(x_{t-1})-F^\star\big) \overset{(a)}{\le} (a_t-1)Z_{t-1}, \]
where (a) follows by \eqref{eq:mom-lyap}. This controls $B$ and completes the proof.
\end{proof}
\begin{proof}[Proof of Theorem~\ref{thm:mom-coverage}]
We verify the conditions of Lemma~\ref{lem:observable-envelope}, with the oracle boundary furnished by the proof of Theorem~\ref{thm:obs-comp}. By \S\ref{se:mom-opt}, $Z_0\le R_0$ surely, and by Proposition~\ref{prop:mom-rec}, \eqref{eq:key-recursion} holds almost surely for every $t\ge1$, with $a_t, b_t$ and $v_t$ as in \eqref{eq:mom-abv}. Write $\wt\bv:=(v_s/A_s)_{s\ge1}$, and let $\Mcal_\alpha(t;\wt\bv)$ denote the oracle boundary $B^\diamond_t$ in the proof of Theorem~\ref{thm:obs-comp} matching the chosen $\diamond\in\{\mathrm{mix},\mathrm{dm},\St\}$ respectively, evaluated at $\wt\bv$ and $(\Delta V_s)_{s\ge1}$. These results give $\IP\big(Y_t(\wt\bv)\le\Mcal_\alpha(t;\wt\bv)\text{ for all }t\ge1\big)\ge1-\alpha$.

It remains to verify \eqref{eq:star}. Fix $t\ge1$ and work on $\bigcap_{s<t}\{Z_s\le R_s\}$. For $s\le t$, $$\|v_s\|/A_s=4\sqrt{\eta_s/s}\,\|\varphi_s\|/A_s\le4\sqrt{\eta_s/sZ_{s-1}}/A_s\le M_s.$$ Therefore, $\mathsf E_t(M)$ of Theorem~\ref{thm:obs-comp} holds for $\wt\bv$. On $\mathsf E_t(M)$, \eqref{eq:stitching-obsv}--\eqref{eq:mixing-obsv} give $\Mcal_\alpha(t;\wt\bv)\le\mathcal U^\diamond_\alpha(t;M,\nu)\le U_t$, which is \eqref{eq:star}. Consequently, Lemma~\ref{lem:observable-envelope}, with $\alpha_0=0$, yields $\IP(Z_t\le R_t\text{ for all }t\ge0)\ge1-\alpha$. Finally, $F(x_t)-F^\star\le Z_t/\bigl[4\sqrt{(t+1)\eta_t}\bigr]$ by \eqref{eq:mom-lyap}, which transfers the envelope to $\wh\Delta_t$ on the same event.
\end{proof}
\begin{proof}[Proof of Proposition~\ref{prop:mom-quad}]
The proof of Proposition~\ref{prop:mom-quad} follows those of Propositions~\ref{prop:ls-quadratic-betting} and~\ref{prop:ls-hybrid}, with one difference. In \S\ref{se:ls-betting}, the bound $\|v_s\|\le\eta_sG$ held surely, whereas here $\|v_s\|/A_s\le M_s$ only holds until the first crossing of the envelope. Accordingly, the comparison between the oracle and the observable mixtures is carried out on the event $\mathsf E_t(M)$, exactly as in Theorem~\ref{thm:obs-comp}. In particular, following the proof of Proposition~\ref{prop:convex-hybrid}, we construct an oracle boundary from a mixture of quadratic product-betting supermartingales, dominate it by $U^{\mathrm Q}_t$ on $\mathsf E_t(M)$, and conclude via Lemma~\ref{lem:observable-envelope}. 

For notational ease, let $X_s:=\langle\widetilde v_s,\xi_s\rangle$, and $Y_t(\bv)= Y_t=\sum_{s=1}^t X_s$. For $\lambda>0$, let
\[
K^{\mathrm Q}_t(\lambda):=\prod_{s=1}^t\frac{1+\lambda X_s+\lambda^2X_s^2/3}{1+\lambda^2\sigma^2\|\wt v_s\|^2/3}, \  K^{\mathrm Q}_0(\lambda)=1.
\]
Since $\wt v_s$ is predictable, $\IE_{s-1}X_s=0$ and $\IE_{s-1}X_s^2\le\|\wt v_s\|^2\,\IE_{s-1}\|\xi_s\|^2\le\sigma^2\|\wt v_s\|^2$. Therefore, as in the proof of Proposition~\ref{prop:ls-quadratic-betting}, $K^{\mathrm Q}_t(\lambda)$ is a nonnegative supermartingale and so is $K_t:=\sum_{k\ge0}p_kK^{\mathrm Q}_t(\lambda_k)$, with $K_0\le1$. Ville's inequality (Lemma~1 of \citealp{howard2018}) yields $\IP(\Omega_\alpha)\ge1-\alpha$, where $\Omega_\alpha:=\{K_t<1/\alpha\text{ for all }t\ge1\}$. 

Write $\log K^{\mathrm Q}_t(\lambda)=\lambda Y_t-\wt J_t(\lambda)$, where $\wt J_t(\lambda):=\sum_{s\le t}\big[\chi(\lambda X_s)+\log(1+\lambda^2\sigma^2\|\wt v_s\|^2/3)\big]$, and put
\[
w_t(y):=\sum_{k\ge0}p_k\exp\bigl[\lambda_ky-\wt J_t(\lambda_k)\bigr],\ \Mcal_\alpha(t;\wt\bv):=\sup\{y\in\R:w_t(y)<1/\alpha\}.
\]
Since $w_t(Y_t)=K_t$, from the definition of $\Omega_{\alpha}$, it holds that $$\IP\big(Y_t(\wt\bv)\le\Mcal_\alpha(t;\wt\bv)\text{ for all }t\ge1\big)\ge1-\alpha.$$ This establishes the first condition for an application of Lemma \ref{lem:observable-envelope}. Now we establish \eqref{eq:star}.

Fix $t\ge1$ and work on $\bigcap_{s<t}\{Z_s\le R^{\mathrm Q}_s\}$. As in the proof of Theorem~\ref{thm:mom-coverage}, $\|\wt v_s\|\le M_s$ for $s\le t$. Moreover, $\|\xi_s\|\le\|\nabla F(x_s)\|+\|g_s\|\le G_s+\|g_s\|$, so that $$|\lambda X_s|\le\lambda M_s(G_s+\|g_s\|) \implies \chi(\lambda X_s)\le j(\lambda M_s(G_s+\|g_s\|)).$$
Moreover, $\log(1+\lambda^2\sigma^2\|\wt v_s\|^2/3)\le\log(1+\lambda^2\sigma^2M_s^2/3)$. Therefore, $\wt J_t\le J^{\mathrm Q}_t$, which yields $w_t\ge W^{\mathrm Q}_t$, and in turn,  $$\{y:w_t(y)<1/\alpha\}\subseteq\{y:W^{\mathrm Q}_t(y)<1/\alpha\}, \ \text{and } \Mcal_\alpha(t;\wt\bv)\le U^{\mathrm Q}_t.$$ This establishes \eqref{eq:star}, and Lemma~\ref{lem:observable-envelope} establishes that $R_t^\mathrm{Q}$ is an anytime-valid envelope of $Z_t$.

Finally, using $j(b)\le b^2/6$ and $\log(1+u)\le u$,
\[
J^{\mathrm Q}_t(\lambda)\le\sum_{s=1}^t\lambda^2M_s^2\biggl[\frac{(G_s+\|g_s\|)^2}6+\frac{\sigma^2}3\biggr]=\frac{\lambda^2}2V^{\mathrm{SN}}_t,
\]
whence $W^{\mathrm Q}_t(y)\ge\sum_kp_k\exp(\lambda_ky-\lambda_k^2V^{\mathrm{SN}}_t/2)$, and $U^{\mathrm Q}_t\le U^{\mathrm{SN,DM}}_t$ at common $(M_s)_{s\le t}$. Furthermore, $U^{\mathrm{SN,DM}}_t\le U^{\mathrm{SN,St}}_t$, since a discrete mixture dominates each of its components (\S\ref{app:stitching}). Induction on $t$ now yields \eqref{eq:mom-ordering}.
\end{proof}
\begin{proof}[Proof of Theorem~\ref{thm:mom-rates}]
Since $\eta_t$ is nonincreasing, $A_t=1$ for all $t$. Recall $c_t$ from Lemma~\ref{lem:mom-aux}. We proceed along the lines of the proof of \citet[Lemma~5]{feng2025breaking}, equivalently \citet[Lemma~6]{feng2023anytime}.

Let $\phi_d$ denote the standard Gaussian density on $\R^d$. As \[
\exp(\lambda\|x\|^2)=\int_{\R^d}\exp(\sqrt{2\lambda}\langle u,x\rangle)\phi_d(u)\,du
\] for $x\in\R^d$ and $\lambda>0$, Tonelli's theorem gives, for every $t\ge1$,
\[
\IE\exp\Big(\frac{\|\xi_t\|^2}{4\sigma_N^2}\Big)=\int_{\R^d}\IE\Big[\IE_{t-1}\exp\Big\langle\frac{u}{\sqrt2\,\sigma_N},\xi_t\Big\rangle\Big]\phi_d(u)\,du\overset{(a)}{\le}\int_{\R^d}e^{\|u\|^2/4}\phi_d(u)\,du=2^{d/2},
\]
where (a) follows from the sub-Gaussian condition with $\theta=u/(\sqrt2\,\sigma_N)$. Now fix $T\ge1$, and put $w_t:=\eta_t/(tH_p)$. It is immediate that 
\[
\IE\exp\Big(\sum_{t\le T}\frac{w_t\|\xi_t\|^2}{4\sigma_N^2}\Big)\le2^{d/2}\sum_{t\le T}w_t+1-\sum_{t\le T}w_t\le2^{d/2}.
\]
where we use $\sum_{t\le T}w_t\le1$ by Lemma~\ref{lem:mom-constants}. 
Therefore, by Markov's inequality, for every $T\ge1$,
\[
\IP\Big(\sum_{t\le T}\frac{\eta_t}{t}\|\xi_t\|^2>2\sigma_N^2H_p\Big(d\log2+2\log\frac1\delta\Big)\Big)\le\delta,
\]
and continuity from below of $\IP$ gives $\IP(\Omega_\xi)\ge1-\delta$, where
\[
\Omega_\xi:=\Big\{\sum_{t\ge1}\frac{\eta_t}{t}\|\xi_t\|^2\le2\sigma_N^2H_p\Big(d\log2+2\log\frac1\delta\Big)\Big\}.
\]
On the other hand, $(\xi_t,\sigma_N^2I_d)_{t\ge0}$ is a sub-$\psi_N$ process, so that Theorem~\ref{thm:mom-coverage} shows that $\Omega_R:=\{Z_t\le R_t\text{ for all }t\ge0\}$ satisfies $\IP(\Omega_R)\ge1-\alpha$. Consequently, $\IP(\Omega_R\cap\Omega_\xi)\ge1-\alpha-\delta$, and we work on $\Omega_R\cap\Omega_\xi$ for the rest of the proof.

We now bound the two components $\sum_{s=1}^t\frac{4\eta_s\|g_s\|^2}{sA_s}$ and $U_t$ from \eqref{eq:mom-envelope}. Firstly, $V^N_t=16\sigma_N^2\sum_{s\le t}(\eta_s/s)R_{s-1}$. For both the $\mathrm{dm}$ and \St\ choices of $U_t$,  we obtain
\begin{align}
    U_t\le\frac{\log\bigl[1/(\alpha p_0)\bigr]}{\lambda_0}+\frac{\lambda_0V^N_t}2=8\sigma_N^2H_p\log\frac1{\alpha p_0}+\frac1{H_p}\sum_{s=1}^t\frac{\eta_s}{s}R_{s-1}. \label{eq:msgd-u}
\end{align}
Secondly, on $\Omega_R$, Lemma~\ref{lem:mom-aux}(ii) gives $\|\nabla F(x_s)\|^2\le c_sZ_{s-1}\le c_sR_{s-1}$, so that $\|g_s\|^2\le2\|\nabla F(x_s)\|^2+2\|\xi_s\|^2\le2c_sR_{s-1}+2\|\xi_s\|^2$. Together with the definition of $\Omega_\xi$, this yields
\begin{align}
    \sum_{s=1}^t\frac{4\eta_s}{s}\|g_s\|^2\le8\sum_{s=1}^t\frac{c_s\eta_s}{s}R_{s-1}+16\sigma_N^2H_p\Big(d\log2+2\log\frac1\delta\Big). \label{eq:msgd-g}
\end{align}
Plugging \eqref{eq:msgd-u}-\eqref{eq:msgd-g} back into \eqref{eq:mom-envelope}, we obtain, for all $t\ge1$,
\[
R_t\le C+\sum_{s=1}^td_sR_{s-1},\ C:=e^{-8C_\circ-1}K_{\alpha,\delta},\ d_s:=\Big(8c_s+\frac1{H_p}\Big)\frac{\eta_s}{s},
\]
together with $R_0\le C$. Hence, the discrete Grönwall inequality \citep{clark1987short} yields, for all $t\ge1$,
\[
R_t\le C\prod_{s\le t}(1+d_s)\overset{(b)}{\le}C\exp\Big(\sum_{s\ge1}d_s\Big)\overset{(c)}{\le}K_{\alpha,\delta},
\]
where (b) uses $1+u\le e^u$, and (c) follows from Lemma \ref{lem:mom-constants}. Finally, $\wh\Delta_t=R_t/\bigl[4\sqrt{(t+1)\eta_t}\bigr]\le K_{\alpha,\delta}\log^{p/2}(t+2)/\bigl[4\sqrt{\eta_\circ(t+1)}\bigr]$, while $Z_t\le R_t$ and $F(x_t)-F^\star\le\wh\Delta_t$ hold on $\Omega_R$ by Theorem~\ref{thm:mom-coverage}. This completes the proof.
\end{proof}
\begin{proof}[Proof of Corollary~\ref{cor:mom-fv}]
As before, $A_t=1$ and $\sum_t \eta_t/t<\infty$ by Lemma \ref{lem:mom-constants}. Consider $R_t=R^{\mathrm{SN,St}}_t$. As $M_s^2=16\eta_s/sR_{s-1}$, similar to the proof of Theorem \ref{thm:mom-rates} we obtain $U_t\le\log\bigl[1/(\alpha p_0)\bigr]/\lambda_0+8\lambda_0\sum_{s\le t}\eta_s/s\nu^{\mathrm{SN}}_sR_{s-1}$, so that
\[
R_t\le C+\sum_{s=1}^t8\lambda_0\eta_s/s\nu^{\mathrm{SN}}_sR_{s-1},\ C:=R_0+4\sum_{s\ge1}\eta_s/s\|g_s\|^2+\frac{\log\bigl[1/(\alpha p_0)\bigr]}{\lambda_0}<\infty\ \text{ a.s.}
\]
Then, the discrete Grönwall inequality \citep{clark1987short} yields $$\sup_tR_t\le C\exp\biggl[8\lambda_0\sum_{s\ge1}\eta_s/s\nu^{\mathrm{SN}}_s\biggr]<\infty \ \text{almost surely.}$$ By \eqref{eq:mom-ordering}, the same holds for $R^{\mathrm{SN,DM}}_t$ and $R^{\mathrm Q}_t$. Observing that $$\wh\Delta_t\le\sup_sR_s\cdot\log^{p/2}(t+2)/\bigl[4\sqrt{\eta_\circ(t+1)}\bigr],$$ completes the proof.
\end{proof}

The proof of Proposition~\ref{prop:q-rec} requires the following contracting lemma.
\begin{lem}\label{lem:q-lyap}
For every $Q\in\R^N$, with $e:=Q-Q^\star$, and every $i\in[N]$,
\[
e_i(Q(i)-(\mathcal TQ)(i))
\ge
\Big(1-\frac\gamma2\Big)e_i^2
-\frac\gamma2\|e\|_\infty^2.
\]
\end{lem}

\begin{proof}[Proof of Lemma~\ref{lem:q-lyap}]
Let $b:=\mathcal TQ-\mathcal TQ^\star$, so that
$Q-\mathcal TQ=e-b$ and $\|b\|_\infty\le\gamma\|e\|_\infty$. Therefore,
\[
e_i(e_i-b_i)
\ge e_i^2-\gamma|e_i|\|e\|_\infty
\ge
\Big(1-\frac\gamma2\Big)e_i^2
-\frac\gamma2\|e\|_\infty^2,
\]
where the last inequality uses
$2|e_i|\|e\|_\infty\le e_i^2+\|e\|_\infty^2$.
\end{proof}

\begin{proof}[Proof of Proposition~\ref{prop:q-rec}]
Fix $t\ge1$ and $i\in[N]$. Expanding the square gives
\[
Z_t(i)=Z_{t-1}(i)
+\eta_te_{t-1}(i)Y_t(i)
+\frac{\eta_t^2}{2}Y_t(i)^2.
\]
Substituting, 
$Y_t(i)=d_t(i)((\mathcal TQ_{t-1})(i)-Q_{t-1}(i))+\xi_t(i)$, we invoke Lemma~\ref{lem:q-lyap} to bound the conditional-mean term by
\[
\eta_te_{t-1}(i)\IE_{t-1}Y_t(i)
\le
-(2-\gamma)\eta_td_t(i)Z_{t-1}(i)
+\gamma\eta_td_t(i)\mathbf{Z}_{t-1}.
\]
Since $\xi_t(i)=d_t(i)\xi_t(i)$, the remaining linear term is
$v_t(i)\xi_t(i)$. Combining these observations proves
\eqref{eq:q-coordinate-rec}.
\end{proof}

\begin{proof}[Proof of Theorem~\ref{thm:q}]
The bound $\|Q^\star\|_\infty\le B^\star$ gives
$Z_0(i)\le R_0(i)$ for every $i$. Put
\[
S_t(i):=\sum_{s=1}^t\frac{v_s(i)\xi_s(i)}{A_s(i)},
\ 
\mathsf E_t(i):=
\bigcap_{s=1}^t
\biggl\{\frac{|v_s(i)|}{A_s(i)}\le M_s(i)\biggr\}.
\]
The weights $v_s(i)/A_s(i)$ and their observable bounds $M_s(i)$ are predictable. Applying Theorem~\ref{thm:obs-comp} to the scalar process $\xi_s(i)$, or Theorem~\ref{thm:obs-betting} with $g_s=Y_s(i)$ and $H_s\equiv0$, yields
\[
\IP\Big(
\bigcup_{t\ge1}
\big[\mathsf E_t(i)\cap\{S_t(i)>U_t(i)\}\big]
\Big)\le\alpha_i.
\]
A union bound over $i$ gives an event $\Omega_\alpha$ of probability at least $1-\alpha$ on which, simultaneously for every $t$ and $i$, $\mathsf E_t(i)$ implies $S_t(i)\le U_t(i)$.

Subsequently, we work on $\Omega_\alpha$ and suppose
$\tau:=\inf\{t\ge1:Z_t(i)>R_t(i)\text{ for some }i\}<\infty$.
Choose a coordinate $i$ such that $Z_\tau(i)>R_\tau(i)$. Note that \eqref{eq:q-observable-coefficients} implies $\mathsf E_\tau(i)$. Moreover, Proposition~\ref{prop:q-rec}, the choice of $a_s(i)$ and $\mathbf{Z}_{s-1}\le\mathbf{R}_{s-1}$ give
\[
Z_s(i)\le
a_s(i)Z_{s-1}(i)+b_s(i)+v_s(i)\xi_s(i),
\  1\le s\le\tau,
\]
which implies
\[
\begin{split}
\frac{Z_\tau(i)}{A_\tau(i)}
&\le
Z_0(i)+\sum_{s=1}^\tau\frac{b_s(i)}{A_s(i)}+S_\tau(i)\le
R_0(i)+\sum_{s=1}^\tau\frac{b_s(i)}{A_s(i)}+U_\tau(i)
=\frac{R_\tau(i)}{A_\tau(i)},
\end{split}
\]
a contradiction. Consequently, all coordinate bounds hold simultaneously, and taking their maximum gives
$\|Q_t-Q^\star\|_\infty\le\sqrt{2\mathbf{R}_t}$.
For $\gamma>0$, the policy bound follows from
\citet[Corollary~15]{evendar2006action}, with their approximation
$\widetilde Q=Q_t$ and error tolerance $\sqrt{2\mathbf{R}_t}$.
For $\gamma=0$, the same conclusion follows from
$\Vcal^\star(s)-\Vcal^{\pi_t}(s)
=\max_aQ^\star(s,a)-Q^\star(s,\pi_t(s))
\le2\|Q_t-Q^\star\|_\infty$.
\end{proof}

\section{Proofs of results in \S\ref{se:prob}} \label{app:se2proofs}

\begin{proof}[Proof of Lemma~\ref{lem:observable-envelope}]
Let
\[
\Omega_\alpha:=\bigcap_{t\ge1}\{Y_t(\widetilde{\bv})\le\Mcal_\alpha(t;\widetilde{\bv})\},\  \IP\big(\Omega_\alpha\cap\{Z_0\le R_0\}\big)\ge1-\alpha-\alpha_0.
\]
Subsequently, we work inside $\Omega_\alpha\cap\{Z_0\le R_0\}$. Suppose that $\tau:=\inf\{t\ge1:Z_t>R_t\}<\infty$. Then $Z_s\le R_s$ for $0\le s<\tau$ and $Z_\tau>R_\tau$. Therefore, using $a_s/A_s=1/A_{s-1}$,
\[
\frac{Z_s}{A_s}\le\frac{Z_{s-1}}{A_{s-1}}+\frac{b_s}{A_s}+\langle v_s/A_s,\xi_s\rangle.
\]
Summing over $1\le s\le\tau$ and subtracting $R_\tau/A_\tau=R_0+\sum_{s=1}^\tau b_s/A_s+U_\tau$, we get
\[
\frac{Z_\tau-R_\tau}{A_\tau}\le Z_0-R_0+Y_\tau(\widetilde{\bv})-U_\tau\le Y_\tau(\widetilde{\bv})-U_\tau,
\]
as $Z_0 \le R_0$. 
Next, \eqref{eq:star} at $t=\tau$ gives $\Mcal_\alpha(\tau;\widetilde{\bv})\le U_\tau$. Consequently,
\[
0<\frac{Z_\tau-R_\tau}{A_\tau}\le Y_\tau(\widetilde{\bv})-U_\tau\le\Mcal_\alpha(\tau;\widetilde{\bv})-U_\tau\le0,
\]
a contradiction. Therefore
\[
\IP\big(Z_t\le R_t\text{ for all }t\ge0\big)\ge\IP\big(\Omega_\alpha\cap\{Z_0\le R_0\}\big)\ge1-\alpha-\alpha_0.
\]
\end{proof}

\begin{proof}[Proof of Theorem~\ref{thm:obs-comp}]
Fix $t\ge1$ and write $Q_t(\bv,\lambda):=\sum_{s=1}^t\Gamma_\psi(\Delta V_s,\lambda v_s)$. Since $\psi$ is convex with $\psi'(0)=0$, it is nondecreasing on $[0,\lambda_{\max})$. Note that on $\mathsf E_t(M)$, $\sup_{s\le t}\|v_s\|\le\max_{s\le t}M_s$, which entails
\begin{align*}
    \Lambda_t(M)\subseteq\Lambda_t(\bv), \  \lambda\|v_s\|\le\lambda M_s<\lambda_{\max}\ \text{ for all }\lambda\in\Lambda_t(M),\ s\le t.
\end{align*}
Therefore, in view of $\Delta V_s\preceq\nu_sI$, on $\mathsf E_t(M)$ and for every $\lambda\in\Lambda_t(M)$,
\begin{align}\label{eq:obs-comp-pf1}
    Q_t(\bv,\lambda)=\sum_{s=1}^t\psi(\lambda\|v_s\|)\frac{\langle v_s,\Delta V_sv_s\rangle}{\|v_s\|^2}\le\sum_{s=1}^t\nu_s\psi(\lambda\|v_s\|)\le\sum_{s=1}^t\nu_s\psi(\lambda M_s)=K_t(\lambda).
\end{align}
Consider the mixing boundaries $\mathcal M_{\alpha}$ and $m_{\alpha}$ from Theorem \ref{thm:prob}, the discrete mixture boundary $\DMcal_{\alpha}$ from Corollary \ref{corol:dm}, and the stitching boundary $\St_{\alpha}$ from Proposition \ref{prop:stitch}.
Plugging \eqref{eq:obs-comp-pf1} into each construction, and noting that the integrands are nonnegative, it follows that on $\mathsf E_t(M)$, for every $x\in\R$,
\begin{align}
&\begin{aligned}
    \int_{\Lambda_t(M)}e^{\lambda x-K_t(\lambda)}\,dF(\lambda)&\le\int_{\Lambda_t(\bv)}e^{\lambda x-Q_t(\bv,\lambda)}\,dF(\lambda)=m_\alpha(x,t;\bv,F),
\end{aligned} \nonumber\\
&\begin{aligned}
    \sum_{k:\lambda_k\in\Lambda_t(M)}p_ke^{\lambda_kx-K_t(\lambda_k)}&\le\sum_{k:\lambda_k\in\Lambda_t(\bv)}p_ke^{\lambda_kx-Q_t(\bv,\lambda_k)},
\end{aligned} \nonumber\\
&\begin{aligned}
B_t^{\St}:= \St_\alpha(t;\bv)&=\inf_{k:\lambda_k\in\Lambda_t(\bv)}\frac{\log\frac{1}{\alpha p_k}+Q_t(\bv,\lambda_k)}{\lambda_k}\\
&\le\inf_{k:\lambda_k\in\Lambda_t(M)}\frac{\log\frac{1}{\alpha p_k}+K_t(\lambda_k)}{\lambda_k}\\
&=\mathcal U^{\St}_\alpha(t;M,\nu).
\end{aligned} \label{eq:stitching-obsv}
\end{align}
Inverting the first two displays yield 
\begin{align}
    B_t^{\mathrm{mix}}:=\Mcal_{\alpha}(t; \bv, F)\le\mathcal U^{\mathrm{mix}}_\alpha(t;M,\nu)\ \text{ and } B_t^{\mathrm{dm}}:=\DMcal_{\alpha}(t; \bv)\le\mathcal U^{\mathrm{dm}}_\alpha(t;M,\nu) \text{ on } \mathsf E_t(M). \label{eq:mixing-obsv}
\end{align}
Since $t\ge1$ was arbitrary, it follows that \eqref{eq:stitching-obsv}-\eqref{eq:mixing-obsv} hold for all $t\ge1$ simultaneously. Consequently,
\begin{align*}
    \bigcup_{t\ge1}\big[\mathsf E_t(M)\cap\{Y_t(\bv)>U_t^\diamond\}\big]\subseteq\bigcup_{t\ge1}\{Y_t(\bv)>B^\diamond_t\},
\end{align*}
which completes the proof in light of Theorem~\ref{thm:prob}, Corollary~\ref{corol:dm} and Proposition~\ref{prop:stitch}, for $\diamond=\mathrm{mix},\mathrm{dm},\St$, respectively. 
\end{proof}
\begin{proof}[Proof of Corollary~\ref{cor:closed-form}]
Recall $\Lambda_t(M)$ and $K_t(\lambda)$ from Theorem \ref{thm:obs-comp}. For (i) and (ii), since $\lambda_{\max}=\infty$ for $\psi_N$, hence $\Lambda_t(M)=[0,\infty)$ and $K_t(\lambda)=\lambda^2V_t/2$. 

(i) The function \[
x\mapsto\int e^{\lambda x-\lambda^2V_t/2}\,dF(\lambda;\rho)
\] is the mixture in Theorem~\ref{thm:prob} with $\sum_{s\le t}\Gamma_{\psi_N}(\Delta V_s,\lambda v_s)$ replaced by $\lambda^2V_t/2$, i.e., with $Q_t(\bv)$ replaced by $V_t$. Hence (i) follows from Proposition~6 of \citet{howard2021time}, which applies with $Q_t(\bv)=V_t$. 

(ii) Put $a:=\log(2/\alpha)$ and $L_k:=\log(1/(\alpha p_k))$, $k\ge0$. Recall that $K_t(\lambda)=\lambda^2V_t/2$; therefore, the definition of $\mathcal U^{\St}_\alpha$ in Theorem~\ref{thm:obs-comp} lets us rewrite
$$\mathcal U^{\St}_\alpha(t;M,\nu)=\inf_{k\ge0}\Big\{\frac{L_k}{\lambda_k}+\frac{\lambda_kV_t}{2}\Big\}.$$
Consequently, it suffices to sharply control the above infimum by an index $k$. To that end, fix $x\ge0$, and put
$$h:=a+\log\log(e+q_\alpha(x)),\ \text{ and, } k_x:=\lfloor\log(q_\alpha(x)/h)\rfloor.$$
Observe that $q_\alpha(x)\ge2a+2$ and $\log\log(e+q_\alpha(x))\le q_\alpha(x)/2$ entail $1/2<h\le q_\alpha(x)$, so that $k_x\ge0$. Therefore, by the definition of $\lambda_k$,
$$\sqrt{h/q_\alpha(x)}\le\lambda_{k_x}<\sqrt{eh/q_\alpha(x)},\ \text{and}\ k_x+1\le1+\log(2q_\alpha(x))\overset{(a)}{\le}2\log(e+q_\alpha(x)),$$
where (a) follows from $2eq_\alpha(x)\le(e+q_\alpha(x))^2$. This, in turn, provides
$$L_{k_x}=\log\frac{\pi^2}{6\alpha}+2\log(k_x+1)\overset{(b)}{\le}a+2\log2+2\log\log(e+q_\alpha(x))\overset{(c)}{\le}3h,$$
where (b) uses $\pi^2/6<2$, and (c) uses $a\ge\log2$ along with $\log\log(e+q_\alpha(x))\ge0$. Evaluating the infimum at $k=k_x$, we obtain
$$\mathcal U^{\St}_\alpha(t;M,\nu)\le\frac{L_{k_x}}{\lambda_{k_x}}+\frac{\lambda_{k_x}V_t}{2}\le3h\sqrt{\frac{q_\alpha(x)}h}+\frac{V_t}2\sqrt{\frac{eh}{q_\alpha(x)}}=\Big(3+\frac{\sqrt e\,V_t}{2q_\alpha(x)}\Big)\sqrt{q_\alpha(x)h}.$$
Finally, note that $\sqrt{q_\alpha(x)h}=\mathfrak S_\alpha(x)$ and $q_\alpha(x)\ge x$. Choosing $x=V_t$ yields $\mathcal U^{\St}_\alpha(t;M,\nu)\le(3+\sqrt e/2)\mathfrak S_{\alpha}(V_t)\le4\mathfrak S_{\alpha}(V_t)$, whereas choosing $x=V_t/4$, for which $V_t/q_\alpha(V_t/4)\le4$, yields $$\mathcal U^{\St}_{\alpha}(t;M,\nu)\le(3+2\sqrt e)\mathfrak S_{\alpha}(V_t/4)\le7\mathfrak S_\alpha(V_t/4).$$ Combining the two bounds proves (ii).

(iii) Fix $t\ge1$, and put $T_k:=L_k/\lambda_k+\lambda_kV_t$
for $k\ge0$ and $k_c:=\lceil2\log_+(2c_t)\rceil$.
For $k\ge k_c$, $\lambda_kc_t\le1/2$, so
$\lambda_k\in\Lambda_t(M)$ and
$K_t(\lambda_k)\le
\sum_{s=1}^t\nu_s\lambda_k^2M_s^2/\bigl[2(1-G\lambda_kM_s)\bigr]
\le\lambda_k^2V_t$.
Hence $\mathcal U_\alpha^{\mathrm{St}}(t;M,\nu)
\le\min_{k\ge k_c}T_k$.

We first control the unrestricted minimum. Take $x=V_t$ in the
notation of (ii). Since $h>1$ and
$k_{V_t}+1\le1+\log q\le(3/2)\log(e+q)$, we have
$$L_{k_{V_t}}\le
\log\bigl[3\pi^2/(8\alpha)\bigr]+2\log\log(e+q)\le2h.$$
Together with
$\sqrt{h/q}\le\lambda_{k_{V_t}}<\sqrt{eh/q}$ and $V_t\le q$,
this gives
$$T_{k_{V_t}}\le(2+\sqrt e\,V_t/q)\sqrt{qh}
\le4\mathfrak S_\alpha(V_t).$$
Since $T_k\to\infty$, its minimum is attained at some $k^\star\ge0$.
If $k^\star\ge k_c$, the preceding bound proves the claim.
Otherwise, $k_c\ge1$,
$\lambda_{k_c}>e^{-1/2}/(2c_t)$, and
$k_c+1\le2+2\log(2c_t)$. Consequently,
$$L_{k_c}\le
\log\bigl[\pi^2/(6\alpha)\bigr]+2\log\bigl[2+2\log_+(2c_t)\bigr].$$
Moreover,
$\lambda_{k_c}V_t\le\lambda_{k^\star}V_t\le T_{k^\star}$.
Therefore
$\mathcal U_\alpha^{\mathrm{St}}(t;M,\nu)
\le T_{k_c}
\le2\sqrt e\,c_tL_{k_c}+T_{k^\star}
\le\mathfrak G_\alpha(V_t,c_t)$,
which completes the proof.
\end{proof}

\begin{proof}[Proof of Theorem~\ref{thm:obs-betting}]
Note that $Y_t(\bv)=\sum_{s=1}^t\langle v_s,g_s-\IE_{s-1}g_s\rangle$. Therefore Corollary~\ref{corol:betting}, with $c=G$ and the predictable sequence $\bv$, yields
\begin{align}\label{eq:obs-bet-pf0}
    \IP\big(Y_t(\bv)\le\mathcal B^1_\alpha(t;\bv)\text{ for all }t\ge1\big)\ge1-\alpha, \  \Lambda_t(\bv)=\Big[0,G^{-1}\wedge\big(G\sup_{s\le t}\|v_s\|\big)^{-1}\Big).
\end{align}
Write $\Psi_t(\lambda):=\sum_{s=1}^t\psi_{E,1}\big(-\lambda\langle v_s,\IE_{s-1}g_s\rangle\big)$. Subsequently, we work inside the set
\begin{align*}
    \mathcal A:=\bigcap_{s\ge1}\{\|g_s\|\le G\}\cap\bigcap_{t\ge1}\Big[\mathsf E_t(M)^c\cup\Big\{H_t(\lambda,z)\le\Psi_t(\lambda)\ \text{for all }\lambda\in\Lambda_t^G(M),\ 0\le z\le Z_t\Big\}\Big].
\end{align*}
It follows from $\|g_s\|\le G$ almost surely, \eqref{eq:obs-bet-mean-correction}, and countable subadditivity that $\IP(\mathcal A)=1$.

 Fix $t\ge1$. On $\mathsf E_t(M)$, $\sup_{s\le t}\|v_s\|\le\max_{s\le t}M_s$, which entails $\Lambda_t^G(M)\subseteq\Lambda_t(\bv)$ and, for every $\lambda\in\Lambda_t^G(M)$ and $s\le t$,
\begin{align*}
    |\lambda\langle v_s,g_s\rangle|\le\lambda\|v_s\|\|g_s\|\le\lambda M_s\|g_s\|\le\lambda G\max_{r\le t}M_r<1.
\end{align*}
Recall $\psi_{E,1}(u)=-\log(1-u)-u$ for $u<1$. It is nondecreasing on $[0,1)$ and, for $u\in[0,1)$, $\psi_{E,1}(u)\le\psi_{E,1}(|u|)$ for $|u|<1$. Therefore, it follows that, on $\mathsf E_t(M)$, for every $\lambda\in\Lambda_t^G(M)$,
\begin{align}\label{eq:obs-bet-pf1}
    \sum_{s=1}^t\psi_{E,1}\big(-\lambda\langle v_s,g_s\rangle\big)\le\sum_{s=1}^t\psi_{E,1}\big(|\lambda\langle v_s,g_s\rangle|\big)\le\sum_{s=1}^t\psi_{E,1}(\lambda M_s\|g_s\|)=J_t(\lambda).
\end{align}
Combining \eqref{eq:obs-bet-pf1} with \eqref{eq:obs-bet-mean-correction}, on $\mathsf E_t(M)$, for every $\lambda\in\Lambda_t^G(M)$ and $0\le z\le Z_t$,
\begin{align*}
    \begin{aligned}
&0=H_t(\lambda,0)\le H_t(\lambda,z),\text{ and, }\\
&-J_t(\lambda)+H_t(\lambda,z)\le-\sum_{s=1}^t\psi_{E,1}\big(-\lambda\langle v_s,g_s\rangle\big)+\sum_{s=1}^t\psi_{E,1}\big(-\lambda\langle v_s,\IE_{s-1}g_s\rangle\big),
\end{aligned}
\end{align*}
where the first relation uses monotonicity of $z\mapsto H_t(\lambda,z)$. Since all summands are nonnegative and $\Lambda_t^G(M)\subseteq\Lambda_t(\bv)$, it follows that, on $\mathsf E_t(M)$, for every $y\in\R$ and $0\le z\le Z_t$,
\begin{align*}
 b_t^H(y,z)\le b_\alpha^1(y;t,\bv),
\end{align*}
where we recall $b_{\alpha}^1$ from Corollary \ref{corol:betting}.
Inverting the above thresholds, one obtains,
\begin{align*}
    \Big\{y:b_\alpha^1(y;t,\bv)<\frac1\alpha\Big\}\subseteq\Big\{y:b_t^H(y,z)< 1/\alpha\Big\}
\end{align*}
which directly yields 
\begin{equation}\label{eq:obs-bet-corrected-comparison}
    \mathcal B_\alpha^1(t;\bv)
    \le
    \mathcal U_\alpha^{\mathrm{bet}}(t;M,H,z)
\end{equation}
Finally, owing to \eqref{eq:obs-bet-pf0} and \eqref{eq:obs-bet-corrected-comparison}, it holds
\begin{align*}
   \IP\Big(\bigcup_{t\ge1}\Big[\mathsf E_t(M)\cap\bigcup_{0\le z\le Z_t}\big\{Y_t(\bv)>\mathcal U_\alpha^{\mathrm{bet}}(t;M,H,z)\big\}\Big]\Big)\le\IP\big(\exists\,t\ge1:Y_t(\bv)>\mathcal B^1_\alpha(t;\bv)\big)\le\alpha,
\end{align*}
where the final inequality results from a direct application of Corollary \ref{corol:betting}. This completes the proof.
\end{proof}

\section{Proofs of results in \S\ref{se: anytime-valid-sgd}}\label{app:se3proofs}
\subsection{Proofs of results in \S\ref{se:rsi}}\label{app:se3.1proofs}
\begin{proof}[Proof of Proposition~\ref{prop:betting-SGD}]
We verify the conditions of Lemma~\ref{lem:observable-envelope}, where the oracle boundary is furnished by  Theorem~\ref{thm:obs-betting}. Write $R_t:=R_t^w$, and recall from \S\ref{se:opt-rsi} that $D_t=R_0+\sum_{s=1}^tA_s^{-1}\eta_s^2\|g_s\|^2$, and $\wt\bv=(\wt v_s)_{s\ge1}$ with $\wt v_s:=v_s/A_s$. Finally recall the key recursion \eqref{eq:stronglyconvex-recursion}. Corollary~\ref{corol:betting}, applied with $c=G$ and the predictable sequence $-\wt\bv$, shows that $\Mcal_\alpha(t;\wt\bv):=\mathcal B^1_\alpha(t;-\wt\bv)$ satisfies $\IP\big(Y_t(\wt\bv)\le\Mcal_\alpha(t;\wt\bv)\text{ for all }t\ge1\big)\ge1-\alpha$. Moreover, putting $U_t:=R_t/A_t-D_t$ and $U_0:=0$, we have $R_t=A_t(D_t+U_t)$, as required by Lemma~\ref{lem:observable-envelope}.

Next, we verify \eqref{eq:obs-bet-mean-correction}. Clearly, $H_t^{(w)}(\lambda,0)=0$, and $z\mapsto H_t^{(w)}(\lambda,z)$ is nondecreasing. On $\bigcap_{s<t}\{Z_{s,1}\le R_s\}$, $\|\wt v_s\|=2\eta_s\sqrt{Z_{s-1,1}}/A_s\le M_s$ for $s\le t$, i.e., $\mathsf E_t(M)$ holds for $\wt\bv$. Therefore, for $0\le z\le Z_{t,1}$ and $1\le s\le t$,
\[
\frac{2\mu\eta_s}{A_s}\big(\sqrt z-\|x_t-x_{s-1}\|\big)_+^2\overset{(a)}{\le}\frac{2\mu\eta_s}{A_s}\|x_{s-1}-x^\star\|^2\overset{(b)}{\le}\langle\wt v_s,\nabla F(x_{s-1})\rangle\overset{(c)}{\le}M_sG,
\]
where $(a)$ follows from the reverse triangle inequality $\|x_{s-1}-x^\star\|\ge\sqrt z-\|x_t-x_{s-1}\|$, $(b)$ from \RSI, and $(c)$ from the Cauchy--Schwarz inequality along with $\|\nabla F(x_{s-1})\|=\|\IE_{s-1}g_s\|\le G$. The monotonicity of $\psi_{E,1}$ on $[0,1)$ implies 
\[\psi_{E,1}\Big(\frac{2\lambda\mu\eta_s}{A_s}\big(\sqrt z-\|x_t-x_{s-1}\|\big)_+^2\Big) \le \psi_{E,1}\!\Bigl(
        -\lambda\langle v_s,\IE_{s-1}g_s
        \rangle
    \Bigr). \]
This establishes \eqref{eq:obs-bet-mean-correction}, and Theorem \ref{thm:obs-betting} applies. 

It remains to verify \eqref{eq:star}. On $\bigcap_{s<t}\{Z_{s,1}\le R_s\}\cap\{Z_{t,1}>R_t\}$, fix $z\in(R_t,Z_{t,1}]$. Since $R_t<z\le R_\star$, the definition of $R_t$ and $W_t^{(w)}(z)$ gives
\begin{align*}
\mathcal U^{\mathrm{bet}}_\alpha(t;M,H,z)\le\frac{z}{A_t}-D_t
\ {\implies}\ \Mcal_\alpha(t;\wt\bv)\le\frac{z}{A_t}-D_t,
\end{align*}
where \eqref{eq:obs-bet-corrected-comparison} in the proof of Theorem~\ref{thm:obs-betting} drives the implication. Letting $z\downarrow R_t$ yields $\Mcal_\alpha(t;\wt\bv)\le R_t/A_t-D_t=U_t$, which is \eqref{eq:star}. Lemma~\ref{lem:observable-envelope} now completes the proof.
\end{proof}
 \subsection{Proofs of results in \S\ref{se:Lipschitz-sgd}}\label{app:se3.2proofs}
\begin{proof}[Proof of Proposition~\ref{prop:ls-quadratic-betting}]
Define \begin{equation}
    K_t^{\mathrm Q}(\bv,\lambda):=\prod_{s=1}^t\frac{1+\lambda\langle v_s,\xi_s\rangle+\lambda^2\langle v_s,\xi_s\rangle^2/3}{1+\lambda^2\sigma^2\|v_s\|^2/3},\  K_0^{\mathrm Q}(\bv,\lambda)=1.\nonumber
\end{equation} Fix $\lambda>0$ and put $X_s:=\langle v_s,\xi_s\rangle$. By \eqref{eq:ls-moment-assumption},  $\IE_{s-1}X_s^2=\langle v_s,\IE_{s-1}[\xi_s\xi_s^\top]v_s\rangle\le\sigma^2\|v_s\|^2$, and 
\[
    \IE_{s-1}\Big[\frac{1+\lambda X_s+\lambda^2X_s^2/3}{1+\lambda^2\sigma^2\|v_s\|^2/3}\Big]\le1.
\]
Therefore $K_t^{\mathrm Q}(\bv,\lambda)$ is a nonnegative supermartingale with $K_0^{\mathrm Q}=1$. Monotone convergence applies to imply $$K_t:=\sum_{k\ge0}p_kK_t^{\mathrm Q}(\bv,\lambda_k), \ K_0=1$$ is a non-negative supermartingale as well. Ville's inequality (Lemma~1 of \citealp{howard2018}) yields $\IP(\Omega_\alpha)\ge1-\alpha$, where $\Omega_\alpha:=\{K_t<1/\alpha\text{ for all }t\ge1\}$.
 
We next lower-bound $K_t$ by the observable $W_t^{\mathrm Q}(Y_t(\bv))$. Recall $\chi$ from \eqref{eq:chi}, and write,
\[
    \log K_t^{\mathrm Q}(\bv,\lambda)=\lambda Y_t(\bv)-\sum_{s=1}^t\chi(\lambda X_s)-\sum_{s=1}^t\log\Big(1+\frac{\lambda^2\sigma^2\|v_s\|^2}{3}\Big).
\]
As $\|v_s\|=\eta_s\|\nabla F(x_{s-1})\|\le\eta_sG$ and $\|\xi_s\|\le G+\|g_s\|$, we have $|\lambda X_s|\le\lambda\eta_sG(G+\|g_s\|)$, and hence $\chi(\lambda X_s)\le j(\lambda\eta_sG(G+\|g_s\|))$ by the definition and monotonicity of $j$. Also, $\log(1+\lambda^2\sigma^2\|v_s\|^2/3)\le\log(1+\lambda^2\sigma^2G^2\eta_s^2/3)$. Consequently, by definition of $J_t^Q$ and $W_t^Q$ in \eqref{eq: jt-quad}, \[\log K_t^{\mathrm Q}(\bv,\lambda)\ge\lambda Y_t(\bv)-J_t^{\mathrm Q}(\lambda)\ , \text{and }
    W_t^{\mathrm Q}(Y_t(\bv))\le K_t,\  t\ge1.
\]
Finally, inverting $W_t^Q$ it is immediate that on $\Omega_\alpha$, $W_t^{\mathrm Q}(Y_t(\bv))<1/\alpha$ leads to $Y_t(\bv)\le U_t^{\mathrm Q}$ for all $t\ge1$. Thus $\Mcal_\alpha(t;\bv):=U_t^{\mathrm Q}$ satisfies the requirement of Lemma~\ref{lem:observable-envelope}, and \eqref{eq:star} holds trivially with $U_t=U_t^{\mathrm Q}$. Lemma~\ref{lem:observable-envelope} with $a_t=1$ and $b_t=L\eta_t^2\|g_t\|^2/2$ completes the proof.
\end{proof}
 
\begin{proof}[Proof of Proposition~\ref{prop:ls-hybrid}]
 Recall that $K_t^{\mathrm Q}(\bv,\lambda)$ is a nonnegative supermartingale by the proof of Proposition~\ref{prop:ls-quadratic-betting}. Moreover, by Definition \ref{def:psi} and Lemma~\ref{lem:suff}, it follows that with $\theta=\lambda v_s$, $E_t(\lambda):=\exp\bigl[\lambda Y_t(\bv)-\lambda^2\sigma^2\sum_{s\le t}\|v_s\|^2/2\bigr]$ is a nonnegative supermartingale with $E_0(\lambda)=1$. Consequently,
\[
    K_t^{\mathrm H}:=\frac12\sum_{k\ge0}p_k\big[E_t(\lambda_k)+K_t^{\mathrm Q}(\bv,\lambda_k)\big]
\]
is a nonnegative supermartingale with $K_0^{\mathrm H}=1$. Since $\sigma^2\sum_{s\le t}\|v_s\|^2\le Q_t^G$, $E_t(\lambda)\ge\exp\bigl[\lambda Y_t(\bv)-\lambda^2Q_t^G/2\bigr]$; together with \[
K_t^{\mathrm Q}(\bv,\lambda)\ge\exp\bigl[\lambda Y_t(\bv)-J_t^{\mathrm Q}(\lambda)\bigr],
\] this gives $W_t^{\mathrm H}(Y_t(\bv))\le K_t^{\mathrm H}$. The rest of the proof follows along the lines of Proposition~\ref{prop:ls-quadratic-betting}, with $W_t^{\mathrm Q}$, $K_t$ and $U_t^{\mathrm Q}$ replaced by $W_t^{\mathrm H}$, $K_t^{\mathrm H}$ and $U_t^{\mathrm H}$.
 
\paragraph{Proof of \eqref{eq:ls-hybrid-comparison}.} Fix $t\ge1$, and observe that $q(Q_t^G)\ge Q_t^G$. Let  $L_\alpha:=\log(2/\alpha)$ and $\ell:=L_\alpha+\log\log(e+q(Q_t^G))$. Note $L_\alpha>1$ since $\alpha<2/e$. For every $k$, $W_t^{\mathrm H}(y)\ge\frac{p_k}2\exp(\lambda_ky-\lambda_k^2Q_t^G/2)$, so that $W_t^{\mathrm H}(y)<1/\alpha$ implies
\[
    y<\frac{\log(2/(\alpha p_k))}{\lambda_k}+\frac{\lambda_kq(Q_t^G)}{2},
\]
Corollary \ref{cor:closed-form} completes the proof by taking an infimum over $k$ in RHS. Also see the proof of Proposition \ref{prop:convex-hybrid}.
\end{proof}

\subsection{Proofs of results in \S\ref{se:pr}}\label{app:pr-proofs}

Throughout this subsection, grant \eqref{eq:pr-sc-ls}, and write $e_t:=x_t-x^\star$ and $m_t:=\nabla F(x_{t-1})=\IE_{t-1}g_t$. In this section, we prove Theorems~\ref{thm:pr-coverage} and~\ref{thm:pr-rates} sequentially.

For the proof of Theorem~\ref{thm:pr-coverage}, we transfer the problem of deriving a time-uniform boundary for $\bar Z_{t}$ to deriving a time-uniform boundary for $\|\Xi_t\|^2$. This is formalized in the following proposition. 
\begin{prop}[Observable transfer to the averaged iterate]\label{prop:pr-transfer}
Grant the assumptions and notations of Theorem~\ref{thm:pr-coverage}. Then almost surely on $\{\|\Xi_t\|\le B_t\text{ for all }t\ge0\}$, it holds $\bar Z_t\le R^{\mathrm{PR}}_t$ simultaneously for all $t\ge 1$.
\end{prop}
\begin{proof}[Proof of Proposition~\ref{prop:pr-transfer}]
Fix $t\ge1$ and work on $\Omega_B:=\{\|\Xi_t\|\le B_t\text{ for all }t\ge0\}$. We first bound $\|\nabla F(\wt x_t)\|$, and then transfer this bound to $\bar x_t$. Since $\sum_{s=1}^t\nabla F(x_{s-1})=\sum_{s=1}^tg_s+\Xi_t$ and $\|\Xi_t\|\le B_t$,
\[
\|\nabla F(\wt x_t)\|\le\frac{\|\sum_{s=1}^tg_s\|+B_t}t+\Big\|\frac1t\sum_{s=1}^t\bigl[\nabla F(x_{s-1})-\nabla F(\wt x_t)\bigr]\Big\|.
\]
We control the last term. By \eqref{eq:pr-sc-ls}, 
\[\Big\|\frac1t\sum_{s=1}^t\bigl[\nabla F(x_{s-1})-\nabla F(\wt x_t)\bigr]\Big\|\le t^{-1}\sum_{s\le t}L\|x_{s-1}-\wt x_t\|=L\Dcal_{t,1}.\]
Moreover, since $\nabla^2F$ is Lipschitz by \eqref{eq:pr-hl}, for $x,z\in\R^d$ via Taylor expansion one obtains
\begin{align}
    \begin{aligned}
&\big\|\nabla F(x)-\nabla F(z)-\nabla^2F(z)(x-z)\big\|\\
&=\Big\|\int_0^1\bigl[\nabla^2F(z+u(x-z))-\nabla^2F(z)\bigr](x-z)\,du\Big\|\\
&\le\frac{L_H}2\|x-z\|^2;
\end{aligned} \label{eq:taylor}
\end{align}
taking $z=\wt x_t$, $x=x_{s-1}$, and noting $\sum_{s=1}^t(x_{s-1}-\wt x_t)=0$, one obtains from \eqref{eq:taylor} that
\[ \Big\|\frac1t\sum_{s=1}^t\bigl[\nabla F(x_{s-1})-\nabla F(\wt x_t)\bigr]\Big\| \le \min\{L\Dcal_{t,1},L_H\Dcal_{t,2}/2\}\]

Moreover, \eqref{eq:pr-sc-ls} with $y=x^\star$ and $\nabla F(x^\star)=0$ gives $\mu\|z-x^\star\|^2\le\langle\nabla F(z),z-x^\star\rangle\le\|\nabla F(z)\|\|z-x^\star\|$, i.e., $\mu\|z-x^\star\|\le\|\nabla F(z)\|$ for $z\in\R^d$. As $\bar x_t=\wt x_t+(x_t-x_0)/t$,
\[
\|\bar x_t-x^\star\|\le\|\wt x_t-x^\star\|+\frac{\|x_t-x_0\|}t\le\frac{\|\sum_{s=1}^tg_s\|+B_t}{\mu t}+\frac{\min\{L\Dcal_{t,1},L_H\Dcal_{t,2}/2\}}\mu+\frac{\|x_t-x_0\|}t.
\]
On the other hand, noting that $\mu\|e_s\|\le\|\IE_sg_{s+1}\|\le G$ almost surely for all $s\ge0$ implies that $\|\bar x_t-x^\star\|\le t^{-1}\sum_{s\le t}\|e_s\|\le G/\mu$. Combining the two bounds above yields $\bar Z_t\le R^{\mathrm{PR}}_t$. 
\end{proof}

\begin{proof}[Proof of Theorem~\ref{thm:pr-coverage}]
First we prove that $\IP\Big(\bar Z_t\le R^{\mathrm{PR}}_t \text{ for all }t\ge1\Big)\ge1-\alpha.$
By Proposition~\ref{prop:pr-transfer}, it suffices to show $\IP(\|\Xi_t\|\le B_t\text{ for all }t\ge0)\ge1-\alpha$. To that end, we apply Lemma~\ref{lem:observable-envelope} to $Z_t:=\|\Xi_t\|^2$ with $R_0:=0=Z_0$, $a_t:=1$, $b_t:=\|g_t\|^2$ and $v_t$ as in \eqref{eq:pr-rec}.  It remains to exhibit $\Mcal_\alpha$ with $\IP(Y_t(\bv)\le\Mcal_\alpha(t;\bv)\text{ for all }t\ge1)\ge1-\alpha$, and to verify \eqref{eq:star}.

We first identify the noise inputs. By the definition, and the Table~\ref{tab:sub-psi-examples}, $(\xi_t,G^2I_d)$ is a sub-$\psi_N$ process, and $(\xi_t,g_tg_t^\top)$ is a sub-$\psi_{E,G}$ process. We will also employ the oft-used inequality $\psi_{E,G}\le \psi_{\mathrm{G},G}$. Accordingly, let $\Mcal_\alpha(t;\bv)$ be
\begin{itemize}
    \item $\Mcal_\alpha(t;\bv,F(\cdot;\rho))$ of Theorem~\ref{thm:prob} for NM;
    \item $\St_\alpha(t;\bv)$ of Proposition~\ref{prop:stitch} with the grid $\{\lambda_k,p_k\}$ for GS and $\Gamma\mathrm S$; and,
    \item $\mathcal B^1_\alpha(t;-\bv)$ of Corollary~\ref{corol:betting} with $c=G$ for Bet, since $Y_t(\bv)=\sum_{s\le t}\langle-v_s,g_s-\IE_{s-1}g_s\rangle$.
\end{itemize}
These results yield the required time-uniform bound.

We next verify \eqref{eq:star}. Fix $t\ge1$ and work on $\bigcap_{s<t}\{Z_s\le R_s\}$. Then $\|v_s\|\le2(\|\Xi_{s-1}\|+\|m_s\|)\le M_s$ for $s\le t$, i.e., $\mathsf E_t(M)$ of Theorem~\ref{thm:obs-comp} holds with the predictable $M=(M_s)_{s\ge1}$; moreover, $\Lambda_t(M)=\Lambda^G_t(M)=[0,1/c_t)$ for $\lambda_{\max}=1/G$. Hence Theorem~\ref{thm:obs-comp}, with
\begin{itemize}
    \item $\nu_s\equiv G^2$ for NM and GS, $\nu_s=\|g_s\|^2$ for $\Gamma\mathrm S$ (as $g_sg_s^\top\preceq\|g_s\|^2I_d$), together with Corollary~\ref{cor:closed-form}(i) for NM; and,
    \item Theorem~\ref{thm:obs-betting} and $H_t\equiv0$ for Bet,
\end{itemize} 
give $\Mcal_\alpha(t;\bv)\le U_t$. Here \eqref{eq:obs-bet-mean-correction} holds for $H_t\equiv0$, since $\psi_{E,1}\ge0$ on $(-\infty,1)$ and $|\lambda\langle v_s,m_s\rangle|\le\lambda GM_s<1$ for $\lambda\in\Lambda^G_t(M)$. Thus \eqref{eq:star} holds, and Lemma~\ref{lem:observable-envelope} yields $\IP(\|\Xi_t\|\le B_t\text{ for all }t\ge0)\ge1-\alpha$.

It remains to compare $R^{\mathrm{PR}, \mathrm{Bet}}$ with $R^{\mathrm{PR}, \Gamma\mathrm S}$. Fix $\{M_s\}_{s=1}^t$, and let $\lambda c_t<1$. By \eqref{eq:betting-quadratic-relaxation}, and $\psi_{E,G}\le\psi_{\mathrm G,G}$,
\begin{align}
    \psi_{E,1}(\lambda M_s\|g_s\|)\le\|g_s\|^2\psi_{E,G}(\lambda M_s)\le\|g_s\|^2\psi_{\mathrm G,G}(\lambda M_s),\  s\le t.\label{eq:E-G-ineq}
\end{align}
For $k\ge 1$, recall $L_k:=\log(1/(\alpha p_k))$ and observe that
\[ p_k\exp\big(\lambda_ky-\sum_{s=1}^t\psi_{E,1}(\lambda_kM_s\|g_s\|)\big) < 1/\alpha \implies  y < \frac{L_k+\sum_{s\le t}\psi_{E,1}(\lambda_kM_s\|g_s\|)}{\lambda_k} \]
Consequently, together with \eqref{eq:E-G-ineq}, we obtain $U^{\mathrm{Bet}}_t\le U^{\Gamma\mathrm S}_t$ for a shared trajectory. By induction on $t$, $B^{\mathrm{Bet}}_t\le B^{\Gamma\mathrm S}_t$ for all $t\ge0$, and this concludes the proof since \eqref{eq:pr-transfer} is nondecreasing in $B_t$.
\end{proof}

We now prove Theorem~\ref{thm:pr-rates}, which rests on two lemmas. The first controls the growth of $B_t$, while the second lemma controls the moments of the iterates and the remaining terms in \eqref{eq:pr-transfer}. 
\begin{lem}\label{lem:pr-noise-growth}
For each $b\in\{\mathrm{NM},\mathrm{GS},\Gamma\mathrm S,\mathrm{Bet}\}$, there exists a constant $C_b$ solely depending on $(\alpha,G)$, and additionally on $\rho$ if $b=\mathrm{NM}$, such that $(B^b_t)^2\le C_bth_b(t)$ almost surely for all $t\ge1$. 
\end{lem}
\begin{proof}
Throughout, $K$ denotes an arbitrary constant depending only on $(\alpha,G)$, and on $\rho$ if $b=\mathrm{NM}$, whose value may change from line to line. When the meaning is clear, we also omit the subscript $b$ from $h_b$ and write $h:=h_b$. 

We first control $B_t$ by a deterministic recursion. Define $\varphi(v,c):=\Ncal_\alpha(v)$ for NM, $\varphi(v,c):=2G\,\mathfrak S_\alpha(v/(4G^2))$ for GS, and $\varphi(v,c):=2G\,\mathfrak G_\alpha(v/(4G^2),c/(2G))$ for $\Gamma\mathrm S$ and Bet, $v,c\ge0$; each $\varphi$ is nondecreasing in $(v,c)$. By the assumption $\|g_s\|\le G$, and $U^{\mathrm{Bet}}_t\le U^{\Gamma\mathrm S}_t$ for shared $\{M_s\}_{s=1}^t$ (proof of Theorem~\ref{thm:pr-coverage}), it holds $U_t\le\varphi(G^2\sum_{s\le t}M_s^2,c_t)$. Hence, by induction on $t$, $B_t\le\bar B_t$, where $\bar B_0:=0$, $\bar M_t:=2(\bar B_{t-1}+G)$ and
\[
\bar B_t^2:=G^2t+\varphi\Big(G^2\sum_{s\le t}\bar M_s^2,\ G\max\{1,\max_{s\le t}\bar M_s\}\Big).
\]

We next bound $\varphi$. Let $a_\alpha:=2\log(2/\alpha)+2$. For $k\ge1$ and $y\ge e$, $\log(ky)\le(1+\log k)\log y$ and $\log\log(ky)\le\log(1+\log k)+\log\log y$. These inequalities, together with 
\[ v+\rho\le(1\vee\rho)(1+v), \ e+q_\alpha(v)\le(e+a_\alpha)(e+v), \ q_\alpha(v)\le a_\alpha(1+v),\]
and $2+2\log_+(2c)\le6\log(e+c)$, applied to the definitions of $\Ncal_\alpha$, $\mathfrak S_\alpha$ and $\mathfrak G_\alpha$ in Corollary~\ref{cor:closed-form}, yield
\begin{equation}\label{eq:pr-phi-growth}
\varphi(v,c)\le K\bigl[\sqrt{(1+v)h(v)}+c\,\ell(c)\bigr],\  v,c\ge0, \text{ where }\ell(t):=1+\log\log(e+t).
\end{equation}

Finally, we proceed via an induction. Fix $C\ge\max\{1,G^2\}$ and suppose $\bar B_s^2\le Csh(s)$ for $1\le s<t$. Since $h\ge1$ is nondecreasing, $\bar M_s^2\le8(\bar B_{s-1}^2+G^2)\le16Cth(t)$ for $s\le t$, so that $$v:=G^2\sum_{s\le t}\bar M_s^2\le16G^2Ct^2h(t), \ \text{and } c:=G\max\{1,\max_{s\le t}\bar M_s\}\le4G\sqrt{Cth(t)}.$$ As $h(t)\le e+t$, both $v$ and $c$ are at most $k_C(e+t)^3$, where $k_C:=(1\vee16G^2)C$, and hence, for $x\in\{v,c\}$ and $\kappa(C):=3+\log(2k_C)$,
\[
\begin{aligned}
&\log(e+x)\le\log(2k_C)+3\log(e+t)\le\kappa(C)\log(e+t) \\
&\implies\log\log(e+x)\le\log\kappa(C)+\log\log(e+t),
\end{aligned}
\]
so that, as $1+\log\kappa(C)\le\kappa(C)$ and $\ell\le\log(e+\cdot)$, $h(v)\le\kappa(C)h(t)$ and $\ell(c)\le\kappa(C)h(t)$. Plugging these bounds into \eqref{eq:pr-phi-growth} and using $h(t)\le2t$,
\[
\bar B_t^2\le\big(G^2+K\sqrt{C\kappa(C)}+K\sqrt C\kappa(C)\big)th(t) \le C t h(t),
\]
where the final inequality follows by choosing $C$ to be large enough. This completes the induction.
\end{proof}

In the following, for a random vector $Y$, write $\|Y\|_{L^2}:=(\IE\|Y\|^2)^{1/2}$.
\begin{lem}\label{lem:pr-moments}
Let $\eta_t=\eta_\circ(t+t_0)^{-\beta}$, where $\eta_\circ>0$, $t_0\ge0$ and $\beta\in(0,1)$. Then, under the assumptions of Theorem~\ref{thm:pr-rates}, there exists a constant $C$, depending only on $(G,\mu,\eta_\circ,t_0,\beta)$, such that, for all $t\ge1$, $\IE\|e_t\|^2\le C(t+t_0)^{-\beta}$, $\IE\|e_t\|^4\le C(t+t_0)^{-2\beta}$. Moreover, it follows that
\begin{equation}\label{eq:pr-remainders}
\|\Dcal_{t,2}\|_{L^2}\le Ct^{-\beta},\ \Big\|\frac1t\sum_{s=1}^tg_s\Big\|_{L^2}\le Ct^{-1+\beta/2},\ \text{and }\frac{\|x_t-x_0\|}t\le\frac{2G}{\mu t}\ \text{ a.s.}
\end{equation}
\end{lem}
\begin{proof}
Write $u_t:=\IE\|e_t\|^2$ and $w_t:=\IE\|e_t\|^4$, so that $u_t\le G^2/\mu^2$ and $w_t\le G^4/\mu^4$, as $\|e_t\|\le G/\mu$ almost surely (proof of Proposition~\ref{prop:pr-transfer}); here $C$ denotes a constant depending only on $(G,\mu,\eta_\circ,t_0,\beta)$, whose value may change from line to line. We first derive recursions for $u_t$ and $w_t$, and then solve them. Since \SC\ implies \RSI, \eqref{eq:stronglyconvex-recursion} holds; taking expectations and using $\IE\langle e_{t-1},\xi_t\rangle=0$,
\begin{equation}\label{eq:pr-u-rec}
u_t\le(1-2\mu\eta_t)u_{t-1}+G^2\eta_t^2,\  t\ge1.
\end{equation}
Next, we expand $\|e_t\|^4=(\|e_{t-1}\|^2-2\eta_t\langle e_{t-1},g_t\rangle+\eta_t^2\|g_t\|^2)^2$ and use $\IE_{t-1}\langle e_{t-1},g_t\rangle=\langle e_{t-1},m_t\rangle\ge\mu\|e_{t-1}\|^2$, $|\langle e_{t-1},g_t\rangle|\le G\|e_{t-1}\|$ and $4G^3\eta_t^3\|e_{t-1}\|\le2G^2\eta_t^2\|e_{t-1}\|^2+2G^4\eta_t^4$; this gives
\begin{equation}\label{eq:pr-w-rec}
\begin{aligned}
&\IE_{t-1}\|e_t\|^4\le(1-4\mu\eta_t)\|e_{t-1}\|^4+8G^2\eta_t^2\|e_{t-1}\|^2+3G^4\eta_t^4 \\
&\implies w_t\le(1-4\mu\eta_t)w_{t-1}+8G^2\eta_t^2u_{t-1}+3G^4\eta_t^4.
\end{aligned}
\end{equation}
We solve these recursions by invoking Lemma 2.4(a) of \cite{jiang2024generalized}. By \eqref{eq:pr-u-rec}, $a_k:=u_{T+k}$ satisfies the recursion therein with $\gamma:=T+1+t_0$, $\nu=q:=\beta$, $c:=2\mu\eta_\circ$ and $d:=G^2\eta_\circ^2$. Choose an integer $T\ge0$ with $$T+1+t_0\ge\max\{c^{1/\beta},(2q/c)^{1/(1-\beta)}\}.$$ Using $u_t\le G^2/\mu^2$ for $t<T$, we obtain $u_t\le C(t+t_0)^{-\beta}$, $t\ge1$. Therefore, in view of $\eta_t^4 \le\eta_\circ^4(t+t_0)^{-3\beta}$, \eqref{eq:pr-w-rec} produces $$w_t\le(1-4\mu\eta_t)w_{t-1}+C(t+t_0)^{-3\beta}.$$ Lemma 2.4(a) of \cite{jiang2024generalized}, now with $a_k:=w_{T'+k}$ for an integer $T'\ge1$, $c:=4\mu\eta_\circ$, $\nu:=\beta$ and $q:=2\beta$, leads to $w_t\le C(t+t_0)^{-2\beta}$, $t\ge1$.

We now control the terms in \eqref{eq:pr-remainders}. Clearly, by the definition of $\tilde x_t$, $\Dcal_{t,2}\le t^{-1}\sum_{s=1}^t\|e_{s-1}\|^2$. By Minkowski's inequality, and $\|e_0\|\le G/\mu$,
\[
\|\Dcal_{t,2}\|_{L^2}\le\frac1t\biggl[\frac{G^2}{\mu^2}+C\sum_{j=1}^{t-1}j^{-\beta}\biggr]\le Ct^{-\beta},
\]
where we use the elementary inequality $\sum_{j=1}^{t-1}j^{-\beta}\le t^{1-\beta}/(1-\beta)$.
Next, as $g_s=(e_{s-1}-e_s)/\eta_s$, summation by parts gives
\[
\sum_{s=1}^tg_s=\frac{e_0}{\eta_1}-\frac{e_t}{\eta_t}+\sum_{s=1}^{t-1}\Big(\frac1{\eta_{s+1}}-\frac1{\eta_s}\Big)e_s,
\]
where $0\le\eta_{s+1}^{-1}-\eta_s^{-1}\le\beta\eta_\circ^{-1}(s+t_0)^{\beta-1}$ by the concavity of $u\mapsto u^\beta$. Hence, yet another application of Minkowski's inequality and $\|e_s\|_{L^2}\le C(s+t_0)^{-\beta/2}$ entails that,
\[
\Big\|\frac1t\sum_{s=1}^tg_s\Big\|_{L^2}\le\frac Ct\biggl[1+(t+t_0)^{\beta/2}+\sum_{s=1}^{t-1}s^{\beta/2-1}\biggr]\le Ct^{-1+\beta/2}.
\]
Finally, $\|x_t-x_0\|\le\|e_t\|+\|e_0\|\le2G/\mu$ almost surely.
\end{proof}

\begin{proof}[Proof of Theorem~\ref{thm:pr-rates}]
Throughout, $C$ denotes a constant whose value may change from line to line. We bound $(\IE R^{\mathrm{PR}}_t)^{1/2}=\|\sqrt{R^{\mathrm{PR}}_t}\|_{L^2}$ termwise. Discarding the cap $G/\mu$ in \eqref{eq:pr-transfer}, bounding $\min\{L\Dcal_{t,1},L_H\Dcal_{t,2}/2\}\le L_H\Dcal_{t,2}/2$, and applying Minkowski's inequality, Lemma~\ref{lem:pr-noise-growth} and \eqref{eq:pr-remainders},
\[
\big(\IE R^{\mathrm{PR}}_t\big)^{1/2}\le\frac1\mu\Big\|\frac1t\sum_{s=1}^tg_s\Big\|_{L^2}+\frac{\|B_t\|_{L^2}}{\mu t}+\frac{L_H\|\Dcal_{t,2}\|_{L^2}}{2\mu}+\frac{2G}{\mu t}\le C\biggl[t^{-1+\beta/2}+\sqrt{\frac{h_b(t)}t}+t^{-\beta}\biggr],
\]
where we used $t^{-1}\le\sqrt{h_b(t)/t}$. Since $\beta\in(1/2,1)$ and $h_b\ge1$, $t^{-1+\beta/2}\le t^{-1/2}\le\sqrt{h_b(t)/t}$ and $t^{-\beta}\le t^{-1/2}\le\sqrt{h_b(t)/t}$. This concludes the proof.
\end{proof}

\section{Proofs of results in \S\ref{se:appl}}\label{app:se4proofs}
\subsection{Proofs of results in \S\ref{se:oja}}\label{app:se4.1proofs}
\begin{proof}[Proof of Proposition~\ref{prop:oja-rec}]
Fix $t\ge1$ and put $\Delta_t:=\|(I+\eta_t\widehat\Sigma_t)u_{t-1}\|^2=1+2\eta_t\langle u_{t-1},\widehat\Sigma_tu_{t-1}\rangle+\eta_t^2\|\widehat\Sigma_tu_{t-1}\|^2$; since $\widehat\Sigma_t\succeq0$, $\Delta_t\ge1$.

We first derive an exact expression for $Z_t-Z_{t-1}$. Since $\widehat\Sigma_t$ is symmetric, $\langle(I+\eta_t\widehat\Sigma_t)u_{t-1},u^\star\rangle=\langle u_{t-1},u^\star\rangle+\eta_t\langle u^\star,\widehat\Sigma_tu_{t-1}\rangle$. Therefore, recalling the definition of $\mathcal H$ in Proposition \ref{prop:oja-rec},
\[
    \langle\mathcal H(u_{t-1}),\widehat\Sigma_t\rangle=2\langle u_{t-1},u^\star\rangle\big\langle\langle u_{t-1},u^\star\rangle u_{t-1}-u^\star,\widehat\Sigma_tu_{t-1}\big\rangle.
\]
Write $Z_t=1-\langle(I+\eta_t\widehat\Sigma_t)u_{t-1},u^\star\rangle^2/\Delta_t$, which leads to
\begin{equation}\label{eq:oja-exact}
    Z_t-Z_{t-1}=\langle u_{t-1},u^\star\rangle^2-\frac{\big(\langle u_{t-1},u^\star\rangle+\eta_t\langle u^\star,\widehat\Sigma_tu_{t-1}\rangle\big)^2}{\Delta_t}=\eta_t\langle\mathcal H(u_{t-1}),\widehat\Sigma_t\rangle+R_t,
\end{equation}
where
\[
    R_t:=-\eta_t\langle\mathcal H(u_{t-1}),\widehat\Sigma_t\rangle\,\frac{\Delta_t-1}{\Delta_t}+\eta_t^2\,\frac{(1-Z_{t-1})\|\widehat\Sigma_tu_{t-1}\|^2-\langle u^\star,\widehat\Sigma_tu_{t-1}\rangle^2}{\Delta_t}:=A+B.
\]
We now show $R_t\le3\eta_t^2\|\widehat\Sigma_tu_{t-1}\|^2$, controlling the two terms of $R_t$ separately. For the first term, $\|\langle u_{t-1},u^\star\rangle u_{t-1}-u^\star\|^2=1-\langle u_{t-1},u^\star\rangle^2=Z_{t-1}$, so that the Cauchy--Schwarz inequality and $2\bigl[(1-Z_{t-1})Z_{t-1}\bigr]^{1/2}\le1$ give $|\langle\mathcal H(u_{t-1}),\widehat\Sigma_t\rangle|\le\|\widehat\Sigma_tu_{t-1}\|$. Moreover, $\langle u_{t-1},\widehat\Sigma_tu_{t-1}\rangle\le\|\widehat\Sigma_tu_{t-1}\|$ gives $1\le\Delta_t\le(1+\eta_t\|\widehat\Sigma_tu_{t-1}\|)^2$, and since $s\mapsto(s-1)/s$ is nondecreasing on $[1,\infty)$,
\[
    \frac{\Delta_t-1}{\Delta_t}\le1-\frac1{(1+\eta_t\|\widehat\Sigma_tu_{t-1}\|)^2}=\frac{\eta_t\|\widehat\Sigma_tu_{t-1}\|\,(2+\eta_t\|\widehat\Sigma_tu_{t-1}\|)}{(1+\eta_t\|\widehat\Sigma_tu_{t-1}\|)^2}\le2\eta_t\|\widehat\Sigma_tu_{t-1}\|.
\]
Hence $A\le 2\eta_t^2\|\widehat\Sigma_tu_{t-1}\|^2$. For the second term, $1-Z_{t-1}\le1$ and $\Delta_t\ge1$ give the bound $B\le \eta_t^2\|\widehat\Sigma_tu_{t-1}\|^2$. Consequently, $R_t\le3\eta_t^2\|\widehat\Sigma_tu_{t-1}\|^2$.

To complete the recursion, \eqref{eq:key-recursion}, it remains to extract the drift term $a_t Z_{t-1}$ from $\eta_t\langle\mathcal H(u_{t-1}),\widehat\Sigma_t\rangle=\eta_t\langle\mathcal H(u_{t-1}),\Sigma\rangle+\langle\eta_t\,\mathrm{vec}\,\mathcal H(u_{t-1}),\xi_t\rangle$. To that end, note that since $\lambda_1(\Sigma)>\lambda_2(\Sigma)$, there exists a spectral decomposition $\Sigma=\sum_{i=1}^d\lambda_i(\Sigma)q_iq_i^\top$ with orthonormal $q_1,\ldots,q_d$ and $q_1=u^\star$. This implies that $$u_{t-1}^\top\Sigma u_{t-1}\le\lambda_1(\Sigma)\langle u_{t-1},u^\star\rangle^2+\lambda_2(\Sigma)\sum_{i\ge2}\langle u_{t-1},q_i\rangle^2=\lambda_1(\Sigma)(1-Z_{t-1})+\lambda_2(\Sigma)Z_{t-1}.$$ Consequently, using the definition of $\mathcal H$ and $\langle u^\star,\Sigma u_{t-1}\rangle=\lambda_1(\Sigma)\langle u_{t-1},u^\star\rangle$,
\begin{align*}
    \langle\mathcal H(u_{t-1}),\Sigma\rangle&=2(1-Z_{t-1})\bigl[u_{t-1}^\top\Sigma u_{t-1}-\lambda_1(\Sigma)\bigr]\\
    &\le-2(1-Z_{t-1})\bigl[\lambda_1(\Sigma)-\lambda_2(\Sigma)\bigr]Z_{t-1}\le-2\delta(1-Z_{t-1})Z_{t-1},
\end{align*}
where the first inequality uses the preceding bound and $1-Z_{t-1}\ge0$, and the second uses $\lambda_1(\Sigma)-\lambda_2(\Sigma)\ge\delta$ and $(1-Z_{t-1})Z_{t-1}\ge0$. Inserting this and the bound on $R_t$ into \eqref{eq:oja-exact} proves \eqref{eq:oja-rec} for every realization, and in particular almost surely.

Finally, on $\{Z_{t-1}\le r_{t-1}\}$, $a_t\ge1-2\delta\eta_t(1-r_{t-1})$ implies that $$\bigl[1-2\delta\eta_t(1-Z_{t-1})\bigr]Z_{t-1}\le\bigl[1-2\delta\eta_t(1-r_{t-1})\bigr]Z_{t-1}\le a_tZ_{t-1}.$$ Substituting this into \eqref{eq:oja-rec} yields \eqref{eq:key-recursion} with $b_t=3\eta_t^2\|\widehat\Sigma_tu_{t-1}\|^2$ and $v_t=\eta_t\,\mathrm{vec}\,\mathcal H(u_{t-1})$. This completes the proof.
\end{proof}
\subsection{Proofs of results in \S\ref{se:rl}}\label{app:se4.2proofs}
\begin{proof}[Proof of Proposition~\ref{prop:td-rec}]
From the TD learning recursion $\theta_t= \theta_{t-1} - \eta_t g_t$, and the definition of $Z_t$, it holds $$Z_t=Z_{t-1}-2\eta_t\langle\theta_{t-1}-\theta^\star,g_t\rangle+\eta_t^2\|g_t\|^2.$$ Substituting $g_t=A_\pi(\theta_{t-1}-\theta^\star)-\xi_t$ and using $e^\top A_\pi e\ge\kappa\|e\|^2$ with $e=\theta_{t-1}-\theta^\star$ yields \eqref{eq:td-rec}.
\end{proof}
\begin{proof}[Proof of Theorem~\ref{thm:td-last}]
 In view of Proposition~\ref{prop:td-rec}, Theorem~\ref{thm:obs-comp} and Lemma~\ref{lem:observable-envelope} give $\|\theta_t-\theta^\star\|^2\le R_t$ for all $t$ with probability at least $1-\alpha$. On this event, $C\preceq I$ yields $\|\Vcal_{\theta_t}-\Vcal_{\theta^\star}\|^2_{d_\pi}=(\theta_t-\theta^\star)^\top C(\theta_t-\theta^\star)\le R_t$.
\end{proof}

\subsection{Background on stochastic variational inference and proofs of \S\ref{se:bw-opt}}\label{app:bw-proof}
 
Throughout this subsection, $\mathcal N_d$ denotes the set of nondegenerate Gaussian distributions on $\R^d$, and, for $q=N(m,\Sigma)\in\mathcal N_d$, $\IE_q$ denotes expectation under $X\sim q$. Recall that $\pi\propto e^{-F}$ with $F\in C^2$ and $\mu I\preceq\nabla^2F\preceq I$, and that $W_2$ denotes the $2$-Wasserstein distance,
\[
W_2^2(q,q'):=\inf_{\gamma\in\Gamma(q,q')}\int\|x-y\|^2\,d\gamma(x,y),
\]
where $\Gamma(q,q')$ is the set of couplings of $q$ and $q'$. A measurable map $T:\R^d\to\R^d$ is called an \emph{optimal transport map} from $q$ to $q'$ if $T_\#q=q'$ and $W_2^2(q,q')=\int\|x-T(x)\|^2\,dq(x)$.
 
\paragraph{Optimal transport between Gaussians.} Between Gaussians, the optimal transport map is affine and available in closed form.
 
\begin{lem}[{\citealp{dowson1982frechet}; \citealp{takatsu2011wasserstein}; \citealp{bhatia2019bures}}]\label{lem:gauss-ot}
Let $q=N(m,\Sigma)$ and $q'=N(m',\Sigma')$ belong to $\mathcal N_d$, and put
\[
T_{\Sigma\to\Sigma'}:=\Sigma^{-1/2}\big(\Sigma^{1/2}\Sigma'\Sigma^{1/2}\big)^{1/2}\Sigma^{-1/2}.
\]
Then $T_{\Sigma\to\Sigma'}$ is symmetric positive definite, the affine map $T(x):=m'+T_{\Sigma\to\Sigma'}(x-m)$ is the optimal transport map from $q$ to $q'$, and
\[
W_2^2(q,q')=\int\|x-T(x)\|^2\,dq(x)=\|m-m'\|^2+\Tr\Bigl[\Sigma+\Sigma'-2(\Sigma^{1/2}\Sigma'\Sigma^{1/2})^{1/2}\Bigr].
\]
\end{lem}
 
In the notation of \S\ref{se:bw-opt}, $T_\Sigma=T_{\Sigma\to\Sigma^\star}$, and \eqref{eq:bw-w2} is the last display of Lemma~\ref{lem:gauss-ot} with $q'=q^\star$.
 
\paragraph{Wasserstein and Bures - Wasserstein gradients.} Following \citet{otto2001geometry}, the space of probability measures with finite second moments, equipped with $W_2$, can be viewed as a formal Riemannian manifold: the tangent space at $q$ is a subspace of the vector fields in $L^2(q)$, with inner product $\langle h_1,h_2\rangle_q:=\int\langle h_1(x),h_2(x)\rangle\,dq(x)$. In this geometry, the gradient of $q\mapsto\mathrm{KL}(q\|\pi)$ at $q$ is the vector field $\nabla\log(q/\pi)=\nabla\log q+\nabla F$, and the associated gradient flow is the Fokker--Planck equation of the Langevin diffusion targeting $\pi$ \citep{jordan1998variational,ambrosio2008gradient}. Gaussian variational inference restricts this flow to $\mathcal N_d$. Since the optimal transport maps between Gaussians are affine (Lemma~\ref{lem:gauss-ot}), the tangent space of $\mathcal N_d$ at $q=N(m,\Sigma)$ consists of the affine vector fields $x\mapsto b+A(x-m)$ with $b\in\R^d$ and $A$ symmetric. The induced geometry on $\mathcal N_d$ is the \emph{Bures - Wasserstein} geometry \citep{takatsu2011wasserstein,bhatia2019bures}. The \emph{Bures - Wasserstein gradient} of $\mathrm{KL}(\cdot\|\pi)$ at $q$ is the $L^2(q)$-orthogonal projection of $\nabla\log(q/\pi)$ onto this subspace \citep[\S3]{lambert2022variational}, and it admits the following explicit form.
 
\begin{defn}[Bures - Wasserstein gradient of the KL divergence]\label{def:bw-grad}
For $q=N(m,\Sigma)\in\mathcal N_d$, define the affine vector field
\[
f_q(x):=\IE_q[\nabla F]+\big(\IE_q[\nabla^2F]-\Sigma^{-1}\big)(x-m),\  x\in\R^d.
\]
\end{defn}
 
The standing Hessian bound gives
$\|\nabla F(x)\|\le\|\nabla F(0)\|+\|x\|$,
ensuring that all vector fields in the following identity
belong to $L^2(q)$.
\begin{lem}[{\citealp[Appendix~C.1, equations~(36)-(37)]{lambert2022variational}}]
\label{lem:bw-proj}
Let $q=N(m,\Sigma)\in\mathcal N_d$, and let
$h(x):=b+A(x-m)$ for some $b\in\R^d$ and
$A=A^\top\in\R^{d\times d}$. Then
\[
\int\langle\nabla\log(q/\pi)(x),h(x)\rangle\,dq(x)
=
\int\langle f_q(x),h(x)\rangle\,dq(x),
\]
where $f_q$ is given in Definition~\ref{def:bw-grad}.
\end{lem}

Subsequently we will also use the strong log-concavity of $\pi$,
which makes the KL divergence strongly convex along
Wasserstein geodesics.
 
\begin{lem}[{\citealp[Chapters~9--10]{ambrosio2008gradient} and \citealp[\S3]{lambert2022variational}}]\label{lem:bw-convex}
Suppose $F\in C^2$ and $\nabla^2F\succeq\mu I$. Let $q,q'\in\mathcal N_d$, and let $T$ be the optimal transport map from $q$ to $q'$. Then
\[
\mathrm{KL}(q'\|\pi)\ge\mathrm{KL}(q\|\pi)+\int\big\langle\nabla\log(q/\pi)(x),T(x)-x\big\rangle\,dq(x)+\frac\mu2W_2^2(q,q').
\]
\end{lem}
 
\begin{lem}[{\citealp{opper2009variational}; \citealp[\S3.3]{lambert2022variational}}]\label{lem:bw-opt}
Suppose $F\in C^2$ and $\mu I\preceq\nabla^2F\preceq I$, and let $q^\star=N(m^\star,\Sigma^\star)$ minimize $\mathrm{KL}(\cdot\|\pi)$ over $\mathcal N_d$. Then $\IE_{q^\star}[\nabla F]=0$ and $(\Sigma^\star)^{-1}=\IE_{q^\star}[\nabla^2F]$. Consequently, $f_{q^\star}\equiv0$ and $I\preceq\Sigma^\star\preceq\mu^{-1}I$.
\end{lem}
 Finally, we recall the contraction property of the clipping operator. 
 \begin{lem}[{\citealp[Lemma~E.3]{lambert2022variational}}]\label{lem:clip}
Let $\tau>0$, $\Sigma^\star\preceq\tau I$, $m,m^\star\in\R^d$ and $\Sigma\succ0$. Then
\[
W_2^2\big(N(m,\mathrm{clip}_\tau(\Sigma)),N(m^\star,\Sigma^\star)\big)\le W_2^2\big(N(m,\Sigma),N(m^\star,\Sigma^\star)\big).
\]
\end{lem}
 
\begin{proof}[Proof of Proposition~\ref{prop:bw-rec}]
Fix $t\ge1$, and define $T_{q_{t-1}}(x):=m^\star+T_{\Sigma_{t-1}}(x-m_{t-1})$ and
\[
M_t:=I-\eta_t(J_t-\Sigma_{t-1}^{-1}),\  \hat f_t(x):=g_t+(J_t-\Sigma_{t-1}^{-1})(x-m_{t-1}),\  x\in\R^d.
\]
Throughout the proof, integrals are taken with respect to a dummy variable $x\sim q_{t-1}$, with $(g_t,J_t)$ held fixed.

Note that it trivially holds that $\Sigma_t \succ 0$. Next, we compare $q_t$ with $q^\star$ through a coupling. Let $\Phi_t(x):=x-\eta_t\hat f_t(x)=m_{t-1}-\eta_tg_t+M_t(x-m_{t-1})$. If $X\sim q_{t-1}$, then $\Phi_t(X)$ is Gaussian with mean $m_{t-1}-\eta_tg_t=m_t$ and covariance $M_t\Sigma_{t-1}M_t=\Sigma^+_t$, so that the unclipped update $q^+_t:=N(m_t,\Sigma^+_t)$ satisfies $q^+_t=(\Phi_t)_\#q_{t-1}$, and $q_t=N(m_t,\mathrm{clip}_{1/\mu}(\Sigma^+_t))$. Since $\Sigma^\star\preceq\mu^{-1}I$ by Lemma~\ref{lem:bw-opt}, Lemma~\ref{lem:clip} with $\tau=1/\mu$ gives $Z_t\le W_2^2(q^+_t,q^\star)$. Moreover, $(T_{q_{t-1}})_\#q_{t-1}=q^\star$ by Lemma~\ref{lem:gauss-ot}, so that, for $X\sim q_{t-1}$, the law of $(\Phi_t(X),T_{q_{t-1}}(X))$ is a coupling of $q^+_t$ and $q^\star$. Therefore,
\begin{align}
Z_t\le W_2^2(q^+_t,q^\star)&\le\int\big\|x-T_{q_{t-1}}(x)-\eta_t\hat f_t(x)\big\|^2\,dq_{t-1}(x)\\*
&=Z_{t-1}-2\underbrace{\eta_t\int\big\langle\hat f_t(x),x-T_{q_{t-1}}(x)\big\rangle\,dq_{t-1}(x)}_{\text{first-order term}}\nonumber\\*
&\ +\underbrace{\eta_t^2\int\|\hat f_t(x)\|^2\,dq_{t-1}(x)}_{\text{second order term}}, \label{eq:bw-couple}
\end{align}
where we used $\int\|x-T_{q_{t-1}}(x)\|^2\,dq_{t-1}(x)=W_2^2(q_{t-1},q^\star)=Z_{t-1}$ from Lemma~\ref{lem:gauss-ot}. We treat the second-order and first-order terms seriatim.

Firstly, the second-order term is observable. Indeed, since $\int(x-m_{t-1})\,dq_{t-1}(x)=0$, $\int(x-m_{t-1})^\top B(x-m_{t-1})\,dq_{t-1}(x)=\Tr(B\Sigma_{t-1})$ for any matrix $B$, and $J_t-\Sigma_{t-1}^{-1}$ is symmetric,
\[
\int\|\hat f_t(x)\|^2\,dq_{t-1}(x)=\|g_t\|^2+\Tr\Bigl[(J_t-\Sigma_{t-1}^{-1})\Sigma_{t-1}(J_t-\Sigma_{t-1}^{-1})\Bigr]=\|g_t\|^2+\big\|(J_t-\Sigma_{t-1}^{-1})\Sigma_{t-1}^{1/2}\big\|^2.
\]

For the first-order term, note that $X_t\mid\Fcal_{t-1}\sim q_{t-1}$ gives $\IE_{t-1}g_t=\IE_{q_{t-1}}[\nabla F]$ and $\IE_{t-1}J_t=\IE_{q_{t-1}}[\nabla^2F]$, so that $\IE_{t-1}\hat f_t(x)=f_{q_{t-1}}(x)$ for every $x$, with $f_{q_{t-1}}$ as in Definition~\ref{def:bw-grad}. We accordingly write $$\hat f_t=\underbrace{f_{q_{t-1}}}_{\text{drift}}+\underbrace{(\hat f_t-f_{q_{t-1}})}_{\text{fluctuation}},$$ and treat the drift and the fluctuation separately. For the drift, since $x\mapsto T_{q_{t-1}}(x)-x$ is affine, Lemmas~\ref{lem:bw-proj} and~\ref{lem:bw-convex}, applied with $(q,q')=(q_{t-1},q^\star)$, give
\[
\begin{aligned}
&\int\big\langle f_{q_{t-1}}(x),x-T_{q_{t-1}}(x)\big\rangle\,dq_{t-1}(x)\\
&=-\int\big\langle\nabla\log(q_{t-1}/\pi)(x),T_{q_{t-1}}(x)-x\big\rangle\,dq_{t-1}(x)\\
&\ge\mathrm{KL}(q_{t-1}\|\pi)-\mathrm{KL}(q^\star\|\pi)+\frac\mu2Z_{t-1}.
\end{aligned}
\]
To control the difference of the KL divergences, let $T^\star$
be the optimal transport map from $q^\star$ to $q_{t-1}$.
By Lemma~\ref{lem:gauss-ot}, $x\mapsto T^\star(x)-x$
is also affine with symmetric linear part.
Lemmas~\ref{lem:bw-convex} and~\ref{lem:bw-proj},
applied with $(q,q')=(q^\star,q_{t-1})$, together with
$f_{q^\star}\equiv0$ from Lemma~\ref{lem:bw-opt}, give
\[
\mathrm{KL}(q_{t-1}\|\pi)-\mathrm{KL}(q^\star\|\pi)\ge\int\big\langle f_{q^\star}(x),T^\star(x)-x\big\rangle\,dq^\star(x)+\frac\mu2W_2^2(q^\star,q_{t-1})=\frac\mu2Z_{t-1}.
\]
Combining the two displays, $\int\langle f_{q_{t-1}}(x),x-T_{q_{t-1}}(x)\rangle\,dq_{t-1}(x)\ge\mu Z_{t-1}$. 

For the fluctuation, write $\xi_t=(\xi^g_t:\xi^J_t)$ with $\xi^g_t=\IE_{t-1}g_t-g_t$ and $\xi^J_t=\IE_{t-1}J_t-J_t$. Then $\hat f_t(x)-f_{q_{t-1}}(x)=-\bigl[\xi^g_t+\xi^J_t(x-m_{t-1})\bigr]$ and $x-T_{q_{t-1}}(x)=(m_{t-1}-m^\star)+(I-T_{\Sigma_{t-1}})(x-m_{t-1})$. Integrating against $q_{t-1}$, it follows that
\begin{align*}
&-2\eta_t\int\big\langle\hat f_t(x)-f_{q_{t-1}}(x),x-T_{q_{t-1}}(x)\big\rangle\,dq_{t-1}(x)\\
&\ =2\eta_t\Bigl[\langle\xi^g_t,m_{t-1}-m^\star\rangle+\Tr\big((\xi^J_t)^\top (I-T_{\Sigma_{t-1}})\Sigma_{t-1}\big)\Bigr]\\
&\ \overset{(a)}{=}2\eta_t\Bigl[\langle\xi^g_t,m_{t-1}-m^\star\rangle+\big\langle\xi^J_t,\mathrm{sym}\big((I-T_{\Sigma_{t-1}})\Sigma_{t-1}\big)\big\rangle\Bigr]\overset{(b)}{=}\langle v_t,\xi_t\rangle,
\end{align*}
where $(a)$ holds because $\xi^J_t$ is symmetric, and $(b)$ uses \S\ref{se:bw-opt}. Finally, substituting the three terms into \eqref{eq:bw-couple} yields
\[
Z_t\le(1-2\mu\eta_t)Z_{t-1}+\eta_t^2\bigl[\|g_t\|^2+\|(J_t-\Sigma_{t-1}^{-1})\Sigma_{t-1}^{1/2}\|^2\bigr]+\langle v_t,\xi_t\rangle,
\]
which completes the proof. 
\end{proof}
\begin{proof}[Proof of Theorem~\ref{thm:bw}]
We first verify $Z_0\le R_0$. By strong monotonicity of $\nabla F$, $\mu\|m_0-m^\star\|\le\|\nabla F(m_0)\|+\|\nabla F(m^\star)\|$, and, since $\IE_{q^\star}\nabla F=0$ and $\nabla F$ is $1$-Lipschitz, $\|\nabla F(m^\star)\|\le\IE_{q^\star}\|X-m^\star\|\le(\Tr(\Sigma^\star))^{1/2}\le\sqrt{d/\mu}$; the trace term in \eqref{eq:bw-w2} is at most $\Tr(\Sigma_0)+\Tr(\Sigma^\star)\le\Tr(\Sigma_0)+d/\mu$. The claim follows from Proposition~\ref{prop:bw-rec}, Theorem~\ref{thm:obs-comp} and Lemma~\ref{lem:observable-envelope}.
\end{proof}
\begin{proof}[Proof of Proposition~\ref{prop:bw-stat}]
Let $(X,X^\star)$ be an optimal coupling of $(q_t,q^\star)$. Then $|\int f\,dq_t-\int f\,dq^\star|\le L_f\IE\|X-X^\star\|\le L_fW_2(q_t,q^\star)\le L_f\sqrt{R_t}$, simultaneously for all $f$ and $t$. For the quantiles, $(w^\top X,w^\top X^\star)$ couples $N(w^\top m_t,\sigma_t^2)$ and $N(w^\top m^\star,\sigma_\star^2)$, with $\sigma_t^2:=w^\top\Sigma_tw$ and $\sigma_\star^2:=w^\top\Sigma^\star w$, so that, by \eqref{eq:bw-w2} with $d=1$,
\[
\bigl[w^\top(m_t-m^\star)\bigr]^2+(\sigma_t-\sigma_\star)^2=W_2^2\big(N(w^\top m_t,\sigma_t^2),N(w^\top m^\star,\sigma_\star^2)\big)\le\|w\|^2R_t.
\]
The difference of $p$-quantiles is $w^\top(m_t-m^\star)+z_p(\sigma_t-\sigma_\star)$, and the Cauchy--Schwarz inequality proves the claim. Finally, let $\bar u\in\arg\min_u\int\mathcal L(u,x)\,dq^\star(x)$. Applying the first assertion to $\mathcal L(u_t,\cdot)$ and $\mathcal L(\bar u,\cdot)$, and using the minimality of $u_t$,
\[
\begin{aligned}
\int\mathcal L(u_t,x)\,dq^\star&\le\int\mathcal L(u_t,x)\,dq_t+L_{\mathcal L}\sqrt{R_t}\\
&\le\int\mathcal L(\bar u,x)\,dq_t+L_{\mathcal L}\sqrt{R_t}\\
&\le\int\mathcal L(\bar u,x)\,dq^\star+2L_{\mathcal L}\sqrt{R_t}.
\end{aligned}
\]
\end{proof}

\section{Helper results and proofs}\label{app:lemma-proofs}
\begin{lem}\label{lem:self-bounding}
If $\|\nabla F(x)\|^2\le K_\bullet(F(x)-F^\star)$ for all $x\in\R^d$, then $F(x)-F^\star\le K_\bullet\|x-x^\star\|^2/4$ and $\|\nabla F(x)\|\le K_\bullet\|x-x^\star\|/2$ for all $x\in\R^d$.
\end{lem}
\begin{proof}
For $\epsilon>0$, $\|\nabla(F-F^\star+\epsilon)^{1/2}\|=\|\nabla F\|/\bigl[2(F-F^\star+\epsilon)^{1/2}\bigr]\le\sqrt{K_\bullet}/2$ leads to $(F(x)-F^\star+\epsilon)^{1/2}\le\sqrt\epsilon+\sqrt{K_\bullet}\|x-x^\star\|/2$. Letting $\epsilon\downarrow0$ yields the first claim, and $\|\nabla F(x)\|\le\bigl[K_\bullet(F(x)-F^\star)\bigr]^{1/2}$ yields the second.
\end{proof}
\begin{lem}
\label{lem:GE-explicit}
Recall $m_{\rho,c}$, $\mathfrak{a}(v,b)$  and $\mathfrak{L}(v,b)$ from \eqref{eq:GE-root}, \eqref{eq:GE-defn}, and \eqref{eq:GE-level}. 
Then,it follows $$\sup\{x\in\mathbb R:m_{\rho,c}(x;v,b)<1/\alpha\}
\le u_{\alpha, \rho, c}(v,b):=
c(b-1)+cbD_{\mathfrak{a}(v,b)}(\mathfrak{L}(v,b)),$$ where $D_a(L)$ is as in \eqref{eq:GE-D-defn}.
\end{lem}

\begin{proof}
Let $h(z; v,b):=z-\mathfrak{a}(v,b)\log(1+z/\mathfrak{a}(v,b))$ for $z\ge0$.
By Proposition~2.7 of \cite{pinelis2005},
$\nu(a,y)\le y^ae^{-y}/(y-a+1)$ for $a\ge1$ and $y>a-1$,
where $\nu(a,y):=\int_y^\infty t^{a-1}e^{-t}\,dt$.
Consequently, \eqref{eq:GE-root} gives, for every $z\ge0$,
\begin{equation}\label{eq:GE-lower}
\begin{aligned}
m_{\rho,c}(c(b-1)+cbz;v,b)
&=
\frac{\kappa}{b}
\frac{e^{\mathfrak{a}(v,b)+z}}{(\mathfrak{a}(v,b)+z)^{\mathfrak{a}(v,b)}}
\big(\Gamma(\mathfrak{a}(v,b))-\nu(\mathfrak{a}(v,b),\mathfrak{a}(v,b)+z)\big)\\
&\ge
\frac{\kappa}{b}
\biggl[\frac{e^{h(z; v,b)}}q-\frac1{1+z}\biggr].
\end{aligned}
\end{equation}
Now we control $h(z;v,b)$ explicitly. In view of the monotonicity of $h$ over $z \ge 0$, applying Lemma 2 of \cite{chatzigeorgiou2013bounds} yields
\begin{equation}\label{eq:GE-inversion}
h\biggl(\sqrt{2\mathfrak{a}(v,b)\ell}+\frac23\ell; v,b\biggr)
\le \ell
\le h\Bigl(\sqrt{2\mathfrak{a}(v,b)\ell}+\ell; v,b\Bigr).
\end{equation}
Write $a:=\mathfrak{a}(v,b)$, $L:=\mathfrak{L}(v,b)$,
and $d:=\sqrt{2aL}+\frac23L$.
For $L>0$, let
$d_1:=d+[L-h(d;v,b)]/h'(d;v,b)$
be the Newton update for $h(z;v,b)=L$.
Since $h'(d;v,b)=d/(a+d)>0$,
\eqref{eq:GE-D-defn} reads
$D_a(L)=\min\{\sqrt{2aL}+L,d_1\}$.
By \eqref{eq:GE-inversion}, both entries are at least $d$,
and the first has $h$-value at least $L$.
For the second, convexity gives
$h(d_1;v,b)\ge h(d;v,b)+h'(d;v,b)(d_1-d)=L$.
Consequently, using $\mathfrak{L}(v,b)\ge\mathfrak{L_0}(v,b)$,
\[
D_a(L)\ge d\ge
\sqrt{2a\mathfrak{L_0}(v,b)}+\frac23\mathfrak{L_0}(v,b),
\qquad
h(D_a(L);v,b)\ge L.
\]
Both conclusions also hold when $L=0$, since then
$D_a(0)=d=\mathfrak{L_0}(v,b)=0$.
Combining the above, we obtain
\begin{align*}
&m_{\rho,c}(c(b-1)+cbD_{\mathfrak{a}(v,b)}(\mathfrak{L}(v,b));v,b)\\
&\ge
\frac{\kappa}{b}\Biggl[\frac{\exp(\mathfrak{L}(v,b))}{q(v,b)}-\frac1{1+D_{\mathfrak{a}(v,b)}(\mathfrak{L}(v,b))}\Biggr]\\
&\ge
\frac1\alpha+
\frac{\kappa}{b}
\Biggl[
\frac{1}{1+\sqrt{2a\mathfrak{L_0}(v,b)}+\frac23\mathfrak{L_0}(v,b)}
-\frac1{1+D_{\mathfrak{a}(v,b)}(\mathfrak{L}(v,b))}
\Biggr] \nonumber\\
&\ge\frac1\alpha.
\end{align*}
The conclusion follows since \eqref{eq:GE-root} is strictly
increasing in $x$.
\end{proof}
\begin{lem}\label{lem:mom-constants}
Let $\eta_t=\eta_\circ\log^{-p}(t+2)$, $t\ge0$, with $\eta_\circ>0$ and $p>1$. Recall $H_p$ and $C_\circ$ from \eqref{eq:mom-constants}, and $c_t$ from Lemma~\ref{lem:mom-aux}. Then it follows that
\begin{equation}
\sum_{t\ge1}\frac{\eta_t}{t}\le H_p,\ \text{and}\ \sum_{t\ge1}\frac{c_t\eta_t}{t}\le C_\circ. \nonumber
\end{equation}
\end{lem}
\begin{proof}
For the first claim, $\eta_1=\eta_\circ(\log3)^{-p}$, and, for $t\ge2$, $1/t\le3/(t+2)$ gives $\eta_t/t\le3\eta_\circ f(t)$ with the decreasing function $f(x):=\bigl[(x+2)\log^p(x+2)\bigr]^{-1}$. The integral test then yields $\sum_{t\ge2}f(t)\le\int_1^\infty f(x)\,dx=(\log3)^{1-p}/(p-1)$, and adding the contribution of $t=1$ gives $H_p$. For the second claim, $\eta_t\le\eta_\circ$ and $\eta_t\le\eta_{t-1}$ for $t\ge1$ give $\eta_t/\sqrt{\eta_{t-1}}\le\sqrt{\eta_\circ}$, so that
\[
\frac{c_t\eta_t}{t}\le\frac{L^2\eta_\circ}{t(t+1)}+\frac{L\sqrt{\eta_\circ}}{2(t+1)\sqrt t}\le\frac{L^2\eta_\circ}{t(t+1)}+\frac{L\sqrt{\eta_\circ}}2\,t^{-3/2};
\]
summing over $t$, and using $\sum_{t\ge1}\bigl[t(t+1)\bigr]^{-1}=1$ along with $\sum_{t\ge1}t^{-3/2}=\zeta(3/2)$, proves the claim.
\end{proof}

\end{document}